\documentclass{article}

\usepackage{booktabs}
\usepackage{subfigure}

\usepackage{graphicx}
\usepackage{multirow}
\usepackage{enumitem}
\usepackage{amsmath}
\usepackage{amssymb}
\usepackage{amsthm}
\usepackage{stmaryrd}
\usepackage{algorithm}
\usepackage{algorithmic}
\usepackage{footmisc}
\usepackage{hyperref}
\usepackage{xspace}
\usepackage{color}
\usepackage{wrapfig}

\newcommand{\model}{\textsc{Grand}}
\newcommand{\full}{\textsc{Graph Random Neural Networks}}

\newtheorem{thm}{Theorem} 
\newcommand{\tabincell}[2]{\begin{tabular}{@{}#1@{}}#2\end{tabular}}

\newcommand{\vpara}[1]{\vspace{0.0in}\noindent\textbf{#1}\xspace}

\newcommand{\reminder}[1]{\textbf{\color{red}[** #1 **]}} 

\newcommand{\yd}[1]{\textit{{\color{red}#1 }}}

\makeatletter

\usepackage{footnote}
\usepackage{footmisc}
\makesavenoteenv{tabular}
\makesavenoteenv{table}

\newcommand{\astfootnote}[1]{
\let\oldthefootnote=\thefootnote
\setcounter{footnote}{0}
\footnote{#1}
\let\thefootnote=\oldthefootnote
}

\usepackage{threeparttable}

\newcommand{\hide}[1]{} 

\usepackage{caption}
\setlist[itemize]{leftmargin=3.5mm}
\usepackage[final,nonatbib]{neurips_2020}
\usepackage{amsthm}
\usepackage[utf8]{inputenc} 
\usepackage[T1]{fontenc}    
\usepackage{hyperref}       
\usepackage{url}            
\usepackage{booktabs}       
\usepackage{amsfonts}       
\usepackage{nicefrac}       
\usepackage{microtype}      
\usepackage[page]{appendix}

\title{Graph Random Neural Networks for Semi-Supervised Learning on Graphs}

\author{%
Wenzheng Feng$^{1*}$, Jie Zhang$^{2*\ddagger}$, Yuxiao Dong$^3$, Yu Han$^1$, Huanbo Luan$^1$, Qian Xu$^2$,\\
\textbf{Qiang Yang$^2$, Evgeny Kharlamov$^4$, Jie Tang$^1$$^\S$} \\
$^1$ Department of Computer Science and Technology, Tsinghua University 
\\
  $^2$WeBank Co., Ltd \quad $^3$Microsoft Research \quad
  $^4$ Bosch Center for Artificial Intelligence\\
{fwz17@mails.tsinghua.edu.cn, \{zhangjie.exe, ericdongyx, yhanthu,  luanhuanbo\}@gmail.com}
\\
{\{qianxu, qiangyang\}@webank.com, evgeny.kharlamov@de.bosch.com, 
jietang@tsinghua.edu.cn}}

\begin{document}

\maketitle

\renewcommand{\thefootnote}{\fnsymbol{footnote}}
\footnotetext[1]{Equal contribution.}
\footnotetext[3]{Work performed while at Tsinghua University.}
\footnotetext[4]{Corresponding author.}
\renewcommand{\thefootnote}{\arabic{footnote}}
\begin{abstract}
\hide{
Graph neural networks (GNNs) have generalized deep learning methods into graph-structured data with promising performance on graph mining tasks. 
However, existing GNNs often meet complex graph structures with scarce labeled nodes and suffer from the limitations of 
non-robustness~\cite{zugner2018adversarial,ZhuRobust}, over-smoothing~\cite{chen2019measuring,li2018deeper,Lingxiao2020PairNorm}, and overfitting~\cite{goodfellow2016deep,Lingxiao2020PairNorm}. 
To address these issues, we propose a simple yet effective GNN framework---Graph Random Neural Network (\model). 
Different from the deterministic propagation in existing GNNs, \model\ adopts a random propagation strategy to enhance model robustness. 
This strategy also naturally enables \model\ to decouple the propagation from feature transformation, reducing the risks of over-smoothing and overfitting. 
Moreover, random propagation acts as an efficient method for graph data augmentation. 
Based on this, we propose the consistency regularization for \model\ by leveraging the distributional consistency of unlabeled nodes in multiple augmentations, improving the generalization capacity of the model. 
Extensive experiments on graph benchmark datasets suggest that \model\ significantly outperforms state-of-the-art GNN baselines on semi-supervised graph learning tasks. 
Finally, we show that \model\ mitigates the issues of over-smoothing and overfitting, and its performance is married with robustness. 

  We study the problem of semi-supervised node classification on graphs. Recent graph neural networks (GNNs) 
 for this task suffer from the limitations of non-robustness~\cite{zugner2018adversarial,ZhuRobust} and over-smoothing~\cite{chen2019measuring,li2018deeper,Lingxiao2020PairNorm}, and have a weak generalization ability when labeled nodes are scarce. In this paper, we propose \full\ (\model) to address these problems.  
 \model\ adopts a simple random propagation strategy to perform graph data augmentation, and utilize consistency regularization to optimize model's prediction consistency of unlabeled nodes in multiple augmentations. With a simple MLP model as classifier, \model\ significantly outperforms state-of-the-art GNNs baselines on semi-supervised graph learning tasks, and mitigates the non-robustness and over-smoothing issues empirically. This indicates that complex GNN structures might not be necessary for this task, we could achieve better performance with a simple prediction model and some regularization techniques.
 We also provide a theorerical framework to understand the effects of random propagation and consistency regularization.
  }

 We study the problem of semi-supervised learning on graphs, for which graph neural networks (GNNs) have been extensively explored. 
 However, most existing GNNs inherently suffer from the limitations of  over-smoothing~\cite{chen2019measuring,li2018deeper,Lingxiao2020PairNorm,oono2020graph}, non-robustness~\cite{zugner2018adversarial,ZhuRobust}, and weak-generalization when labeled nodes are scarce. 
 In this paper, we propose a simple yet effective framework---\full\ (\model)---to address these issues.  
 In \model,  we first design a random propagation strategy to perform graph data augmentation.
 Then we leverage consistency regularization to optimize the prediction consistency of unlabeled nodes across different data augmentations. 
 Extensive experiments on graph benchmark datasets suggest that \model\ significantly outperforms state-of-the-art GNN baselines on semi-supervised node classification.  
 Finally, we show that \model\ mitigates the issues of over-smoothing and non-robustness, exhibiting better generalization behavior than existing GNNs. The source code of \model\ is publicly available at \url{https://github.com/Grand20/grand}.

\end{abstract}

\section{Introduction}

Graphs serve as a common language for modeling structured and relational data~\cite{leskovec2016snap}, such as  social networks, knowledge graphs, and the World Wide Web. 
Mining and learning graphs can benefit various real-world problems and applications. 
The focus of this work is on the problem of semi-supervised learning on graphs~\cite{zhu2003semi,kipf2016semi,ding2018semi}, which aims to predict the categories of unlabeled nodes of a given graph with only a small proportion of labeled nodes. 
Among its solutions, graph neural networks (GNNs)~\cite{kipf2016semi,hamilton2017inductive, Velickovic:17GAT,abu2019mixhop} have recently emerged as powerful approaches. 
The main idea of GNNs lies in a deterministic feature propagation process to learn expressive node representations. 

However, recent studies show that such propagation procedure brings some inherent issues: 
First, most GNNs suffer from \textit{over-smoothing}~\cite{li2018deeper,chen2019measuring,Lingxiao2020PairNorm,oono2020graph}. 
Li et al. show that the graph convolution operation is a special form of Laplacian smoothing~\cite{li2018deeper}, and consequently, stacking many GNN layers tends to make nodes' features indistinguishable. In addition, a very recent work~\cite{oono2020graph} suggests that the coupled non-linear transformation in the propagation procedure can further aggravate this issue.
Second, GNNs are often \textit{not robust} to graph attacks~\cite{zugner2018adversarial,ZhuRobust}, due to the deterministic propagation adopted in most of them. 
Naturally, the deterministic propagation makes each node highly dependent with its (multi-hop) neighborhoods, leaving the nodes to be easily misguided by potential data noise and susceptible to adversarial perturbations. 

The third issue lies in the general setting of semi-supervised learning, wherein standard training methods (for GNNs) can easily \textit{overfit} the scarce label information~\cite{chapelle2009semi}. 
Most efforts to addressing this broad issue are focused on how to fully leverage the large amount of unlabeled data. 
In computer vision, recent attempts, e.g. MixMatch~\cite{berthelot2019mixmatch}, UDA~\cite{xie2019unsupervised}, have been proposed to solve this problem by designing data augmentation methods for consistency regularized training, which have achieved great success in the semi-supervised image classification task. This inspires us to apply this idea into GNNs to facilitate semi-supervised learning on graphs.

In this work, we address these issues by designing graph data augmentation and consistency regularization strategies for semi-supervised learning. 
Specifically, we present the \full\ (\model), a simple yet powerful  graph-based semi-supervised learning framework.

To effectively augment graph data, we propose random propagation in \model, wherein each node's features can be randomly dropped either partially (dropout) or entirely,  
after which the perturbed feature matrix is propagated over the graph. 
As a result, each node is enabled to be insensitive to specific neighborhoods, \textit{increasing the robustness of \model.} 
Further, the design of random propagation can naturally separate feature propagation and transformation, which are commonly coupled with each other in most GNNs. 
This empowers \model\ to safely perform  higher-order feature propagation without increasing the complexity, \textit{reducing the risk of over-smoothing for \model.} 
More importantly, random propagation enables each node to randomly pass messages to its neighborhoods. 
Under the assumption of homophily of graph data~\cite{mcpherson2001birds}, we are able to stochastically generate different augmented representations for each node. 
We then utilize consistency regularization to enforce the prediction model, e.g., a simple Multilayer Perception (MLP), to output similar predictions on different augmentations of the same unlabeled data, \textit{improving \model's generalization behavior under the semi-supervised setting.} 

\hide{
In \model, we propose random propagation to effectively augment graph data. 
In random propagation, each node's features can be randomly dropped either partially (dropout) or entirely,  
after which the perturbed feature matrix are propagated over the graph. 

In doing so, each node is enabled to randomly pass messages to its neighborhoods. 
Under the assumption of local smoothness of graph data~\cite{dakovic2019local}, we are able to stochastically generate different augmented representations for each node. 
With a simple Multilayer Perception (MLP) as the prediction model, we utilize consistency regularization to enforce the MLP to output similar predictions on different augmentations of the same unlabeled data. 
}


Finally, we theoretically illustrate that \textit{random propagation} and \textit{consistency regularization} can enforce the consistency of classification confidence between each node and its multi-hop neighborhoods.
Empirically, we also show both strategies can improve the generalization of \model, and mitigate the issues of non-robustness and over-smoothing that are commonly faced by existing GNNs. 
Altogether, extensive experiments demonstrate that \model\ achieves state-of-the-art semi-supervised learning results on GNN benchmark datasets. 


\hide{
\begin{figure*}
    \centering
    \includegraphics[width=0.99\linewidth]{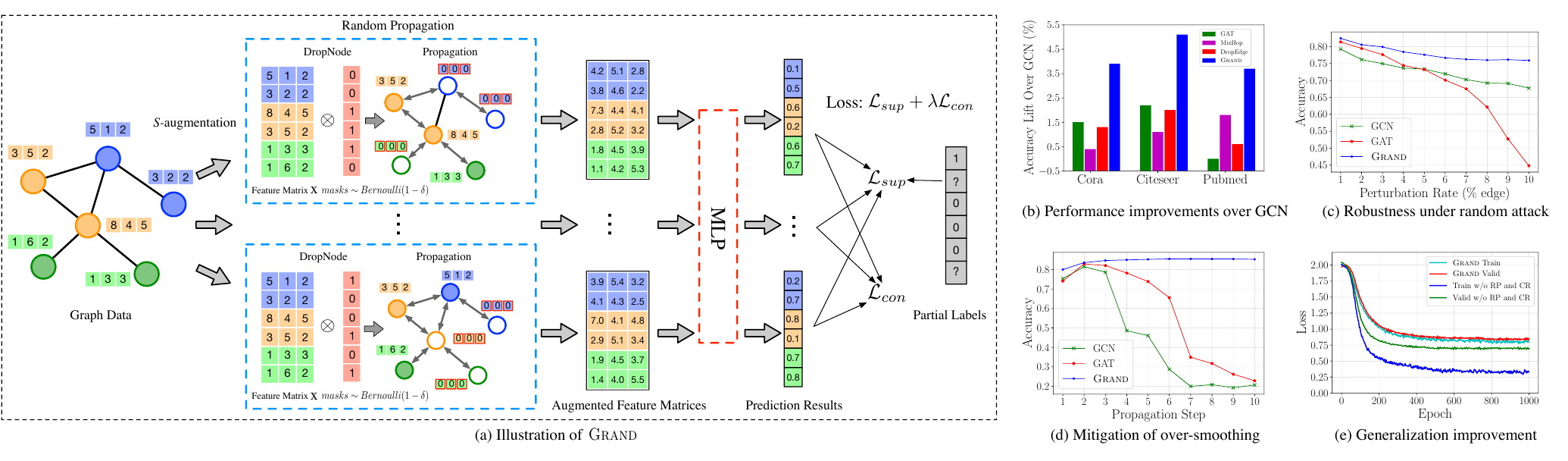}
    \caption{\model's architecture, performance, robustness, and mitigation of over-smoothing \& overfitting demonstrated}
\label{fig:grand_intro}
\end{figure*}
}

\hide{
\begin{figure*}
\centering
\captionsetup[sub]{font=small,labelfont={bf,sf}}

\includegraphics[width=0.59\textwidth]{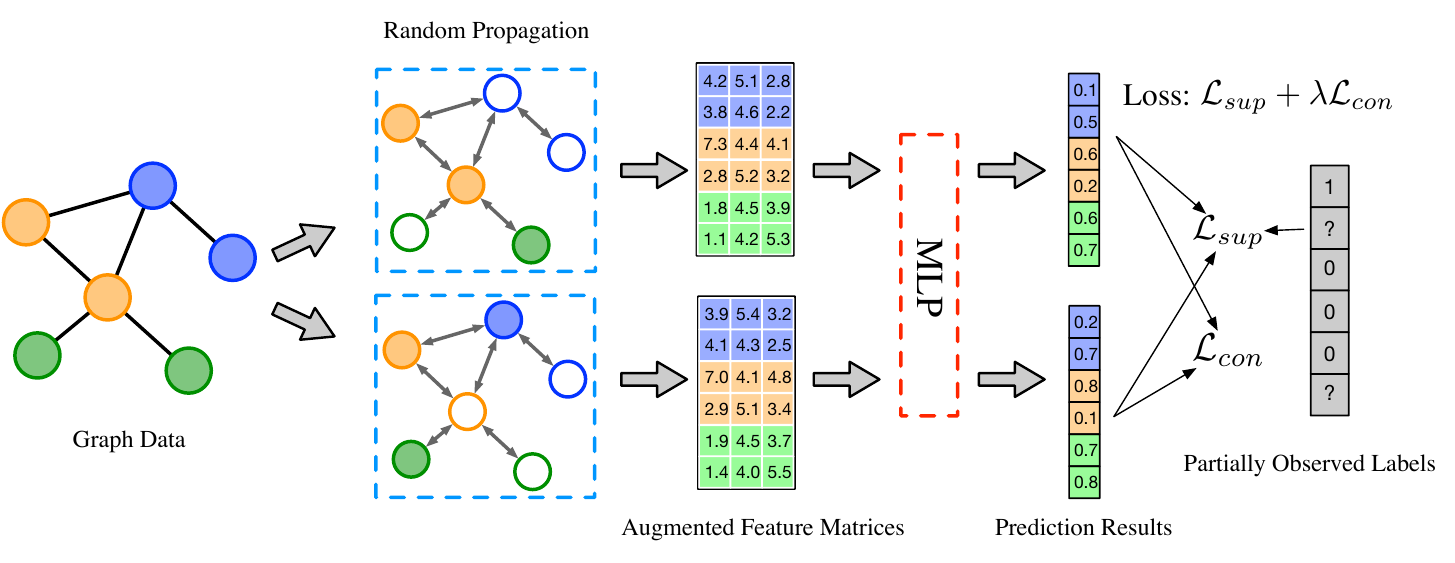}
\hfill
\begin{minipage}[b]{0.19\textwidth}
  \centering
  \includegraphics[width=0.9\textwidth]{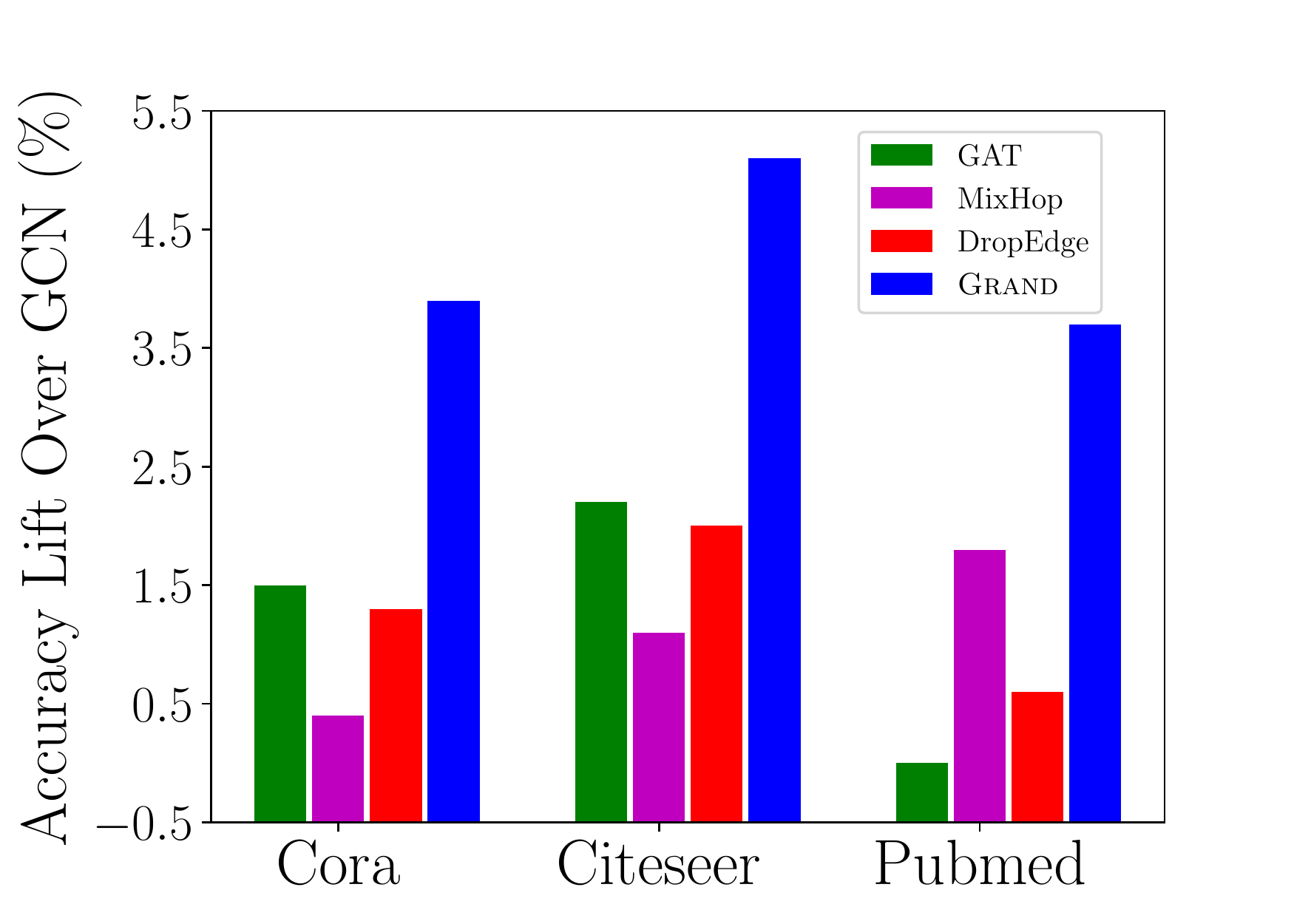}
    \vfill
  \includegraphics[width=0.9\textwidth]{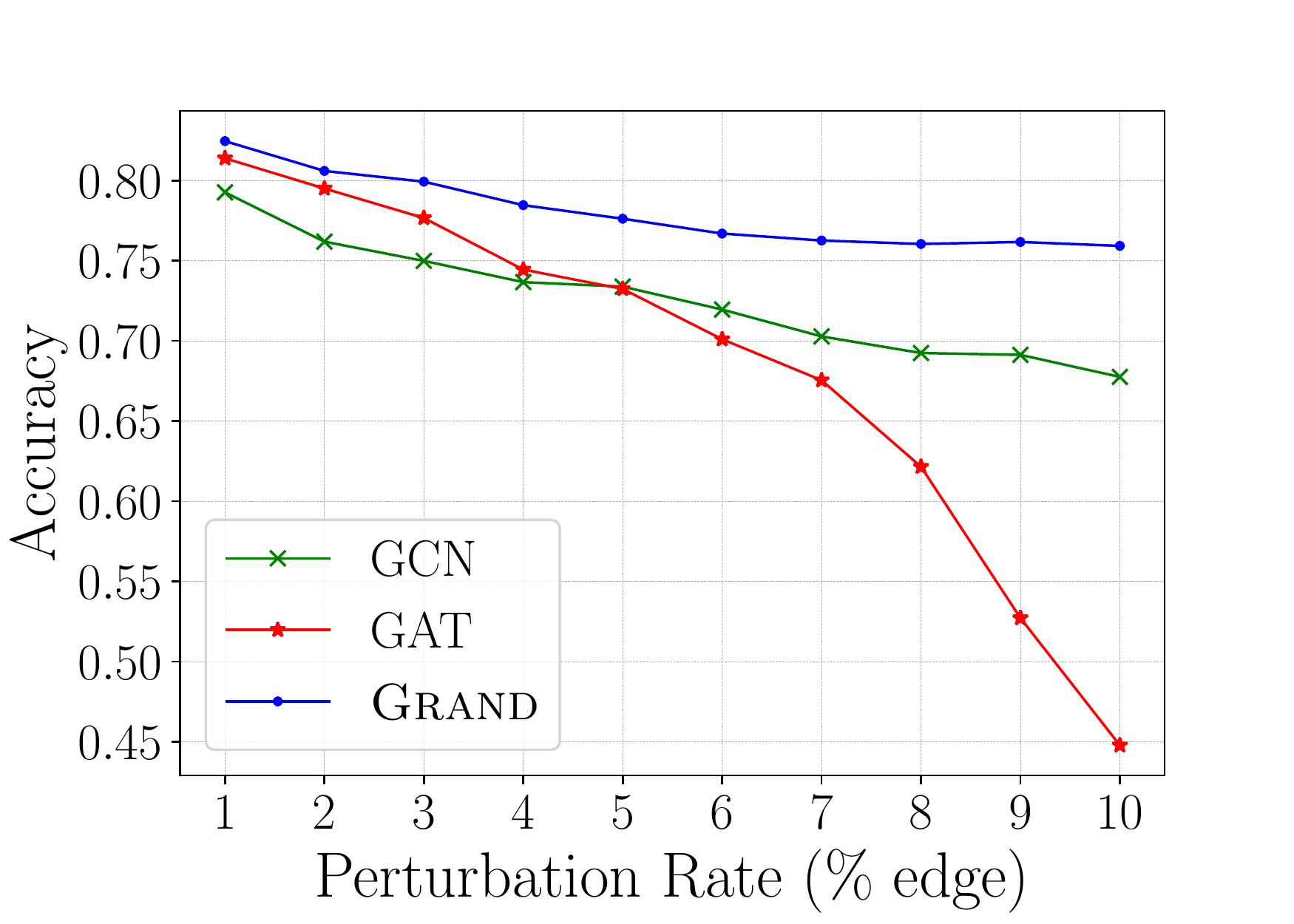}
 \end{minipage}
\hfill
\begin{minipage}[b]{0.19\textwidth}
  \subfigure[Generalization improvement]{
  \includegraphics[width=0.9\textwidth]{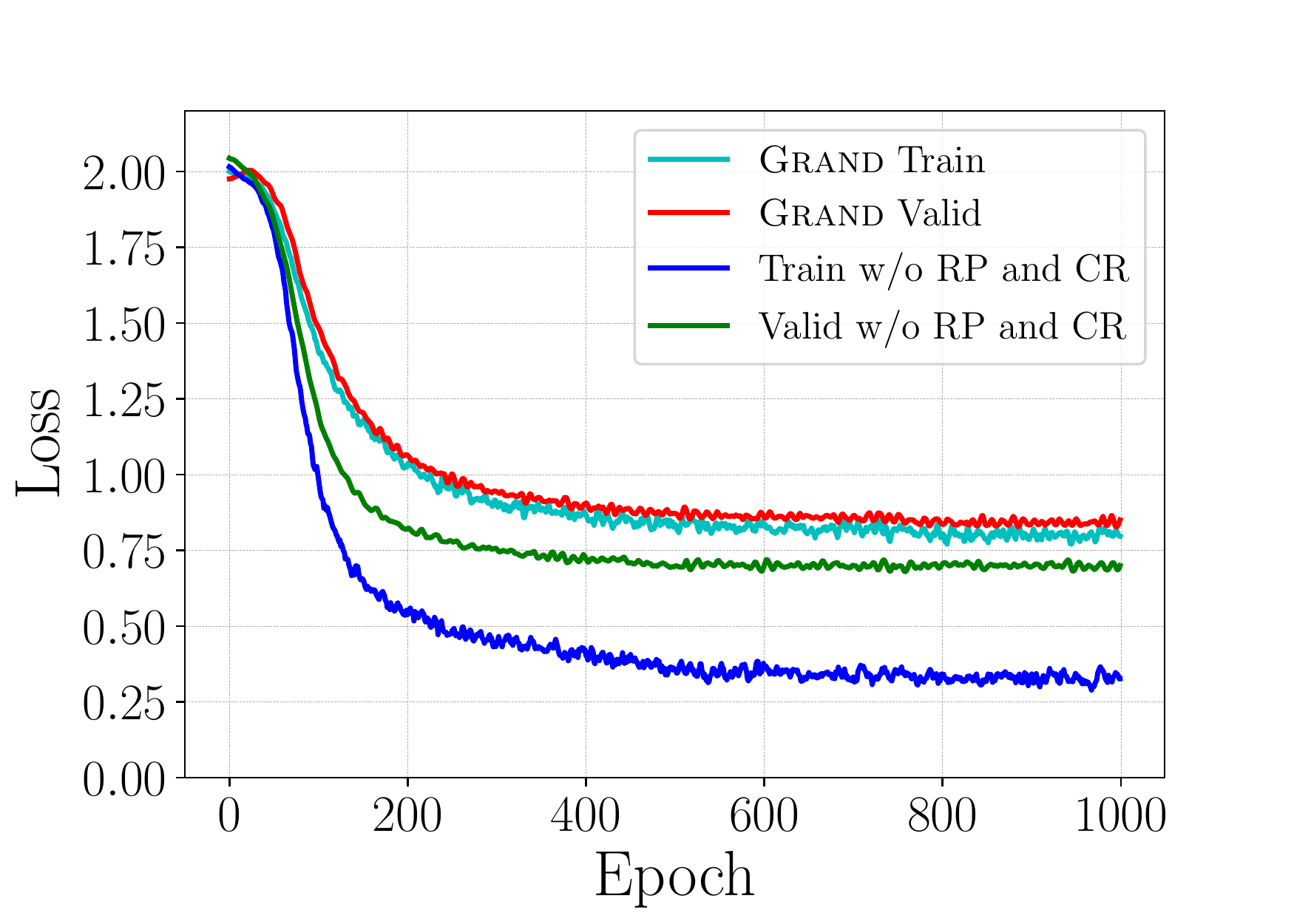}
  }
  \vfill
  \subfigure[Mitigation of over-smoothing]{
  \includegraphics[width=0.9\textwidth]{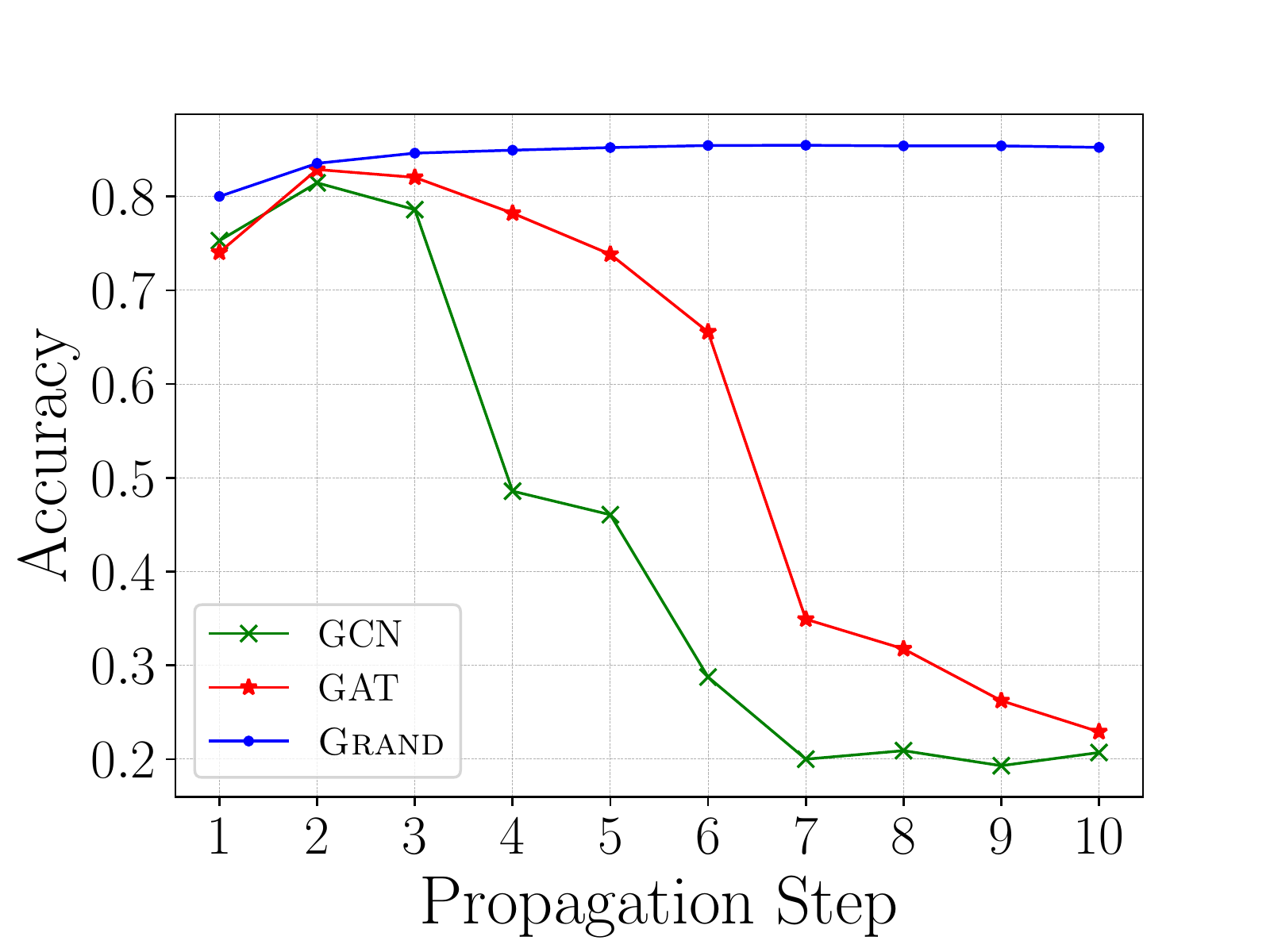}}
  
\end{minipage}
\caption{A caption}
\end{figure*}
}

\hide{

Graph-structured data is ubiquitous and exists in many real applications, such as social networks, the World Wide Web, and the knowledge graphs. In this paper, we consider the problem of semi-supervised learning on graphs, which aims to predict the categories of unlabeled nodes of a given graph, with only a small proportion of labeled nodes. With the developing of deep learning, graph neural networks (GNNs)~\cite{kipf2016semi,Velickovic:17GAT,qu2019gmnn,abu2019mixhop} have been proposed to solve this problem with utilizing a deterministic feature propagation rule to learn more expressive node embeddings. For example, the graph convolutional network (GCN)~\cite{kipf2016semi} propagates information based on the normalized Laplacian of the input graph, which is coupled with the non-linear transformation process.  
Recent study shows that such propagation procedure brings some inherent issues:
1)  \textit{non-robustness}~\cite{zugner2018adversarial,ZhuRobust}. The deterministic propagation in most GNNs naturally makes each node to be highly dependent with its (multi-hop) neighborhoods, leaving the nodes to be easily misguided by potential data noise and susceptible to adversarial perturbations. 
2) \textit{over-smoothing}~\cite{chen2019measuring,li2018deeper,Lingxiao2020PairNorm,oono2020graph}. The convolution operation is essentially a special form of Laplacian smoothing~\cite{li2018deeper}, and stacking many layers tends to make nodes' features indistinguishable. Recent study~\cite{oono2020graph} also shows that the coupled non-linear transformation will aggravate this problem.

On the other hand, under the semi-supervised setting, standard training methods for neural networks are easy
to \textit{overfit} due to the scarce supervised information~\cite{chapelle2009semi}. How to improve model's generalization capacity by fully-leveraging the large amount of unlabeled data is the main challenge in semi-supervised learning. In other domain, such as computer vision, an effective solution to this problem is consistency regularization~\cite{berthelot2019mixmatch}. In that, we first perform stochastic data augmentation~\cite{devries2017dataset,goodfellow2016deep, vincent2008extracting} onto the image via some transformations 
assumed to leave label unaffected (e.g. flipping, rotation), then the classifier is enforced to give similar predictions onto different augmentations of the same data. This kinds of methods have achieved impressive performance improvements on semi-supervised image classification~\cite{berthelot2019remixmatch,sohn2020fixmatch}. However, this idea can not be directly applied onto the graph data, we should design an effective graph data augmentation methods first. Though some related efforts have been devoted to this problem, such as GraphMix~\cite{verma2019graphmix} augmented graph data via performing linear interpolation between two nodes, it does not 
consider underlying structural information and may not be the best option for graph data.


In this paper, we develop Graph Random Neural Network (\model), a simple yet effective graph-based semi-supervised learning framework. In \model, we propose a powerful graph data augmentation method --- random propagation. In that, each node's features are designed to be randomly dropped entirely or partially (analogous to dropout~\cite{srivastava2014dropout}), then the perturbed feature matrix are propagated over the graph. In doing so, each node randomly passes messages to its neighborhoods. Under the assumption of local smoothness of graph data, we are able to stochastically generate different augmented representations for each node. With a simple Multilayer Perception (MLP) as prediction model, we utilize consistency regularization~\cite{berthelot2019mixmatch} to enforce the MLP to output similar predictions on different augmentations of the same unlabeled data. 
Extensive experiments show that \model\ achieves state-of-the-art results on three GNN benchmark datasets (Cf. Section \ref{sec:overall}), as well as six publicly available large datasets (Cf. Appendix \ref{exp:large_data}). We also provide theoretical analyses to understand the regularization effects of the proposed \textit{random propagation} (RP) and \textit{consistency regularization} (CR) strategies (Cf. Section~\ref{sec:theory}), and empirically show that both RP and CR can improve model's generalization capacity~\ref{sec:generalize}. What's  more, the design of random propagation lets each node is enabled to be not sensitive to specific neighborhoods, and naturally   
separate feature propagation and non-linear transformation, mitigating the problem of non-robustness (Cf. Section~\ref{sec:robust}) and over-smoothing (Cf. Section~\ref{sec:oversmoothing}), those common issues faced by most GNNs. 

}

\hide{
In summary, the proposed \model\ leads to the following benefits:
\begin{itemize}
    \item Extensive experiments show that \model\ obtains state-of-the-art results on three GNN benchmark datasets (Cf. Section \ref{sec:exp}), as well as six publicly available large datasets (Cf. Appendix \ref{exp:large_data}). That indicates that   
    \item We provide theoretical analyses to understand the regularization effects of the proposed \textit{random propagation} (RP) and \textit{consistency regularization} (CR) strategies on GNNs (Cf. Section~\ref{sec:theory}), and empirically show that both RP and CR can improve model's generalization capacity~\ref{sec:generalize}.
    \item We demonstrate that \model\ mitigates non-robustness (Cf. Appendix~\ref{sec:robust}) and over-smoothing (Cf. Appendix~\ref{sec:oversmoothing}), those common issues faced by most GNNs. 
\end{itemize}
}

\hide{

\begin{figure*}[ht]
	\centering
	{
		\subfigure[Performance improvements over GCN]{
				\centering
				\includegraphics[width = 0.23\textwidth]{pics/Acc_lift.pdf}
			}
		\subfigure[Robustness under random attack]{
				\centering
				\includegraphics[width = 0.23 \textwidth]{pics/random_attack_intro.pdf}
			}
		\subfigure[Mitigation of over-smoothing]{
				\centering
				\includegraphics[width = 0.23 \textwidth]{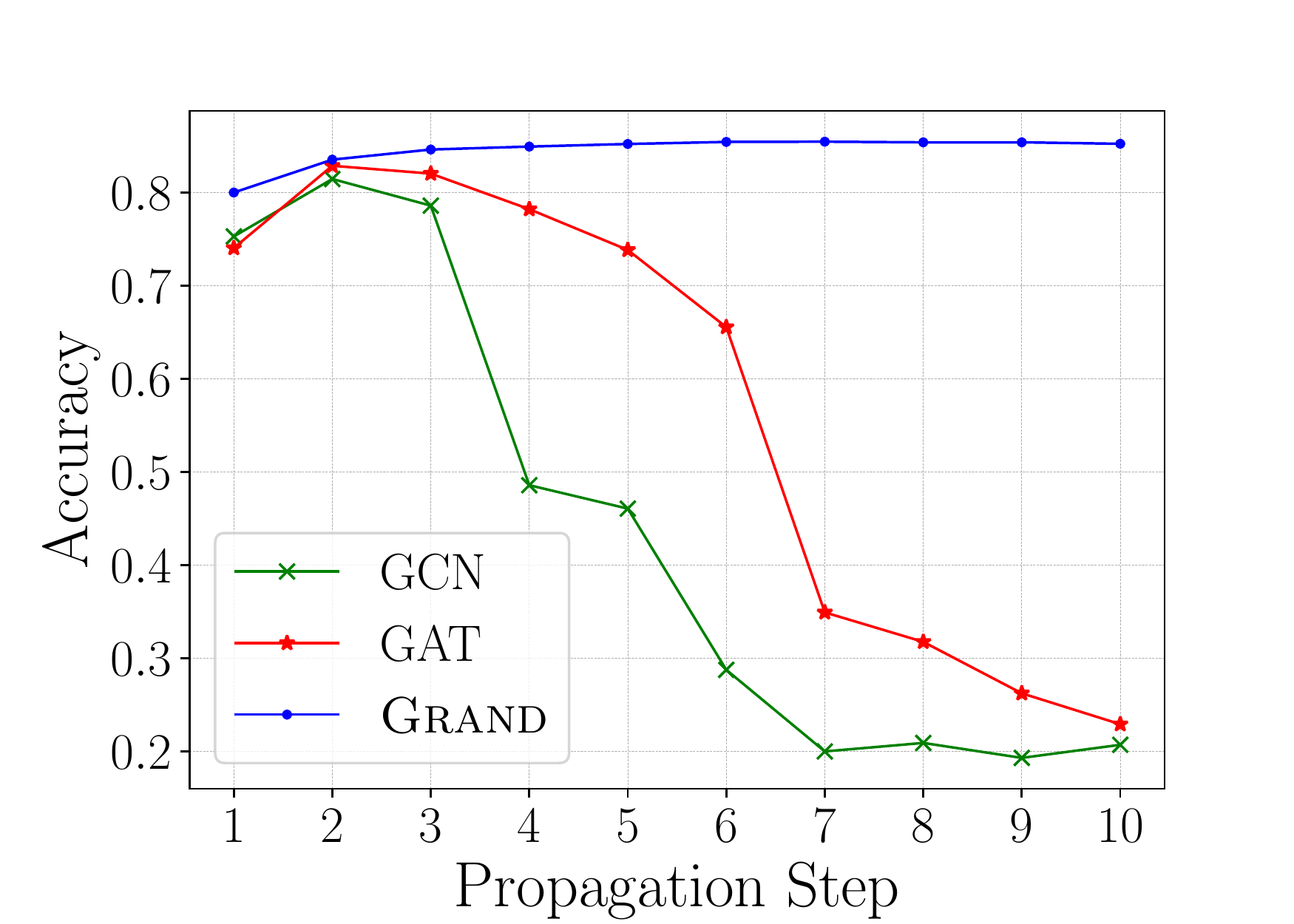}
			}
		\subfigure[Generalization improvement]{
				\centering
				\includegraphics[width = 0.23 \textwidth]{pics/gen_intro.pdf}
			}
	}
	\caption{\model's performance, robustness, and mitigation of over-smoothing \& overfitting demonstrated.}
	\label{fig:grand_intro}
\end{figure*}

The success of deep learning has provoked interest to generalize neural networks to structured data, marking the emergence of graph neural networks (GNNs)~\cite{gori2005new, bruna2013spectral, gilmer2017neural}. 
Over the course of its development, GNNs have been shifting the paradigm of graph mining from  structural explorations to representation learning. 
A wide variety of graph applications have benefited from this shift, such as node
classification~\cite{kipf2016semi,Chiang2019ClusterGCN}, link prediction~\cite{zhang2018link,ying2018graph,Wang2019KGCN}, and  graph classification~\cite{xu2018powerful,Ma2019EigenPooling}.  

The 
essential procedure in GNNs is the feature propagation, which is usually performed by some deterministic propagation rules derived from the graph structures. 
For example,  the graph convolutional network (GCN)~\cite{kipf2016semi} propagates information based on the normalized Laplacian of the input graph, which is coupled with the feature transformation process.  
Such a propagation can be also viewed as modeling each node's neighborhood as a receptive field and enabling a recursive neighborhood propagation process by stacking multiple GCN layers. 
Further, this process can be also unified into the neural message passing framework ~\cite{gilmer2017neural}. 

Recent attempts to advance the propagation based architecture include adding self-attention (GAT)~\cite{Velickovic:17GAT}, integrating with graphical models (GMNN)~\cite{qu2019gmnn}, and neighborhood mixing (MixHop)~\cite{abu2019mixhop}, etc. 
However, while the propagation procedure can enable GNNs to achieve attractive performance, it also brings some inherent issues that have been  recognized recently, including  non-robustness~\cite{zugner2018adversarial,ZhuRobust}, over-smoothing~\cite{chen2019measuring,li2018deeper,Lingxiao2020PairNorm}, and overfitting~\cite{goodfellow2016deep,Lingxiao2020PairNorm}.

\textbf{Non-robustness.}  
The deterministic propagation in most GNNs naturally makes each node to be highly dependent with its (multi-hop) neighborhoods.  
This leaves the nodes to be easily misguided by potential data noise,
making GNNs non-robust~\cite{ZhuRobust}. 
For example, it has been 
shown that GNNs are very susceptible to adversarial  attacks, and the attacker can indirectly attack the target node by manipulating 
long-distance neighbors~\cite{zugner2018adversarial}. 


\textbf{Overfitting.}
To make the node representations more expressive, it is often desirable to increase GNNs' layers so that information can be captured from high-order neighbors. 
However, in each GNN layer, the propagation process is coupled with the non-linear transformation. 
Therefore, stacking many layers for GNNs can bring more 
 parameters to learn and thus easily cause overfitting~\cite{goodfellow2016deep,Lingxiao2020PairNorm}.

\textbf{Over-smoothing.}
In addition, recent studies show that the convolution operation is essentially a special form of Laplacian smoothing~\cite{li2018deeper}, which propagates neighbors' features into the central node. 
As a result, directly stacking many layers tend to make nodes' features over-smoothed~\cite{chen2019measuring,Lingxiao2020PairNorm}. 
In other words, each node incorporates too much information from others but loses the specific information of itself, making them indistinguishable after the propagation.


These challenges are further amplified under the semi-supervised setting, wherein the supervision information is scarce~\cite{chapelle2009semi}. 
Though several efforts have been devoted, such as DropEdge~\cite{YuDropedge}, 
these problems remain largely unexplored and, in particular, unresolved in a systemic way.   
In this work, we propose to systemically address all these fundamental issues for graph neural networks. 
We achieve this by questioning and re-designing GNNs' core procedure---the (deterministic) graph propagation. 
Specifically, we present the \full\ (\model) for semi-supervised learning on graphs. 
\model\ comprises two major components: random propagation (RP) and consistency regularization (CR). 


First, 
we introduce a simple yet effective message passing strategy---random propagation---which allows each node to randomly drop the entire features of some (multi-hop) neighbors during each training epoch. 
As such, each node is enabled to be not sensitive to specific neighborhoods, \textit{increasing the robustness of \model.} 
Second, the design of random propagation can naturally separate feature propagation and transformation, which are commonly coupled with each other in most GNNs. 
This empowers \model\ to safely perform  higher-order feature propagation without increasing the complexity, \textit{reducing the risk of over-smoothing and overfitting.} 
Finally, we demonstrate that random propagation is an economic data augmentation method on graphs, 
based on which we propose the \textit{consistency regularized training} for \model.
This strategy enforces the model to output similar predictions on different data augmentations of the same data, \textit{further improving \model's generalization capability under the semi-supervised setting.} 

To demonstrate the performance of \model, we conduct extensive experiments for semi-supervised graph learning on three GNN benchmark datasets, as well as six publicly available large datasets (Cf. Appendix \ref{exp:large_data}).  
In addition, we also provide theoretical analyses to understand the effects of the proposed {random propagation} and {consistency regularization} strategies on GNNs. 

Figure \ref{fig:grand_intro} illustrates the \model's advantages in terms of performance, robustness, and mitigation of over-smoothing and overfitting. 
(a) \textbf{Performance}: \model\ achieves a 3.9\% improvement (absolute accuracy gap) over GCN on Cora, while the margins lifted by GAT and DropEdge were only 1.5\% and 1.3\%, respectively. 
(b) \textbf{Robustness}: As more random edges injected into the data, \model\ experiences a weak accuracy-declining trend, while the decline of GCN and GAT is quite sharp. 
(c) \textbf{Over-smoothing}: As more layers stacked, the accuracies of GCN and GAT decrease dramatically---from $~$0.75 to $~$0.2---due to over-smoothing, while \model\ actually benefits from more propagation steps. 
(d) \textbf{Overfitting}: Empowered by the random propagation and consistency regularization, \model's training and validation cross-entropy losses converge close to each other, while a significant gap between the losses can be observed when RP and CR are removed. 

In summary, \model\ outperforms state-of-the-art GNNs in terms of effectiveness and robustness, and  mitigates the over-smoothing and overfitting issues that are commonly faced in existing GNNs.


}

\hide{

\section{Introduction}

Graph-structured data is ubiquitous and exists in many real applications, such as social networks, the World Wide Web, and the knowledge graphs. 
With the remarkable success of deep learning~\cite{lecun1995convolutional}, 
graph neural networks (GNNs)~\cite{bruna2013spectral,henaff2015deep, defferrard2016convolutional,kipf2016semi,duvenaud2015convolutional,niepert2016learning, monti2017geometric}  generalize neural networks to graphs and offer powerful graph representations for a wide variety of applications, such as node
classification~\cite{kipf2016semi}, link prediction~\cite{zhang2018link}, graph classification~\cite{ying2018hierarchical} and knowledge graph completion~\cite{schlichtkrull2018modeling}.

As a pioneer work of GNNs, Graph Convolutional Network (GCN)~\cite{kipf2016semi} first formulate the convolution operation on a graph as feature propagation in neighborhoods, and achieves promising performance on semi-supervised learning over graphs. 
They model a node's neighborhood as a receptive field and enable a recursive neighborhood propagation process by stacking multiple graph convolutional layers, and thus information from multi-hop neighborhoods is utilized. 
After that, much more advanced GNNs have been proposed by designing more complicated propagation patterns, e.g., the multi-head self-attention in GAT~\cite{Velickovic:17GAT}, differentiable cluster in DIFFPOOL~\cite{ying2018hierarchical} and Neighborhood Mixing~\cite{abu2019mixhop}. Despite strong successes achieved by these models, existing GNNs still suffer from some inherent issues:

First, nodes in graphs are highly tangled with each other, and existing GNNs employ a deterministic propagation strategy, where each node interact with its whole neighborhoods iteratively. This results in the \textbf{non-robustness} problem, i.e., node is highly dependent with its multi-hop neighborhoods and can be easily misguided by potential noises from other nodes. Some works also reveal that GNNs are very susceptible to adversarial  attacks~\cite{zugner2018adversarial,dai2018adversarial}. More seriously, the
attacker can even indirectly attack the target node by manipulating other long-distance nodes~\cite{zugner2018adversarial}. 

Second, to make the node representations more expressive, it is desirable to increase GNN's layers so that node can capture information from high-order neighbors. However, in each layer of GNN, the propagation process is coupled with non-linear transformation. Thus increasing  layers might bring unnecessary learnable parameters and causes \textbf{overfitting}. On the other hand, recent studies show that propagation scheme of GCN is essentially a type of Laplacian smoothing, and directly stacking many layers will lead to \textbf{over-smoothing}~\cite{li2018deeper}, i.e., node incorporates too much messages from remote nodes but losses the specific information of itself.

Under the semi-supervised setting, where supervision information is scarce, these problems become more challenging. Though several efforts to overcome these challenges, such as enhancing robustness~\cite{ZhuRobust}, mitigating over-smoothing~\cite{YuDropedge, chen2019measuring} and improving generalization~\cite{verma2019graphmix}, no research has resolved these problems systematically, to the best of our knowledges. In this paper, we propose a novel GNN framework, namely \emph{Graph Random Network (\model)}, for semi-supervised learning on graphs. \model\ addresses aforementioned problems from the following aspects:

First, \model\ utilizes \textit{random propagation} strategy. Different from deterministic propagation scheme used in other GNNs,  random propagation allows each node randomly interacts with its different subset of multi-hop neighborhoods during training, and thus mitigates the risk of being misguided by specific nodes and edges.
Second, \model\ performs propagation and transformation separately. That is, node features are first manipulated through the random propagation procedure, and then fed into a Multi-Layer Perceptron (MLP) to perform non-linear transformation and make prediction. 
This design allows us to perform much higher feature propagation without increasing the complexity of neural network, reducing the risk of overfitting and over-smoothing. Finally, we demonstrate that random propagation is an economic data augmentation method on graphs. Based on that, we propose \textit{consistency regularized training} method in \model , which enforces model to output similar predictions on different data augmentations of the same data. This strategy further improves model's generalization capability under semi-supervised setting. Furthermore, this training method can be also generalized to enhance other GNNs, e.g., GCN, GAT.

To demonstrate the effectiveness of \model , we conduct thorough experiments on benchmark datasets for semi-supervised learning on graphs. Experimental results show that the proposed methods achieve significant improvements compared with state-of-the-art models.
Extensive experiments show \model\ has better robustness and generalization, and lower risk of over-smoothing. What's more, we also provide theoretical analyses to explain the effects of proposed \textit{random propagation} strategy and \textit{consistency regularized} training methods from the view of regularization.

\hide{
 To our best knowledge, our work may be the first to propose a general graph-based data augmentation paradigm to alleviate the reliance on labeled data for GCNs training, and it's also the key to address the learning problems mentioned above on graphs.\reminder{} 
 }
 \hide{
 To summarize, we propose \model, a general framework for semi-supervised learning on graphs, which makes the following contributions:
  \begin{itemize}[leftmargin=*]
	\item \model\ explores graph-structured data in a random way and the random propagation method offers an efficient graph data augmentation methods, leading to the implicit ensemble training style. Consistency regularization further mitigates the reliance on labeled data and lowers the generalization loss.
	\item  We reveal the effects of graph dropout and consistency regularization in theory. Extensive experiments suggest that \model\ can even offer the simplest prediction model on graphs, MLP, in a simple way, a more expressive power than state-of-the-art GNNs, even including GraphNAS~\cite{gao2019graphnas}, a optimal GCN architecture designed by using neural architecture searching. 
\end{itemize}

}

\hide{
Another observation is that, different from the i.i.d. assumption in non-structured data, graph objects (e.g., nodes) are highly tangled and correlated with each other. 
Current GCNs leverage the structural dependency to improve  learning tasks, but have not realized and leveraged the potential information redundancy resulting from the structural correlations~\cite{macarthur2009spectral}. 
As an illustrative example, we found, surprisingly, that by randomly removing all the features for half of the nodes in a graph, the original GCN model~\cite{kipf2016semi} actually achieves similar or even better results than GraphSAGE, FastGCN, and GCN with the full features from all of the nodes.  
This 
implies that the missing information carried by part of the graph can actually be complemented 
during GCNs' neighborhood aggregation. 
More importantly, the random node set split offers multiple (redundant) views of the graph, which implies an equivalence to the condition of co-training, if each view is sufficient~\cite{dasgupta2002pac, goldman2000enhancing}. 
Inspired by this, we propose to minimize the disagreement of multiple GCNs on different network views in order to achieve a low generalization error, that is, the \textit{\nsgcn} model.

The final issue lies in the number of network splits in \nsgcn, which is exponential for exploring the optimal one. 
Previous research has shown that the dropout technique in convolutional neural networks is capable of implicitly ensembling exponential number of models~\cite{srivastava2014dropout, ghiasi2018dropblock}. 
Therefore, different from existing sampling-based GCNs, we propose a graph dropout-based network sampling strategy for an efficient and sufficient exploitation of information redundancy in GCNs. 
Finally, we combine the two ideas of node ranking and graph dropout together by incorporating the \rank\ model into \nsgcn's disagreement minimization framework. 

In this work, we revisit two important components in GCNs---\textit{neighborhood aggregation} and \textit{network sampling}. 
To summarize, this work makes the following contributions: 
\begin{itemize}[leftmargin=*]
    \item We present a node ranking-aware graph convolutional network (\rank), which incorporates the information diffusion process
    into the neighborhood aggregation step in GCNs.  \srank can adaptively learn the importance of each node and automatically rescale the signal.
    More importantly, \srank  can be connected with graph attention with different attention mechanisms, including node attention, (multi-hop) edge attention, and path attention, providing a new perspective to understand graph attention. 

	\item We present a network sampling-based graph convolution framework (\nsgcn). It leverages a graph dropout-based sampling technique to efficiently model exponential number of network views in an implicit manner for co-training a GCN ensemble. 
	Moreover, the proposed graph-dropout sampling method also offers a new perspective for the potential improvements of existing sampling based GCNs. 
	Straightforwardly, we also present \rank+\nsgcn\ by embedding the \rank\ model into the sampling based ensemble learning framework \nsgcn. 

	\item We conduct extensive experiments on several graph
	benchmarks datasets of different types, including Cora, Citeseer, Pubmed, and the protein-protein interaction network (PPI). 
	Experiments show that the proposed frameworks can consistently achieve performance superiority over existing state-of-art GCN and generative adversarial nets based baselines across all datasets. 
\end{itemize}

\hide{
Another more challenging issue is that, different from i.i.d. assumption in
traditional machine learning,  the information of nodes in the network is highly tangled. 
The tangled dependency may cause information redundancy.
Existing GCNs 
mainly focus on leveraging the dependency to improve learning performance, 
but lacks of deep thought about the redundancy.
The empirical evidence we have found (C.f. \shalf in Table~\ref{tab:res_2}) is that 
GCNs
with features from randomly selected half of the nodes in the network (Cora, Citeseer, and Pubmed) can 
result in a slightly better performance than of traditional methods.
This gives us a hint that the missing node information might be complemented by other nodes via propagation. 
Think more deeply, the random node set split gives us multiple  redundant views of the network data, which implies an equivalence to the condition of co-training, if each view is sufficient~\cite{dasgupta2002pac, goldman2000enhancing}.
Thus, it might be possible to achieve a lower generalization error by minimizing the disagreement of  GCNs on different data views. 
However, the number of splits in our problem is exponential for exploring the optimal one. The concept of dropout in neural networks also 
shows that 
it is possible to ensemble exponential number of models implicitly~\cite{srivastava2014dropout, ghiasi2018dropblock}. 
This inspires us to propose a dropout-like network sampling method, with a more efficient way to exploit the redundancy sufficiently for GCNs.

In this paper, we conduct a systematic investigation to this problem. The main contributions can be summarized as follows:

\begin{itemize}
	\item We begin with studying the node ranking in the network and propose a novel propagation mechanism which can adaptively learn the ranking of nodes, and automatically rescale the signal and also  choose the suitable encoder for transmitting the signal. 
	The most important thing is that, mathematically, we only add a diagonal matrix and node-wise activation functions into the vanilla GCN architectures. 
	The simple microcosmic node model can even interpret the complex propagation mechanism and cover many network attention mechanisms, including node attention, edge attention, multiple-hop edge attention and path attention. Some existing complicated  attention-based methods, e.g. GAT~\cite{Velickovic:17GAT} can also be reformulated in our framework.

	\item We then investigate 
	the redundancy of graph data and  present a solution of 
	ensemble for GCNs with different sufficient and redundant data views. 
	We design a  dropout-like sampling technique for graphs by co-training with  splitting subgraph.
This network sampling method also gives a new perspective for several existing network sampling GCNs, e.g., GraphSAGE~\cite{hamilton2017inductive} and FastGCN~\cite{chen2018fastgcn} and may help them optimize the training and improve the inference performance.

	\item We did extensive experiments on several
	benchmarks, including the Cora, Citeseer, Pubmed, as well as a protein-protein interaction dataset.
	Experimental results show that the proposed framework can significantly improve the graph classification accuracy.
	 
\end{itemize}

}

\hide{
However, there still remain a lot for exploration in graph convolutional networks  as the graph is often sophisticated and convolutional neural architecture should consider the typical characteristics of graphs, including the following observations.
 
\begin{itemize}
\item \textbf{node role in propagation mechanism. }	
	While GCNs  obey a relatively rough propagation mechanism of recursive neighborhood aggregation, there are lots of propagation model in complex network analysis, especially for social networks~\cite{kempe2003maximizing, ziegler2005propagation, cha2009measurement, yang2011like}. 
	In microcosmic view, sophisticated propagation is driven by the diversity of nodes' influence, or their roles in the network. 
	For example, information from opinion leaders is more likely to be exposed to the public and the structural holes are more likely to retweet. Besides, some nodes may be more sensitive for information from indirect nodes. Similar phenomenons holds for other networks, like 
	protein-protein interaction networks. Therefore, to depict the mechanism of information sending and receiving for nodes of different roles may help to elaborate GCNs.
	 
	\item \textbf{information dependency and redundancy. } Different from i.i.d. data requirement for traditional machine learning,  the information of nodes is highly tangled. However, the dependency also causes information redundancy, while GCNs are proposed to tackle the former but ignore the latter. An empirical evidence we found (the experiment results of \shalf in Table~\ref{tab:res_2}) is that leverage GCNs and complete network structure, even with features from random half of nodes in Cora we can achieve the node classification accuracy no worse than non-GCNs methods, which supports the redundancy assumption that the missing node information can be complemented by others via propagation. The random node set split gives us multiple sufficient and redundant views for network data, which meets the condition of co-training~\cite{dasgupta2002pac, goldman2000enhancing} and thus could make few generalization error by maximizing the agreement of  GCNs on differentdata views. However, the number of split is exponential for explore the optimal one. The experience from deep learning that dropout could ensemble exponential models implicitly~\cite{srivastava2014dropout, ghiasi2018dropblock} inspires us to propose a network dropout method, a more economic but easier approach to exploit the redundancy sufficiently, which can be integrated as a new part of GCNs.

\end{itemize}
}

\hide{

\vpara{Organization.} 
\S~\ref{sec:problem} formulates the problem. \S~\ref{sec:approach1} and \S~\ref{sec:approach2} detail the proposed framework.
~\S~\ref{sec:exp} presents experimental results to validate the redundancy of graph data and effectiveness of our framework.
\S~~\ref{sec:related} reviews the related work. Finally, \S~\ref{sec:conclusion} concludes the paper.
}

\hide{
The rest of the paper is organized as follows. Section 2 formulates the problem. Section 3 introduces the proposed framework in detail. Section 4 presents experimental results. Section 5 reviews related works and Section 6 concludes the paper. 
}
}

}
\vspace{-0.05in}
\section{Problem and Related Work}
\label{sec:problem}


Let $G=(V, E)$ denote a graph, where $V$ is a set of $|V|=n$ nodes and $E \subseteq V \times V$ is a set of $|E|$ edges between nodes. 
$\mathbf{A}\in \{0,1\}^{n \times n}$ denotes the adjacency matrix of $G$, with each element $\mathbf{A}_{ij} = 1$ indicating there exists an edge between $v_i$ and $v_j$, otherwise $\mathbf{A}_{ij} = 0$.

\vpara{Semi-Supervised Learning on Graphs.}This work focuses on semi-supervised graph learning, in which each node $v_i$ is associated with 
1) a feature vector $\mathbf{X}_i \in \mathbf{X} \in \mathbb{R}^{n\times d}$ 
and 
2) a label vector $\mathbf{Y}_i \in \mathbf{Y} \in \{0,1\}^{n \times C}$ with $C$ representing the number of classes. 
For semi-supervised classification, $m$ nodes ($ 0 < m \ll n$) 
have observed their labels $\mathbf{Y}^L$ and the labels $\mathbf{Y}^U$ of the remaining $n-m$ nodes are missing. 
The objective is to learn a predictive function 
$f: G, \mathbf{X}, \mathbf{Y}^L \rightarrow \mathbf{Y}^U$ 
to infer the missing labels $\mathbf{Y}^U$ for unlabeled nodes. Traditional approaches to this problem are mostly based on graph Laplacian regularizations~\cite{zhu2003semi,zhou2004learning,mei2008general,weston2012deep,belkin2006manifold}. 
Recently, graph neural networks (GNNs) have emerged as a powerful approach for semi-supervised graph learning,  
which are reviewed below.




\vpara{Graph Neural Networks.}
GNNs~\cite{gori2005new, scarselli2009graph, kipf2016semi} generalize neural techniques into graph-structured data.
The core operation in GNNs is graph propagation, in which information is propagated from each node to its neighborhoods with some deterministic propagation rules. 
For example, the graph convolutional network (GCN) \cite{kipf2016semi} adopts the  propagation rule 
$\mathbf{H}^{(l+1)} = \sigma (\mathbf{\hat{A}}\mathbf{H}^{(l)}\mathbf{W}^{(l)})$, 
\hide{
\begin{equation}
\small
\label{equ:gcn_layer}
	\mathbf{H}^{(l+1)} = \sigma (\mathbf{\hat{A}}\mathbf{H}^{(l)}\mathbf{W}^{(l)}),
\end{equation}
}
where $\mathbf{\hat{A}}$ is the symmetric normalized adjacency matrix, $\sigma(.)$ denotes the ReLU function, and $\mathbf{W}^{(l)}$ and $\mathbf{H}^{(l)}$ are the weight matrix and the hidden node representation in the $l^{th}$ layer 
with $\mathbf{H}^{(0)}=\mathbf{X}$. 

The GCN propagation rule 
could be explained via 
the 
approximation of the spectral graph convolutions~\cite{bruna2013spectral,henaff2015deep, defferrard2016convolutional}, 
neural message passing~\cite{gilmer2017neural}, 
and 
convolutions on direct neighborhoods~\cite{monti2017geometric, hamilton2017inductive}. 
Recent attempts to advance this architecture include GAT~\cite{Velickovic:17GAT}, GMNN~\cite{qu2019gmnn}, MixHop~\cite{abu2019mixhop},  and GraphNAS~\cite{gao2019graphnas}, etc. 
\hide{Often, these models face the challenges of non-robustness and over-smoothing due to the deterministic graph propagation process~\cite{li2018deeper,chen2019measuring,Lingxiao2020PairNorm}. 
Differently, we propose  random propagation 
for GNNs, which decouples  feature propagation and non-linear transformation in Eq. \ref{equ:gcn_layer}, reducing the risk of over-smoothing. }
In addition, sampling based techniques have also been developed for 
fast and scalable GNN training, such as GraphSAGE~\cite{hamilton2017inductive}, FastGCN~\cite{FastGCN}, AS-GCN~\cite{huang2018adaptive}, and LADIES~\cite{ladies}. The sampling based propagation used in those models may also be used as a graph augmentation method. However, its potential effects under semi-supervised setting have not been well-studied, which we try to explore in future work.

\hide{Different from these work, in this paper, a new sampling strategy DropNode, is proposed for improving the robustness and generalization of GNNs for semi-supervised learning.

Compared with GraphSAGE's node-wise sampling, DropNode 1) enables the decoupling of feature propagation and transformation, and 2) is more efficient as it does not require recursive sampling of neighborhoods for every node. Finally, it drops each node based an i.i.d. Bernoulli distribution, differing from the importance sampling in FastGCN, AS-GCN, and LADIES. }


\vpara{Regularization Methods for GNNs.}
Another line of work has aimed to design powerful regularization methods for GNNs, such as 
VBAT~\cite{deng2019batch}, 
GraphVAT~\cite{feng2019graph},
$\text{G}^3$NN~\cite{ma2019flexible}, 
GraphMix~\cite{verma2019graphmix}, 
and 
DropEdge~\cite{YuDropedge}. 
For example,
VBAT~\cite{deng2019batch} and GraphVAT~\cite{feng2019graph} first apply consistency regularized training into GNNs via 
virtual adversarial training~\cite{miyato2018virtual}, which is highly time-consuming in practice.
GraphMix~\cite{verma2019graphmix} introduces the MixUp strategy~\cite{zhang2017mixup} for training GNNs. Different from \model, GraphMix augments graph data by performing linear interpolation between two samples in the hidden space, and regularizes GNNs by encouraging the model to predict the same interpolation of corresponding labels. 

\hide{Broadly, a popular regularization method in deep learning is data augmentation, which expands the training samples by applying  transformations or injecting noises  \cite{devries2017dataset,goodfellow2016deep, vincent2008extracting}. 
Based on data augmentation, we can further leverage consistency regularization~\cite{berthelot2019mixmatch,laine2016temporal} for semi-supervised learning, which enforces the model to output the same distribution on different augmentations of the input data.
Following this idea, a line of work has aimed to design powerful regularization methods for GNNs, such as 
VBAT~\cite{deng2019batch},
$\text{G}^3$NN~\cite{ma2019flexible}, 
GraphMix~\cite{verma2019graphmix}, 
and 
DropEdge~\cite{YuDropedge}. 
For example, GraphMix~\cite{verma2019graphmix} introduces the MixUp strategy~\cite{zhang2017mixup} for training GCNs. 
GraphMix augments graph data by performing linear interpolation between two samples in the hidden space, and regularizes GCNs by encouraging the model to predict the same interpolation of corresponding labels. 
DropEdge~\cite{YuDropedge} is designed for militating the over-smoothing problem faced by GNNs. However, it can not improve model's generalization capacity with fully leveraging unlabeled data and does not bring significant performance gains for semi-supervised learning task.
However, \yd{to add one sentence what dropedge can't do but grand can do for semi-s learning wihtout mentioning grand/dropnode}}






\hide{

\section{Definition and Preliminaries}
\label{sec:problem}

In this section, we first present the problem formalization of graph-based semi-supervised learning. Then we briefly introduce the concept of graph convolutional networks and consistency regularization. 

Let $G=(V, E)$ denote a graph, where $V$ is a set of $|V|=n$ nodes, $E \subseteq V \times V$ is a set of $|E|=k$ edges between nodes. Let $A\in \{0,1\}^{n \times n}$ denote the adjacency matrix of the graph $G$, with an element $A_{ij}>0$ indicating node $v_i \in V$ has a link to $v_j \in V$. We further assume each node $v_i$ is associated with a $d-$dimensional feature vector $x_i$ and a class label $y_i \in \mathcal{Y} = \{1,...,C\}$. 


In this work, we focus on semi-supervised learning on graph. In this setting, only a part of nodes have observed their labels and the labels of other samples are missing. Without loss of generality
, we use $V^L=\{v_1, v_2, ..., v_{m}\}$ to represent the set of labeled nodes, and $V^U=\{v_{m+1}, v_{m+2}, ..., v_{n}\}$ refers to unlabeled nodes set, where $ 0 < m \ll n$. Denoting the feature matrix of all nodes as $X \in \mathbf{R}^{n\times d}$, the labels of labeled nodes as $Y^L$, 
the goal of semi-supervised learning on graph is to infer the missing labels $Y^U$ based on $(X, Y^L, G)$. 



\subsection{Graph Propagation}
Almost all the graph models, whether based on probabilistic graph theory or neural networks, have a very important phase, i.e., graph propagation, which is usually performed by some deterministic rules derived from the graph structures. Here we take the propagation mechanism in GCN \cite{kipf2016semi}, a popular graph neural model, as an example to introduce it.
GCN performs information propagation by the following rule:
\begin{equation}
\label{equ:gcn_layer}
	H^{(l+1)} = \sigma (\hat{A}H^{(l)}W^{(l)}),
\end{equation}
\noindent where $\hat{A}=\widetilde{D}^{-\frac{1}{2}}\widetilde{A}~\widetilde{D}^{-\frac{1}{2}}$ is the symmetric normalized adjacency matrix, $\widetilde{A} = A + I_{n}$ is the adjacency matrix with augmented self-connections, $I_{n}$ is the identity matrix, $\widetilde{D}$ is the diagonal degree matrix with $\widetilde{D}_{ii} = \sum_j \widetilde{A}_{ij}$, and $W^{(l)}$ is a layer-specific learnable weight matrix. Function $\sigma_l(.)$ denotes a nonlinear activation function, e.g., $\mathrm{ReLu}(.)=\max(0,.)$. $H^{(l)}\in R^{n \times D^{(l)}}$ is the matrix of activations, or the $D^{(l)}$-dimensional hidden node representation in the $l^{th}$ layer 
with $H^{(0)}=X$. 

The form of propagation rule in Equation \ref{equ:gcn_layer} could be motivated via a first-order approximation of the spectral graph convolutional operations~\cite{bruna2013spectral,henaff2015deep, defferrard2016convolutional}, and it also obeys neural message passing~\cite{gilmer2017neural}, similar to convolutions on direct neighborhoods~\cite{monti2017geometric, hamilton2017inductive}. Hence, in the rest of the paper, we will not distinguish between spectral or non-spectral convolutional neural networks, and between graph convolutions and neural message passing, strictly, for the  convenience of narration.

\subsection{Regularization  in Semi-supervised Learning.} 
\subsubsection{Data Augmentation}
A widely used regularization method in deep learning is data augmentation, which expands the training samples by applying some transformations on input data \cite{devries2017dataset}. In practical, the transformations used here can be some domain-specific methods, such as rotation, shearing on images. What's more, a general kind of data augmentation method is injecting noise to the input data~\cite{goodfellow2016deep}, this idea has been adopted in a variety of deep learning methods, e.g., denoising auto-encoders~\cite{vincent2008extracting}. In this way, the model is encouraged to be invariant to transformations and noises, leading to better generalization and robustness. 
\subsubsection{Consistency Regularization}

Consistency regularization is based on data augmentation and applies it into semi-supervised learning. 
Roughly speaking, it enforces the model to output the same class distribution for the same unlabeled example under different augmentations \cite{berthelot2019mixmatch}. Let $f_{aug}(x)$ denote a stochastic augmentation function for unlabeled data $x$, e.g., randomly transformation or adding noise. One alternative method to perform consistency regularization is adding a regularization loss term \cite{laine2016temporal}:
\begin{equation}
\label{eq_consis}
    \|P(y|\widetilde{x}^{(1)}, \theta) - P(y|\widetilde{x}^{(2)}, \theta) \|^2,
\end{equation}
where $\theta$ refers to model parameters, $\widetilde{x}^{(1)}$ and $\widetilde{x}^{(2)}$ are two random augmentations of $x$.

}

\hide{
\begin{figure*}
    \centering
    \includegraphics[width= \linewidth]{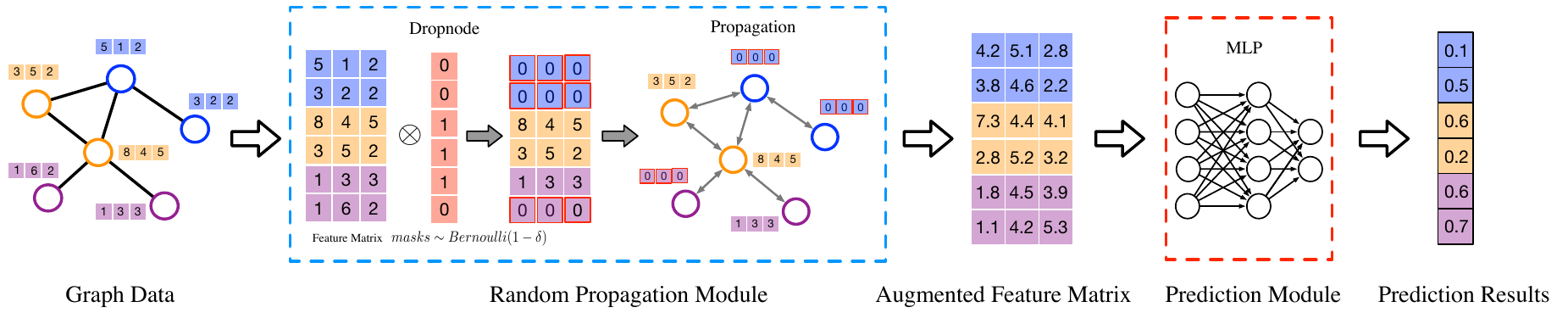}
    \caption{Basic \model\ Operations. \yd{to remove this figure.} }
    \label{fig:arch1}
\end{figure*}
}

\label{sec:consis}  
\begin{figure*}[t]
	\centering
	\includegraphics[width= 1.0 \linewidth]{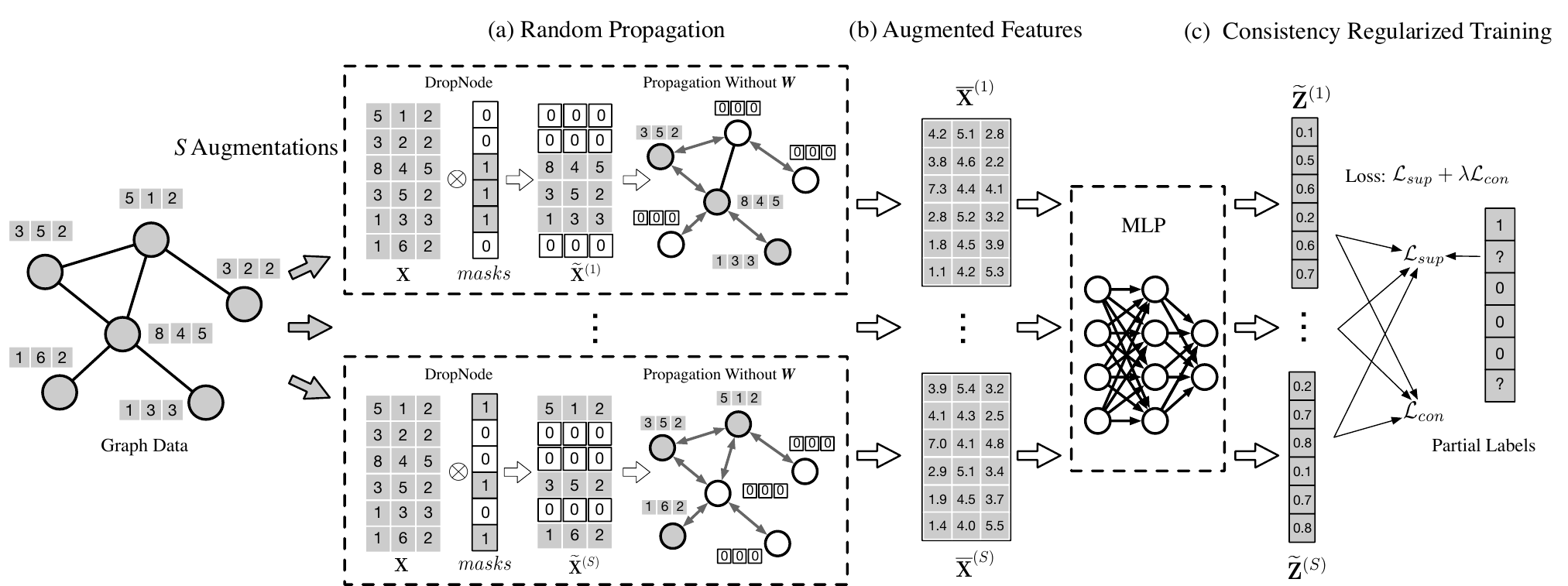}
	\caption{Illustration of \model\ with DropNode as the perturbation method. 
	\textmd{\model\ designs random propagation (a) to generate multiple graph data augmentations (b), which are further used as consistency regularization (c) for semi-supervised learning. 
	\hide{
	\model\ consists of two mechanisms---random propagation and consistency regularized training. In random propagation, some nodes' feature vectors are randomly dropped with DropNode. 
	The resultant perturbed feature matrix $\widetilde{\mathbf{X}}$ is then used to perform propagation without parameters to learn. 
	Further, random propagation is used for stochastic graph data augmentation. 
	After that, the augmented feature matrices are fed into a two-layer MLP model for prediction. With applying consistency regularized training, \model\ generates $S$ data augmentations by performing random propagation $S$ times, and leverages both supervised classification loss $\mathcal{L}_{sup}$ and consistency regularization loss $\mathcal{L}_{con}$ in optimization.
	}
	}
	}
	\label{fig:arch2}
\end{figure*}

\section{\full}

We present the \full\ (\model) for semi-supervised learning on graphs, as illustrated in Figure  \ref{fig:arch2}. 
The idea is to design a propagation strategy (a) to stochastically generate multiple graph data augmentations (b),  
based on which we present a consistency regularized training (c) for improving the generalization capacity under the semi-supervised setting. 


\subsection{Random Propagation for Graph Data Augmentation}
\label{sec:randpro}

Given an input graph $G$ with its adjacency matrix $\mathbf{A}$ and feature matrix $\mathbf{X}$, the random propagation module generates multiple data augmentations. 
For each augmentation $\overline{\mathbf{X}}$, it is then fed into the classification model, a two-layer MLP, for predicting node labels. 
The MLP model can also be replaced with more complex and advanced GNN models, such as GCN and GAT.



\vpara{Random Propagation.} 
There are two steps in random propagation. 
First, we generate a perturbed feature matrix $\widetilde{\mathbf{X}}$ by randomly dropping out elements in $\mathbf{X}$.  
Second, we leverage $\widetilde{\mathbf{X}}$ to perform feature propagation for generating the augmented features $\overline{\mathbf{X}}$. 

In doing so, each node's features are randomly mixed with signals from its neighbors. Note that the homophily assumption suggests that adjacent nodes tend to have similar features and labels~\cite{mcpherson2001birds}. 
Thus, the dropped information of a node could be compensated by its neighbors, forming an approximate representation for it in the corresponding augmentation. In other words, random propagation allows us to stochastically generate multiple augmented representations for each node.




In the first step, there are different ways to perturb the input data $\mathbf{X}$. 
Straightforwardly, we can use the dropout strategy ~\cite{srivastava2014dropout}, which has been widely used for regularizing neural networks. 
Specifically, dropout perturbs the feature matrix by randomly setting some elements of $\mathbf{X}$ to 0 during training, i.e., $\widetilde{\mathbf{X}}_{ij} = \frac{\epsilon_{ij}}{1-\delta} \mathbf{X}_{ij}$, where $\epsilon_{ij}$ draws from $Bernoulli(1-\delta)$. 
In doing so, dropout makes the input feature matrix $\mathbf{X}$ noisy by randomly dropping out its elements without considering graph structures.

To account for the structural effect, we can simply remove some nodes' entire feature vectors---referred to as DropNode, instead of dropping out single feature elements. 
In other words, DropNode enables each node to only aggregate information from a subset of its (multi-hop) neighbors by completely ignoring some nodes' features, which reduces its dependency on particular neighbors and thus helps increase the model's robustness (Cf. Section~\ref{sec:robust}). 
Empirically, it generates more stochastic data augmentations and achieves better performance than dropout (Cf. Section~\ref{sec:overall}). 

Formally, in DropNode, we first 
randomly sample a binary mask $\epsilon_i \sim Bernoulli(1-\delta)$ for each node $v_i$.
Second, we obtain the perturbed feature matrix $\widetilde{\mathbf{X}}$ by multiplying each node's feature vector with its corresponding mask, i.e., $\widetilde{\mathbf{X}}_i=\epsilon_i \cdot \mathbf{X}_i$ where $\mathbf{X}_i$ denotes the $i^{th}$ row vector of $\mathbf{X}$. 
Finally,  we scale $\widetilde{\mathbf{X}}$ with the factor of $\frac{1}{1-\delta}$ to guarantee the perturbed feature matrix is in expectation equal to $\mathbf{X}$. 
Note that the sampling procedure is only performed during training. 
During inference, we directly set $\widetilde{\mathbf{X}}$ as the original feature matrix $\mathbf{X}$.

In the second step of random propagation, we adopt the mixed-order propagation, i.e., $\overline{\mathbf{X}} = \overline{\mathbf{A}} \widetilde{\mathbf{X}}$, where  $\overline{\mathbf{A}} =  \sum_{k=0}^K\frac{1}{K+1}\hat{\mathbf{A}}^k$ is the average of the power series of $\hat{\mathbf{A}}$ from order 0 to order $K$. 
This propagation rule enables the model to incorporate more local information, reducing the risk of over-smoothing when compared with directly using $\hat{\mathbf{A}}^K$~\cite{abu2019mixhop,xu2018representation}. Note that calculating the dense matrix $\overline{\mathbf{A}}$ is computationally inefficient, thus we compute $\overline{\mathbf{X}}$ by iteratively calculating and summing up the product of  sparse matrix $\hat{\mathbf{A}}$ and $\hat{\mathbf{A}}^{k}\widetilde{\mathbf{X}}$ ($0\leq k \leq K-1$) in implementation.

With this propagation rule, we could observe that DropNode (dropping the $i^{th}$ row of $\mathbf{X}$) is equivalent to dropping the $i^{th}$ column of $\overline{\mathbf{A}}$. This is similar to DropEdge~\cite{YuDropedge}, which aims to address over-smoothing by randomly removing some edges. In practice, DropEdge could also be adopted as the perturbation method here. Specifically, we first generate a corrupted adjacency matrix $\tilde{\mathbf{A}}$ by dropping some elements from $\hat{\mathbf{A}}$, and then use $\tilde{\mathbf{A}}$ to perform mix-order propagation as the substitute of $\hat{\mathbf{A}}$  at each epoch. We empirically compare the effects of different perturbation methods in Section~\ref{sec:overall}. By default, we use DropNode as the perturbation method. 

\hide{
Differently, DropEdge is originally designed to be directly manipulating the adjacency matrix $\mathbf{A}$, and extending it to drop the elements of $\overline{\mathbf{A}}$ is time-consuming since it needs to calculate the exact form of the dense matrix $\overline{\mathbf{A}}$.
}

\hide{
In random propagation,  each node's features are randomly mixed with signals from its neighbors. 
Note that the local smoothness assumption suggests that adjacent nodes often tend to have similar features and labels~\cite{dakovic2019local}. 
Therefore, the information of a dropped node could be compensated by its neighbors, forming an approximate representation for it in the corresponding augmentation. 
In other words, random propagation allows us to stochastically generate multiple augmented representations for each node. 
}


\hide{

\vpara{Dropping Function.} As for the perturbation function, a naive choice is dropout~\cite{srivastava2014dropout}. 

In that, a proportion of feature elements of each nodes are randomly set to $0$ during training. Formally, dropout perturbs the feature matrix by randomly setting some elements of $\mathbf{X}$ to 0, i.e., $\widetilde{\mathbf{X}}_{ij} = \frac{\epsilon_{ij}}{1-\delta} \mathbf{X}_{ij}$, where $\epsilon_{ij}$ draws from $Bernoulli(1-\delta)$. 
It has been widely-used as a general regularization method for neural network. However, each node still interacts with their deterministic neighbors despite dropout, though with random partial feature information. Inspired by DropBlock applied in CNNs, here we provide a more structured form of perturbation method for graphs---DropNode.

Different with dropout, DropNode is designed to randomly remove some nodes' entire feature vector instead of dropping single feature element. By doing so, DropNode enables each node only aggregates information from a subset of its (multi-hop) neighborhoods in random propagation, which will perform better in reducing the dependency between nodes to increase model's robustness. Meanwhile, it can also generate more stochastic data augmentations and achieves better performance than dropout (Cf. Section~\ref{sec:overall}). 
The formal DropNode operation is shown in Algorithm \ref{alg:dropnode}. 
First, we randomly sample a binary mask $\epsilon_i \sim Bernoulli(1-\delta)$ for each node $v_i$.
Second, we obtain the perturbed feature matrix $\widetilde{\mathbf{X}}$ by multiplying each node's feature vector with its corresponding mask, i.e., $\widetilde{\mathbf{X}}_i=\epsilon_i \cdot \mathbf{X}_i$\footnote{$\mathbf{M}_i$ denotes the $i^{th}$ row vector of $\mathbf{M}$.} . 
Finally,  we scale $\widetilde{\mathbf{X}}$ with the factor of $\frac{1}{1-\delta}$ to guarantee the perturbed feature matrix is in expectation equal to $\mathbf{X}$. 
Note that the sampling procedure is only performed during training. 
During inference, we directly set $\widetilde{\mathbf{X}}$ with the original feature matrix $\mathbf{X}$. 


\begin{algorithm}[h]
\small
\caption{DropNode}
\label{alg:dropnode}
\begin{algorithmic}[1]
\REQUIRE ~~\\
Feature matrix $\mathbf{X} \in \mathbb{R}^{n \times d}$, DropNode probability 
$\delta \in (0,1)$. \\
\ENSURE ~~\\
Perturbed feature matrix  $\mathbf{\widetilde{X}}\in \mathbb{R}^{n \times d}$.
\STATE Randomly sample $n$ masks: $\{\epsilon_i \sim Bernoulli(1-\delta)\}_{i=0}^{n-1}$.
\STATE Obtain deformity feature matrix by  multiplying each node's feature vector with the corresponding  mask: $\widetilde{\mathbf{X}}_{i} = \epsilon_i \cdot \mathbf{X}_{i} $.
\STATE Scale the deformity features: $\widetilde{\mathbf{X}} = \frac{\widetilde{\mathbf{X}}}{1-\delta}$.
\end{algorithmic}
\end{algorithm}

}



\vpara{Prediction.}
After performing random propagation for $S$ times, we generate $S$ augmented feature matrices $\{\overline{\mathbf{X}}^{(s)}|1\leq s \leq S\}$. 
Each of these augmented data is fed into a two-layer MLP to get the corresponding outputs:
$$
 \small
     \widetilde{\mathbf{Z}}^{(s)} = f_{mlp}(\overline{\mathbf{X}}^{(s)}, \Theta), 
$$
where $\widetilde{\mathbf{Z}}^{(s)} \in [0,1]^{n\times C}$ denotes the prediction probabilities on $\overline{\mathbf{X}}^{(s)}$ and $\Theta$ are the model parameters.

Observing the data flow from random propagation to the prediction module, it can be realized that \model\ actually separates the feature propagation step, i.e., $\overline{\mathbf{X}} = \overline{\mathbf{A}} \widetilde{\mathbf{X}}$, and transformation step, i.e., $f_{mlp}(\overline{\mathbf{X}} \mathbf{W}, \Theta)$. 
Note that these two steps are commonly coupled with each other in standard GNNs, that is, $\sigma(\mathbf{AX} \mathbf{W})$. 
This separation allows us to perform the high-order feature propagation 
without conducting 
non-linear transformations, reducing the risk of over-smoothing (Cf. Section~\ref{sec:oversmoothing}). 
A similar idea has been adopted by Klicpera et al.~\cite{klicpera2018predict}, with the difference that they first perform the prediction for each node and then propagate the prediction probabilities over the graph. 


\hide{
each of which $\overline{\mathbf{X}}$ can be then fed into any classification models for predicting node labels. 
In \model, we simply employ a two-layer MLP as the classifier, 
\begin{equation}
\small
\label{equ:mlp}
    p(\mathbf{Y}|\overline{\mathbf{X}};\Theta) = f_{mlp}(\overline{\mathbf{X}}, \Theta),
\end{equation}
where $\Theta$ denotes the model's parameters. 
The MLP classification model can be also replaced with more complex and advanced GNN models, such as GCN and GAT. 
The experimental results show that the replacements result in consistent performance drop across different datasets due to GNNs' over-smoothing problem (Cf. Appendix~~\ref{sec:oversmoothing_grand} for details). 
}

%

\subsection{Consistency Regularized Training}
\label{sec:consis}

In graph based semi-supervised learning, the objective is usually to smooth the label information over the graph with regularizations~\cite{zhu2003semi,weston2012deep,kipf2016semi}, i.e., its loss function is a combination of the supervised loss on the labeled nodes and the graph regularization loss. 
Given the $S$ data augmentations generated in random propagation, 
we can naturally design a consistency regularized loss for \model's semi-supervised learning.  

\hide{
Employing random propagation as stochastic graph data augmentation, it's natural to design a consistency regularized training algorithm for \model. 
In specific, we perform the random propagation operation for $S$ times to generate $S$ augmented feature matrices $\{\overline{\mathbf{X}}^{(s)}|1\leq s \leq S\}$. 
Each of these augmented feature matrices is fed into the MLP prediction module to get the corresponding outputs:
$
 \small
     \widetilde{\mathbf{Z}}^{(s)} = f_{mlp}(\overline{\mathbf{X}}^{(s)}, \Theta), 
$
where $\widetilde{\mathbf{Z}}^{(s)} \in [0,1]^{n\times C}$ denotes the prediction probabilities on $\overline{\mathbf{X}}^{(s)}$. 


Feature propagation in existing GNNs and Weisfeiler-Lehman Isomorphism test~\cite{shervashidze2011weisfeiler} has been proven to be an effective method for enhancing node representation by aggregating information from neighborhoods. 
We discuss the additional implications that \model's random propagation brings into feature propagation. 
Random propagation randomly drops some nodes' entire features before propagation. 
As a result, each node only aggregates information from a random subset of its (multi-hop) neighborhood. 
In doing so, we are able to stochastically generate different representations for each node, which can be considered as a stochastic graph augmentation method. 
In addition, random propagation can be seen as injecting random noise 
into the propagation procedure.

To empirically examine this data augmentation idea, we generate a set of augmented node representations $\overline{\mathbf{X}}$ with different drop rates in random propagation and use each $\overline{\mathbf{X}}$ to train a GCN for node classification on commonly used datasets---Cora, Citeseer, and Pubmed.  The results show that the decrease in GCN's classification accuracy is less than $3\%$ even when the drop rate is set to $0.5$. In other words, with half of rows in the input $\mathbf{X}$ removed (set to $\vec{0}$), random propagation is capable of generating augmented node representations that are sufficient for prediction.

Though one single $\overline{\mathbf{X}}$ is relatively inferior to the original $\mathbf{X}$ in performance, in practice, multiple augmentations---each per epoch---are utilized for training the \model\ model. 
Similar to bagging~\cite{breiman1996bagging}, \model's random data augmentation scheme makes the final prediction model implicitly assemble models on exponentially many augmentations, yielding much better performance than the deterministic propagation used in GCN and GAT.}


\vpara{Supervised Loss.}
With $m$ labeled nodes among $n$ nodes, the supervised objective of the graph node classification task in each epoch is defined as the average cross-entropy loss over $S$ augmentations:
\begin{equation}
\small
\label{equ:loss}
	\mathcal{L}_{sup} = -\frac{1}{S}\sum_{s=1}^{S}\sum_{i=0}^{m-1}\mathbf{Y}_{i}^\top \log \widetilde{\mathbf{Z}}_{i}^{(s)} .
\end{equation}

 

\vpara{Consistency Regularization Loss.} 
In the semi-supervised setting, we propose to optimize the prediction consistency among $S$ augmentations for unlabeled data. 
Considering a simple case of $S=2$, we can minimize the squared $L_2$ distance between the two outputs, i.e.,
 $
\min \sum_{i=0}^{n-1} \|\widetilde{\mathbf{Z}}^{(1)}_i - \widetilde{\mathbf{Z}}^{(2)}_i\|_2^2$.
To extend this idea into the multiple-augmentation situation, 
we first 
calculate the label distribution center by taking the average of all distributions, i.e., 
 $ \overline{\mathbf{Z}}_i = \frac{1}{S}\sum_{s=1}^{S} \widetilde{\mathbf{Z}}_i^{(s)}$. Then we utilize the \textit{sharpening}~\cite{berthelot2019mixmatch} trick to ``guess'' the labels based on the average distributions. 
 Specifically, the $i^{th}$ node's guessed probability on the $j^{th}$ class is calculated by:
\hide{
\begin{equation}
    \min \sum_{s=1}^{S}\sum_{i=1}^n \mathcal{D}(\overline{\mathbf{Z}}_i, \widetilde{\mathbf{Z}}^{(s)}_i).
\end{equation}
}
\begin{equation}
\small
\label{equ:sharpen}
\overline{\mathbf{Z}}^{'}_{ij} = \overline{\mathbf{Z}}_{ij}^{\frac{1}{T}} ~\bigg/~\sum_{c=0}^{C-1}\overline{\mathbf{Z}}_{ic}^{\frac{1}{T}}, (0\leq j \leq C-1),
\end{equation}
where $0< T\leq 1$ acts as the ``temperature'' that controls the sharpness of the categorical distribution. 
As $T \to 0$, the sharpened label distribution will approach a one-hot distribution. 
We minimize the distance between  $\widetilde{\mathbf{Z}}_i$ and $\overline{\mathbf{Z}}^{'}_i$ in \model:
\begin{equation}
\small
\label{equ:consistency}
    \mathcal{L}_{con} =   \frac{1}{S}\sum_{s=1}^{S}\sum_{i=0}^{n-1} \|\overline{\mathbf{Z}}^{'}_i- \widetilde{\mathbf{Z}}^{(s)}_i\|_2^2.
\end{equation}

Therefore, by setting $T$ as a small value, we can enforce the model to output low-entropy predictions. 
This can be viewed as adding an extra entropy minimization regularization into the model, which assumes that the classifier's decision boundary should not pass through high-density regions of the marginal data distribution~\cite{grandvalet2005semi}. 


\vpara{Training and Inference.}
In each epoch, we employ both the supervised classification loss in Eq. \ref{equ:loss} and the consistency regularization loss in Eq. \ref{equ:consistency} on $S$ augmentations. 
The final loss of \model\ is:
\begin{equation}
\small
\label{equ:inf}
	\mathcal{L} = \mathcal{L}_{sup} + \lambda \mathcal{L}_{con},
\end{equation}
where $\lambda$ is a hyper-parameter that controls the balance between the two losses. 
Algorithm~\ref{alg:2} outlines \model's training process. 
During inference, as mentioned in Section \ref{sec:randpro}, we directly use the original feature $\mathbf{X}$  for propagation. 
This is justified because we scale  the perturbed feature matrix $\widetilde{\mathbf{X}}$ during training to guarantee its expectation to match $\mathbf{X}$. Hence the inference formula is $\mathbf{Z} = f_{mlp}(\overline{\mathbf{A}}\mathbf{X}, \Theta)$.

\vpara{Complexity.}
 The complexity of random propagation is  $\mathcal{O}(Kd(n+|E|))$, where $K$ denotes propagation step, $d$ is the dimension of node feature, $n$ is the number of nodes and $|E|$ denotes edge count. The complexity of its prediction module (two-layer MLP) is $\mathcal{O}(nd_h(d+ C))$, where $d_h$ denotes its hidden size and $C$ is the number of classes. 
By applying consistency regularized training, the total computational complexity of \model\ is $\mathcal{O}(S(Kd(n + |E|)+ nd_h(d + C)))$, which is linear with the sum of node and edge counts. 

\vpara{Limitations.}
\model\ is based on the homophily assumption~\cite{mcpherson2001birds}, i.e., ``birds of a feather flock together'', a basic assumption in the literature of graph-based semi-supervised learning~\cite{zhu2003semi}. Due to that, however, \model\ may not succeed on graphs with less homophily. 

 \begin{algorithm}[t]
\caption{\model}
\small
\label{alg:2}
\begin{algorithmic}[1] 
\REQUIRE ~~\\
 Adjacency matrix $\hat{\mathbf{A}}$,
feature matrix $\mathbf{X} \in \mathbb{R}^{n \times d}$, 
times of augmentations in each epoch $S$, DropNode/dropout probability $\delta$, learning rate $\eta$, an MLP model: $f_{mlp}(\mathbf{X}, \Theta)$.\\
\ENSURE ~~\\
Prediction $\mathbf{Z}$. 
\WHILE{not convergence}
\FOR{$s=1:S$} 
\STATE Pertube the input:  
$\widetilde{\mathbf{X}}^{(s)} \sim \text{DropNode}(\mathbf{X},\delta)$. 
\STATE Perform propagation: $\overline{\mathbf{X}}^{(s)} = \frac{1}{K+1}\sum_{k=0}^K\hat{\mathbf{A}}^k \widetilde{\mathbf{X}}^{(s)}$.
\STATE Predict class distribution using MLP: 
$\widetilde{\mathbf{Z}}^{(s)} = f_{mlp}(\mathbf{\overline{X}}^{(s)}, \Theta)$
\ENDFOR
\STATE Compute supervised classification loss $\mathcal{L}_{sup}$ via Eq. 1 and consistency regularization loss via Eq. 3.
\STATE Update the parameters $\Theta$ by gradients descending:
$\Theta = \Theta - \eta \nabla_\Theta (\mathcal{L}_{sup} + \lambda \mathcal{L}_{con})$
\ENDWHILE
\STATE Output prediction $\mathbf{Z}$ via: $\mathbf{Z}= f_{mlp}(\frac{1}{K+1}\sum_{k=0}^K\hat{\mathbf{A}}^k \mathbf{X}, \Theta)$.

\end{algorithmic}
\end{algorithm}




\hide{

\section{Graph Random Networks}
\hide{
\begin{figure*}
  		\centering
  		\includegraphics[width=0.8 \linewidth,height =  0.25\linewidth]{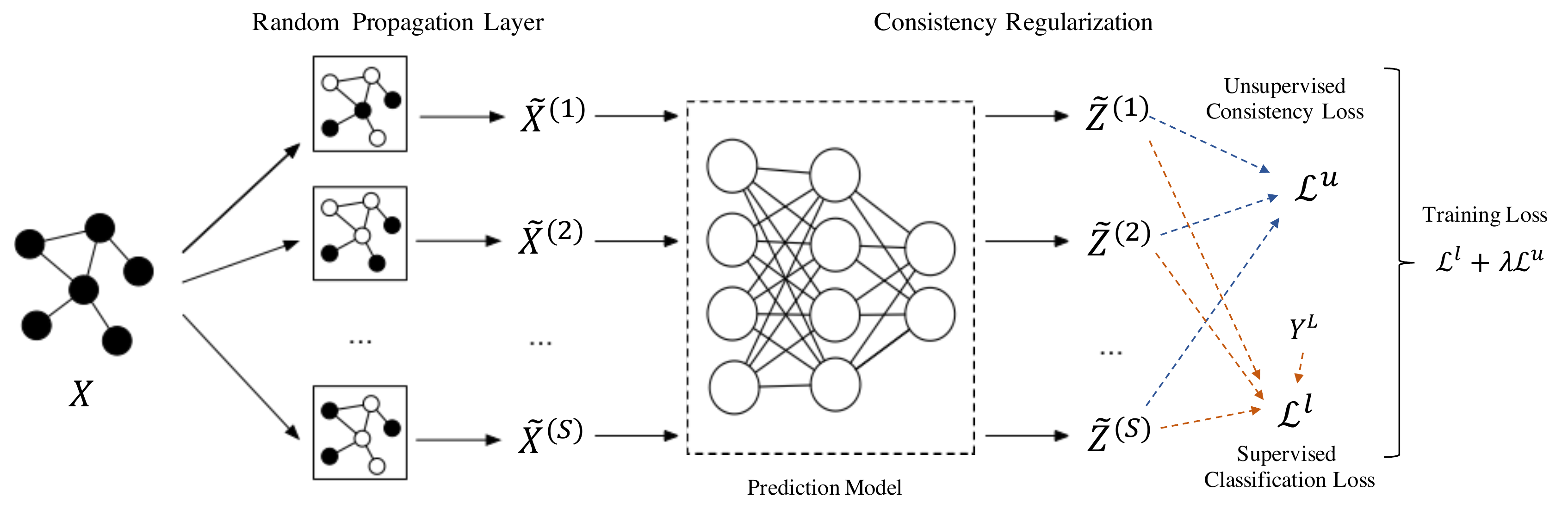}
 	\caption{Diagram of the training process of \model.  
 	As a general semi-supervised learning framework on graphs, \model\ provides two mechanisms---random propagation and consistency regularization----to enhance the prediction model's robustness and generalization. In each epoch, given the input graph, \model\ first generates $S$ graph data augmentations $\{\widetilde{X}^{(s)}| 1\leq s \leq S\}$ via random propagation layer. Then, each $\widetilde{X}^{(s)}$ is fed into a prediction model, leading to the corresponding prediction distribution $\widetilde{Z}^{(s)}$. In optimization, except for optimizing the supervised classification loss $\mathcal{L}^l$ with the given labels $Y^L$, we also minimize the prediction distance among different augmentations of unlabeled nodes via unsupervised consistency loss $\mathcal{L}^u$. }
  	\label{fig:model}
\end{figure*}
}
To achieve a better model for semi-supervised learning on graphs, we propose Graph Random Networks (\model). In \model , different with other GNNs, each node is allowed to randomly interact with different subsets of neighborhoods in different training epochs. This random propagation mechanism is shown to be an economic way for stochastic graph data augmentation. Based on that, we also design a consistency regularized training method to improve model's generalization capacity by encouraging predictions invariant to different augmentations of the same node. 

\hide{
Graph is complex with highly tangled nodes and edges, while previous graph models, e.g., GCN and GAT, take the node neighborhood as a whole and follow a determined aggregation process recursively. These determined models, where nodes and edges interact with each other in a fixed way, suffer from overfitting and risk of being misguiding by small amount of potential noise, mistakes, and malevolent attacks. To address it, we explore graph-structured data in a random way. In the proposed framework, we train the graph model on massive augmented graph data, generated by random sampling and propagation. In different training epochs, nodes interact with different random subsets of graph, and thus mitigate the risk of being misguiding by specific nodes and edges. On the other hand, the random data augmentation alleviates the overfitting and the implicit ensemble style behind the process improves the capacity of the representation. As the model should generalize well and have a similar prediction on unlabeled data on different data augmentations, despite the potential divergence from sampling randomness, we can further introduce an unsupervised graph-based regularization to the framework.
}

\hide{
In a graph neural model, if any given node's  representation (including unlabeled nodes') has a sufficient and consistent performance in any random subgraphs, that is, any node representation disentangles with specific nodes or edge links, which can be swapped by others, the graph neural model can effectively alleviate the  overfitting to specific graph structure and being misguided by noise, mistakes, and malevolent attacks. One the other hand, the graph neural model trained on exponentially many random subgraphs will also explore graph structures in different levels as sufficiently as possible, while previous models, e.g., GCN and GAT, model the neighborhood as a whole. Based on it, the proposed model, \model, utilizes a series of random sampling strategies to explore graph structures. We found that sampling strategy coupled with graph propagation is an efficient way of graph data augmentation. By leveraging the consistency of abundant unlabeled data on different random augmentations, we can further lower the generalization loss.
}

%
\hide{
\subsection{Over-smoothing Problem }
The over-smoothing issue of GCNs was first studied in \cite{li2018deeper}, which indicates that node features will converge to a fixed vector as the network depth increases. This undesired convergence heavily restricts the expressive power of deep GCNs. Formally speaking, suppose $G$ has $P$ connected components $\{C_i\}^{P}_{i=1}$. Let $\mathbbm{1}^{(i)} \in \mathbb{R}^n$ denote the indication vectors for the $i^{th}$ component $C_i$, which indicates whether a node belongs to $C_i$  i.e.,
	\begin{equation*}
\mathbbm{1}_j^{(i)} = \left\{ \begin{aligned}
1, &v_j \in C_i \\
0, &v_j \notin C_i  \end{aligned}\right..
\end{equation*}
 The over-smoothing phenomenon of GCNs is formulated as the following theorem:
\begin{theorem}
	Given a graph $G$ which has $P$ connected components $\{C_i\}^{P}_{i=1}$, for any $\mathbf{x} \in \mathbb{R}^n$, we have:
	\begin{equation*}
		\lim_{k\rightarrow +\infty} \hat{A}^k x = \widetilde{D}^{\frac{1}{2}}[\mathbbm{1}^{(1)}, \mathbbm{1}^{(2)}, ..., \mathbbm{1}^{(P)}] \hat{x}
	\end{equation*}
	where $\hat{x} \in \mathbb{R}^P$ is a vector associated with $x$.  

\end{theorem}
From this theorem, we could notice that after repeatedly performing the propagation rule of GCNs many times, node features will converge to a linear combination of $\{D^{\frac{1}{2}}\mathbbm{1}^{(i)}\}$. And the nodes in the same connected component only distinct by their degrees. In the above analyses, the activation function used in GCNs is assumed to be a linear transformation, but the conclusion can also be generalized to non-linear case\cite{oono2019graph}.
}

\begin{figure*}
    \centering
    \includegraphics[width= \linewidth]{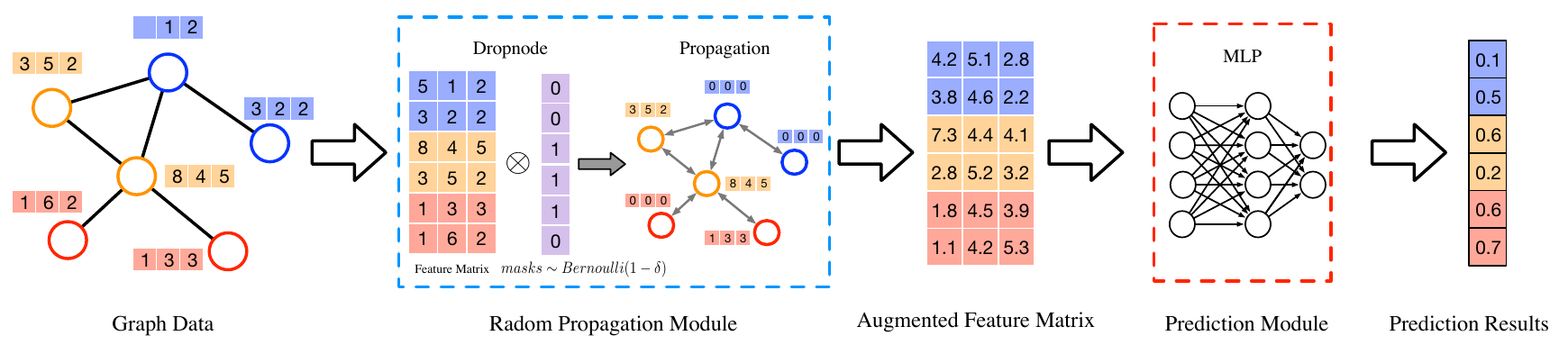}
    \caption{Architecture of \model.}
    \label{fig:arch1}
\end{figure*}

\subsection{Architecture}
The architecture of \model\ is illustrated in Figure~\ref{fig:arch1}. Overall, the model includes two components, i.e., Random propagation module and classification module.

\subsubsection{Random Propagation Module.}
\label{sec:randpro}
\begin{figure}
    \centering
    \includegraphics[width=0.8
    \linewidth]{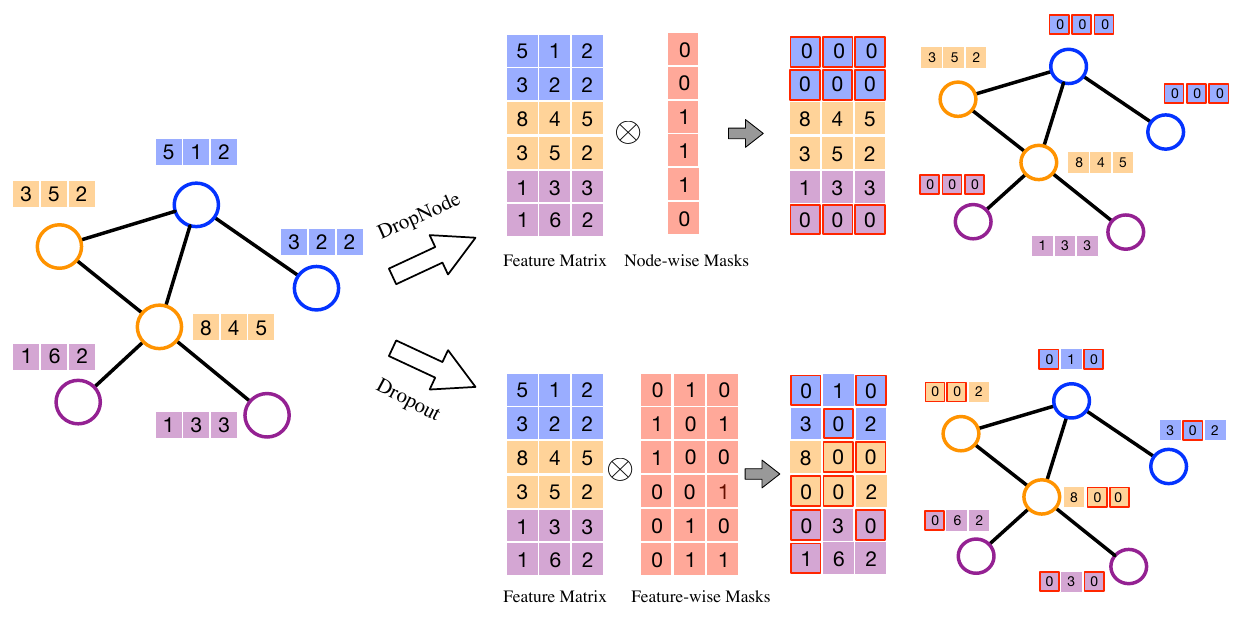}
    \caption{Difference between DropNode and dropout. Dropout drops  elements of $X$ independently. While dropnode drops feature vectors of nodes (row vectors of $X$) randomly.}
    \label{fig:dropnode_vs_dropout}
\end{figure}

\begin{algorithm}[tb]
\caption{Dropnode}
\label{alg:dropnode}
\begin{algorithmic}[1] 
\REQUIRE ~~\\
Original feature matrix $X \in R^{n \times d}$, DropNode probability 
$\delta \in (0,1)$. \\
\ENSURE ~~\\
Perturbed feature matrix  $X\in R^{n \times d}$.
\IF{mode == Inference}
\STATE $\widetilde{X} = X$.
\ELSE
\STATE Randomly sample $n$ masks: $\{\epsilon_i \sim Bernoulli(1-\delta)\}_{i=1}^n$.
\STATE Obtain deformity feature matrix by  multiplying each node's feature vector with the corresponding  mask: $\widetilde{X}_{i} = \epsilon_i \cdot X_{i} $.
\STATE Scale the deformity features: $\widetilde{X} = \frac{\widetilde{X}}{1-\delta}$.
\ENDIF
\end{algorithmic}
\end{algorithm}

\label{sec:randpro}

In random propagation, we aim to perform message passing in a random way during model training. To achieve that, we add an extra node sampling operation called ``DropNode'' in front of the propagation layer.

\vpara{DropNode.} In DropNode, feature vector of each node is randomly removed (rows of $X$ are randomly set to $\vec{0}$) with a pre-defined probability $\delta \in (0,1)$ at each training epoch. The resultant perturbed feature matrix $\widetilde{X}$ is fed into the propagation layer later on.
More formally, we first randomly sample a binary mask $\epsilon_i \sim Bernoulli(1-\delta)$ for each node $v_i$, and obtain the perturbed feature matrix $\widetilde{X}$ by multiplying each node's feature vector with the corresponding mask, i.e., $\widetilde{X}_i=\epsilon_i \cdot X_i$. Furthermore, we scale $\widetilde{X}$ with a factor of $\frac{1}{1-\delta}$ to guarantee the perturbed feature matrix is equal to $X$ in expectation. Please note that the sampling procedure is only performed during training. In inference time, we directly let $\widetilde{X}$ equal to the original feature matrix $X$. The algorithm of DropNode is shown in Algorithm \ref{alg:dropnode}. 

DropNode is similar to dropout\cite{srivastava2014dropout}, a more general regularization method used in deep learning to prevent overfitting. 
In dropout, the elements of $X$ are randomly dropped out independently. While DropNode drops a node's whole features together, serving as a node sampling method. 
Figure \ref{fig:dropnode_vs_dropout} illustrates their differences.
Recent study suggests that a more structured form of dropout, e.g., dropping a contiguous region in DropBlock\cite{ghiasi2018dropblock}, is required for better regularizing CNNs on images. From this point of view, dropnode can be seen as a structured form of dropout on graph data.
We also demonstrate that dropnode is more suitable for graph data than dropout both theoretically (Cf. Section \ref{sec:theory}) and experimentally (Cf. Section \ref{sec:ablation}). 


After dropnode, the perturbed feature matrix $\widetilde{X}$ is fed into a propagation layer to perform message passing. Here we adopt mixed-order propagation, i.e.,

\begin{equation}
\label{equ:kAX}
	\overline{X} = \overline{A} \widetilde{X}.
\end{equation}
Here we define the propagation matrix as $\overline{A} =  \frac{1}{K+1}\sum_{k=0}^K\hat{A}^k$, that is, the average of the power series of $\hat{A}$ from order 0 to order $K$. This kind of propagation rule enables model to incorporate multi-order neighborhoods information, and have a lower risk of over-smoothing compared with using $\hat{A}^K$ when $K$ is large. Similar ideas have been adopted in previous works~\cite{abu2019mixhop,abu2018n}. We compute the Equation \ref{equ:kAX} by iteratively calculating the product of sparse  matrix $\hat{A}^k$ and $\widetilde{X}$. The corresponding time complexity is $\mathcal{O}(Kd(n+|E|))$.

\hide{
In other words, the original features of about $\delta |V|$ nodes are removed (set as $\vec{0}$).
Obviously, this random sampling strategy may destroy the information carried in nodes and the resultant corrupted feature matrix is insufficient for prediction. To compensate it, we try to recover the information in a graph signal propagation process, and get the recovery feature matrix $\widetilde{X}$. The propagation recovery process is:
\begin{equation}
\label{equ:kAX}
	\widetilde{X} = \frac{1}{K+1}\sum_k^K\hat{A}^k X^{'}.
\end{equation}

It also offers a general way for implicit model ensemble on exponentially many augmented data.
}
\hide{
In each epoch, given the input graph, \model\ first generates $S$ graph data augmentations $\{\widetilde{X}^{(s)}| 1\leq s \leq S\}$ via random propagation layer. Then, each $\widetilde{X}^{(s)}$ is fed into a prediction model, leading to the corresponding prediction distribution $\widetilde{Z}^{(s)}$. In optimization, except for optimizing the supervised classification loss $\mathcal{L}^l$ with the given labels $Y^L$, we also minimize the prediction distance among different augmentations of unlabeled nodes via unsupervised consistency loss $\mathcal{L}^u$.}



\hide{
\subsection{Motivation}
Our motivation is that GCNs can not sufficiently leverage unlabeled data in this task. Specifically,

\subsection{Overview of Graph Random Network}
    In this section, we provide a brief introduction of the proposed graph random network. The basic idea of \model is to promote GCNs' generalization and robustness by taking the full advantage of unlabeled data in the graph.

%
Random propagation layer consists of graph dropout and propagation.
We first introduce how to generate a subgraph by random sampling nodes.
Formally, we randomly sample the nodes without replacement at a probability $1-\delta$  and drop the rest.
The deformity feature matrix $X^{'}$ is formed in the following way,
\begin{align}
\label{equ:samplingX1}
\left\{
\begin{aligned}
	& Pr(X^{'}_{i}=\vec{0}) = \delta, \\
	&Pr(X^{'}_{i}=X_i) = 1-\delta. 
\end{aligned}
\right.
\end{align}

In other words, the original features of about $\delta |V|$ nodes are removed (set as $\vec{0}$).
Obviously, this random sampling strategy may destroy the information carried in nodes and the resultant corrupted feature matrix is insufficient for prediction. To compensate it, we try to recover the information in a graph signal propagation process, and get the recovery feature matrix $\widetilde{X}$. The propagation recovery process is:
\begin{equation}
\label{equ:kAX}
	\widetilde{X} = \frac{1}{K}\sum_k^K\hat{A}^k X^{'}.
\end{equation}

\hide{
\begin{equation}
\label{Equ:denoise}
\widetilde{X} = \arg\min_{\widetilde{X}} \frac{1}{2}\left\| \widetilde{X}- X^{'}\right\|^2_2 + \alpha \frac{1}{2}\left \| \widetilde{X} - D^{-\frac{1}{2}}AD^{-\frac{1}{2}} \widetilde{X}\right\|^2_2
\end{equation}  
\begin{small}
\begin{equation}
\label{Equ:denoise}
\widetilde{X}=\arg\min_{\widetilde{X}} 
   \frac{1}{2}\left(\sum_{i, j=1}^{n} A_{i j}\left\|\frac{1}{\sqrt{D_{i i}}} \widetilde{X}_{i}-\frac{1}{\sqrt{D_{j j}}} \widetilde{X}_{j}\right\|^{2}+\mu \sum_{i=1}^{n}\left\|\widetilde{X}_{i}-X^{'}_{i}\right\|^{2}\right)
\end{equation}  
\end{small}
where the first term keeps the denoised signal similar to the measurement and the second term enforces the smoothing of the solution. By setting the derivate of $\widetilde{X}$ to zero, we can obtain the solution to the Equation \ref{Equ:denoise}:
\begin{equation}
\label{Equ:denoise2}
\widetilde{X} = (I+\alpha \tilde{L})^{-1}X^{'}
\end{equation}

To avoid the inversion operation in Equation \ref{Equ:denoise2}, we derive an approximate solution of Equation \ref{Equ:denoise2} using Taylor Extension:
\begin{equation}
\label{Equ:denoise3}
\widetilde{X} = (I + \alpha \tilde{L}^2)^{-1}X^{'} \approx \sum_{i=0}^K (-\alpha\tilde{L}^2)^i X^{'} 
\end{equation}
}

Combining Equation \ref{equ:samplingX1} and \ref{equ:kAX}, the original feature matrix $X$ is firstly \hide{heavily} corrupted by sampling and then smoothed by graph propagation. In fact, the whole procedure consisting of two opposite operations provides a new representation $\widetilde{X}$ for the original $X$. Later we will verify that $\widetilde{X}$ is still a good representation substitute with a sufficient prediction ability, while the randomness inside the process helps the model avoid the over-dependence on specific graph structure and 
explore different graph structure in a more robust way.
We treat these opposite operations as a whole procedure and call them \textit{random propagation layer} in the proposed model \model. 
\hide{After filtered by random sampling operation in random propagation layer, the subsequent model parts only involve interaction among nearly half of the original node features and thus graph-structure data are disentangled somehow.}

Here, we generalize the random propagation layer with another sampling strategies. Instead of removing the feature of a node entirely by sampling once, we drop the element of the feature in each dimension one by one by multiple sampling. This is a multi-channel version of node sampling. Furthermore, we scale the remaining elements with a factor of $\frac{1}{1-\delta}$ to 
 guarantees the deformity feature matrix or adjacency matrix are the same as the original in expectation. Similar idea of dropping and rescaling has been also used in Dropout~\cite{srivastava2014dropout}. Hence we called our three graph sampling strategy graph dropout (\textit{dropnode} and \textit{dropout}), and formulate them following:
 
\vpara{Dropnode.} In dropnode, the feature vector of each node is randomly dropped with a pre-defined probability $\delta \in (0,1)$ during the propagation. More formally, we first form a deformity feature matrix $X^{'}$ in the following way:
\begin{align}
\label{equ:nodedropout}
\left\{
\begin{aligned}
& Pr(X^{'}_{i}=\vec{0}) = \delta,& \\
&Pr(X^{'}_{i}= \frac{X_i}{1-\delta} ) = 1- \delta. &
\end{aligned}
\right.
\end{align}
Then let $X^{'}$ propagate in the graph via Equation~\ref{equ:kAX}. 

\vpara{Dropout.} In dropout, the element of the feature matrix is randomly dropped with a probability $\delta \in (0,1)$ during the propagation. Formally, we have:
\begin{align}
\label{equ:featuredropout}
\left\{
\begin{aligned}
& Pr(X^{'}_{ij}=0) = \delta,& \\
&Pr(X^{'}_{ij}= \frac{X_{ij}}{1-\delta} ) = 1- \delta. &
\end{aligned}
\right.
\end{align}
Then we  propagate $X^{'}$ in the graph via Equation~\ref{equ:kAX}. 


Note that dropout in graph dropout shares the same definition of Dropout~\cite{srivastava2014dropout}, a more general method widely used in deep learning to prevent parameters from overfitting. However, dropout in graph dropout, as a multi-channel version of dropnode, is mainly developed to explore graph-structured data in the semi-supervised learning framework. The graph dropout strategy, directly applied to graph objects, is also an efficient way of graph data augmentation and  model ensemble on exponentially many subgraphs. As for the common dropout applied to the input units, the optimal probability of drop is usually closer to 0 to prevent the loss of information~\cite{srivastava2014dropout}, which is not the case in our graph dropout. In this paper we will further analyze theoretically that graph dropout help \model\ leverage unlabeled data, and dropnode and dropout play different roles. 
}

\hide{
Note that dropout is applied to neural network activation to prevent overfitting of parameters while our node dropout and edge dropout are coupled with graph propagation, performing on graph objects without parameters. We will reveal that our graph dropout strategy is an efficient way of graph data augmentation and graph models ensemble on exponentially many subgraphs. We will further analyse theoretically that graph dropout help \model\ leverage unlabeled data. \reminder{}
}

\vpara{Stochastic Graph Data Augmentation.}
 Feature propagation have been proven to be an effective method for enhancing node representation by aggregating information from neighborhoods, and becomes the key step of Weisfeiler-Lehman Isomorphism test~\cite{shervashidze2011weisfeiler} and GNNs. In random propagation, some nodes are first randomly dropped before propagation. That is, each node only aggregates information from a subset of multi-hop neighborhoods randomly. In this way, random propagation module stochastically generates different representations for a node, suggesting an efficient stochastic augmentation method for graph data. From another point of view, random propagation can also be seen as injecting random noise to the propagation procedure by dropping a portion of nodes . 

We also conduct an interesting experiment to examine the validation of this data augmentation method. In this setting, we first sample a set of augmented node representations $\overline{X}$ from random propagation module with different drop rates, then use $\overline{X}$ to train a GCN for node classification. 
By examining the classification accuracy on $\overline{X}$, we can empirically analyze the influence of the injected noise on model performance. The results with different drop rates have been shown in Figure \ref{fig:redundancy}. It can be seen that the performance only decreases by less than $3\%$ as the drop rate increases from $0$ to $0.5$, indicating the augmented node representations is still sufficient for prediction\footnote{In practice, the drop rate is always set to $0.5$}. 
 Though $\overline{X}$  is still inferior to $X$ in performance, in practice, multiple augmentations will be utilized for training prediction model since the sampling will be performed in every epoch. From the view of bagging~\cite{breiman1996bagging}, this random data augmentation scheme makes the final prediction model implicitly ensemble models on exponentially many augmentations, yielding much better performances than using deterministic propagation.


\begin{figure}
  		\centering
  		\includegraphics[width = 0.8 \linewidth, height =  0.58\linewidth]{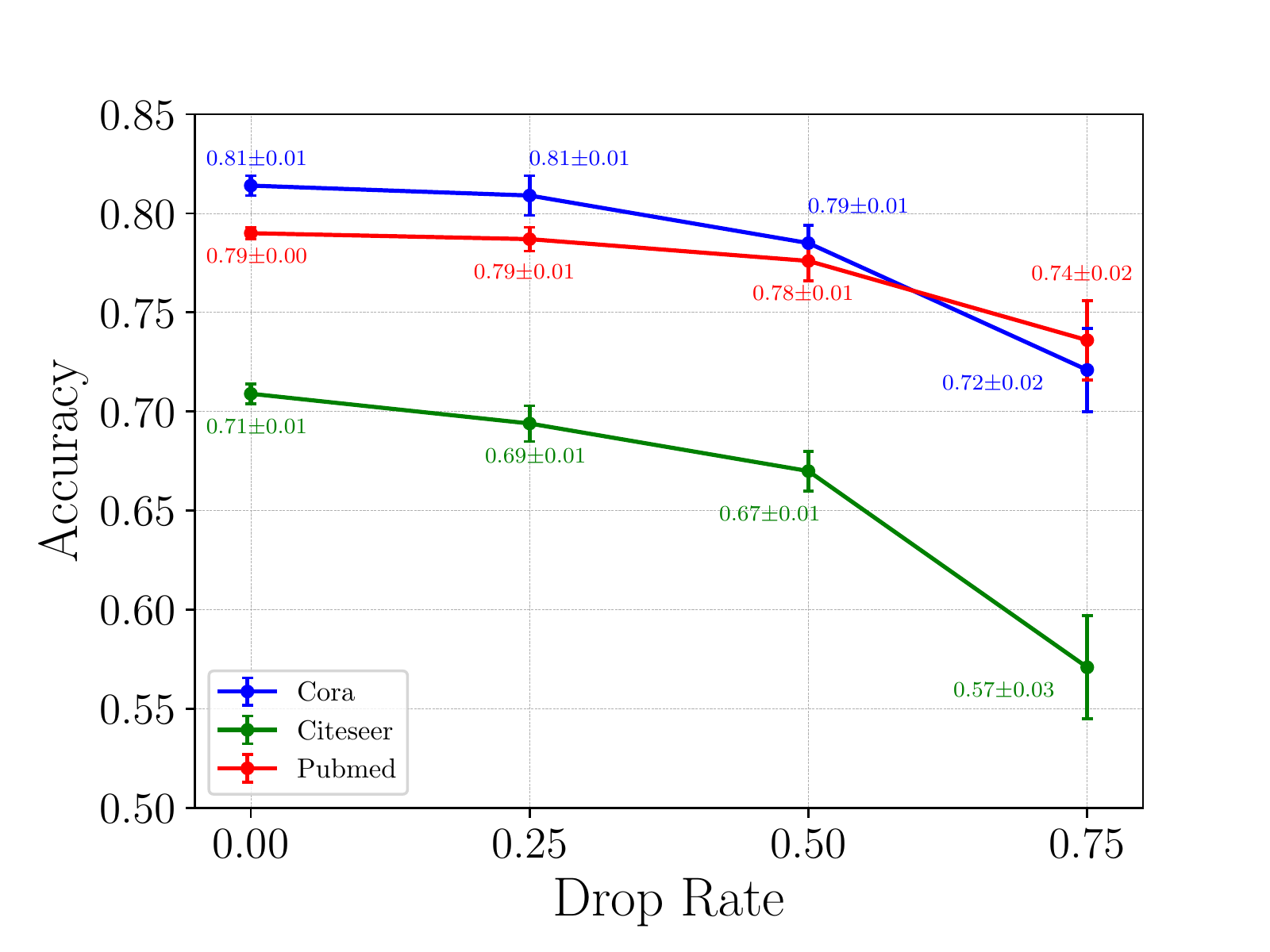}
  	\caption{Classification results of GCN with $\overline{X}$ as input.} 
  	\label{fig:redundancy}
\end{figure}
  



\subsubsection{Prediction Module.}
In prediction module, the augmented feature matrix $\overline{X}$ is fed into a neural network to predict nodes labels. We employ a two-layer MLP as the classifier:
\begin{equation}
\label{equ:mlp}
    P(\mathcal{Y}|\overline{X};\Theta) = \sigma_2(\sigma_1(\overline{X}W^{(1)})W^{(2)})
\end{equation}
where $\sigma_1$ is ReLU function, $\sigma_2$ is softmax function, $\Theta=\{W^{(1)}, W^{(2)}\}$ refers to model parameters. Here the classification model can also adopt more complex GNN based node classification models, e.g., GCN, GAT.  But we find the performance decreases when we replace MLP with GNNs because of the over-smoothing problem. We will explain this phenomenon in Section \ref{sec:oversmoothing}.


\hide{
From another perspective, Equation \ref{equ:samplingX1} and \ref{equ:kAX} in the random propagation layer perform a data augmentation procedure by linear interpolation~\cite{devries2017dataset} in a random subset of the multi-hop neighborhood.
}

\hide{
In a graph, we random sample a fixed proportion of neighborhoods for each node $v_i \in \mathcal{V}$, and let $v_i$ only interact with the sampled neighborhoods in propagation. Using this method, we are equivalent to let each node randomly perform linear interpolation with its neighbors. 

However, generating neighborhood samples for each node always require a time-consuming preprocessing. For example, the sampling method used in GraphSAGE~\cite{hamilton2017inductive} requires $k \times n$ times of sampling operations, where $k$ denotes the size of sampled neighbor set. To solve this problem, here we propose an efficient sampling method to perform random propagation --- graph dropout. Graph dropout is inspired from Dropout\cite{srivastava2014dropout}, a widely used regularization method in deep learning. \reminder{graph dropout is an efficient sampling method}
In graph dropout, we randomly drop a set of nodes or edges in propagation, which is introduced separately. \\

\vpara{Edge Dropout.} The basic idea of edge dropout is randomly dropping a fix proportion of edges during each propagation. 
Specifically, we construct a deformity feature matrix $\hat{A}^{'}$ following:
\begin{align}
\label{equ:nodedropout}
\left\{
\begin{aligned}
& Pr(\hat{A}^{'}_{ij}=\vec{0}) = \delta.& \\
&Pr(\hat{A}^{'}_{ij}= \frac{\hat{A}^{'}_{ij}}{1-\delta} ) = 1- \delta. &
\end{aligned}
\right.
\end{align}
Then we use $\hat{A}^{'}$ as the replacement of $\hat{A}$ in propagation.

\vpara{Node Dropout.} In node dropout, the feature vector of each node is randomly dropped with a pre-defined probability $\delta \in (0,1)$ during propagation. More formally, we first form a deformity feature matrix $X^{'}$ in the following way:
\begin{align}
\label{equ:nodedropout}
\left\{
\begin{aligned}
& Pr(X^{'}_{i}=\vec{0}) = \delta.& \\
&Pr(X^{'}_{i}= \frac{X_i}{1-\delta} ) = 1- \delta. &
\end{aligned}
\right.
\end{align}
Then let $X^{'}$ propagate in graph as the substitute of $X$, i.e., $\tilde{X} = \hat{A}X$. 

Actually, node dropout can be seen as a special form of edge dropout:
}
%

\subsection{Consistency Regularized Training}
\label{sec:consis}
\begin{figure*}
	\centering
	\includegraphics[width= \linewidth]{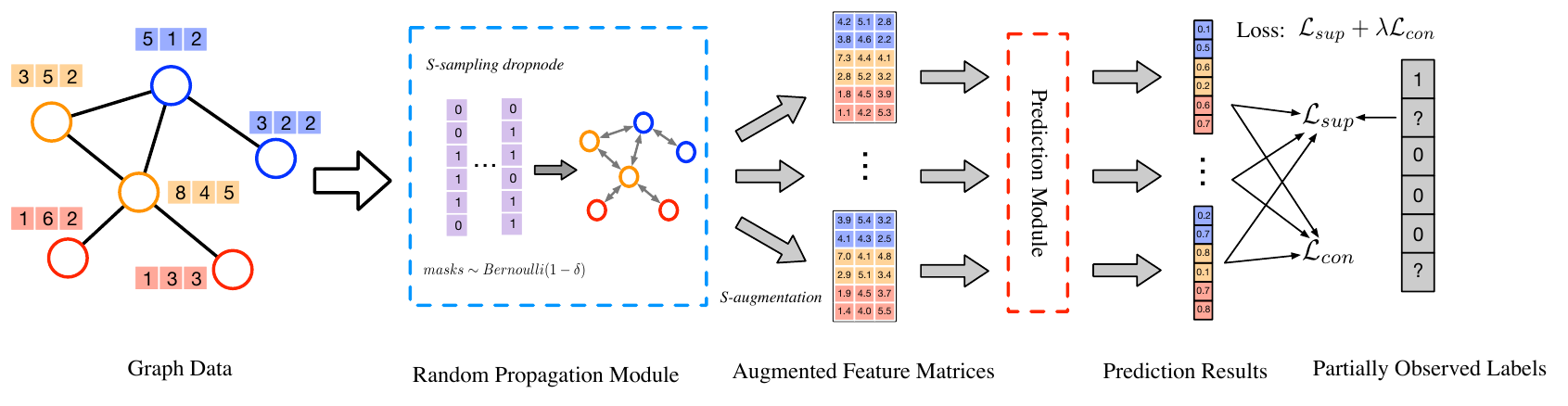}
	\caption{Illustration of consistency regularized training for \model. \yd{how about making the random progagation module into two as well?}}
	\label{fig:arch2}
\end{figure*}
  As mentioned in previous subsection, random propagation module can be seen as an efficient method of stochastic data augmentation. That inspires us to design a consistency regularized training algorithm for optimizing parameters. 
  In the training algorithm, we generate multiple data augmentations at each epoch by performing dropnode multiple times. Besides the supervised classification loss, we also add a consistency regularization loss to enforce model to give similar predictions across different augmentations. This algorithm is illustrated in Figure \ref{fig:arch2}.
  
 \begin{algorithm}[tb]
\caption{Consistency Regularized Training for \model}
\label{alg:2}
\begin{algorithmic}[1] 
\REQUIRE ~~\\
 Adjacency matrix $\hat{A}$,
labeled node set $V^L$,
 unlabeled node set $V^U$,
feature matrix $X \in R^{n \times d}$, 
times of dropnode in each epoch $S$, dropnode probability $\delta$.\\
\ENSURE ~~\\
Prediction $Z$.
\WHILE{not convergence}
\FOR{$s=1:S$} 
\STATE Apply dropnode via Algorithm \ref{alg:dropnode}: $
\widetilde{X}^{(s)} \sim \text{dropnode}(X,\delta)$. 
\STATE Perform propagation: $\overline{X}^{(s)} = \frac{1}{K}\sum_{k=0}^K\hat{A}^k \widetilde{X}^{(s)}$.
\STATE Predict class distribution using MLP: 
$\widetilde{Z}^{(s)} = P(\mathcal{Y}|\overline{X}^{(s)};\Theta)$.
\ENDFOR
\STATE Compute supervised classification loss $\mathcal{L}_{sup}$ via Equation \ref{equ:loss} and consistency regularization loss via Equation \ref{equ:consistency}.
\STATE Update the parameters $\Theta$ by gradients descending:
$$\nabla_\Theta \mathcal{L}_{sup} + \lambda \mathcal{L}_{con}$$
\ENDWHILE
\STATE Output prediction $Z$ via Equation \ref{equ:inference}
\end{algorithmic}
\end{algorithm}

 \subsubsection{$S-$augmentation Prediction}
At each epoch, we aim to generate $S$ different data augmentations for graph data $X$. To achieve that, we adopt $S-$sampling dropnode strategy in random propagation module. That is, we performing $S$ times of random sampling in dropnode to generate $S$ perturbed feature matrices $\{\widetilde{X}^{(s)}|1\leq s \leq S\}$. Then we propagate these features according to Equation \ref{equ:kAX} respectively, and hence obtain $S$ data augmentations $\{\overline{X}^{(s)}|1 \leq s \leq S\}$. Each of these augmented feature matrice are fed into the prediction module to get the corresponding output:

 \begin{equation}
     \widetilde{Z}^{(s)} = P(\mathcal{Y}|\overline{X}^{(s)}; \Theta).
 \end{equation}
Where $\widetilde{Z}^{(s)} \in (0,1)^{n\times C}$ denotes the classification probabilities on the $s^{th}$ augmented data. 
\subsubsection{Supervised Classification Loss.}
The supervised objective of graph node classification in an epoch is the averaged cross-entropy loss over $S$ times sampling:
\begin{equation}
\label{equ:loss}
	\mathcal{L}_{sup} = -\frac{1}{S}\sum_{s=1}^{S}\sum_{i=1}^m \sum_{j=1}^{C}Y_{i,j} \ln \widetilde{Z}_{i,j}^{(s)} ,
\end{equation}

\noindent where $Y_{i,l}$ is binary, indicating whether node $i$ has the label $l$. By optimizing this loss, we can also enforce the model to output the same predictions on multiple augmentations of a labeled node.
 However, in the semi-supervised setting, labeled data is always very rare. In order to make full use of unlabeled data, we also employ consistency regularization loss in \model. 
 
 \subsubsection{Consistency Regularization Loss.} How to optimize the consistency among $S$ augmentations of unlabeled data? 
 Let's first consider a simple case, where the random propagation procedure is performed only twice in each epoch, i.e., $S=2$. Then a straightforward method is to minimize the distributional distance between two outputs:
 \begin{equation}
 \label{equ:2d}
     \min \sum_{i=1}^n \mathcal{D}(\widetilde{Z}^{(1)}_i, \widetilde{Z}^{(2)}_i),
 \end{equation}
where $ \mathcal{D}(\cdot,\cdot)$ is the distance function. Then we extend it into multiple augmentation situation. We can first calculate the label distribution center by taking average of all distributions:
\begin{equation}
    \overline{Z} = \frac{1}{S}\sum_{s=1}^{S} \widetilde{Z}^{(s)}.
\end{equation}

And we can minimize the distributional distance between $\widetilde{Z}^{(s)}$ and $\overline{Z}$, i.e., $\min \sum_{s=1}^{S}\sum_{i=1}^n \mathcal{D}(\overline{Z}_i, \widetilde{Z}^{(s)}_i)$.
\hide{
\begin{equation}
    \min \sum_{s=1}^{S}\sum_{i=1}^n \mathcal{D}(\overline{Z}_i, \widetilde{Z}^{(s)}_i).
\end{equation}
}
However, the distribution center calculated in this way is always inclined to have more entropy value, which indicates to be more ``uncertain''. 
Thus this method will bring extra uncertainty into model's predictions. To avoid this problem, We utilize the label sharpening trick in \model. Specifically, we apply a sharpening function onto the averaged label distribution to reduce its entropy:
\begin{equation}
 \overline{Z}^{'}_{i,l} = \frac{\overline{Z}_{i,l}^{\frac{1}{T}}}{\sum_{j=1}^{|\mathcal{Y}|}\overline{Z}_{i,j}^{\frac{1}{T}}},
\end{equation}
where $0< T\leq 1$ acts as the ``temperature'' which controls the sharpness of the categorical distribution. As $T \to 0$, the sharpened label distribution will approach a one-hot distribution. As the substitute of $\overline{Z}$, we minimize the distance between  $\widetilde{Z}^i$ and $\overline{Z}^{'}$ in \model:
\begin{equation}
\label{equ:consistency}
    \mathcal{L}_{con} =   \frac{1}{S}\sum_{s=1}^{S}\sum_{i=1}^n \mathcal{D}(\overline{Z}^{'}_i, \widetilde{Z}^{(s)}_i).
\end{equation}

By setting $T$ as a small value, we could enforce the model to output low-entroy predictions. This can be seen as adding an extra entropy minimization regularization into the model, which assumes that classifier's decision boundary should not pass through high-density regions of the marginal data distribution\cite{grandvalet2005semi}. As for the  distance function $\mathcal{D}(\cdot, \cdot)$, we adopt squared $L_2$ loss, i.e., $\mathcal{D}(x, y)=\|x-y\|^2$, in our model. It has been proved to be less sensitive to incorrect predictions\cite{berthelot2019mixmatch}, and is more suitable to the semi-supervised setting compared to cross-entropy. 

\subsubsection{Training and Inference}
In each epoch, we employ both supervised classification loss (Cf. Equation \ref{equ:loss}) and consistency regularization loss (Cf. Equation \ref{equ:consistency}) on $S$ times of sampling. Hence, the final loss is:
\begin{equation}
\label{equ:inf}
	\mathcal{L} = \mathcal{L}_{sup} + \lambda \mathcal{L}_{con}.
\end{equation}
Here $\lambda$ is a hyper-parameter which controls the balance between supervised classification loss and consistency regularization loss.
In the inference phase, as mentioned in Section \ref{sec:randpro}, we directly use original feature $X$ as the output of dropnode instead of sampling. Hence the inference formula is:
\begin{equation}
\label{equ:inference}
Z= P\left(\mathcal{Y}~\bigg|~\frac{1}{K+1}\sum_{k=0}^K\hat{A}^k X;\hat{\Theta}\right).
\end{equation}
Here $\hat{\Theta}$ denotes the optimized parameters after training. We summarize our algorithm in Algorithm \ref{alg:2}.

\hide{Note that $X$ is equal to the expectation of $\widetilde{X}$s from data augmentation by multiple sampling. From the view of model bagging~\cite{breiman1996bagging}, the final graph model implicitly aggregates models trained on exponentially many subgraphs, and performs a plurality vote among these models in the inference phase, resulting in a lower generalization loss. }
 
\hide{
\begin{equation}
\label{equ:model}
	\widetilde{Z}^{(s)}=p_{\text{model}}\left(\mathcal{Y}~\bigg|~\frac{1}{K}\sum_k^K\hat{A}^k X^{'}\right) \in R^{n \times |\mathcal{Y}|}.
\end{equation}


Hence the supervised objective of graph node classification in an epoch is the averaged cross-entropy loss over $S$ times sampling:
\begin{equation}
\label{equ:loss}
	\mathcal{L}^l = -\frac{1}{S}\sum_{s}^{S}\sum_{i\in V^L } \sum_l^{|\mathcal{Y}|}Y_{i,l} \ln \widetilde{Z}_{i,l}^s ,
\end{equation}

\noindent where $Y_{i,l}$ is binary, indicating whether node $i$ has the label $l$. 

In each epoch, we employ both supervised cross-entropy loss (Cf. Eq.\ref{equ:loss}) and unsupervised consistency loss (Cf. Eq.\ref{equ:consistency}) on $S$ times of sampling. Hence, the final loss is:
\begin{equation}
\label{equ:inf}
	\mathcal{L} = \mathcal{L}^l + \lambda \mathcal{L}^u.
\end{equation}
Here $\lambda$ is a hyper-parameter which controls the balance between supervised classification loss and unsupervised consistency loss.
In the inference phase, the output is achieved by averaging  over the results on exponentially many augmented test data. This can be economically realized by inputting $X$ without sampling:
\begin{equation}
\label{equ:inference}
Z= p_{\text{model}}\left(\mathcal{Y}~\bigg|~\frac{1}{K}\sum_k^K\hat{A}^k X\right).
\end{equation}

Note that $X$ is the average of $X^{'}$s from data augmentation by multiple sampling. From the view of model bagging~\cite{breiman1996bagging}, the final graph model implicitly aggregates models trained on exponentially many subgraphs, and performs a plurality vote among these models in the inference phase, resulting in a lower generalization loss.  
}


\hide{
\subsection{graph propagation for smoothing}

In this section, we introduce the random propagation layer, an efficient method to perform stochastic data augmentation on the graph. We first demonstrate the propagation layer in GCNs is actually a special form of data augmentation on graph-structured data. Based on this discovery, we propose random propagation strategy, which augments each node with a part of randomized selected neighborhoods \reminder{}. Finally we show that this strategy can be efficiently instantiated with a stochastic sampling method --- graph dropout.

\subsection{Feature Propagation as Data Augmentation}
\label{sec:aug}
An elegant data augmentation technique used in supervised learning is linear interpolation \cite{devries2017dataset}. Specifically, for each example on training dataset, we first find its $K-$nearest neighboring samples in feature space which share the same class label. Then for each pair of neighboring feature vectors, we get the augmented example using interpolation:
\begin{equation}
    x^{'}_i = \lambda x_i + (1-\lambda)x_j
\end{equation}
where $x_j$ and $x_i$ are neighboring pairs with the same label $y$, $x^{'}_i$ is augmented feature vector, and $\lambda \in [0,1]$ controls degree of interpolation\reminder{}. Then the example $(x^{'}_i, y)$ is used as extra training sample to facilitate the learning task. In this way, the model is encouraged to more smooth in input space and more robust against small perturbations. \reminder{why?}

As for the problem of semi-supervised learning on graphs, we assume the graph signals are smooth, i.e., \textit{neighborhoods have similar feature vectors and similar class labels}. Thus a straightforward idea of augmenting graph-structured data is interpolating node features with one of its neighborhoods' features. However, in the real network, there exist small amounts of neighboring nodes with different labels and we can't identify that for samples with non-observable labels \reminder{?}. Thus simple interpretation with one neighbor might bring uncontrollable noise into the learning framework. 
An alternative solution is interpolating samples with multiple neighborhoods, which leads to Laplacian Smoothing:

\begin{equation}
\label{equ:lap}
    x^{'}_i = (1-\lambda) x_i + \lambda \sum_j \frac{\widetilde{A}_{ij}}{\widetilde{D}_{ii}} x_j
\end{equation}.
Here we follow the definition of GCNs which adding a self-loop for each node in graph\reminder{}. Rewriting Eq.~\ref{equ:lap} in matrix form, we have:
\begin{equation}
\label{equ:lap2}
    X^{'}=  (I-\lambda I)X + \lambda \widetilde{D}^{-1} \widetilde{A}X
\end{equation}.
As pointed out by Li et.al.\cite{li2018deeper}\reminder{}, when $\lambda = 1$ and replacing $\Tilde{D}^{-1} \Tilde{A}$ with the symmetric normalized adjacency matrix $\hat{A}$, Eq.~\ref{equ:lap2} is identical to the propagation layer in GCN, i.e., $X^{'}= \hat{A}X$.

}

\hide{
\subsection{Random Propagation with Graph Dropout}
With the insight from last Section~\ref{sec:aug}, we develop a random propagation strategy for stochastic data augmentation on graph.

During each time of random propagation, we random sample a fixed proportion of neighborhoods for each node $v_i \in \mathcal{V}$, and let $v_i$ only interact with the sampled neighborhoods in propagation. Using this method, we are equivalent to let each node randomly perform linear interpolation with its neighbors. 

However, generating neighborhood samples for each node always require a time-consuming preprocessing. For example, the sampling method used in GraphSAGE~\cite{hamilton2017inductive} requires $k \times n$ times of sampling operations, where $k$ denotes the size of sampled neighbor set. To solve this problem, here we propose an efficient sampling method to perform random propagation --- graph dropout. Graph dropout is inspired from Dropout\cite{srivastava2014dropout}, a widely used regularization method in deep learning. \reminder{graph dropout is an efficient sampling method}
In graph dropout, we randomly drop a set of nodes or edges in propagation, which is introduced separately. \\

\vpara{Edge Dropout.} The basic idea of edge dropout is randomly dropping a fix proportion of edges during each propagation. 
Specifically, we construct a deformity feature matrix $\hat{A}^{'}$ following:
\begin{align}
\label{equ:nodedropout}
\left\{
\begin{aligned}
& Pr(\hat{A}^{'}_{ij}=\vec{0}) = \delta.& \\
&Pr(\hat{A}^{'}_{ij}= \frac{\hat{A}^{'}_{ij}}{1-\delta} ) = 1- \delta. &
\end{aligned}
\right.
\end{align}
Then we use $\hat{A}^{'}$ as the replacement of $\hat{A}$ in propagation.

\vpara{Node Dropout.} In node dropout, the feature vector of each node is randomly dropped with a pre-defined probability $\delta \in (0,1)$ during propagation. More formally, we first form a deformity feature matrix $X^{'}$ in the following way:
\begin{align}
\label{equ:nodedropout}
\left\{
\begin{aligned}
& Pr(X^{'}_{i}=\vec{0}) = \delta.& \\
&Pr(X^{'}_{i}= \frac{X_i}{1-\delta} ) = 1- \delta. &
\end{aligned}
\right.
\end{align}
Then let $X^{'}$ propagate in graph as the substitute of $X$, i.e., $\tilde{X} = \hat{A}X$. 

Actually, node dropout can be seen as a special form of edge dropout: \reminder{Dropping a node is equivalent to drop all the edges start from the node.} \\

\reminder{connection between graph dropout and feature dropout. dropblock..}


}

\hide{
\section{Consistency Optimization}
\subsection{\drop: Exponential Ensemble of GCNs}
Benefiting from the information redundancy, the input of \shalf gives a relatively sufficient view for node classification.
It is straightforward to generalize \shalf to ensemble multiple \shalf s, as each set $\mathcal{C}$ provides a unique network view. 
In doing so, we can use $|\mathcal{C}_i \bigcap \mathcal{C}_j|$ to measure the correlation between two network data view $i$ and $j$. In particular, $\mathcal{C}$ and $V-\mathcal{C}$ can be considered independent. However, one thing that obstructs the direct application of traditional ensemble methods is that a network can generate exponential data views when sampling $n/2$ nodes from its $n$ nodes.

\begin{figure}{
		\centering
		\includegraphics[width = 1\linewidth]{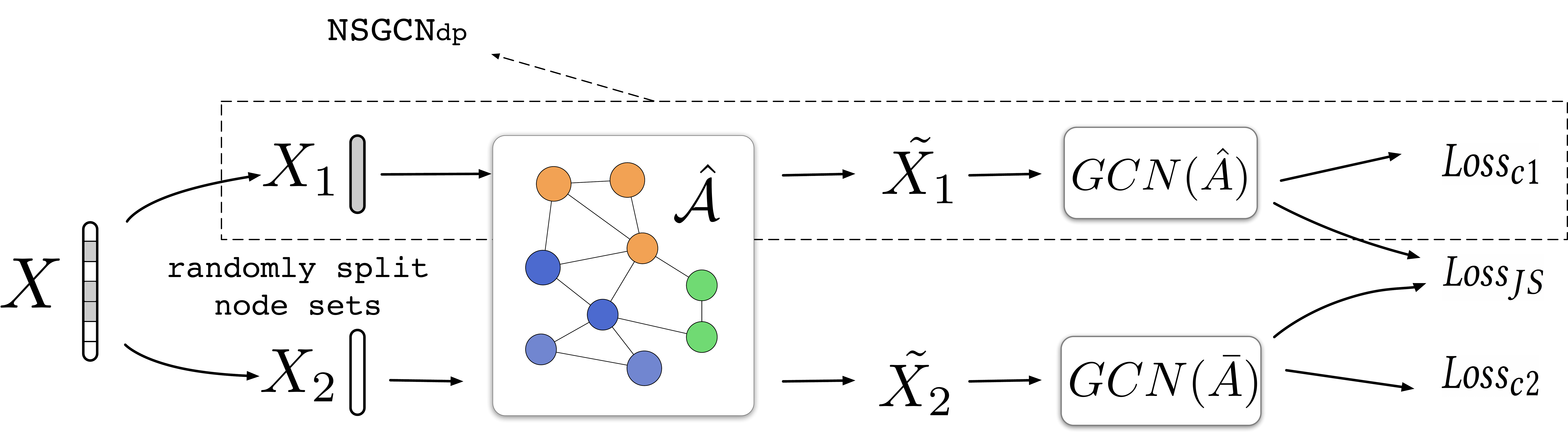}
		\caption{\sdrop and \dm. \drop: In each epoch we randomly drop half of the nodes and do the process of \half. The input in the inference phase is the original feature matrix without dropout, but with a discount factor of $0.5$. This method can be treated as the exponential ensemble of \half.
			\dm: In each epoch of \drop, we consider the half of the nodes that are dropped are independent to the remaining nodes and input them to another \drop. We minimize the disagreement of these two \drop s via the Jensen-Shannon divergence. This method can be treated as an economical co-training of two independent \drop s.}
		\label{fig:cotrain}
	}
\end{figure}

We propose a dropout-style ensemble model \drop.
In \drop, we develop a new network-sampling ensemble method---graph dropout, which samples the node set $\mathcal{C}$ from a network's node set and do the \shalf process in each training epoch. 
In the inference phase, we use the entire feature matrix with a discount factor $0.5$ as input~\cite{srivastava2014dropout}. 
The data flow in \drop\ is shown in 
Figure~\ref{fig:cotrain}.

Specifically, given the recovery feature matrix $\widetilde{X}$ in a certain epoch, the softmax output of GCN in the training phase is $Z = GCN(\widetilde{X}) \in R^{n \times |\mathcal{Y}|}$. Hence the  objective of node classification in the network is

\begin{equation}
\label{equ:loss}
	Loss_{c1} = -\sum_{i\in V^L } \sum_l^{|\mathcal{Y}|}Y_{i,l} \ln Z_{i,l}
\end{equation}

\noindent where $Y_{i,l}$ is binary, indicating whether node $i$ has the label $l$.  

To make the model practical, in the inference phase, the output is achieved by using the entire feature matrix $X$ and a discount factor $0.5$~\cite{srivastava2014dropout}, that is, 

\begin{equation}
\label{equ:inf}
	\hat{Z} = GCN\left(\frac{0.5}{k}\sum_i^k\hat{A}^i X\right)
\end{equation}

The discount factor guarantees the input of GCNs in the inference phase is the same as the expected input in the training phase. Similar idea has been also used in~\cite{srivastava2014dropout}. 

\subsection{\dm: Disagreement Minimization}
In this section, we present the \dm\ model, which leverages the independency between the chosen node set $\mathcal{C}$ and the dropout node set $V-\mathcal{C}$ in each epoch of \drop. 

In \drop, the network data view provided by the chosen set $\mathcal{C}$ is trained and the view provided by the remaining nodes $V-\mathcal{C}$ at each epoch is completely ignored. 
The idea behind \drop\ aims to train over as many as network views as possible. 
Therefore, in \dm, we propose to simultaneously train two \drop s with complementary inputs---$\mathcal{C}$ and $V-\mathcal{C}$---at every epoch. 
In the  training phase, we improve the two GCNs' prediction ability separately by minimizing the disagreement of these two models. 


The objective to minimize the disagreement of the two complementary \sdrop at a certain epoch can be obtained by Jensen-Shannon-divergence, that is, 

\begin{equation}
\label{equ:jsloss}
	Loss_{JS} = \frac{1}{2}\sum_{i\in V} \sum_l^{|\mathcal{Y}|} \left(Z'_{i,l} \ln \frac{Z'_{i,l}}{Z''_{i,l}} + Z''_{i,l} \ln \frac{Z''_{i,l}}{Z'_{i,l}}\right)
\end{equation}

\noindent where $Z'$ and $Z''$ are the outputs of these two \drop s in the training phase, respectively. 

The final objective of \sdm is

\begin{equation}
\label{equ:loss}
	Loss_{dm} = (Loss_{c1} + Loss_{c2})/2 + \lambda Loss_{JS}
\end{equation}

In the inference phase, the final prediction is made by the two GCNs together, i.e., 

\begin{equation}
\label{equ:inf}
	\hat{Z} = (\hat{Z}^{'} + \hat{Z}{''})/2
\end{equation}

Practically, the correlation of two \drop s and their individual prediction ability jointly affect the performance of \dm. 
The inputs to the two models are complementary  and usually uncorrelated, 
empowering \dm\ with more representation capacity than \drop, which is also evident from the empirical results in Section \ref{sec:exp}.

\subsection{ Graph Dropout for Network Sampling}
In GraphSAGE~\cite{hamilton2017inductive} and FastGCN~\cite{chen2018fastgcn}, 
nodes or edges are sampled in order to accelerate training, avoid overfitting, and make the network data regularly like grids. However in the inference phase, these models may miss information when the sampling result is not representative. 

Different from previous works, 
our \drop\ and \dm\ models are based on the network redundancy observation and the proposed dropout-like ensemble enable the (implicit) training on exponential sufficient network data views, making them outperform not only existing sampling-based methods, but also the original GCN with the full single-view~\cite{kipf2016semi}. 
This provides 
insights into the network sampling-based GCNs and offers a new way to sample networks for GCNs.

In addition, there exist connections between the original dropout mechanism in CNNs and our dropout-like network sampling technique. 
Similarly, both techniques use a discount factor in the inference phase. 
Differently, the dropout operation in the fully connected layers in CNNs results in the missing values of the activation (hidden representation) matrix $H^{(l)}$, while our graph dropout sampling method is applied in network nodes, which removes network information in a structured way. 

Notice that very recently, DropBlock~\cite{ghiasi2018dropblock} suggested that a more structured form of dropout is required for better regularizing CNNs, and from this point of view, our graph dropout based network sampling is more in line with the idea of DropBlock. In \drop\ and \dm, dropping network information, instead of removing values from the activation functions, brings a structured way of removing information from the network. 
}


}
    \subsection{Theoretical Analysis}
\label{sec:theory}

We theoretically discuss the regularization effects brought by \textit{random propagation} and \textit{consistency regularization} in \model.
\hide{
The main theoretical conclusions include: 
\begin{itemize}
    \item The DropNode regularization with consistency regularization (or supervised classification) loss can enforce the consistency of the classification confidence between each node and its all (or labeled) multi-hop neighborhoods. 
    \item Dropout is actually an adaptive $L_2$ regularization for $\mathbf{W}$ in GNNs, and its regularization term is the upper bound of DropNode's. 
    By minimizing this term, dropout can be regarded as an approximation of DropNode.
\end{itemize}
}
For analytical simplicity, we assume that the MLP used in \model\ has one single output layer, and the task is binary classification. 
Thus we have 
$\widetilde{\mathbf{Z}} = \text{sigmoid}(\overline{\mathbf{A}} \widetilde{\mathbf{X}}\cdot \mathbf{W} ),$ 
where 
$\mathbf{W} \in \mathbb{R}^{d}$ is the learnable parameter vector. 
For the $i^{th}$ node, the corresponding conditional distribution is $\tilde{z}_i^{y_i}(1-\tilde{z}_i)^{1-y_i},$ 
in which  $\tilde{z}_i \in \widetilde{\mathbf{Z}} $ 
and  $y_i \in \{0,1\}$ denotes the corresponding label. 

As for the consistency regularization loss, we consider the simple case of generating $S=2$ augmentations. 
Then the loss  
   $
   \mathcal{L}_{con} =\frac{1}{2}\sum_{i=0}^{n-1} \left(\tilde{z}_i^{(1)}-\tilde{z}_i^{(2)}\right)^2
   $,
where $\tilde{z}^{(1)}_i$ and $\tilde{z}^{(2)}_i$ represent the model's two outputs on node $i$ corresponding to the two augmentations, respectively. 

With these assumptions, we have the following theorem with proofs in Appendix \ref{them:proof1}. 
\begin{thm}
\label{thm1}
In expectation, optimizing the unsupervised consistency loss $\mathcal{L}_{con}$ is approximate to optimize a regularization term: $\mathbb{E}_\epsilon\left(\mathcal{L}_{con}\right) \approx \mathcal{R}^c(\mathbf{W}) = \sum_{i=0}^{n-1}z^2_i(1-z_i)^2 \textnormal{Var}_\epsilon \left(\overline{\mathbf{A}}_i \widetilde{\mathbf{X}}\cdot \mathbf{W} \right)$.
\end{thm}

\vpara{DropNode Regularization.} With DropNode as the perturbation method, we can easily check that
$\small
\label{equ:dropnodevar}
\text{Var}_\epsilon(\overline{\mathbf{A}}_i \widetilde{\mathbf{X}}\cdot \mathbf{W}) = \frac{\delta}{1-\delta} \sum_{j=0}^{n-1}(\mathbf{X}_j \cdot \mathbf{W})^2 (\overline{\mathbf{A}}_{ij})^2, 
$
where $\delta$ is drop rate.
Then the corresponding regularization term $\mathcal{R}^c_{DN}$ can be expressed as:
\begin{equation}
\small
\label{equ:Rc}
\begin{aligned}
\mathcal{R}_{DN}^c(\mathbf{W}) = \frac{\delta}{1-\delta} \sum_{j=0}^{n-1} \left[(\mathbf{X}_j \cdot \mathbf{W})^2  \sum_{i=0}^{n-1} (\overline{\mathbf{A}}_{ij})^2 z^2_i(1-z_i)^2 \right].
\end{aligned}
\end{equation}

Note that $z_i(1-z_i)$ (or its square) is an indicator of the classification uncertainty for the $i^{th}$ node, as $z_i(1-z_i)$ (or its square) reaches its maximum at $z_i=0.5$ and minimum at $z_i=0$ or $1$. 
Thus $\sum_{i=0}^{m-1}  (\overline{\mathbf{A}}_{ij})^2z^2_i(1-z_i)^2$ can be viewed as the weighted average classification uncertainty over the $j^{th}$ node's multi-hop neighborhoods with the weights as the square values of $\overline{\mathbf{A}}$'s elements, which is related to graph structure. 
On the other hand, $(\mathbf{X}_j \cdot \mathbf{W})^2$---as the square of the input of sigmoid---indicates the classification confidence for the $j^{th}$ node. 
In optimization, in order for a node to earn a higher classification confidence $(\mathbf{X}_j \cdot \mathbf{W})^2$, it is required that the node's neighborhoods have lower classification uncertainty scores. 
Hence, \textit{the random propagation with the consistency regularization loss can enforce the consistency of the classification confidence between each node and its multi-hop neighborhoods}.

\hide{
\vpara{Dropout Regularization.} For dropout, the corresponding variance term is:
\begin{equation}
Var_\epsilon(\hat{A}^k_i \widetilde{\mathbf{X}}\cdot W) = \frac{\delta}{1-\delta} \sum_{j=1}^n \sum_{k=1}^d X_{jk}^2 W_k^2 (\hat{A}^k_{ij})^2.
\end{equation}
Then the corresponding quadratic regularization term is:
\begin{equation}
\begin{aligned}
\mathcal{R}^q(W)
& = \frac{1}{2}\frac{\delta}{1-\delta} \sum_{h=1}^d W_h^2  \sum_{j=1}^n \left[X_{jh}^2  \sum_{i=1}^m z_i(1-z_i) (\hat{A}^k_{ij})^2\right].  \\
\end{aligned}
\end{equation}

Similar to the dropnode case, the classification uncertainty $z_i(1-z_i)$ of labeled nodes are also propagated in the graph and all nodes update their uncertainty with the smoothness of the graph. However, \textit{differently from dropnode, dropout plays an adaptive l2 regularization role; and differently from the general dropout, dropout on the graph takes unlabeled data, prediction confidence on labeled data, and the smoothness of the graph into account.}

By applying Cauchy-Buniakowsky-Schwarz Inequality, we have:
\begin{equation}
\begin{aligned}
 &~\frac{1}{2}\frac{\delta}{1-\delta} \sum_{j=1}^n \left[(X_j \cdot W)^2   \left(\sum_{i=0}^{m-1} \sqrt{z_i(1-z_i)}(\hat{A}^k_{ij})\right)^2 \right].\\
 \leq &~\frac{1}{2}\frac{\delta}{1-\delta} \sum_{h=1}^d W_h^2  \sum_{j=1}^n \left[X_{jh}^2  \sum_{i=0}^{m-1} z_i(1-z_i) (\hat{A}^k_{ij})^2\right].
\end{aligned}
\end{equation}

Hence, the regularization term of dropout is the upper bound of that of dropnode.
}

\hide{
\vpara{Dropout Regularization.}
\begin{figure}
    \centering
    \includegraphics[width=0.9
    \linewidth]{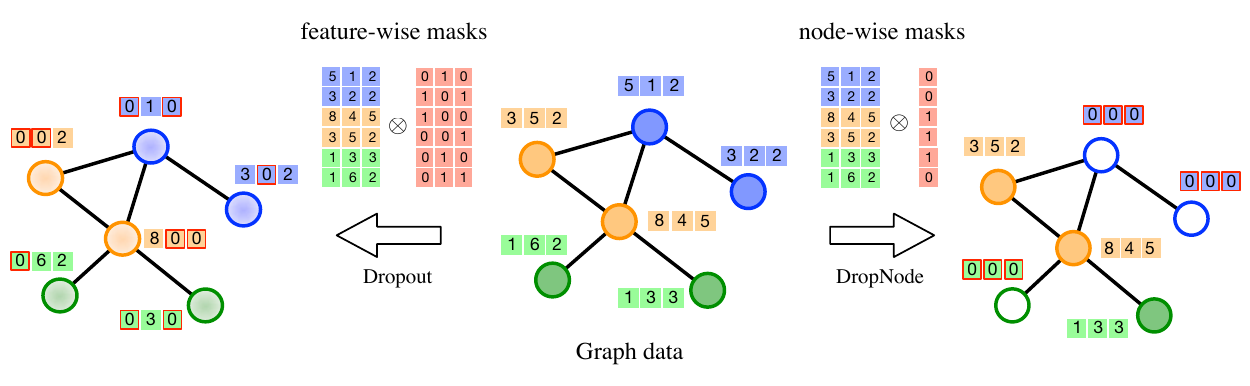}
    \vspace{-0.13in}
    \caption{DropNode vs. dropout. 
    \textmd{Dropout drops each element in $\mathbf{X}$ independently, while DropNode drops the entire features of selected nodes, i.e., the row vectors of $\mathbf{X}$, randomly. }} 
    \label{fig:dropnode_vs_dropout}
    \vspace{-0.15in}
\end{figure}
}
 
\vpara{Dropout Regularization.}
With $\mathbf{X}$ perturbed by dropout, the variance term 
$ 
\small
\text{Var}_{\epsilon}(\overline{\mathbf{A}}_i \widetilde{\mathbf{X}}\cdot \mathbf{W}) = \frac{\delta}{1-\delta} \sum_{j=0}^{n-1} \sum_{k=0}^{d-1} \mathbf{X}_{jk}^2 \mathbf{W}_k^2 (\overline{\mathbf{A}}_{ij})^2.$ The corresponding regularization term $\mathcal{R}^c_{Do}$ is
\begin{equation}
\small
\begin{aligned}
\label{equ:dropoutreu}
{\mathcal{R}^c_{Do}}(\mathbf{W})
& = \frac{\delta}{1-\delta} \sum_{h=0}^{d-1} \mathbf{W}_h^2  \sum_{j=0}^{n-1} \left[\mathbf{X}_{jh}^2  \sum_{i=0}^{n-1} z^2_i(1-z_i)^2 (\overline{\mathbf{A}}_{ij})^2\right].  \\
\end{aligned}
\end{equation}
Similar to DropNode, this extra regularization term also includes the classification uncertainty  $z_i(1-z_i)$ of neighborhoods. 
However, \textit{we can observe that different from the DropNode regularization, dropout is actually an adaptive $L_2$ regularization for $\mathbf{W}$, where the regularization coefficient is associated with unlabeled data, classification uncertainty, and the graph structure}. 

Previous work~\cite{wager2013dropout} has also drawn similar conclusions for the case of applying dropout in generalized linear models. 

\hide{
\begin{equation}
\small
\widetilde{\mathcal{R}}^c(\mathbf{W}) \geq \frac{\delta}{1-\delta} \sum_{j=0}^{n-1} \left[(\mathbf{X}_j \cdot \mathbf{W})^2  \sum_{i=0}^{n-1} (\overline{\mathbf{A}}_{ij})^2 z^2_i(1-z_i)^2 \right] = \mathcal{R}^c(\mathbf{W})
\end{equation}

\begin{equation}
    \begin{aligned}
&~\frac{\delta}{1-\delta} \sum_{h=0}^{d-1} \mathbf{W}_h^2  \sum_{j=0}^{n-1} \left[\mathbf{X}_{jh}^2  \sum_{i=0}^{n-1} z_i(1-z_i) (\overline{\mathbf{A}}_{ij})^2\right] \\
\geq &~\frac{\delta}{1-\delta} \sum_{j=0}^{n-1} \left[(\mathbf{X}_j \cdot \mathbf{W})^2  \sum_{i=0}^{n-1} (\overline{\mathbf{A}}_{ij})^2 z_i(1-z_i) \right].
    \end{aligned}
\end{equation}

\begin{equation}
\begin{aligned}
&~\frac{\delta}{1-\delta} \sum_{j=0}^{n-1} \left[(\mathbf{X}_j \cdot \mathbf{W})^2   \left(\sum_{i=0}^{n-1} \sqrt{z_i(1-z_i)}(\overline{\mathbf{A}}_{ij})\right)^2 \right]\\
\leq &~\frac{\delta}{1-\delta} \sum_{h=0}^{d-1} \mathbf{W}_h^2  \sum_{j=0}^{n-1} \left[\mathbf{X}_{jh}^2  \sum_{i=0}^{n-1} z_i(1-z_i) (\overline{\mathbf{A}}_{ij})^2\right].
\end{aligned}
\end{equation}
}

 \hide{
 \begin{equation}
 	\label{equ:dropout}
 	\left\{
 	\begin{aligned}
 		& Pr(\epsilon_{i,j}=0) = \delta,& \\
 		&Pr(\epsilon_{i,j} = 1) = 1 - \delta. &
 	\end{aligned}
 	\right.
 \end{equation}

\begin{equation}
\begin{aligned}
\mathcal{R}^u(W)
& = \frac{\delta}{1-\delta} \sum_{h=0}^{d-1} W_h^2  \sum_{j=0}^{n-1} \left[X_{jh}^2  \sum_{i=0}^{n-1} z_i(1-z_i) (\hat{A}^k_{ij})^2\right].  \\
\end{aligned}
\end{equation}
}

\vpara{Random propagation w.r.t supervised classification loss.}
\label{sec:suploss}
We also discuss the regularization effect of random propagation with respect to the supervised classification loss. 

With the previous assumptions, the supervised classification loss is:
$
\small
\mathcal{L}_{sup} =\sum_{i=0}^{m-1} -y_i\log(\tilde{z}_i) - (1-y_i)\log(1-\tilde{z}_i).
$ 
Note that $\mathcal{L}_{sup}$ refers to the perturbed classification loss with DropNode on the node features. 
By contrast, the original (non-perturbed) classification loss is defined as:
$
\small
\mathcal{L}_{org} =\sum_{i=0}^{m-1} -y_i\log(z_i) - (1-y_i)\log(1-z_i),
$
where $z_i = \text{sigmoid}(\overline{\mathbf{A}}_i \mathbf{X} \cdot \mathbf{W})$ is the output with the original feature matrix $\mathbf{X}$. 
Then we have the following theorem with proof in Appendix \ref{them:proof2}.

\begin{thm}
\label{thm2}
	In expectation, optimizing the perturbed classification loss $\mathcal{L}_{sup}$ is equivalent to optimize the original loss $ \mathcal{L}_{org}$ with an extra  regularization term $ \mathcal{R}(\mathbf{W})$, which has a quadratic approximation form  $  \mathcal{R}(\mathbf{W})\approx \mathcal{R}^q(\mathbf{W})= \frac{1}{2}\sum_{i=0}^{m-1} z_i(1-z_i) \textnormal{Var}_\epsilon \left(\overline{\mathbf{A}}_i \widetilde{\mathbf{X}} \cdot \mathbf{W} \right)$.
\end{thm}

This theorem suggests that DropNode brings an extra regularization loss to the optimization objective. Expanding the variance term, this extra quadratic regularization loss can be expressed as: 
\begin{equation}
\small
\label{equ:Rq}
\begin{aligned}
\mathcal{R}^q_{DN}(\mathbf{W}) 
& = \frac{1}{2}\frac{\delta}{1-\delta} \sum_{j=0}^{n-1} \left[(\mathbf{X}_j \cdot \mathbf{W})^2  \sum_{i=0}^{m-1} (\overline{\mathbf{A}}_{ij})^2\ z_i(1-z_i)\right].
\end{aligned}
\end{equation}

Different from $\mathcal{R}^c_{DN}$ in Eq. \ref{equ:Rc}, the inside summation term in Eq. \ref{equ:Rq} only incorporates the first $m$ nodes, i.e, the labeled nodes.

\hide{

\section{Theoretical Analysis for \model}
\label{sec:theory}
In this section, we perform some theoretical analyses for \model. 
Specifically, we discuss how \textit{random propagation} and \textit{consistency regularized training} can help enhance model's generalization from the view of regularization. 
The key idea of our analyses is to explore the additional effect that dropnode brings to the model optimization. We observe that \textit{dropnode is a special kind of regularization by enforcing the consistency of classification confidence between each node and its neighborhoods}.  Here we first give an analysis of dropnode only with  supervised classification loss ,i.e., without consistency regularization training (Cf. Section \ref{sec: suploss}). Then we extend it to the setting of consistency  regularized training (Cf. Section \ref{sec: conloss}). What's more, we also distinguish the differences between dropnode and dropout in theory.

\subsection{Supervised Classification Loss}
\label{sec: suploss}
Recall the definition of dropnode (Cf. Algorithm \ref{alg:dropnode}). It can be seen as injecting perturbations to the nodes by multiplying scaled Bernoulli random variable with node feature vector.  i.e., $\widetilde{X}_{i} = \frac{\epsilon_{i}}{1-\delta} \cdot X_{i}$. Where $\epsilon_{i}$ draws from $Bernoulli(1-\delta)$.

\hide{
\begin{align}
\label{equ:nodedropout2}
\left\{
\begin{aligned}
& Pr(\epsilon_i=0) = \delta,& \\
&Pr(\epsilon_i = 1) = 1 - \delta. &
\end{aligned}
\right.
\end{align}
}
For analytical simplicity, we assume the MLP used in \model\  only has a single output layer, and the task is binary classification. Then the output of \model\ in Equation \ref{equ:mlp} can be rewritten as:
\begin{equation}
\widetilde{Z} = \text{sigmoid}(\overline{A} \widetilde{X}\cdot W ).
\end{equation}
Where $\overline{A} = \frac{1}{K+1}\sum_{k=0}^K\hat{A}^k$, and $W \in \mathbb{R}^{d}$ is learnable parameters and $\widetilde{Z} \in \mathbb{R}^n$ is output. For the $i^{th}$ node, the corresponding conditional distribution is
 $$P(y_i|A,X,W)=\tilde{z}_i^{y_i}(1-\tilde{z}_i)^{1-y_i},$$
where $\tilde{z}_i = \text{sigmoid}(\overline{A}_i \widetilde{X} \cdot W)$, $y_i \in \{0,1\}$ denotes the corresponding label. Then the supervised classification loss is:
\begin{equation}
\mathcal{L}_{sup} =\sum_{i=1}^m -y_i\log(\tilde{z}_i) - (1-y_i)\log(1-\tilde{z}_i).
\end{equation}
Here we refer to the $\mathcal{L}_{sup}$ as perturbed classification loss, that is, node features are perturbed by dropnode. We also define the original (non-perturbed) classification loss:
\begin{equation}
\mathcal{L}_{org} =\sum_{i=1}^m -y_i\log(z_i) - (1-y_i)\log(1-z_i).
\end{equation}
Where $z_i = \text{sigmoid}(\overline{A}_i X \cdot W)$ is model's output with the original feature matrix $X$ as input (without applying dropnode). Then we have the following theorem:

\hide{
Dropout/dropnode can be seen as injecting perturbations to the nodes by multiplying scaled Bernoulli random variable with node feature. For dropout,  the perturbations are performed on each feature element independently, i.e., $\widetilde{X}_{ij} = \epsilon_{ij} X_{ij}$. Where $\epsilon_{ij}$ draws from:
\begin{align}
\label{equ:dropout}
\left\{
\begin{aligned}
& Pr(\epsilon_{ij}=0) = \sigma,& \\
&Pr(\epsilon_{ij} = \frac{1}{1-\sigma}) = 1 - \sigma. &
\end{aligned}
\right.
\end{align}

While dropnode is considered as adding the same noise to all features of the node, which means that we only need to sample one random variable $ \epsilon_{i}$ for each feature vector $X_i$:
\begin{align}
\label{equ:nodedropout2}
\left\{
\begin{aligned}
& Pr(\epsilon_i=0) = \sigma,& \\
&Pr(\epsilon_i = \frac{1}{1-\sigma}) = 1 - \sigma. &
\end{aligned}
\right.
\end{align}

Then the perturbed feature vector $\widetilde{X}_i = \epsilon_i \cdot X_i $. Without loss of generality, we replace the non-linearity $\sigma(\cdot)$ in MLP with linear transformation, and only consider binary classification problem, the output of \model\ in Equation~\ref{equ:model} can be rewritten as:
\begin{equation}
Z = sigmoid(\hat{A}^kXW^{(1)}W^{(2)} \cdots W^{(n)} ) = sigmoid(\hat{A}^kX \cdot W).
\end{equation}

Since $\{W^{(l)}, 1\leq l \leq n\}$ are learnable parameters, thus we can absorb them into a single vector $W \in \mathbb{R}^{d}$. 
}




\hide{
\begin{equation}
W^* = \arg \min_{W^* \in \mathbb{R}^d} \mathcal{L}^l = \arg \min_{W^* \in \mathbb{R}^d} \sum_{i=1}^m \ell_i.
\end{equation}
}


\begin{thm}
	\label{thm2}
	In expectation, optimizing the perturbed classification loss $\mathcal{L}_{sup}$ is equivalent to optimize the original loss $ \mathcal{L}_{org}$ with an extra  regularization term $ \mathcal{R}(W)$, which has a quadratic approximation form  $  \mathcal{R}(W)\approx\mathcal{R}^q(W)= \frac{1}{2}\sum_{i=1}^m z_i(1-z_i) \text{Var}_\epsilon \left[\overline{A}_i \widetilde{X}\cdot W \right]$.
\end{thm}
The proof details of this theorem can be find in Appendix \ref{them:proof1}. This theorem shows that dropnode brings an extra regularization loss to the optimization objective. We will explain the concrete effect of the regularization in the following.
For dropnode with the drop rate $\delta$, we can easily check that:
\begin{equation}
\label{equ:dropnodevar}
Var_\epsilon(\overline{A}_i \widetilde{X}\cdot W) = \frac{\delta}{1-\delta} \sum_{j=1}^n(X_j \cdot W)^2 (\overline{A}_{i,j})^2. 
\end{equation}
So the quadratic dropnode regularization term is:
\begin{equation}
\label{equ:Rq}
\begin{aligned}
\mathcal{R}^q(W) & = \frac{1}{2} \sum_{i=0}^{m-1} z_i(1-z_i) \frac{\delta}{1-\delta} \sum_{j=0}^{n-1}(X_j \cdot W)^2 (\overline{A}_{i,j})^2  \\
& = \frac{1}{2}\frac{\delta}{1-\delta} \sum_{j=0}^{n-1} \left[(X_j \cdot W)^2  \sum_{i=0}^{m-1} (\overline{A}_{i,j})^2\ z_i(1-z_i)\right].
\end{aligned}
\end{equation}
\hide{
\begin{equation}
\begin{aligned}
\mathcal{R}^q(W)
& = \frac{1}{2}\frac{\delta}{1-\delta} \sum_{j=0}^{n-1} \left[(X_j \cdot W)^2  \sum_{i=0}^{m-1} \hat{y}_i(1-\hat{y}_i) (\hat{A}^p_{ij})^2\right]. \\
\end{aligned}
\end{equation}
}

Note that $z_i(1-z_i)$ arrives the maximum at $z_i=0.5$ and minimum at $z_i=0$ or $1$, hence  
we conclude that $z_i(1-z_i)$ indicates the classification uncertainty score of the $i^{th}$ node. Thus $\sum_{i=1}^m  (\overline{A}_{i,j})^2z_i(1-z_i)$  can be considered as the weighted average classification uncertainty score of the $j^{th}$ node's multi-hop neighborhoods. The average weights are square values of $\overline{A}$'s elements, which is related to graph structure. 
Note that the summation term only incorporates the first $m$ nodes, suggesting only labeled nodes are taken into consideration. 
On the other hand, $(X_j \cdot W)^2$, as the square of the input of activation function sigmoid, indicates the classification confidence of the $j^{th}$ node. 
In optimization, in order for a node to earn a higher classification confidence $(X_j \cdot W)^2$, it is required that the node's neighborhoods have lower classification uncertainty scores. Thus, \textit{the dropnode regularization with supervised classification loss can enforce the consistency between classification confidence between a node and its labeled multi-hop neighborhoods}.


\hide{
\vpara{Dropout Regularization.} For dropout, the corresponding variance term is:
\begin{equation}
Var_\epsilon(\hat{A}^k_i \tilde{X}\cdot W) = \frac{\delta}{1-\delta} \sum_{j=1}^n \sum_{k=1}^d X_{jk}^2 W_k^2 (\hat{A}^k_{ij})^2.
\end{equation}
Then the corresponding quadratic regularization term is:
\begin{equation}
\begin{aligned}
\mathcal{R}^q(W)
& = \frac{1}{2}\frac{\delta}{1-\delta} \sum_{h=1}^d W_h^2  \sum_{j=1}^n \left[X_{jh}^2  \sum_{i=1}^m z_i(1-z_i) (\hat{A}^k_{ij})^2\right].  \\
\end{aligned}
\end{equation}

Similar to the dropnode case, the classification uncertainty $z_i(1-z_i)$ of labeled nodes are also propagated in the graph and all nodes update their uncertainty with the smoothness of the graph. However, \textit{differently from dropnode, dropout plays an adaptive l2 regularization role; and differently from the general dropout, dropout on the graph takes unlabeled data, prediction confidence on labeled data, and the smoothness of the graph into account.}

By applying Cauchy-Buniakowsky-Schwarz Inequality, we have:
\begin{equation}
\begin{aligned}
 &~\frac{1}{2}\frac{\delta}{1-\delta} \sum_{j=1}^n \left[(X_j \cdot W)^2   \left(\sum_{i=0}^m \sqrt{z_i(1-z_i)}(\hat{A}^k_{ij})\right)^2 \right].\\
 \leq &~\frac{1}{2}\frac{\delta}{1-\delta} \sum_{h=1}^d W_h^2  \sum_{j=1}^n \left[X_{jh}^2  \sum_{i=1}^m z_i(1-z_i) (\hat{A}^k_{ij})^2\right].
\end{aligned}
\end{equation}

Hence, the regularization term of dropout is the upper bound of that of dropnode.
}
\subsection{Consistency Regularization Loss}
\label{sec: conloss}

Here we discuss the regularization of dropnode with the consistency regularization loss. We follow the assumption settings expressed in Section \ref{sec: suploss}. As for consistency regularization loss, without loss of generality, we only consider the case of $S=2$ (Cf. Equation \ref{equ:2d}). And we adopt squared L2 loss as the distance function $\mathcal{D}(\cdot,\cdot)$. Then the loss can be rewritten as:
   \begin{equation}
   \mathcal{L}_{con} =\sum_{i=1}^n \left(\tilde{z}_i^{(1)}-\tilde{z}_i^{(2)}\right)^2,
   \end{equation}
   where $\tilde{z}^{(1)}_i$ and $\tilde{z}^{(2)}_i$ are model's outputs on node $i$ with different augmentations 
   With these assumptions, we can prove that:
\begin{thm}
\label{thm2}
In expectation, optimizing the unsupervised consistency loss $\mathcal{L}_{con}$ is approximate to optimize a regularization term: $\mathbb{E}_\epsilon\left[\mathcal{L}_{con}\right] \approx \mathcal{R}^c(W) = 2\sum_{i=1}^nz_i(1-z_i) \text{Var}_\epsilon \left[\overline{A}_i \widetilde{X}\cdot W \right]$.
\end{thm}

Similar to the discuss about $\mathcal{R}^q$, we show the concrete expression of $\mathcal{R}^c$ for dropnode. Based on Equation \ref{equ:dropnodevar}, we have:

\begin{equation}
\begin{aligned}
\mathcal{R}^c(W) = \frac{\delta}{1-\delta} \sum_{j=1}^n \left[(X_j \cdot W)^2  \sum_{i=1}^n (\overline{A}_{i,j})^2 z_i(1-z_i) \right].
\end{aligned}
\end{equation}

Compared with  $\mathcal{R}^q$ (Cf. Equation \ref{equ:Rq}), the main difference of $\mathcal{R}^c$ is that the averaged uncertainty score $\sum_{i=1}^n z_i(1-z_i) (\overline{A}_{i,j})^2 $ comes from all the $n$ nodes, indicating both labeled and unlabeled neighborhoods are incorporated into the regularization. Hence, \textit{the dropnode regularization with consistency regularization loss can enforce the consistency between classification confidence between a node and its all multi-hop neighborhoods}.


\subsection{Dropout Regularization}
 Applying dropout on graph features, the perturbed feature matrix $\widetilde{X}$ is obtained via randomly set elements of $X$ to 0, i.e., $\widetilde{X}_{i,j} = \frac{\epsilon'_{i,j}}{1-\delta} X_{i,j}$, where $\epsilon'_{i,j}$ draws from $Bernoulli(1-\delta)$. As for dropout perturbation We can easily check Theorem \ref{thm1} and Theorem \ref{thm2} still holds in this case, and we have:
 
\begin{equation}
Var_{\epsilon'}(\overline{A}_i \widetilde{X}\cdot W) = \frac{\delta}{1-\delta} \sum_{j=1}^n \sum_{k=1}^d X_{j,k}^2 W_k^2 (\overline{A}_{i,j})^2.
\end{equation}
Without loss of generalization, we focus on supervised classification loss here. The corresponding regularization term of dropout is:
\begin{equation}
\begin{aligned}
\label{equ:dropoutreu}
\mathcal{R}^q(W)
& = \frac{1}{2}\frac{\delta}{1-\delta} \sum_{h=1}^d W_h^2  \sum_{j=1}^n \left[X_{j,h}^2  \sum_{i=1}^m z_i(1-z_i) (\overline{A}_{i,j})^2\right].  \\
\end{aligned}
\end{equation}

Similar to the dropnode case, the regularization term also includes the  classification uncertainty  $z_i(1-z_i)$ of labeled neighborhoods. However, \textit{different from dropnode regularization, we can observe that dropout is actually an adaptive $L_2$ regularization for $W$, where the regularization coefficient is associated with unlabeled data, classification uncertainty on labeled data and graph structure}.

By applying Cauchy-Buniakowsky-Schwarz Inequality to Equation \ref{equ:dropoutreu}, we have:
\begin{equation}
\begin{aligned}
&~\frac{1}{2}\frac{\delta}{1-\delta} \sum_{j=1}^n \left[(X_j \cdot W)^2   \left(\sum_{i=0}^m \sqrt{z_i(1-z_i)}(\overline{A}_{i,j})\right)^2 \right].\\
\leq &~\frac{1}{2}\frac{\delta}{1-\delta} \sum_{h=1}^d W_h^2  \sum_{j=1}^n \left[X_{j,h}^2  \sum_{i=1}^m z_i(1-z_i) (\overline{A}_{i,j})^2\right].
\end{aligned}
\end{equation}
Hence, the regularization term of dropout is the upper bound of that of dropnode. By minimizing this term, dropout can be regarded as an approximation of dropnode.
 
 \hide{
 \begin{equation}
 	\label{equ:dropout}
 	\left\{
 	\begin{aligned}
 		& Pr(\epsilon_{i,j}=0) = \delta,& \\
 		&Pr(\epsilon_{i,j} = 1) = 1 - \delta. &
 	\end{aligned}
 	\right.
 \end{equation}

\begin{equation}
\begin{aligned}
\mathcal{R}^u(W)
& = \frac{\delta}{1-\delta} \sum_{h=1}^d W_h^2  \sum_{j=1}^n \left[X_{jh}^2  \sum_{i=1}^n z_i(1-z_i) (\hat{A}^k_{ij})^2\right].  \\
\end{aligned}
\end{equation}
}

}
\section{Experiments}
\label{sec:exp}


\subsection{Experimental Setup}

We follow exactly the same experimental procedure---such as features and data splits---as the standard GNN settings on semi-supervised graph learning~\cite{yang2016revisiting, kipf2016semi, Velickovic:17GAT}. The setup and reproducibility details are covered in Appendix \ref{sec:reproduce}.

\vpara{Datasets.}We conduct experiments on three benchmark graphs~\cite{yang2016revisiting, kipf2016semi, Velickovic:17GAT}---Cora, Citeseer, and Pubmed---and also report results on six publicly available and  large datasets 
 in Appendix \ref{exp:large_data}.  

\vpara{Baselines.}
By default, we use DropNode as the perturbation method in \model\ and compare it with 14 GNN baselines representative of three different categories, as well as its variants: 
\begin{itemize}
	\item 
	{Eight} {graph convolutions:} GCN~\cite{kipf2016semi}, GAT~\cite{Velickovic:17GAT},
	APPNP~\cite{klicpera2018predict},
	Graph U-Net~\cite{gao2019graph}, 
	SGC~\cite{wu2019simplifying},
	MixHop~\cite{abu2019mixhop}, 
	GMNN~\cite{qu2019gmnn} and GrpahNAS~\cite{gao2019graphnas}. 
	
	\item 
	{Two} {sampling based GNNs:}
	GraphSAGE~\cite{hamilton2017inductive} and FastGCN~\cite{FastGCN}.
	
	\item 
	{Four} {regularization based GNNs:} VBAT~\cite{deng2019batch}, $\text{G}^3$NN~\cite{ma2019flexible},  GraphMix~\cite{verma2019graphmix} and Dropedge~\cite{YuDropedge}. We report the results of these methods with GCN as the backbone model.
	
	\item 
	{Four} {\model\ variants:} \model\_dropout, \model\_DropEdge, \model\_GCN and \model\_GAT. In \model\_dropout and \model\_DropEdge, we use dropout and DropEdge as the perturbation method respectively, instead of DropNode. In \model\_GCN and \model\_GAT, we replace MLP with more complex models, i.e., GCN and GAT, respectively. 
\end{itemize}


\hide{
\begin{itemize}
    \item \textbf{GCN}~\cite{kipf2016semi}  uses the propagation rule described in Eq.~\ref{equ:gcn_layer}. 
    \item \textbf{GAT}~\cite{Velickovic:17GAT} propagates information based on  self-attention.
    \item \textbf{Graph U-Net}~\cite{gao2019graph} proposes the graph pooling operations.
    \item \textbf{MixHop}~\cite{abu2019mixhop} employs the mixed-order propagation.
    \item \textbf{GMNN}~\cite{qu2019gmnn} combines GNNs with probabilistic graphical models.
    \item \textbf{GraphNAS}~\cite{gao2019graphnas} automatically generates GNN architectures using reinforcement learning.
    \item \textbf{VBAT}~\cite{deng2019batch} applies virtual adversarial training~\cite{miyato2015distributional} into GCNs.
    \item \textbf{G$^3$NN}~\cite{ma2019flexible} 
    regularizes GNNs with an extra link prediction task.
    \item \textbf{GraphMix}~\cite{verma2019graphmix} adopts MixUp~\cite{zhang2017mixup} for regularizing GNNs.
    \item \textbf{DropEdge}~\cite{YuDropedge} randomly drops some edges in GNNs training.
    \item \textbf{GraphSAGE}~\cite{hamilton2017inductive} proposes node-wise neighborhoods sampling.
    \item \textbf{FastGCN}~\cite{FastGCN} using importance sampling for fast GCNs training.
    \item \textbf{\model\_GCN}. Note that the prediction module in \model\ is the simple MLP model. In \model\_GCN, we replace MLP with GCN. 
    \item \textbf{\model\_GAT} replaces the MLP component in \model\ with GAT.
    \item \textbf{\model\_dropout} substitutes our DropNode technique with the dropout operation in \model's random propagation.
\end{itemize}
}




\subsection{Overall Results}
\label{sec:overall}
Table~\ref{tab:overall} summarizes the prediction accuracies of node classification. 
Following the community convention~\cite{kipf2016semi,Velickovic:17GAT,qu2019gmnn}, the results of baselines are taken from the original works~\cite{kipf2016semi,Velickovic:17GAT,gao2019graph,abu2019mixhop,gao2019graphnas,deng2019batch,ma2019flexible,verma2019graphmix,YuDropedge,klicpera2018predict}. 
The results of \model\  are averaged over \textbf{100} runs with random weight initializations. 

From the top part of Table~\ref{tab:overall}, we can observe that \model\ consistently achieves large-margin outperformance over all baselines 
across all datasets. Note that the improvements of \model\ over other baselines are all {statistically significant} (p-value $\ll$ 0.01 by a t-test).
Specifically, \model\ improves upon GCN by a margin of 3.9\%, 5.1\%, and 3.7\% (absolute differences) on Cora, Citeseer, and Pubmed, while the margins improved by GAT upon GCN were 1.5\%, 2.2\%, and 0\%, respectively.
When compared to the very recent regularization based model---DropEdge, the proposed model achieves 2.6\%, 3.1\%, and 3.1\% improvements, while DropEdge's improvements over GCN were only 1.3\%, 2.0\%, and 0.6\%, respectively. To better examine the effectiveness of \model\ in semi-supervised setting,  we further evaluate \model\ under different label rates in Appendix \ref{sec:diff_label}.

We observe \model\_dropout and \model\_DropEdge also outperform most of baselines, though still lower than \model. This indicates DropNode is the best way to generate graph data augmentations in random propagation. 
Detailed experiments to compare DropNode and dropout under different propagation steps $K$ are shown in Appendix \ref{sec:dropnode_vs_dropout}. 

We interpret the performance of \model\_GAT, \model\_GCN  from two perspectives. 
First, both \model\_GAT and \model\_GCN outperform the original GCN and GAT models, demonstrating the positive effects of the proposed random propagation and consistency regularized training methods. 
Second, both of them are inferior to \model\ with the simple MLP model, suggesting GCN and GAT are relatively easier to over-smooth than MLP. More analyses can be found in Appendix \ref{sec:oversmoothing_grand}.

\begin{minipage}[t]{\linewidth}
\begin{minipage}[b]{0.5\linewidth}
\centering
\scriptsize
\setlength{\tabcolsep}{3mm}
\renewcommand{\arraystretch}{1.1}
\begin{tabular}{c|cccc}
		\toprule\toprule
		Method &Cora & Citeseer & Pubmed \\
		\midrule
	    GCN~\cite{kipf2016semi}   & 81.5 & 70.3 & 79.0  \\

		GAT~\cite{Velickovic:17GAT} & 83.0$\pm$0.7 & 72.5$\pm$0.7 & 79.0$\pm$0.3 \\
		APPNP~\cite{klicpera2018predict} & 83.8$\pm$0.3 & 71.6$\pm$ 0.5 & 79.7 $\pm$ 0.3 \\ 
		Graph U-Net~\cite{gao2019graph} & 84.4$\pm$0.6 & 73.2$\pm$0.5 & 79.6$\pm$0.2 \\
		SGC~\cite{wu2019simplifying} & 81.0 $\pm$0.0 & 71.9 $\pm$ 0.1 &78.9 $\pm$ 0.0 \\
		MixHop~\cite{abu2019mixhop} & 81.9$\pm$ 0.4 & 71.4$\pm$0.8 & 80.8$\pm$0.6 \\
		GMNN~\cite{qu2019gmnn} & 83.7  & 72.9 & 81.8  \\
		GraphNAS~\cite{gao2019graphnas} & 84.2$\pm$1.0 & 73.1$\pm$0.9 & 79.6$\pm$0.4 \\
		 \midrule
		 
	 	GraphSAGE~\cite{hamilton2017inductive}& 78.9$\pm$0.8 & 67.4$\pm$0.7 & 77.8$\pm$0.6 \\
		FastGCN~\cite{FastGCN} & 81.4$\pm$0.5 & 68.8$\pm$0.9 & 77.6$\pm$0.5  \\
        \midrule
        
		VBAT~\cite{deng2019batch} & 83.6$\pm$0.5 & 74.0$\pm$0.6 &79.9$\pm$0.4 \\
		$\text{G}^3$NN~\cite{ma2019flexible}  &82.5$\pm$0.2 &74.4$\pm$0.3 &77.9 $\pm$0.4  \\
		GraphMix~\cite{verma2019graphmix} &83.9$\pm$0.6 & 74.5$\pm$0.6& 81.0$\pm$0.6 \\
		DropEdge~\cite{YuDropedge} &82.8 & 72.3 & 79.6 \\
        \midrule

		\multicolumn{1}{c|}{\model\_{dropout}} & 84.9$\pm$0.4  & 75.0$\pm$0.3 & 81.7$\pm$1.0 \\
		\multicolumn{1}{c|}{\model\_{DropEdge}} & 84.5$\pm$0.3  & 74.4$\pm$0.4 & 80.9$\pm$0.9 \\

		\multicolumn{1}{c|}{\model\_{GCN}}   & 84.5$\pm$0.3 & 74.2$\pm$0.3 & 80.0$\pm$0.3  \\
		 \multicolumn{1}{c|}{\model\_{GAT}}     & 84.3$\pm$0.4 & 73.2$\pm$ 0.4 & 79.2$\pm$0.6\\
		\multicolumn{1}{c|}{\model}  & \textbf{85.4$\pm$0.4} & \textbf{75.4$\pm$0.4} & \textbf{82.7$\pm$0.6}  \\		
		
		\midrule\midrule
		\multicolumn{1}{l|}{\quad w/o CR} & 84.4$\pm$0.5  & 73.1$\pm$0.6 & 80.9$\pm$0.8 \\
		\multicolumn{1}{l|}{\quad w/o mDN} &84.7$\pm$0.4 & 74.8$\pm$0.4 & 81.0$\pm$1.1  \\
		\multicolumn{1}{l|}{\quad w/o sharpening} & 84.6$\pm$0.4  & 72.2$\pm$0.6 & 81.6$\pm$0.8 \\
		\multicolumn{1}{l|}{\quad w/o CR \& DN} & 83.2$\pm$0.5     & 70.3$\pm$0.6    & 78.5$\pm$1.4   \\ 
		\bottomrule \bottomrule
	\end{tabular}
	\captionof{table}{Overall classification accuracy (\%).}
 \label{tab:overall}
\end{minipage}
\hfill
\hspace{0.2in}
\begin{minipage}[b]{0.5\linewidth}
	\includegraphics[width=.90\linewidth]{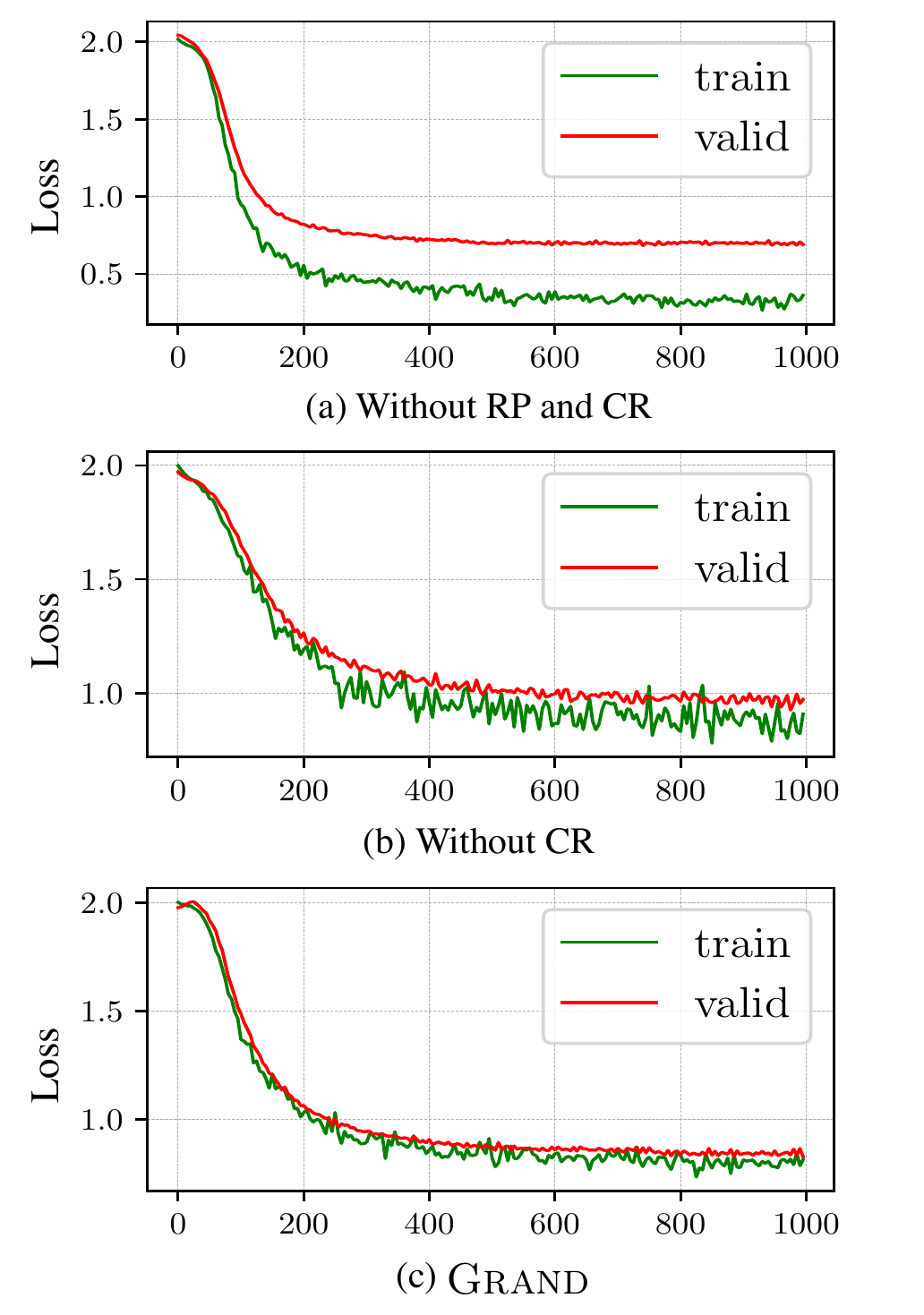}
	\captionof{figure}{\footnotesize{Generalization on Cora  ($x$: epoch).}}
	      \label{fig:loss}

\end{minipage}

\end{minipage}


\subsection{Ablation Study}
\label{sec:ablation}
We conduct an ablation study to examine the contributions of different components in \model. 
\begin{itemize}
\small
	\item \textbf{Without consistency regularization (CR):} We only use the  supervised classification loss, i.e., $\lambda=0$.
	\item \textbf{Without multiple DropNode (mDN):} Do DropNode once at each epoch, i.e., $S=1$, meaning that CR only enforces the model to give low-entropy predictions for unlabeled nodes. 
	\item \textbf{Without sharpening:} The sharpening trick in Eq.~\ref{equ:sharpen} is not used in getting the distribution center, i.e., $T=1$. 
	\item \textbf{Without CR and DropNode (CR \& DN):} Remove DropNode (as a result, the CR loss is also removed), i.e., $\delta=0, \lambda=0$. In this way, \model\ becomes the combination of deterministic propagation and MLP. 
\end{itemize}
In Table \ref{tab:overall}, the bottom part summarizes the results of the ablation study, from which we have two observations. 
First, all \model\ variants with some components removed witness clear performance drops when comparing to the full model, suggesting that each of the designed components contributes to the success of \model. 
Second, \model\ without consistency regularization outperforms almost all eight non-regularization based GCNs and DropEdge in all three datasets, demonstrating the significance of the proposed random propagation technique for semi-supervised graph learning.


\hide{
 
\begin{table}[h]
\footnotesize
	\caption{Ablation study results (\%).}
	\label{tab:ablation}
		\begin{tabular}{lccc}
		\toprule
		Model &Cora & Citeseer & Pubmed\\
		\midrule
		\model  & \textbf{85.4$\pm$0.4} & \textbf{75.4$\pm$0.4} & \textbf{82.7$\pm$0.6} \\
		\midrule
		\quad without consistency regularization  & 84.4 $\pm$0.5  & 73.1 $\pm$0.6 & 80.9 $\pm$0.8 \\ 
		 \quad without multiple DropNode &84.7 $\pm$0.4 & 74.8$\pm$0.4 & 81.0$\pm$1.1  \\
		\quad without sharpening & 84.6 $\pm$0.4  & 72.2$\pm$0.6 & 81.6 $\pm$ 0.8 \\
		\quad without consistency regularization and DropNode & 83.2 $\pm$ 0.5     & 70.3 $\pm$ 0.6    & 78.5$\pm$ 1.4   \\ 
		\bottomrule
	\end{tabular}
	\vspace{-0.1in}
\end{table}
}

\subsection{Generalization Analysis}

\label{sec:generalize}
We examine how the proposed techniques---random propagation and consistency regularization---improve the model's generalization capacity. 
To achieve this, we analyze the model's cross-entropy losses on both training and validation sets on Cora. 
A small gap between the two losses indicates a model with good generalization. Figure \ref{fig:loss} reports the results for \model\ and its two variants. 
We can observe the significant gap between the validation and training losses when without both consistency regularization (CR) and random propagation (RP), indicating an obvious overfitting issue. 
When applying only the random propagation (without CR), the gap becomes much smaller.
Finally, when further adding the CR loss to make it the full \model\ model, the validation loss becomes much closer to the training loss and both of them are also more stable. 
This observation demonstrates both the random propagation and consistency regularization can significantly improve \model's generalization capability.

\subsection{Robustness Analysis}
\label{sec:robust}
\hide{
We study the robustness of the proposed \model\ model. Specifically, we utilize the following adversarial attack methods to generate perturbed graphs, and then examine the model's classification accuracies on them. 
\begin{itemize}
\small
	\item \textbf{Random Attack.} Perturbing the structure by randomly adding fake edges.
\item \textbf{Metattack~\cite{zugner2019adversarial}.} Attacking the graph structure by removing or adding edges based on meta learning.
\end{itemize}
}

We study the robustness of \model\ by generating perturbed graphs with two adversarial attack methods: Random Attack perturbs the graph structure by randomly adding fake edges, and Metattack~\cite{zugner2019adversarial} attacks the graph by removing or adding edges based on meta learning.

\hide{
\begin{wrapfigure}{r}{0.6\textwidth}
\vspace{-0.3cm}
\footnotesize
	\includegraphics[width=\linewidth]{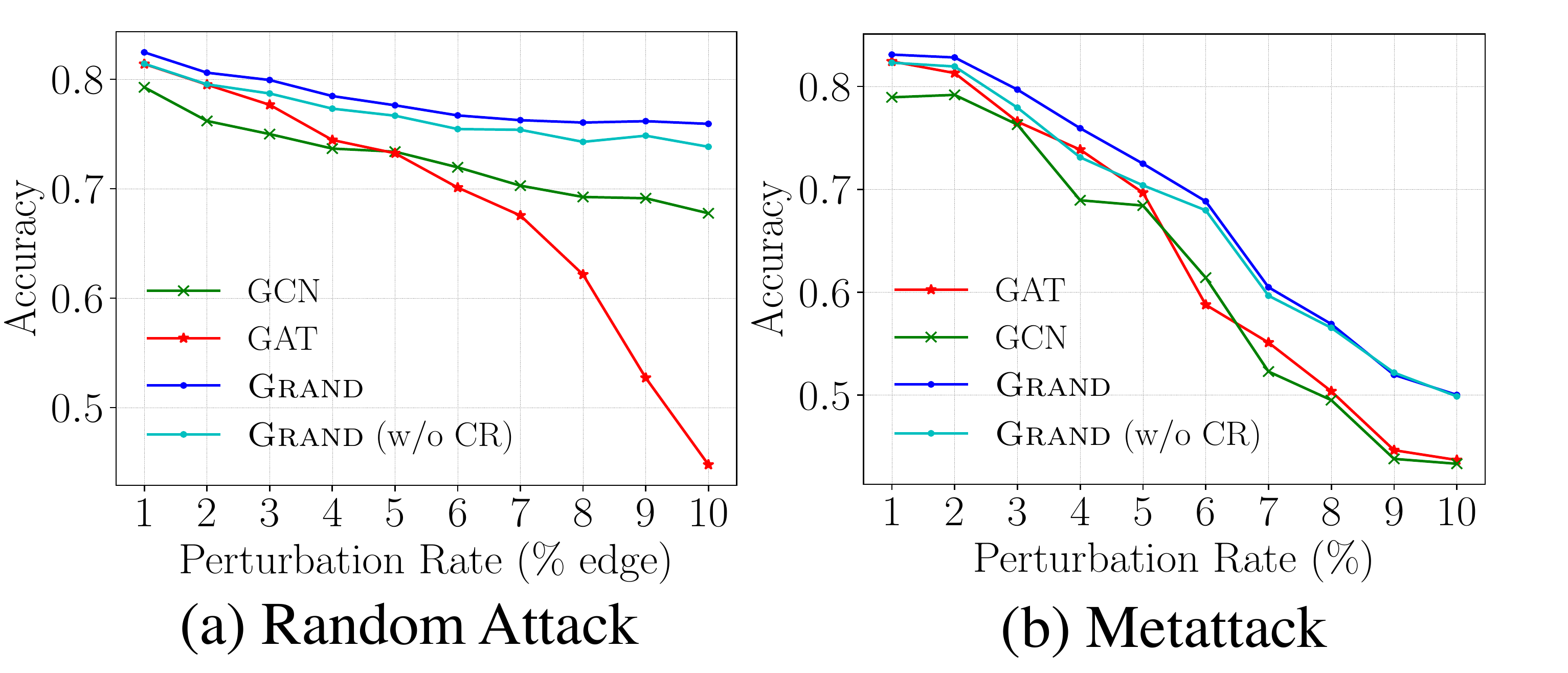}
\captionof{figure}{Robustness Analysis on Cora.}
 \label{fig:robust}
 \vspace{-0.3cm}
\end{wrapfigure}
}

Figure \ref{fig:robust} presents the classification accuracies of different methods with respect to different perturbation rates on the Cora dataset. 
We observe that \model\ consistently outperforms GCN and GAT across all perturbation rates on both attacks. 
When adding 10\% new random edges into Cora, we observe only a 7\% drop in classification accuracy for \model, while 12\% for GCN and 37\% for GAT. Under Metattack, the gap between \model\ and GCN/GAT also enlarges with the increase of the perturbation rate. 
This study suggests the robustness advantage of the \model\ model (with or without) consistency regularization over GCN and GAT. 

\begin{minipage}[b]{0.52\linewidth}
\footnotesize
	\includegraphics[width=\linewidth]{pics/robust_nips20.pdf}
\captionof{figure}{Robustness Analysis on Cora.}
 \label{fig:robust}
\end{minipage}
\hspace{-0.3cm}
\begin{minipage}[b]{0.52\linewidth}
\footnotesize
	\includegraphics[width=\linewidth]{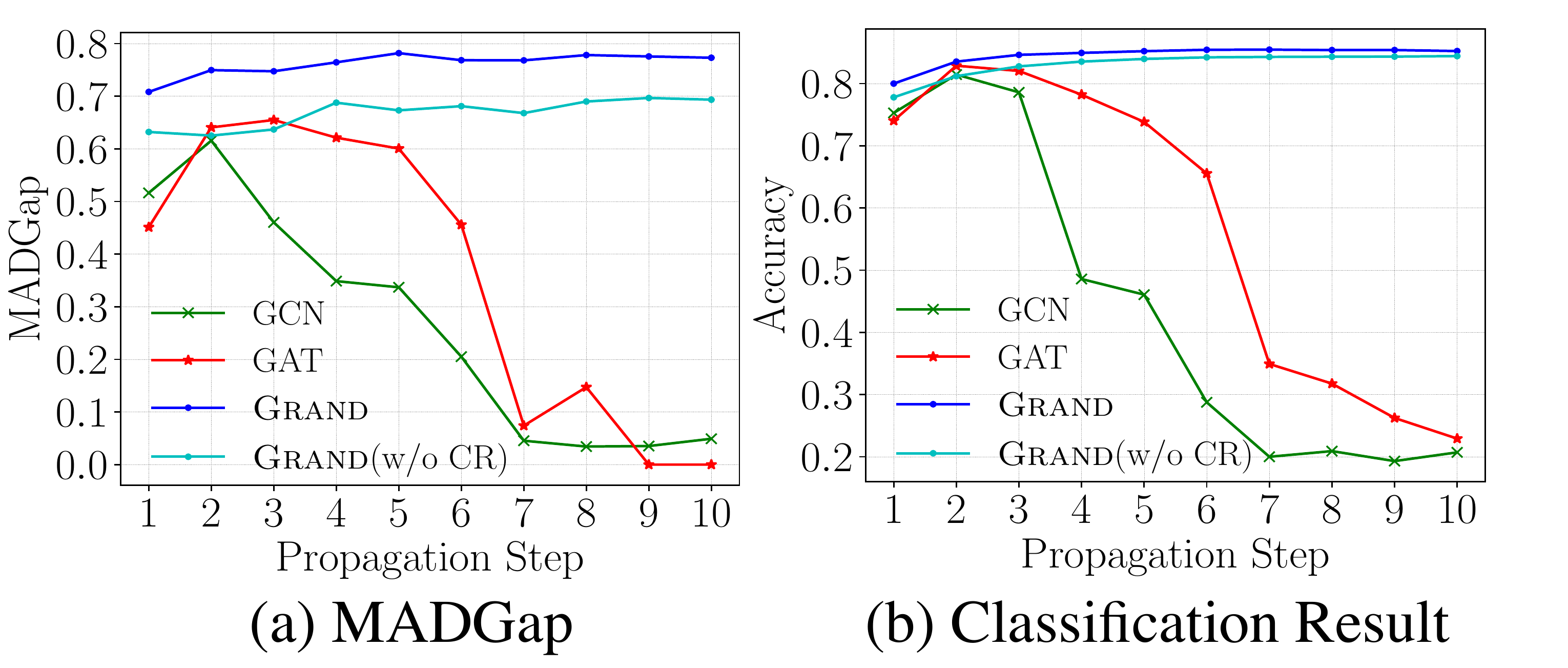}
	\captionof{figure}{{ Over-Smoothing on Cora}}
	      \label{fig:oversmooth}
\end{minipage}

\hide{

\begin{figure}[t]
	\centering
	\mbox
	{
		\begin{subfigure}[Robustness analysis]{
				\centering
				\includegraphics[width = 0.45 \linewidth]{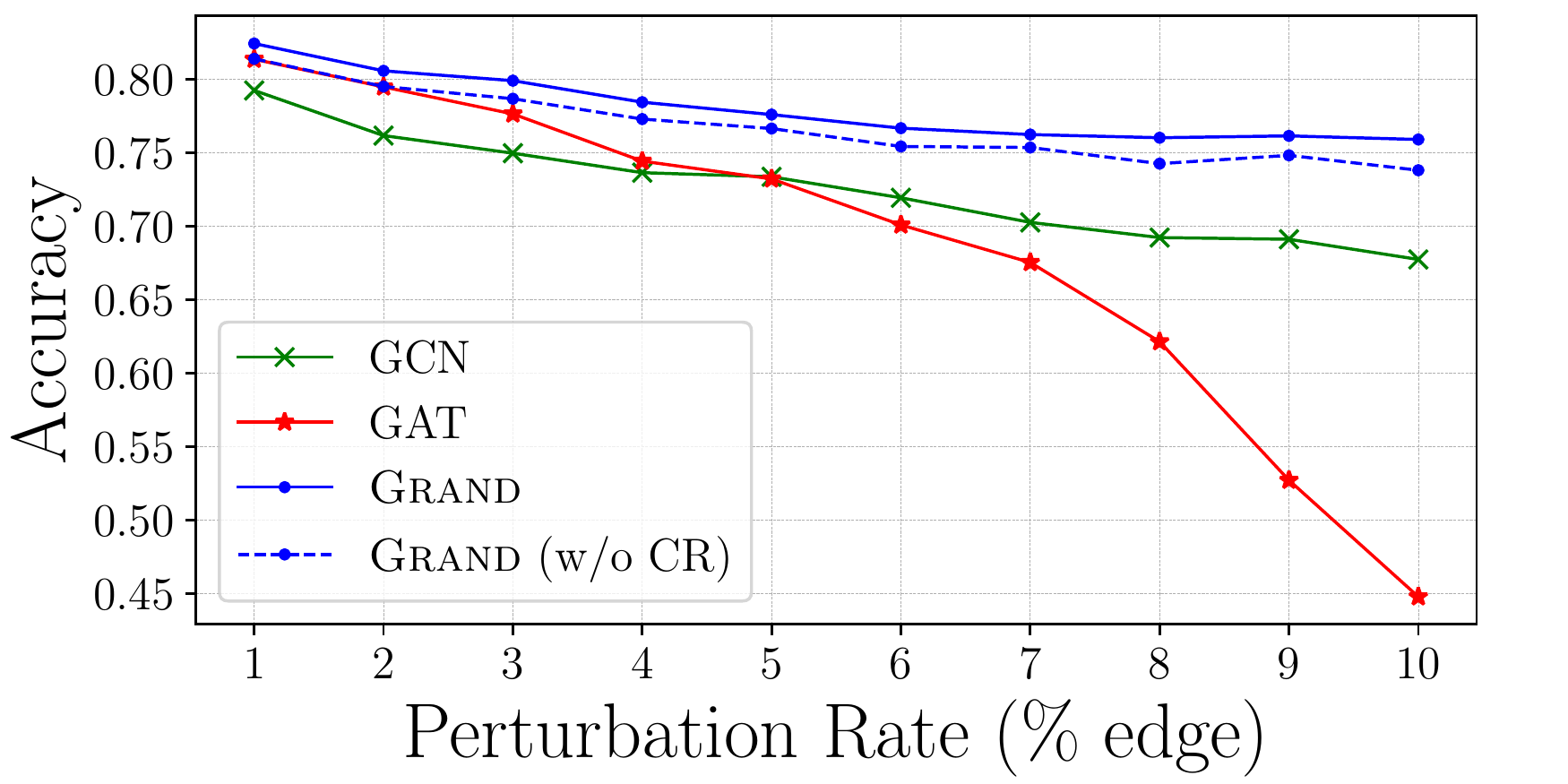}
			}
		\end{subfigure}
		\begin{subfigure}[Mitigation of over-smoothing]{
				\centering
				\includegraphics[width = 0.45 \linewidth]{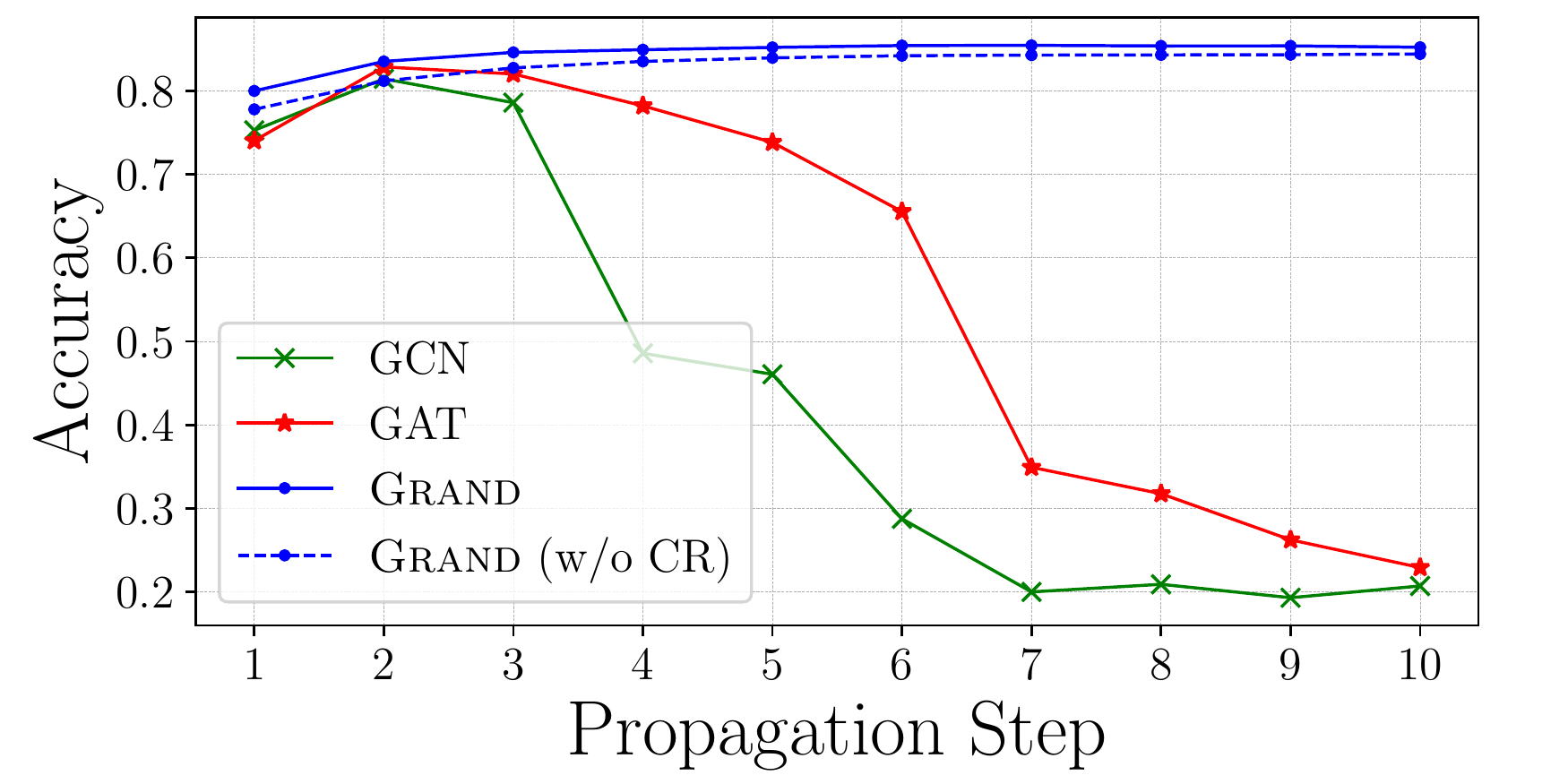}
			}
		\end{subfigure}
	}
	\caption{\model's robustness and mitigation of over-smoothing demonstrated.}
	\label{fig:rob_over}
\end{figure}
}


\hide{
\begin{wrapfigure}{r}{0.6\textwidth}
\vspace{-0.3cm}
\footnotesize
	\includegraphics[width=\linewidth]{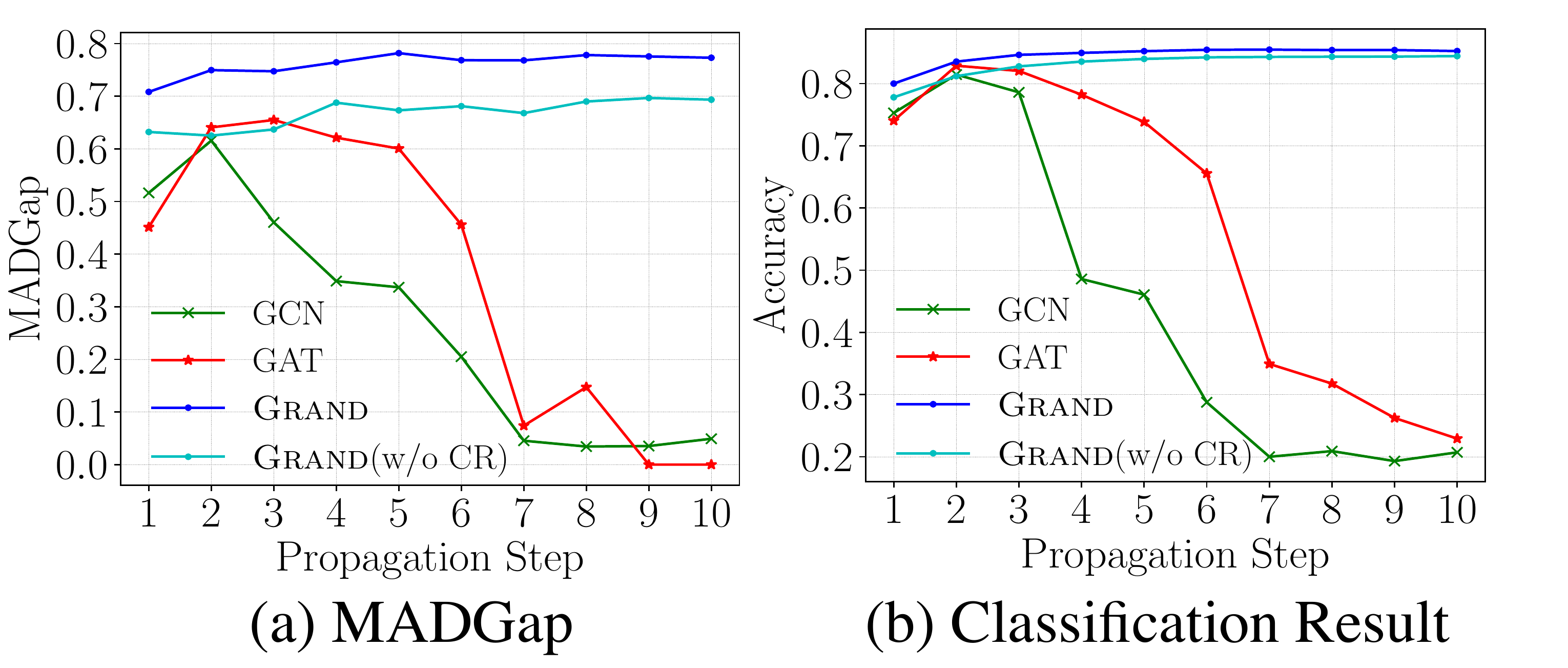}
	\captionof{figure}{{ Over-Smoothing on Cora}}
	      \label{fig:oversmooth}
 \vspace{-0.3cm}
\end{wrapfigure}
}

\subsection{Over-Smoothing Analysis}
\label{sec:oversmoothing}
Many GNNs face the over-smoothing issue---{nodes with different labels  become indistinguishable}---when enlarging the feature propagation steps~\cite{li2018deeper,chen2019measuring}.
We study how vulnerable \model\ is to this issue by using MADGap~\cite{chen2019measuring}, a measure of the over-smoothness of node representations. 
A smaller MADGap value indicates the more indistinguishable node representations and thus a more severe over-smoothing issue. 

Figure \ref{fig:oversmooth} shows both the MADGap values of the last layer's representations and classification results w.r.t. different propagation steps.    
In \model, the propagation step is controlled by the hyperparameter $K$, while for GCN and GAT, it is adjusted by stacking different hidden layers. 
The plots suggest that as the propagation step increases, both metrics of GCN and GAT decrease dramatically---MADGap drops from $\sim$0.5 to 0 and accuracy drops from 0.75 to 0.2---due to the over-smoothing issue. 
However, \model\ behaves completely different, i.e., both the performance and MADGap benefit from more propagation steps. 
This indicates that \model\ is much more powerful to relieve over-smoothing, when existing representative GNNs are very vulnerable to it.

\hide{
\vpara{More Discussions on DropEdge.}
DropEdge~\cite{YuDropedge} is proposed as a regularization method for mitigate over-smoothing of GNNs. 
Table ~\ref{tab:classification} shows that directly applying DropEdge to  GNNs cannot get comparable performance to not only \model\ but also \model\ without consistency regularization. 

Though DropEdge is originally not designed for augmenting graph data
for consistency regularization, it can actually be used as an alternative perturbation method in \model's random propagation. 
During the propagation $\overline{\mathbf{X}} = \overline{\mathbf{A}} \mathbf{X}$, we could observe that dropping $\mathbf{X}$'s $i^{th}$-row (DropNode) is equivalent to dropping the $i^{th}$ column from $\overline{\mathbf{A}}$ (DropEdge). 
This indicates that DropNode on $\mathbf{X}$ can be seen as a special form of DropEdge on $\overline{\mathbf{A}}$, i.e., dropping each node's outcome edges. 
However, in practice, DropEdge is not adopted in \model, as calculating the exact form of $\overline{\mathbf{A}} =  \sum_{k=0}^K\frac{1}{K+1}\hat{\mathbf{A}}^k$ is computationally expensive. 
}



\hide{On the other hand, when applying DropEdge on $\widetilde{\mathbf{A}} = 
\text{DropEdge}(\overline{\mathbf{A}}, \delta)$, the value of the variance term $\text{Var}_{\epsilon}(\widetilde{\mathbf{A}}_i\mathbf{X} \cdot \mathbf{W}) = \frac{\delta}{1-\delta} \sum_{j=0}^{n-1}(\mathbf{X}_j \cdot \mathbf{W})^2 (\overline{\mathbf{A}}_{ij})^2$, which shares the same value as DropNode's variance term. This suggesting that DropEdge shares the same regularization effect with DropNode. Consequently, the DropNode operation used in \model\ can also been totally replaced with DropEdge on $\overline{\mathbf{A}}$.}

\hide{

\section{Experiments}
\label{sec:exp}

In this section, we evaluate the proposed \model\ model on semi-supervised graph learning benchmark datasets from various perspectives, including performance demonstration, ablation study, and the analyses of robustness, over-smoothing, and generalization. 

\subsection{Experimental Setup}

\vpara{Datasets.}
We conduct experiments on three benchmark graphs~\cite{yang2016revisiting, kipf2016semi, Velickovic:17GAT}---Cora, Citeseer, and Pubmed---and also report results on seven relatively new and large datasets---Cora-Full, \yd{add ref}, Coauthor CS, Coauthor Physics, Amazon Computers, Amazon Photo, and AMiner CS---in Appendix \yd{xx.}.  
Tables~\ref{tab:data} and \ref{tab:large_data} summarize the dataset statistics, respectively. 
\textit{We use exactly the same experimental settings---such as features and data splits---on the three benchmark datasets as literature on semi-supervised graph mining~\cite{yang2016revisiting, kipf2016semi, Velickovic:17GAT}.  }

\begin{table}
\caption{Dataset statistics.}
\label{tab:data}
\begin{tabular}{cccc}
	\toprule
	~&Cora & Citeseer & Pubmed\\
	\midrule
	Nodes  & 2,708 & 3,327 & 19,717 \\
	Edges  & 5,429 & 4,732 & 44,338 \\
	Classes & 7     & 6     & 3       \\
	Features& 1,433 & 3,703 & 500   \\
	Training Nodes & 140 & 120 & 60 \\
	Validation Nodes & 500 & 500 & 500 \\
	Test Nodes & 1,000 & 1,000 & 1,000 \\
	Label Rate& 0.052 & 0.036 & 0.003 \\
	\bottomrule
\end{tabular}
\end{table}
\vspace{-0.in}
\hide{
\begin{itemize}[leftmargin=*]
	\item \textbf{Citation network datasets} consist of papers as nodes and citation links as directed edges. Each node has a human annotated topic as the corresponding label and content-based features. The features are Bag-of-Words (BoW) vectors and extracted from the paper abstracts. For Cora and Citeseer, the feature entries indicate the presence/absence of the corresponding word from a dictionary, and for Pubmed, the feature entries indicate the Term Frequency-Inverse Document Frequency (TF-IDF) of the corresponding word from a dictionary. Therefore, the semi-supervised learning task is to predict the topic of papers, given the BoW of their abstracts and the citations to other papers. 
	
	\textit{For fair comparison, we follow exactly the same data splits as
	~\cite{yang2016revisiting, kipf2016semi, Velickovic:17GAT}, with 20 nodes per class for training, 500 overall nodes for validation, and 1000 nodes for evaluation.}
	
	\item \textbf{The PPI dataset}, as processed and described by~\cite{hamilton2017inductive}, consists of 24 disjoint subgraphs, each corresponding to a different human tissue. 20 of those subgraphs are used for training, 2 for validation, and 2 for testing, as partitioned by~\cite{hamilton2017inductive}.
\end{itemize}
}

\vpara{Baselines.}
To comprehensively evaluate the proposed \model\ model, we compare it with both state-of-the-art  graph convolution models and regularization based GNNs. 
\begin{itemize}
	\item \textit{Graph convolution methods:} GCN~\cite{kipf2016semi}, GAT~\cite{Velickovic:17GAT}, Graph U-Net~\cite{gao2019graph}, MixHop~\cite{abu2019mixhop} and GraphNAS~\cite{gao2019graphnas}. 
	\item \textit{Regularization based GNNs:} VBAT~\cite{deng2019batch}, $\text{G}^3$NN~\cite{ma2019flexible},  GraphMix~\cite{verma2019graphmix} and Dropedge~\cite{YuDropedge}. We report the results of these methods with GCN as the backbone model.
	\end{itemize}
	
	\subsection{Overall Results}

The results of these comparison models are taken from the corresponding papers\footnote{Since dropedge~\cite{YuDropedge} did not report the results of semi-supervised node classification in the original paper, we take the results from its githup page \url{https://github.com/DropEdge/DropEdge}.}. As for our methods, besides the original \model, we also reports two additional variants of \model: \model\_GCN and \model\_GAT, which respectively adopt GCN and GAT as the prediction model, instead of MLP. The accuracy results of our methods are averaged over  \textbf{100 runs} with random weight initializations.
We also report the standard deviations. 

\hide{
We first compare against traditional methods without graph convolutions, including Multilayer Perceptron (MLP), manifold regularization (ManiReg)~\cite{belkin2006manifold}, semi-supervised embedding (SemiEmb)~\cite{weston2012deep}, label propagation (LP)~\cite{zhu2003semi}, DeepWalk~\cite{perozzi2014deepwalk}, Iterative Classification Algorithm (ICA)~\cite{lu2003link}, and Planetoid~\cite{yang2016revisiting}. Their results are taken from the Planetoid paper~\cite{yang2016revisiting} and GCN paper~\cite{kipf2016semi}. The last non-graph convolution baseline we considered is GraphSGAN~\cite{ding2018semi}, which uses generative adversarial nets to help semi-supervised learning on graphs. 
We also compare against some state-of-the-art graph convolutional methods, and their experimental results are mainly extracted from their original works:
\begin{itemize}[leftmargin=*]
	\item \textbf{GCN}~\cite{kipf2016semi}.  It uses an efficient layer-wise propagation rule that is based on a first-order approximation of spectral convolutions on graphs.
	\item \textbf{MoNet}~\cite{monti2017geometric}. It parametrizes the edge weight in graph and defines the convolution as weighted sum in neighborhood.
	\item \textbf{DPFCNN}~\cite{monti2018dual}. It generalizes MoNet and defines the convolution operations both on the graph and its dual graph.
	\item \textbf{GAT}~\cite{Velickovic:17GAT}. It uses self-attention mechanism to parametrize the edge weight and do the convolution via the weighted sum in neighborhood.
	\item \textbf{Graph U-Net}~\cite{gao2019graph}. It proposes graph pooling and uppooling operations on graph-structured data, inspired by the architecture of the U-net.
	\item \textbf{GraphNAS}~\cite{gao2019graphnas}. It automatically generates neural architecture for graph neural networks by using reinforcement learning.
	\item  \textbf{}
\end{itemize}
}

\hide{
We compare our sampling methods with network sampling-based methods, \textbf{GraphSAGE} and \textbf{FastGCN}.
\begin{itemize}[leftmargin=*]
	\item \textbf{GraphSAGE}~\cite{hamilton2017inductive}. It samples fixed number of nodes in neighborhood and the architecture is similar to GCN.
	\item \textbf{FastGCN}~\cite{chen2018fastgcn}. It's another sampling version of GCN which uses importance sampling to sample nodes.
\end{itemize}
}

\hide{
The proposed methods are listed below: 
\begin{itemize}[leftmargin=*]
\item \textbf{\model\_{MLP}}: The \model\ with a two-layer MLP as the submodel and dropnode as the graph dropout strategy. 
\item \textbf{\model\_{MLP} (dropout)}: The \model\ with a two-layer MLP as the submodel and dropout as the graph dropout strategy. 
\item \textbf{\model\_{GCN}}: The \model\ with a two-layer GCN as the submodel  and dropnode as the graph dropout strategy. 
\item \textbf{\model\_{GCN} (dropout)}: The \model\ with a two-layer GCN as the submodel and dropout as the graph dropout strategy. 
\end{itemize}
}

\hide{
GraphSAGE, GAT, and our \rank, are also used for inductive learning experiments, where test data is unseen during training. 
As \sdrop and \sdm are proposed especially for semi-supervised learning with scarce labeled data while the proportion of labeled data in PPI is high and thus network sampling is not necessary. Therefore we only apply our \srank to PPI.
}

Table~\ref{tab:classification} summarizes the prediction accuracy of node classification of comparison methods. It shows that \model\ achieves the best results on Cora and Citeseer, and also gets a better performance on Pubmed compared with most of baselines. We also observe that \model\_GAT and \model\_GCN outperform vanilla GCN and GAT, indicating the consistency regularized training method can also generalize to improve other models. More interestingly, \model\_GCN and \model\_GAT are inferior to \model, this is because GCN and GAT are easier to over-smooth than MLP. We will provide detailed analyses for this phenomenon in Section \ref{sec:oversmoothing}.



\begin{table}
	\caption{Summary of classification accuracy (\%).}
	\label{tab:classification}
	\setlength{\tabcolsep}{0.8mm}\begin{tabular}{c|ccccc}
		\toprule
		Category &  Method &Cora & Citeseer & Pubmed \\
		\midrule
		\multirow{5}{*}{{\tabincell{c}{ Graph \\Convolution}} }
		&GCN~\cite{kipf2016semi}   & 81.5 & 70.3 & 79.0  \\
		&GAT~\cite{Velickovic:17GAT} & 83.3$\pm$0.4 & 71.2$\pm$0.6 & 78.4 $\pm$0.4 \\
		
		&Graph U-Net~\cite{gao2019graph} & 84.4$\pm$0.6 & 73.2$\pm$0.5 & 79.6$\pm$0.2 \\
		&MixHop~\cite{abu2019mixhop} & 81.9 $\pm$ 0.4 & 71.4$\pm$0.8 & 80.8$\pm$0.6 \\
		&GMNN~\cite{qu2019gmnn} & 83.7  & 72.9 & 81.8  \\
		&GraphNAS~\cite{gao2019graphnas} & 84.2$\pm$1.0 & 73.1$\pm$0.9 & 79.6$\pm$0.4 \\
		
		\midrule
		\multirow{5}{*}{{\tabincell{c}{ Regularization\\based GCNs}} }
		& VBAT~\cite{deng2019batch} & 83.6 $\pm$ 0.5 & 74.0 $\pm$ 0.6 &79.9 $\pm$ 0.4 \\
		& $\text{G}^3$NN~\cite{ma2019flexible}  &82.5 $\pm$0.2 &74.4$\pm$0.3 &77.9 $\pm$ 0.4  \\
		& GraphMix~\cite{verma2019graphmix} &83.9 $\pm$ 0.6 & 74.5 $\pm$0.6& 81.0 $\pm$ 0.6 \\
		& Dropedge~\cite{YuDropedge} &82.8 & 72.3 & 79.6 \\
		\midrule
		\multirow{3}{*}{{\tabincell{c}{ Our \\ methods}} }

		&\model\   & \textbf{85.4$\pm$0.4} & \textbf{75.4$\pm$0.4} & \textbf{82.7$\pm$0.6}  \\
		&\model\_{GCN}   & 84.5$\pm$0.3 & 74.2$\pm$0.3 & 80.0$\pm$0.3  \\
		& \model\_{GAT} & 84.0$\pm$0.4 & 73.2$\pm$ 0.4 & 78.8$\pm$0.3\\
		\bottomrule
	\end{tabular}
\end{table}

\hide{

\begin{figure*}[t]
	\centering
	\mbox
	{
		\hspace{-0.3in}
		\begin{subfigure}[Cora]{
				\centering
				\includegraphics[width = 0.33 \linewidth]{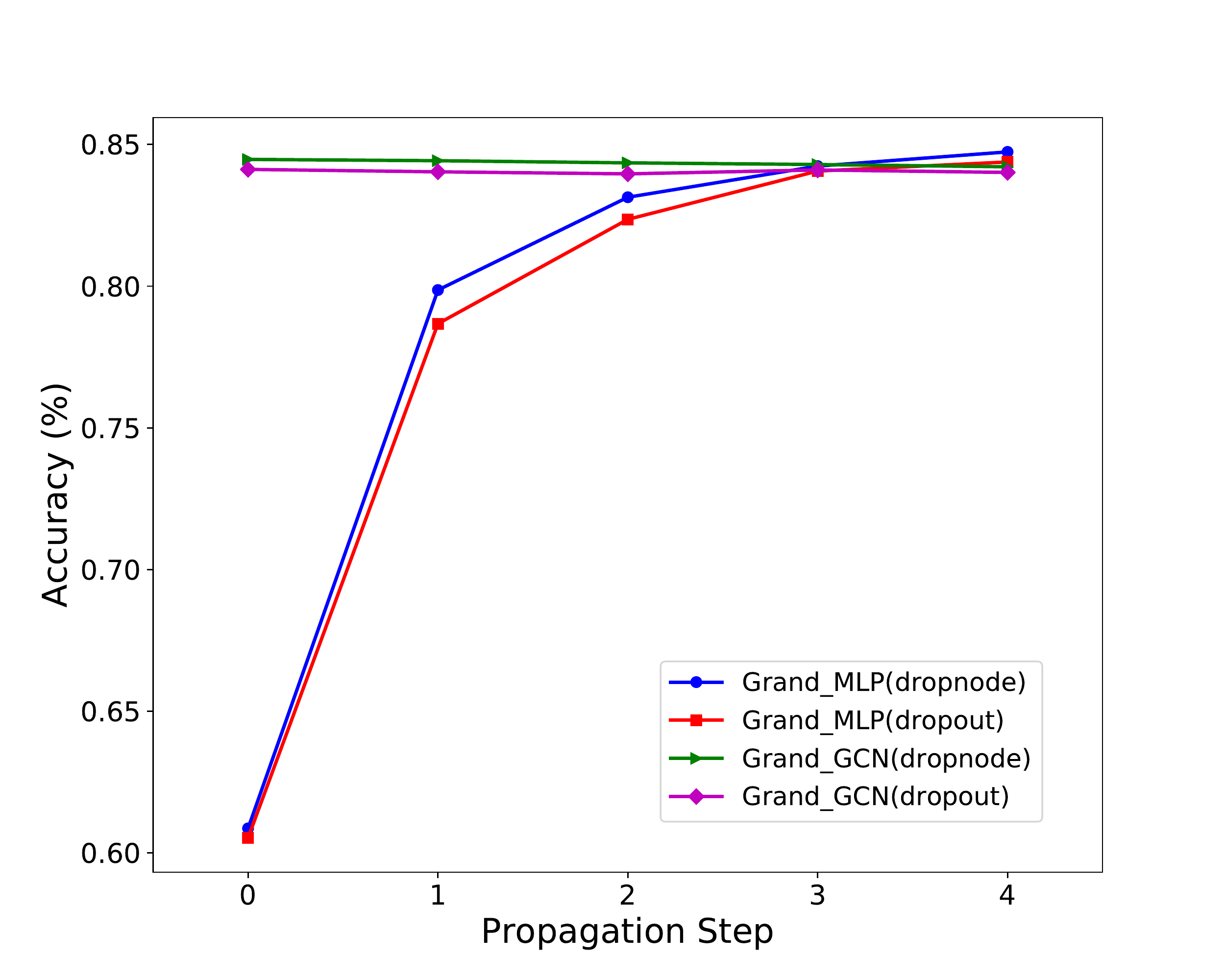}
				\label{fig:step_cora}
			}
		\end{subfigure}
		\hspace{-0.38in}
		\begin{subfigure}[Citeseer]{
				\centering
				\includegraphics[width = 0.33 \linewidth]{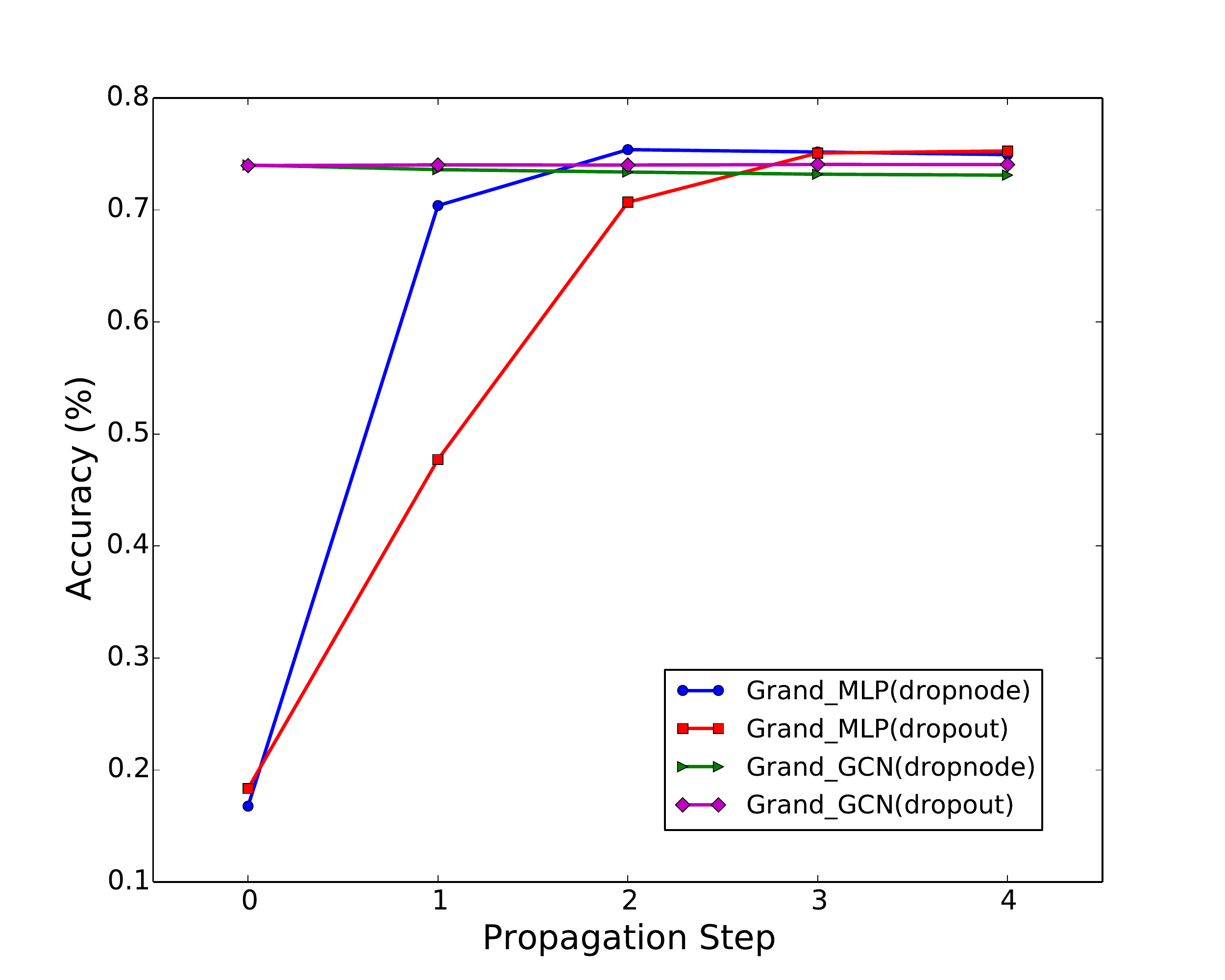}
				\label{fig:step_cite}
			}
		\end{subfigure}
		\hspace{-0.38in}
		\begin{subfigure}[Pubmed]{
				\centering
				\includegraphics[width = 0.33 \linewidth]{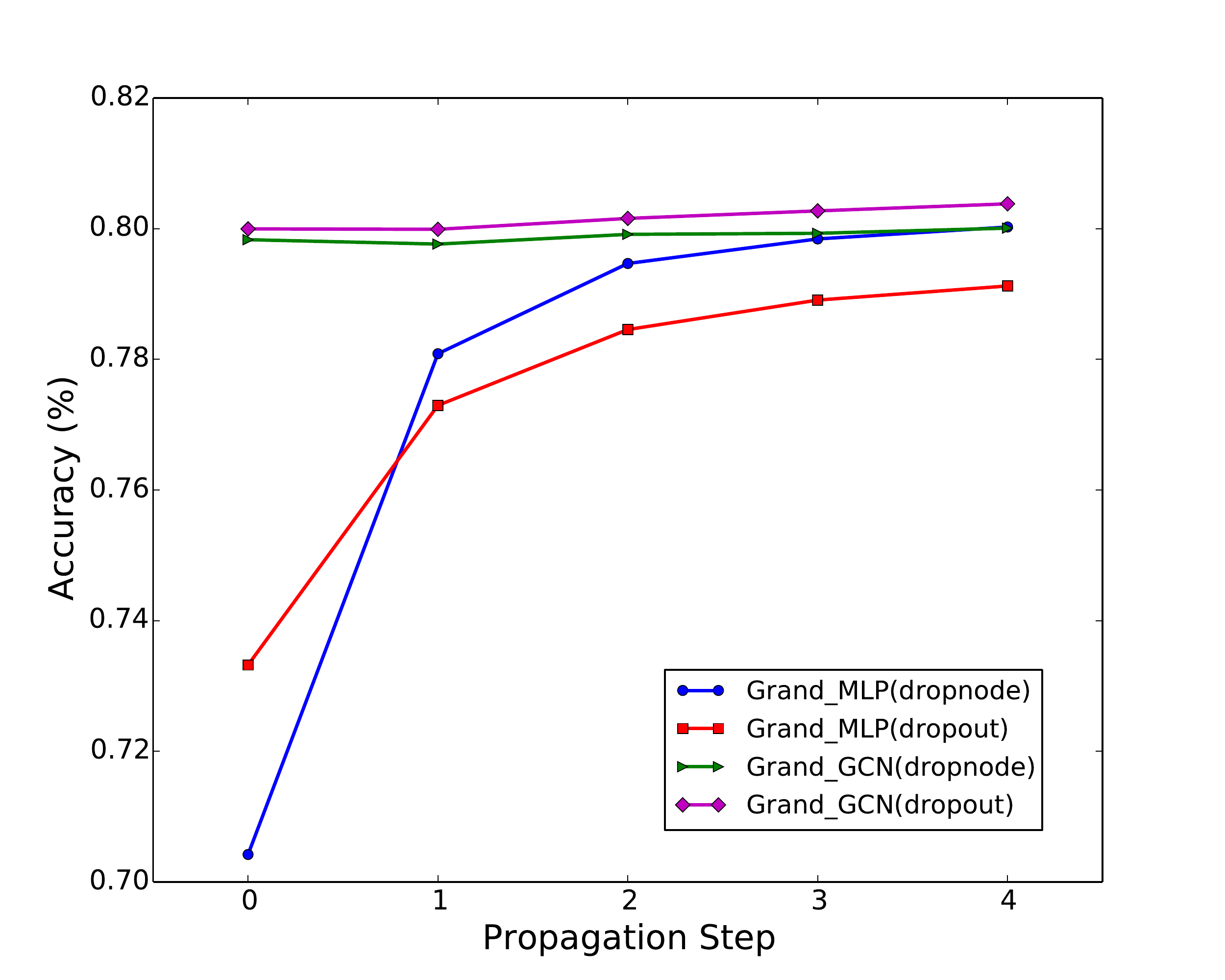}
				\label{fig:step_pub}
			}
		\end{subfigure}
	}
	\caption{Parameter Analysis of Propagation Step.}
	\label{fig:step}
\end{figure*}

\begin{figure*}[t]
	\centering
	\mbox
	{
		\hspace{-0.3in}
		\begin{subfigure}[Cora]{
				\centering
				\includegraphics[width = 0.33 \linewidth]{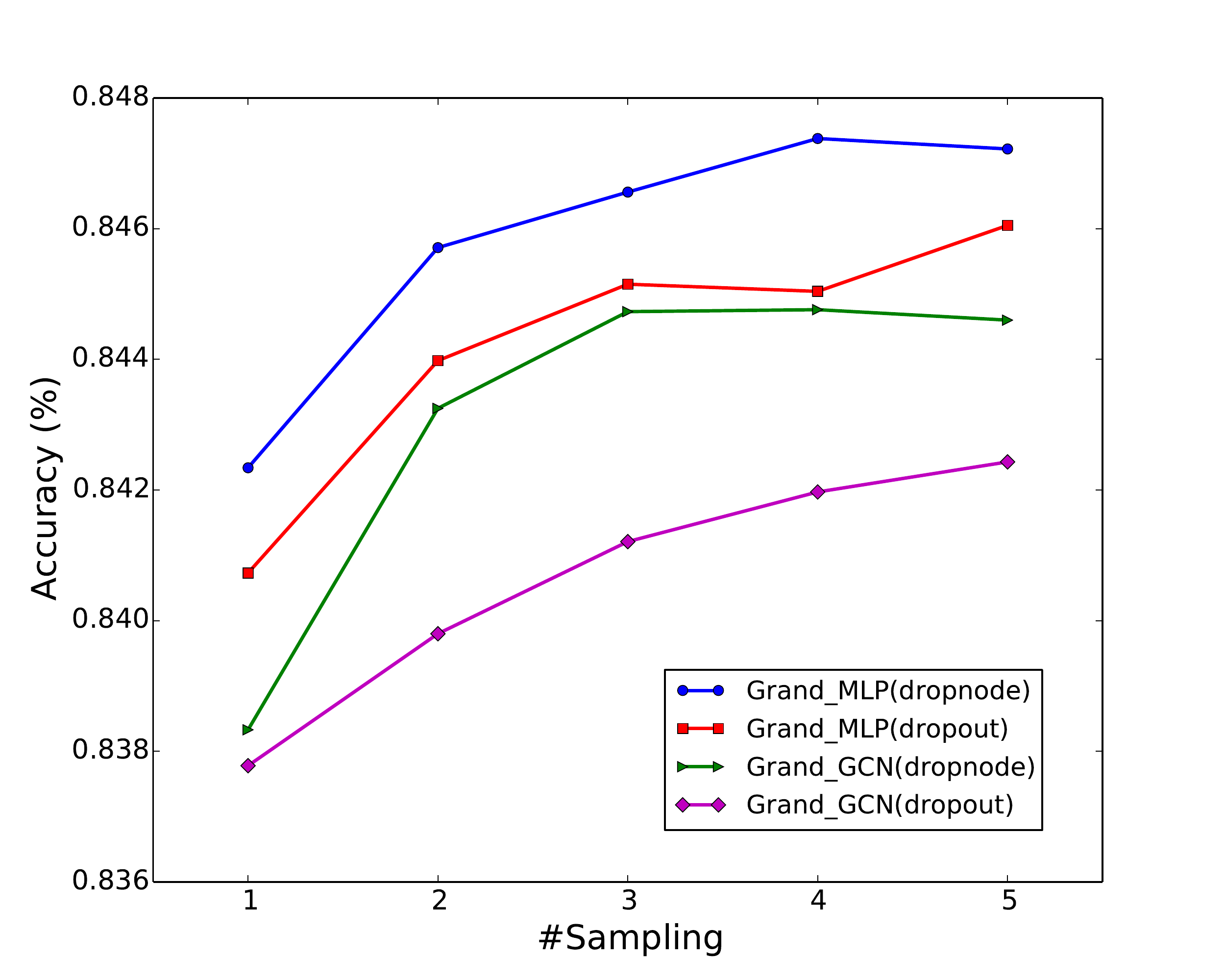}
				\label{fig:step_cora}
			}
		\end{subfigure}
		\hspace{-0.38in}
		\begin{subfigure}[Citeseer]{
				\centering
				\includegraphics[width = 0.33 \linewidth]{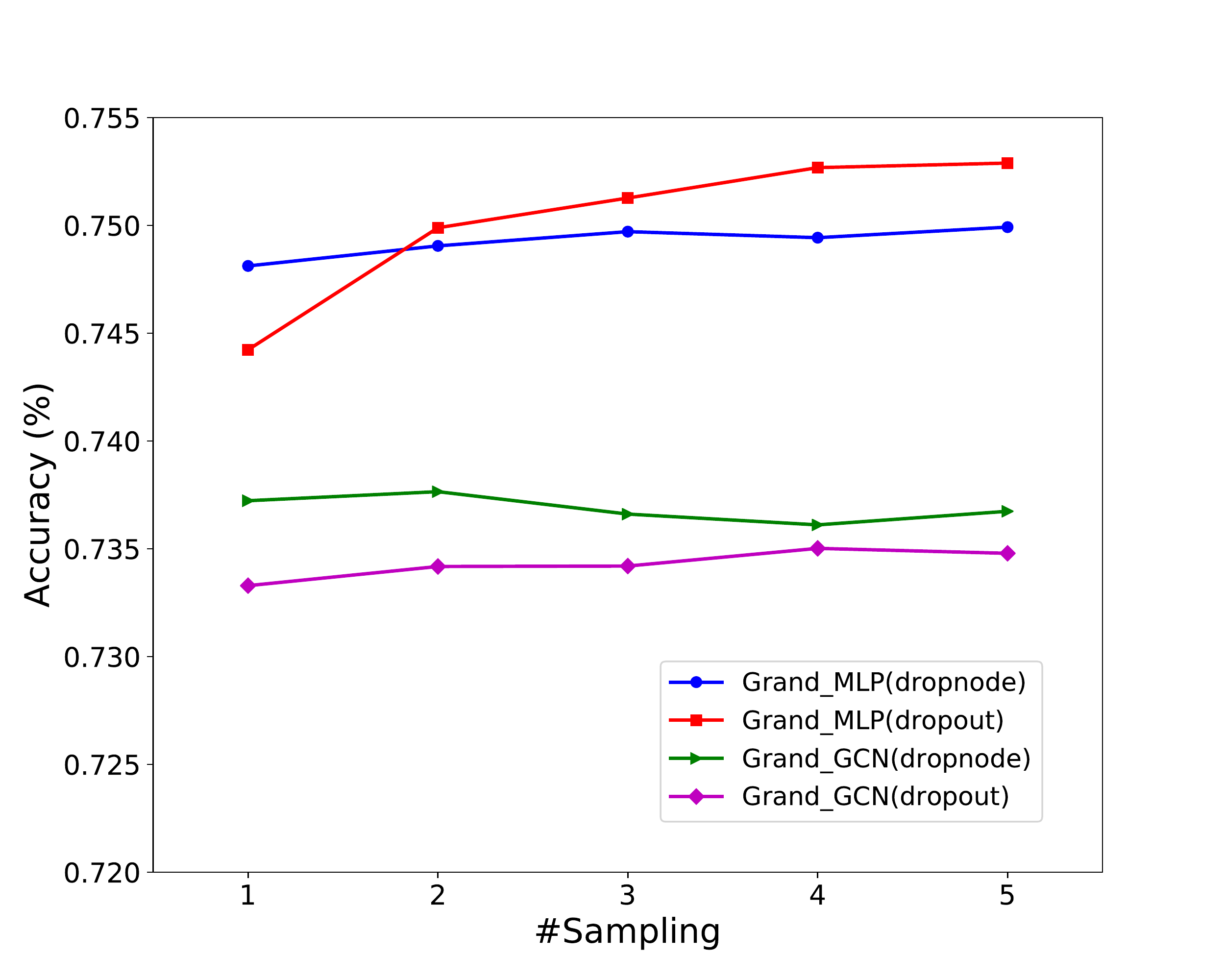}
				\label{fig:step_cite}
			}
		\end{subfigure}
		\hspace{-0.38in}
		\begin{subfigure}[Pubmed]{
				\centering
				\includegraphics[width = 0.33 \linewidth]{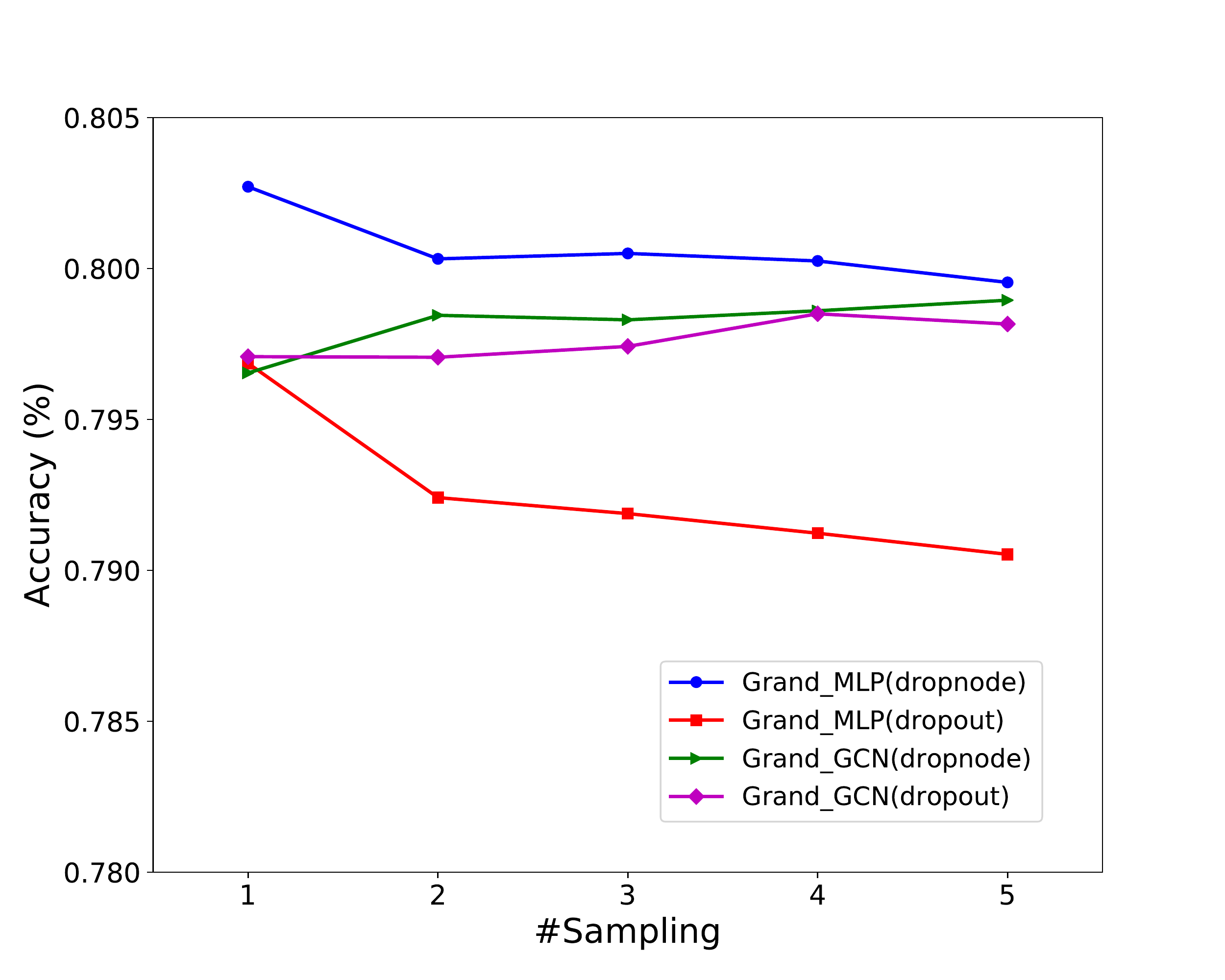}
				\label{fig:step_pub}
			}
		\end{subfigure}
	}
	\caption{Parameter Analysis of Sampling.}
	\label{fig:sampling}
\end{figure*}

\begin{figure*}[t]
\label{Fig:prop}
	\centering
	\mbox
	{
		\hspace{-0.3in}
		\begin{subfigure}[Cora]{
				\centering
				\includegraphics[width = 0.33 \linewidth]{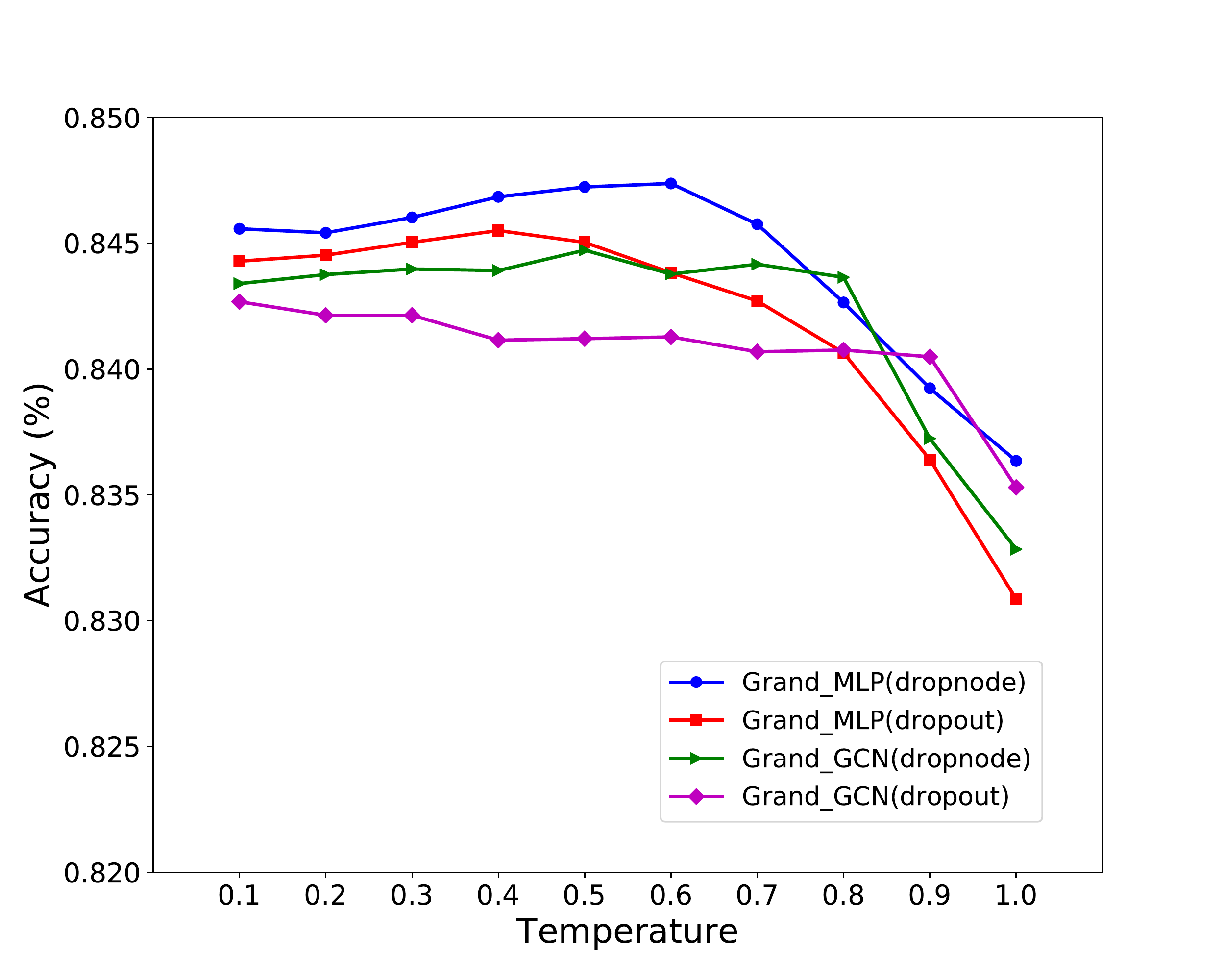}
				\label{fig:step_cora}
			}
		\end{subfigure}
		\hspace{-0.38in}
		\begin{subfigure}[Citeseer]{
				\centering
				\includegraphics[width = 0.33 \linewidth]{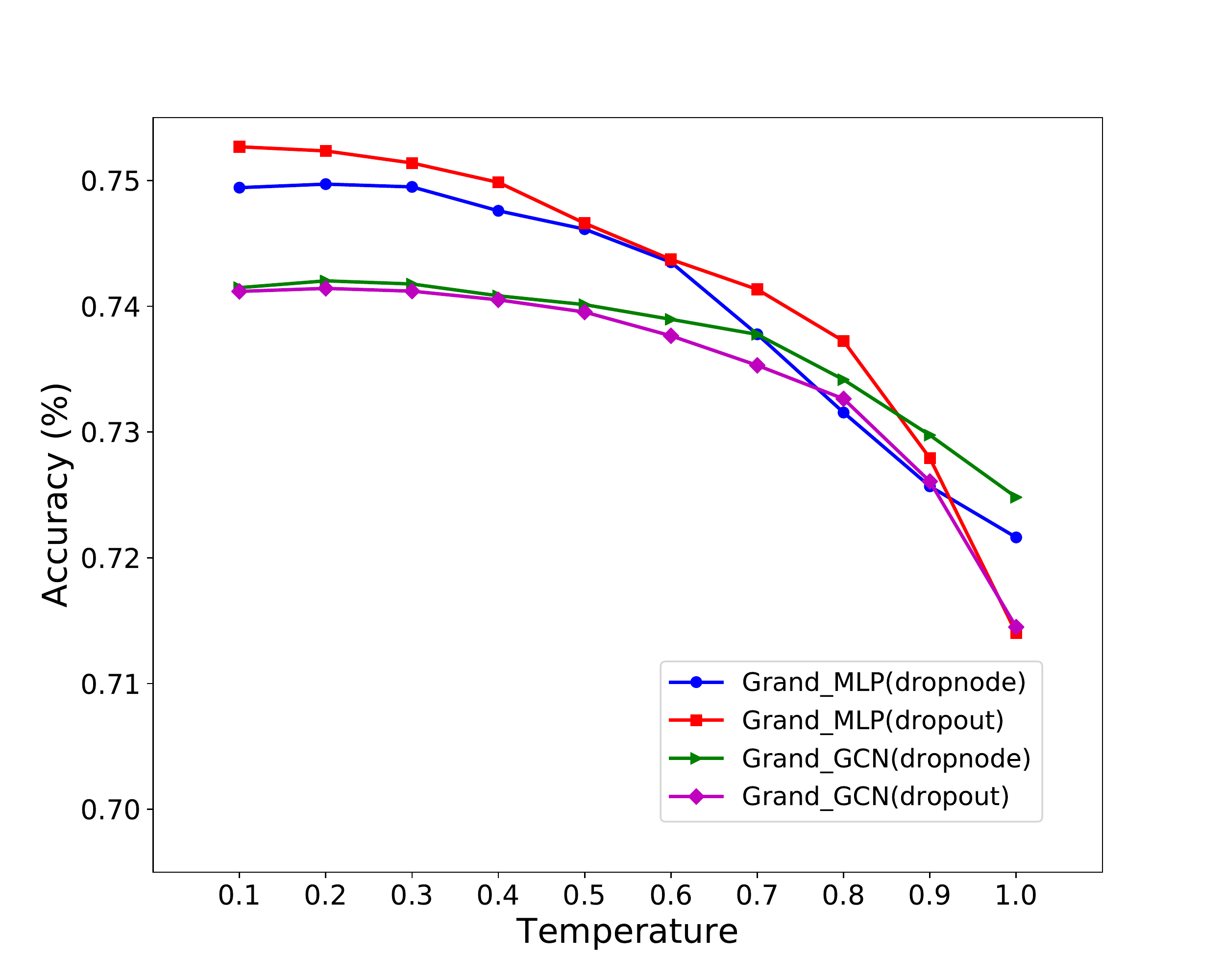}
				\label{fig:step_cite}
			}
		\end{subfigure}
		\hspace{-0.38in}
		\begin{subfigure}[Pubmed]{
				\centering
				\includegraphics[width = 0.33 \linewidth]{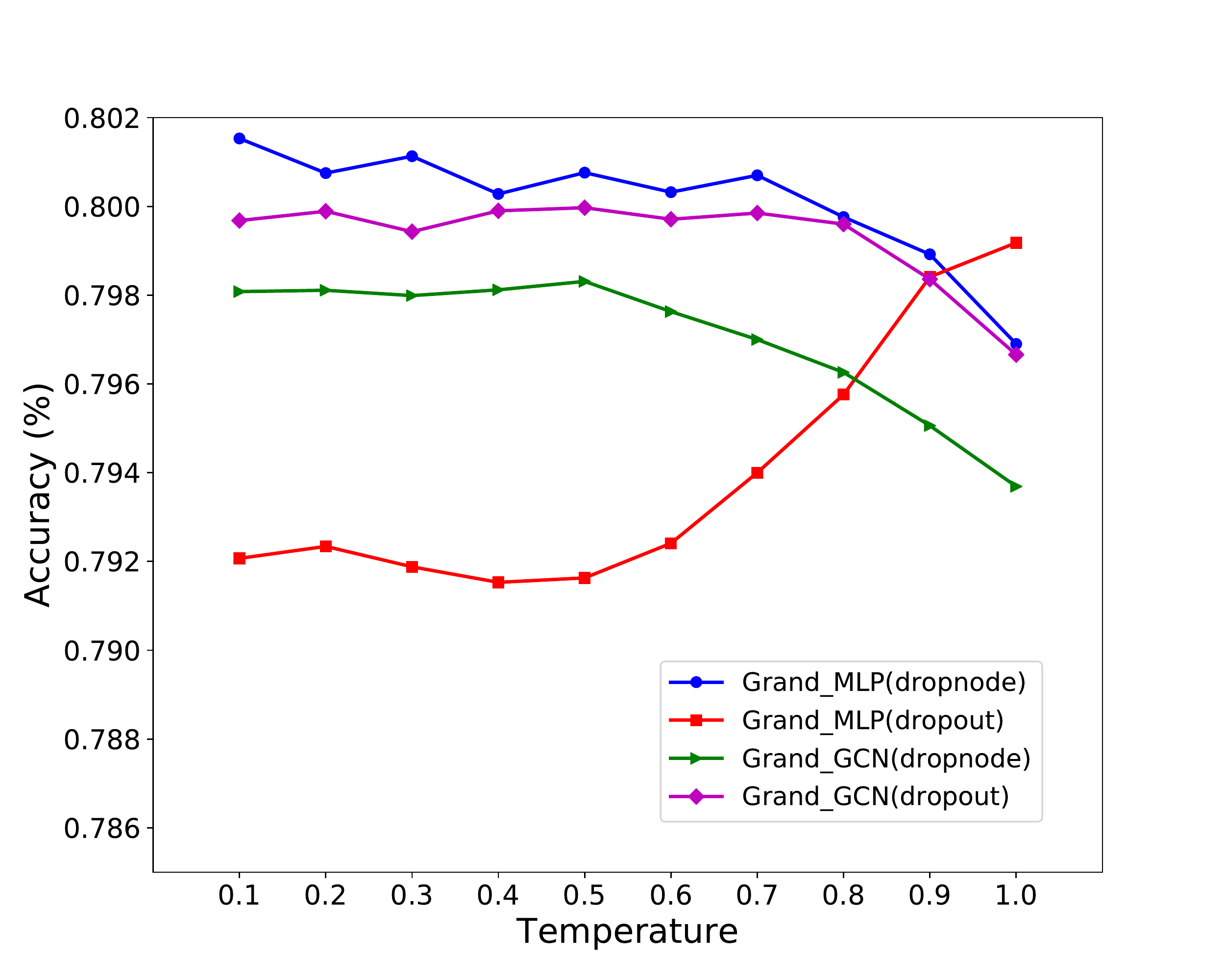}
				\label{fig:step_pub}
			}
		\end{subfigure}
	}
	\caption{Parameter Analysis of Temperature.}
	\label{fig:temp}
\end{figure*}

\begin{figure*}[t]
	\centering
	\mbox
	{
		\hspace{-0.3in}
		\begin{subfigure}[Cora]{
				\centering
				\includegraphics[width = 0.33 \linewidth]{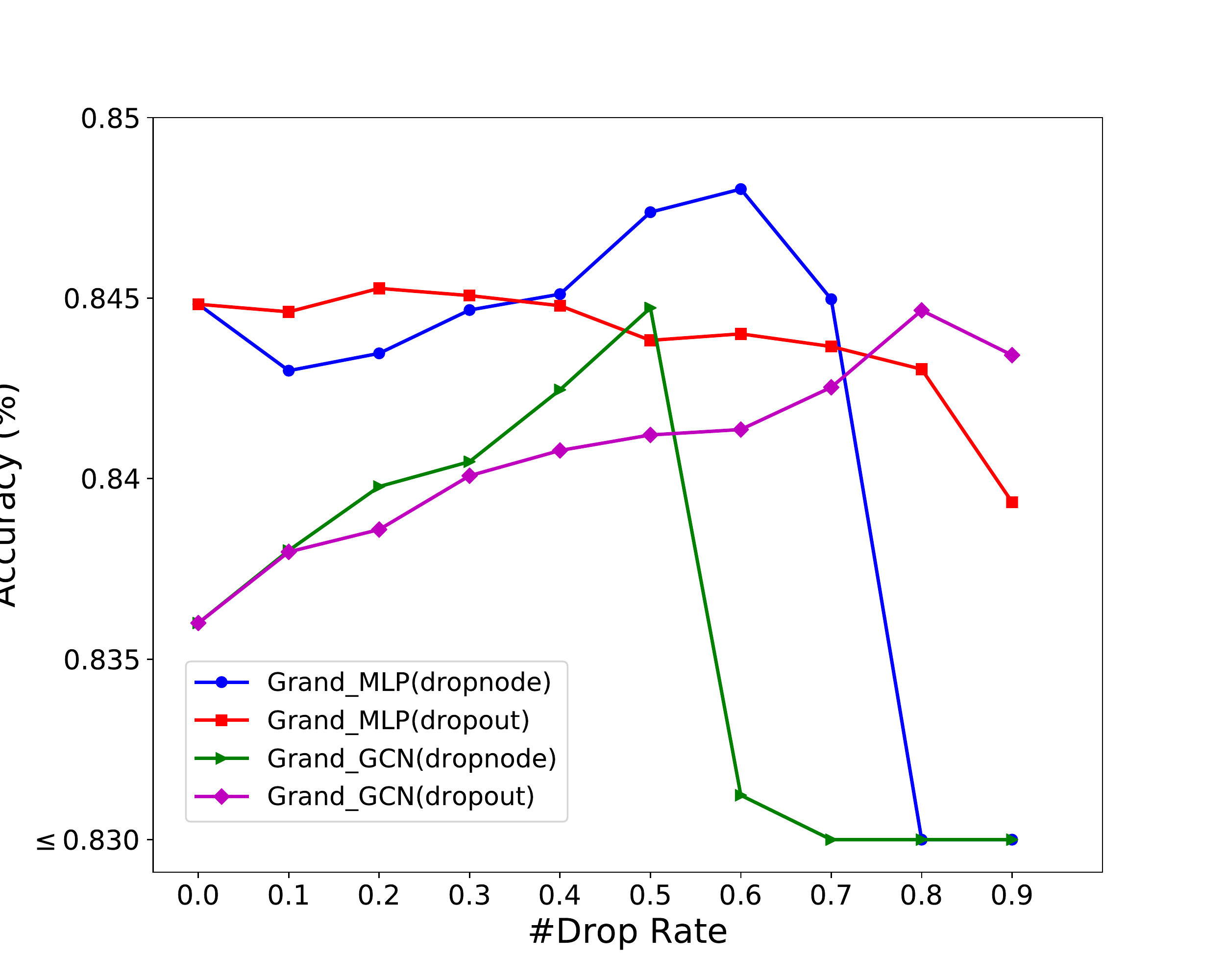}
				\label{fig:step_cora}
			}
		\end{subfigure}
		\hspace{-0.38in}
		\begin{subfigure}[Citeseer]{
				\centering
				\includegraphics[width = 0.33 \linewidth]{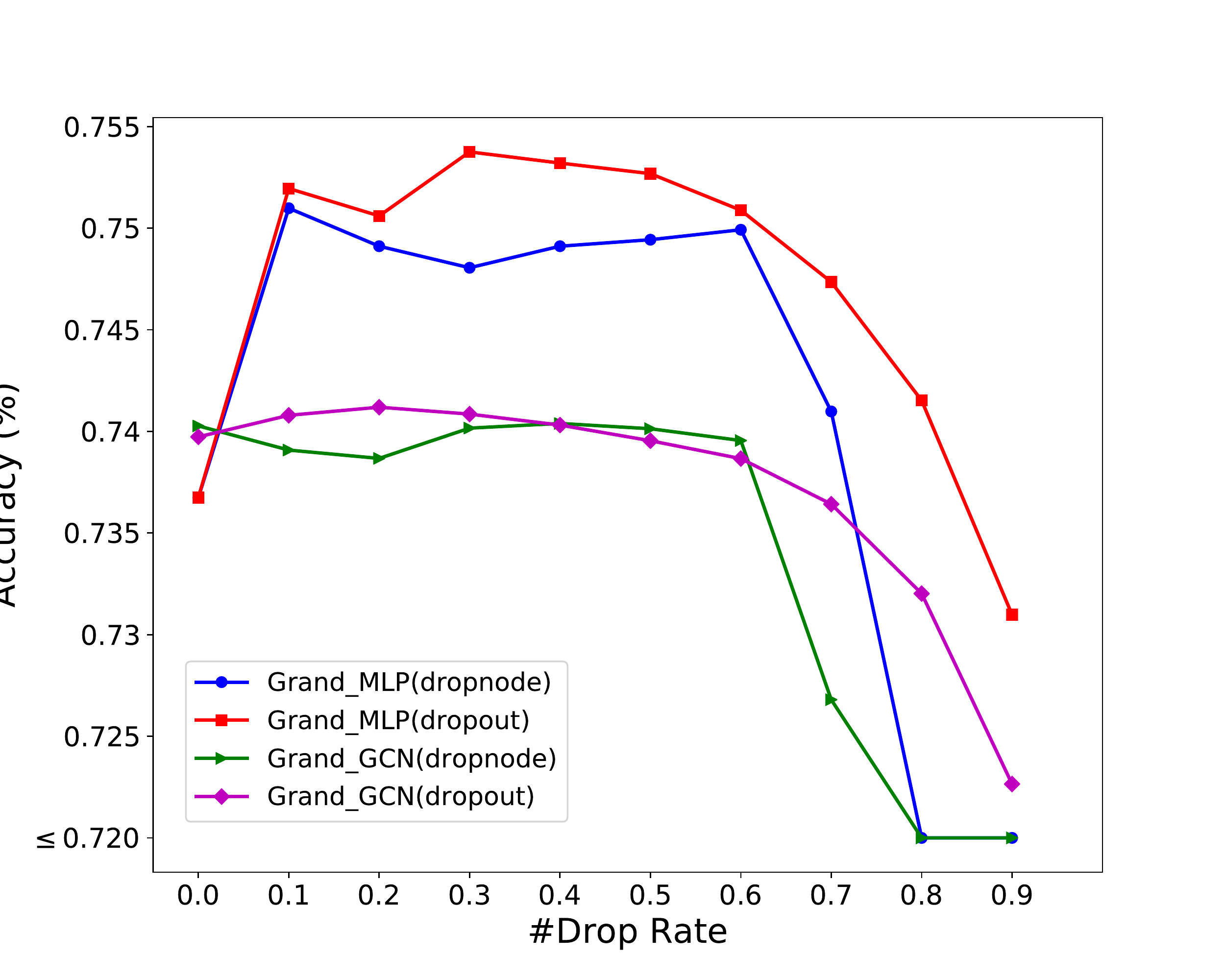}
				\label{fig:step_cite}
			}
		\end{subfigure}
		\hspace{-0.38in}
		\begin{subfigure}[Pubmed]{
				\centering
				\includegraphics[width = 0.33 \linewidth]{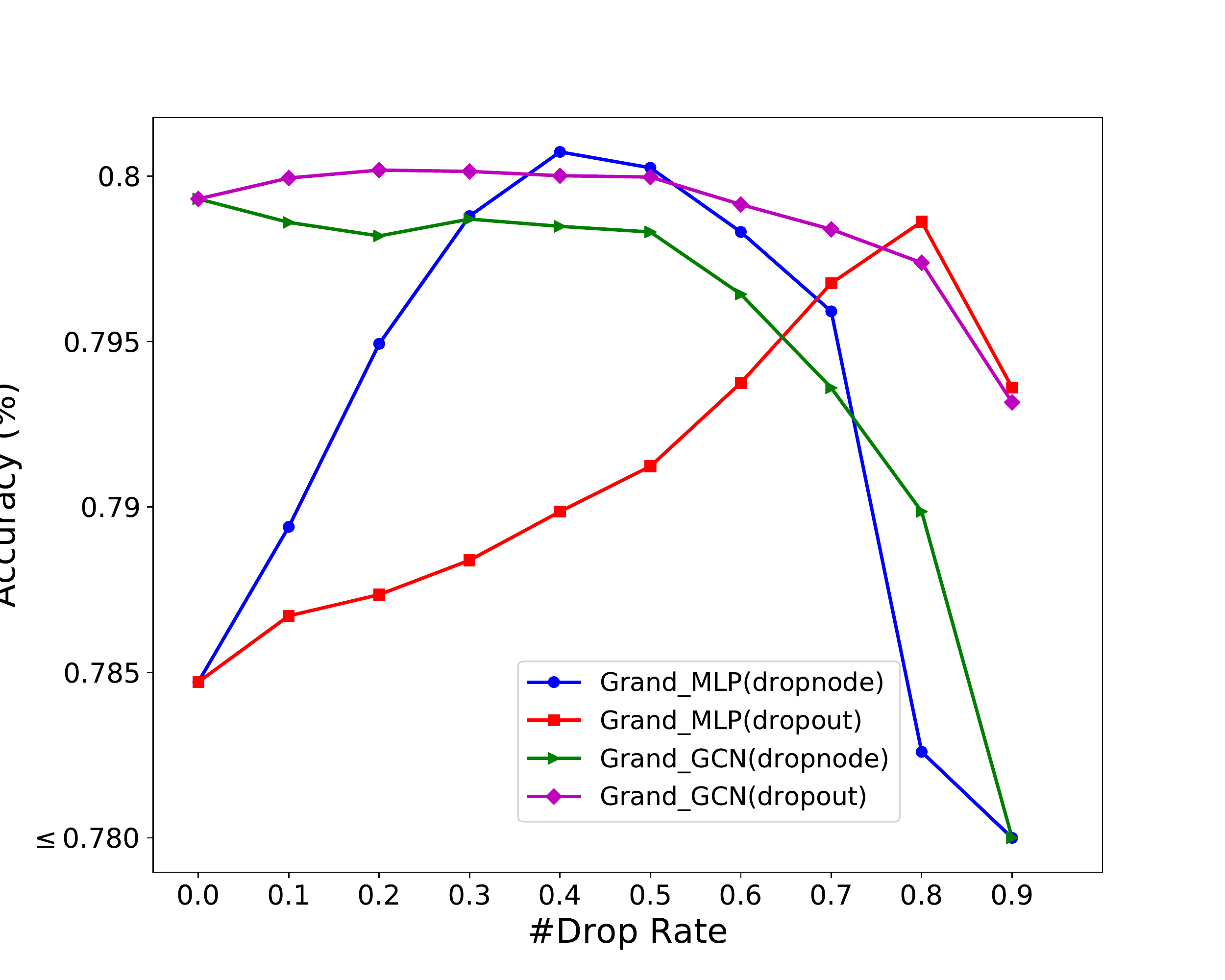}
				\label{fig:step_pub}
			}
		\end{subfigure}
	}
	\caption{Parameter Analysis for Drop Rate.}
	\label{fig:droprate}
\end{figure*}

\begin{figure*}[t]
	\centering
	\mbox
	{
		\hspace{-0.3in}
		\begin{subfigure}[Cora]{
				\centering
				\includegraphics[width = 0.33 \linewidth]{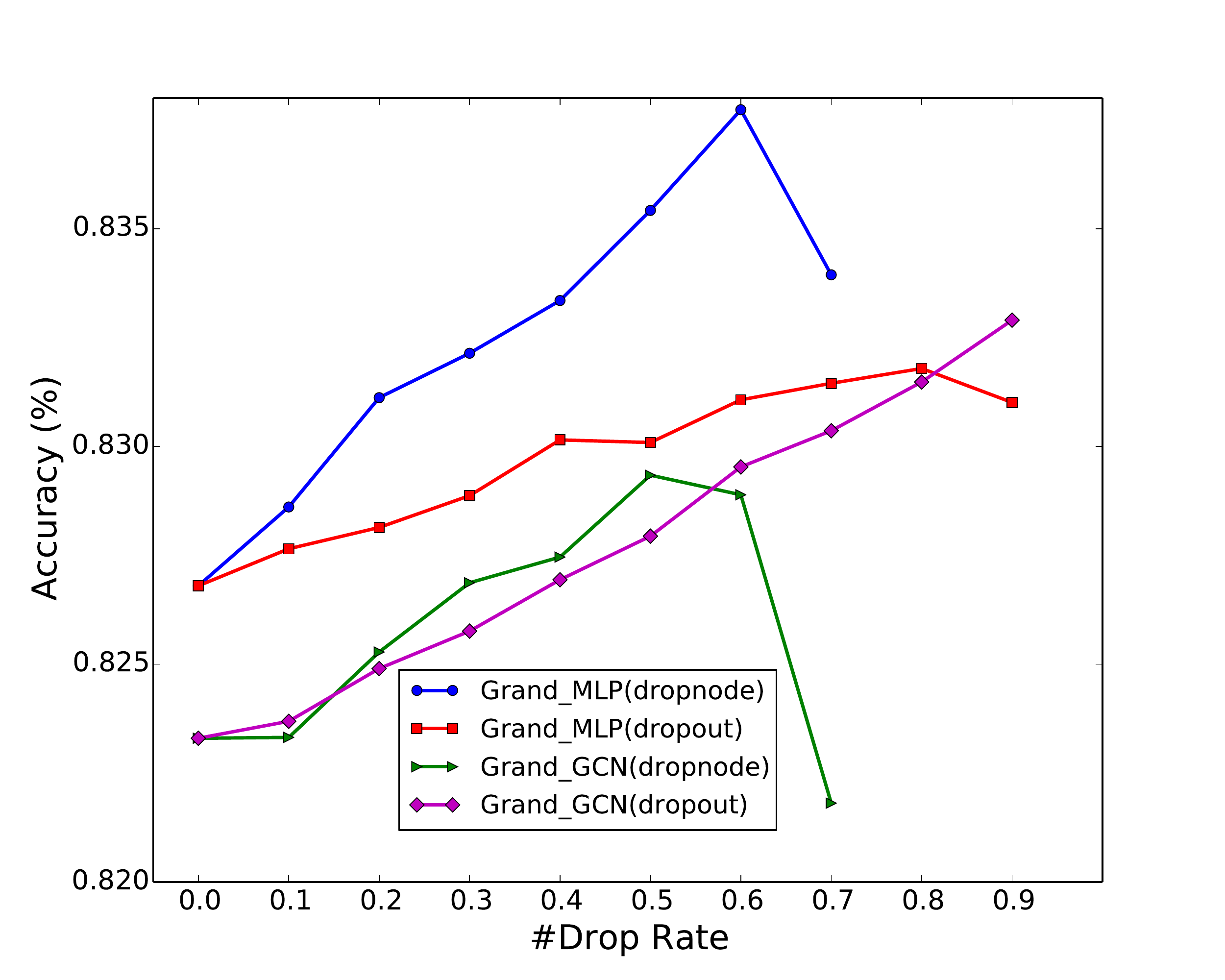}
				\label{fig:step_cora}
			}
		\end{subfigure}
		\hspace{-0.38in}
		\begin{subfigure}[Citeseer]{
				\centering
				\includegraphics[width = 0.33 \linewidth]{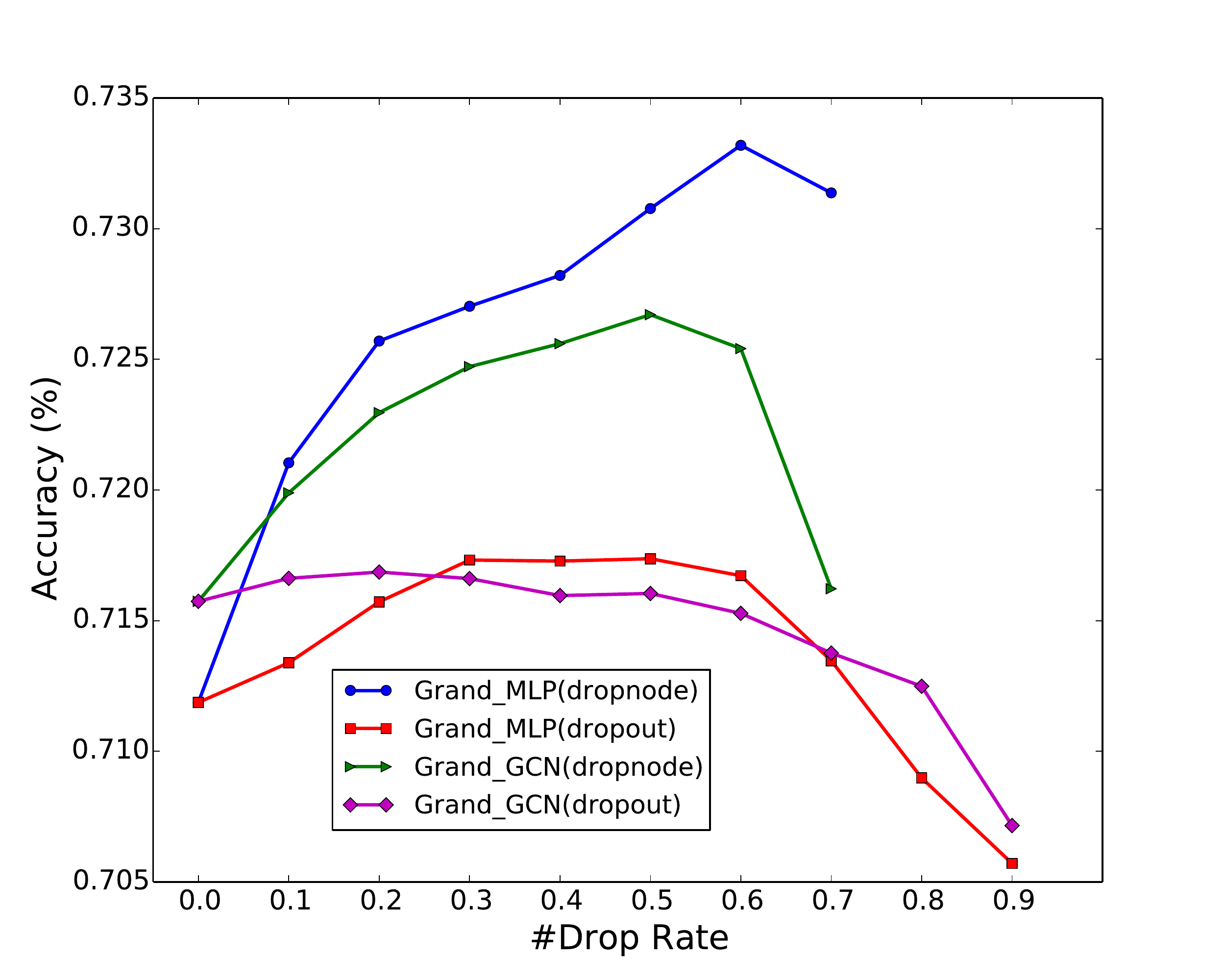}
				\label{fig:step_cite}
			}
		\end{subfigure}
		\hspace{-0.38in}
		\begin{subfigure}[Pubmed]{
				\centering
				\includegraphics[width = 0.33 \linewidth]{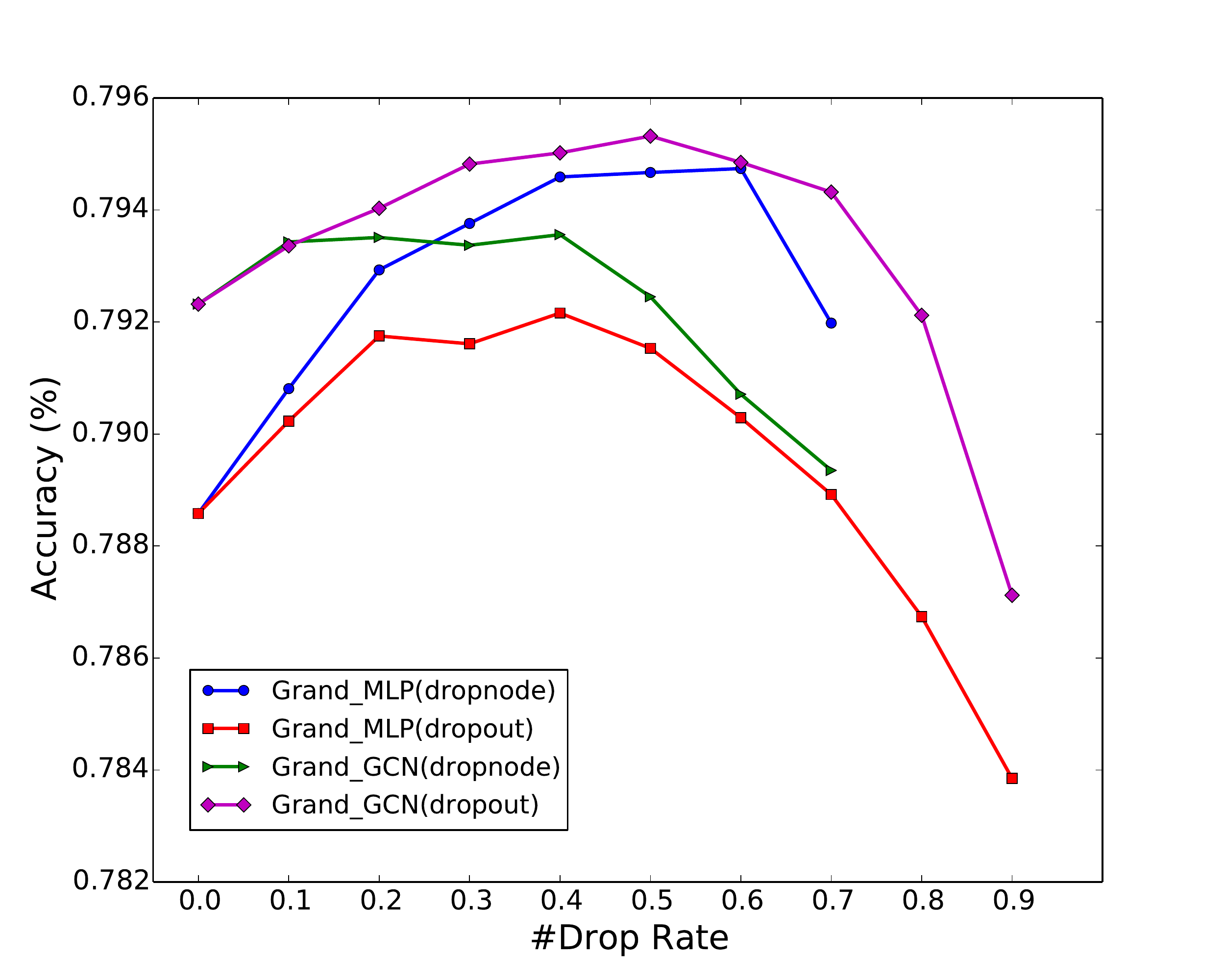}
				\label{fig:step_pub}
			}
		\end{subfigure}
	}
	\caption{Drop rate (without unsupervised loss).}
	\label{fig:step}
\end{figure*}
}

\subsection{Ablation Study}
\label{sec:ablation}
Since \model\ consists of various technicals of semi-supervised learning on graphs. In order to examine the effect of different components, we conduct an ablation study by removing or changing a component from \model. Specifically, we measure the effect of the following:
\begin{itemize}
	\item \textbf{Without CR.} Removing Consistency Regularization (CR) loss in training, i.e., $\lambda=0$. In this way, we only using the supervised classification loss in training.
	\item \textbf{Without multiple dropnode.} Only sampling once in dropnode at each training epoch, i.e., $S=1$. In this way, the consistency regularization only enforcing model to give low-entropy predictions for unlabeled nodes. 
	\item \textbf{Without sharpening.} Not using label sharpening trick (Cf. Eq.~\ref{}) in calculating distribution center for consistency regularization, i.e., $T=1$.
	\item \textbf{Without CR and dropnode.} Removing Consistency Regularization (CR) loss and not using dropnode, i.e., $\lambda=0, \delta=0$. In this way, the model becomes the concatenation of deterministic propagation and MLP.
	\item \textbf{Replace dropnode with dropout.} Replacing the dropnode component with dropout.
\end{itemize}
 In this experiment, we directly use the best performing hyperparameters found for the results in Table \ref{tab:classification} and did not do any hyperparameter tuning. Table \ref{tab:ablation} summarizes the results of ablation study. We find that each component contributes to the performance of \model\ on the three datasets, except for the multiple dropnode on Pubmed. What's more, we observe that our model can also achieve the best performance on Cora ($84.4$) without applying consistency regularization compared with the baseline methods in Table \ref{tab:classification}. When replacing dropnode with dropout, we observe the classification results consistently decrease, indicating dropnode is more suitable for graph data. We also conduct more detailed experiments to compare dropnode and dropout under different propagation steps, the results are shown in Appendix \ref{sec:dropnode_vs_dropout}.

\begin{table}
	\caption{Ablation study results (\%).}
	\label{tab:ablation}
		\begin{tabular}{lccc}
		\toprule
		Model &Cora & Citeseer & Pubmed\\
		\midrule
		\model  & \textbf{85.4$\pm$0.4} & \textbf{75.4$\pm$0.4} & \textbf{82.7$\pm$0.6} \\
		\midrule
		\quad without Consistency Regularization  & 84.4 $\pm$0.5  & 73.1 $\pm$0.6 & 80.9 $\pm$0.8 \\ 
		 \quad without multiple Dropnode &84.7 $\pm$0.4 & 74.8$\pm$0.4 & 81.0$\pm$1.1  \\
		\quad without sharpening & 84.6 $\pm$0.4  & 72.2$\pm$0.6 & 81.6 $\pm$ 0.8 \\
		\quad without Consistency Regularization and dropnode & 83.2 $\pm$ 0.5     & 70.3 $\pm$ 0.6    & 78.5$\pm$ 1.4   \\ 
		\quad replace dropnode with dropout & 84.9 $\pm$ 0.4  & 75.0 $\pm$ 0.3 & 81.7 $\pm$ 1.0 \\
		\bottomrule
	\end{tabular}
\end{table}

\subsection{Robustness Analysis}
\begin{figure}[t]
	\centering
	\mbox
	{
		\hspace{-0.1in}
		\begin{subfigure}[Random Attack]{
				\centering
				\includegraphics[width = 0.5 \linewidth]{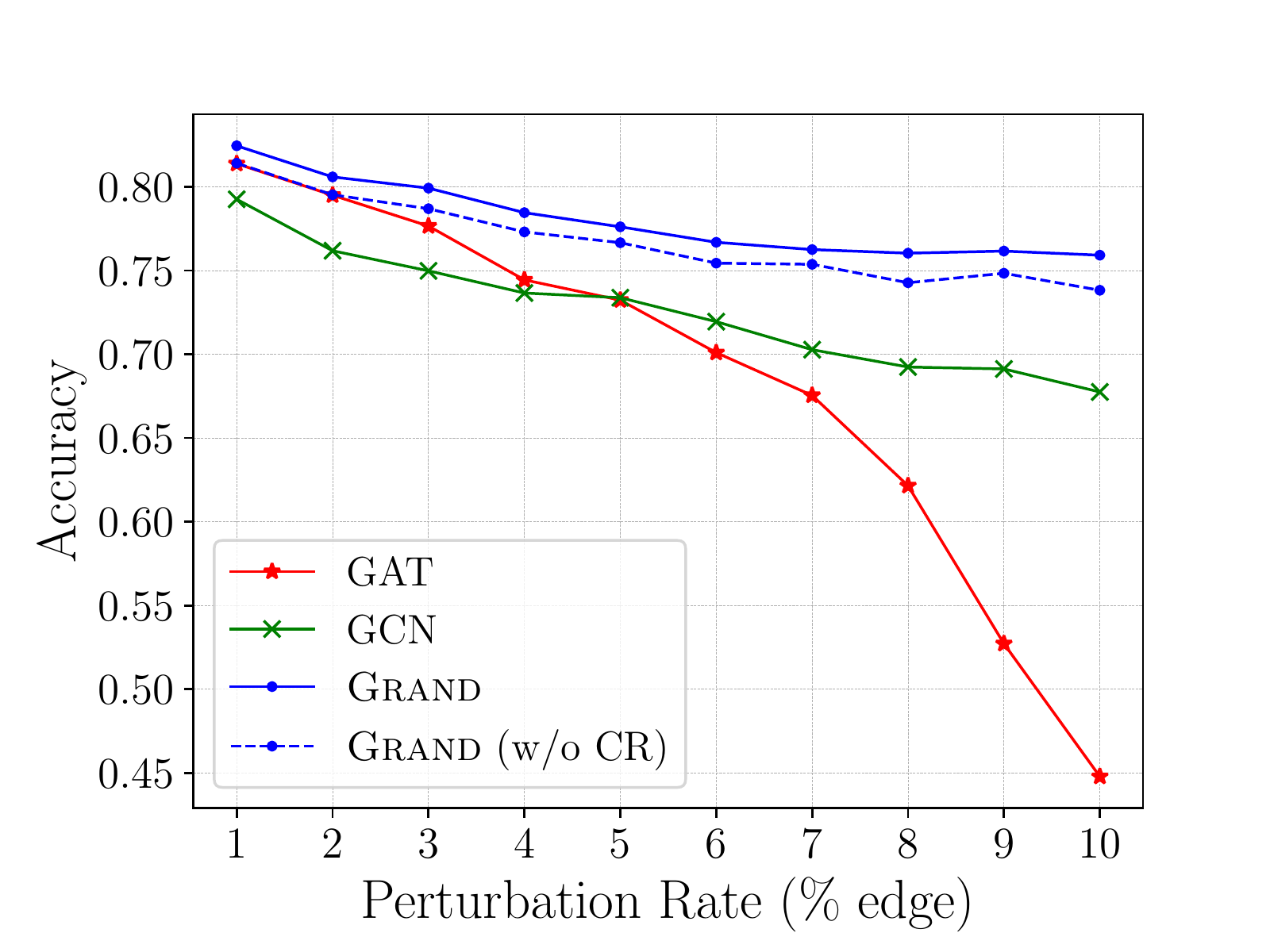}
			}
		\end{subfigure}
		\hspace{-0.1in}
		\begin{subfigure}[Metattack]{
				\centering
				\includegraphics[width = 0.5 \linewidth]{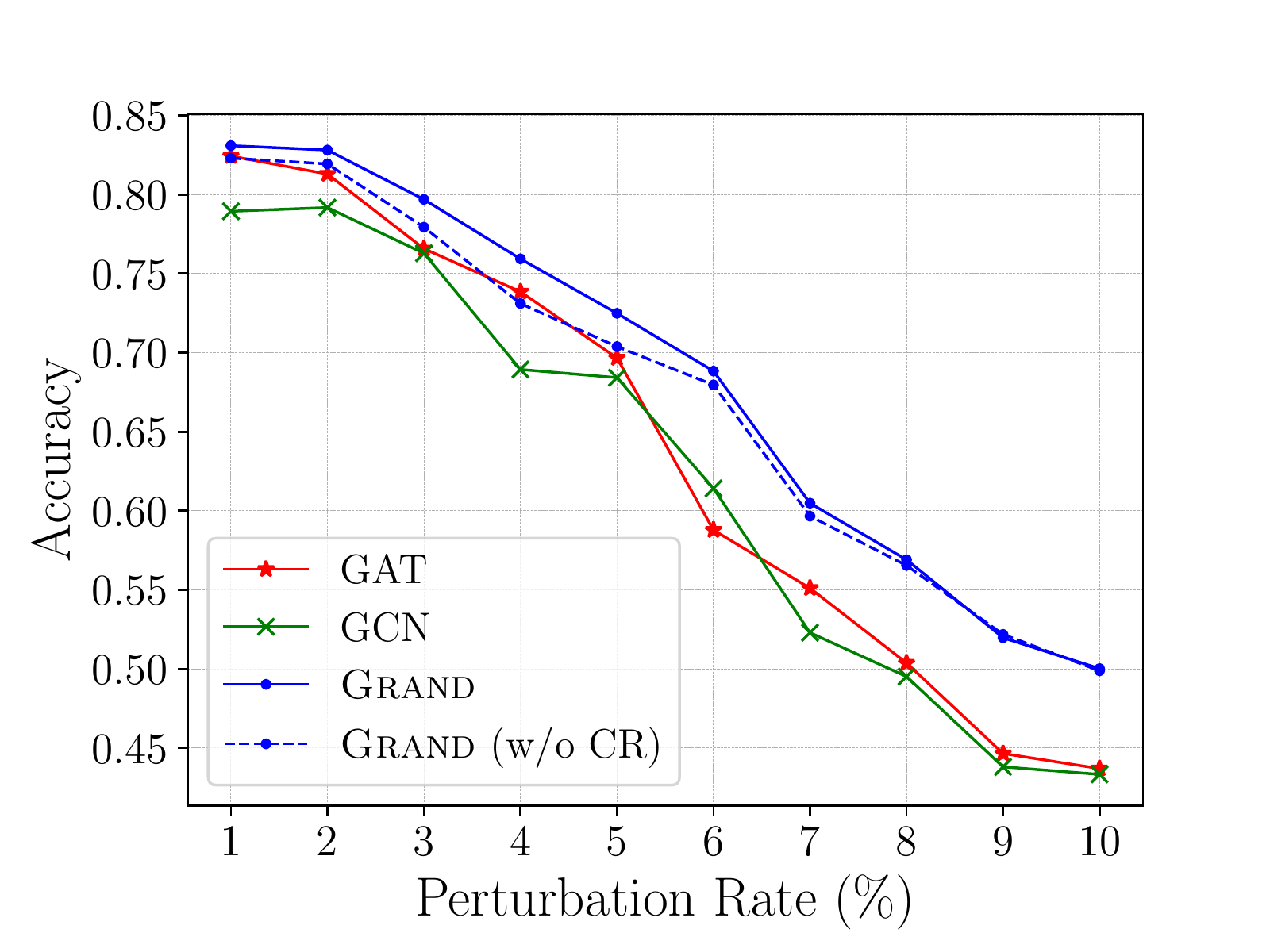}
			}
		\end{subfigure}
	}
	\caption{Results of different methods trained with perturbed graphs.}
	\label{fig:robust}
\end{figure}

In this section, we evaluate the robustness of \model. Specifically, we utilize several attack methods to perturb the graph data, and examine model's classification accuracies on perturbed graphs. Here we adopt two different attack methods:
\begin{itemize}
	\item \textbf{Random Attack.} Perturbing graph structure by randomly adding several fake edges.
\item \textbf{Metattack~\cite{zugner2019adversarial}.} Attacking graph structure by removing or adding some edges based on meta learning.
\end{itemize}
The results of different perturbation rates on Cora dataset are shown in Figure \ref{fig:robust}.We also report the results of \model without using consistency regularization loss.   
 We observe \model\ consistently outperforms GCN and GAT in all settings. Especially for Random Attack, the performance improvements of \model\, including with and without consistency regularization loss, relative to GCN and GAT increases as enlarging the perturbation rate, suggesting both random propagation and consistency regularization can improve model's robustness. 
\subsection{Relieving Over-smoothing}
\label{sec:oversmoothing}
\begin{figure}[t]
	\centering
	\mbox
	{
		\hspace{-0.1in}
		\begin{subfigure}[Classification Results]{
				\centering
				\includegraphics[width = 0.5 \linewidth]{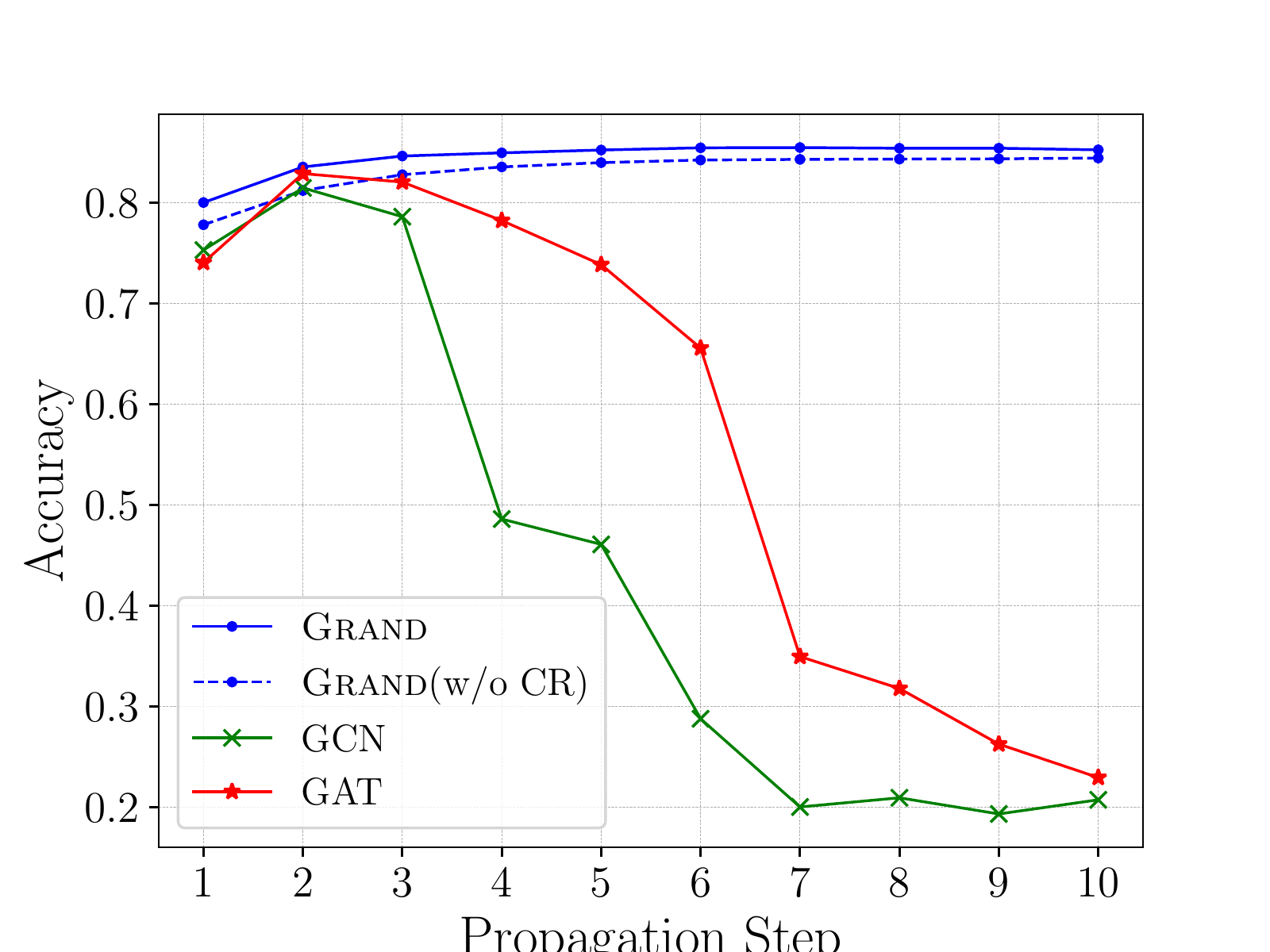}
			}
		\end{subfigure}
		\hspace{-0.1in}
		\begin{subfigure}[MADGap]{
				\centering
				\includegraphics[width = 0.5 \linewidth]{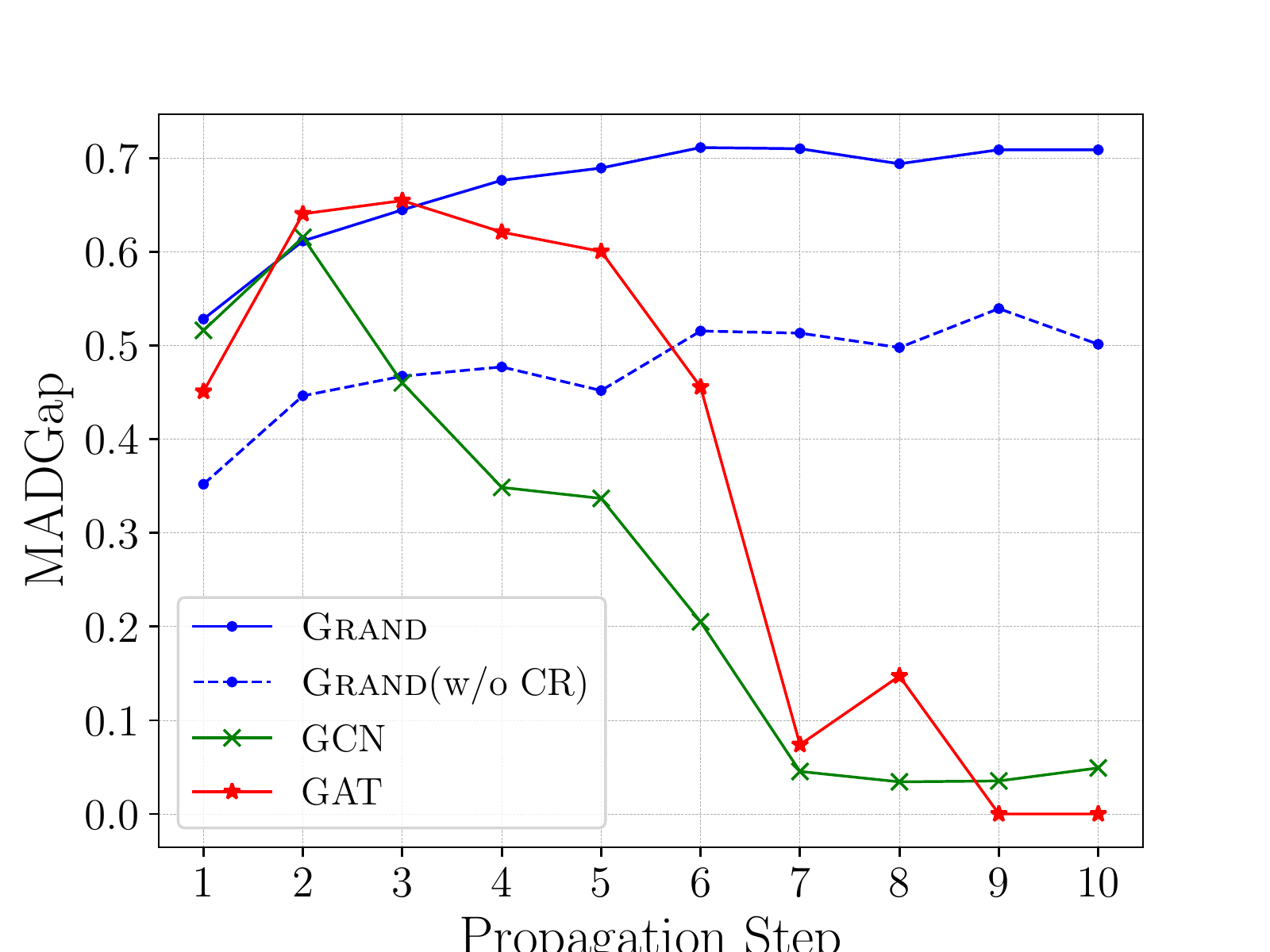}
			}
		\end{subfigure}
	}
	\caption{Comparisons among \model, GCN and GAT under different propagation steps.}
	\label{fig:grandmlp_vs_gcn}
\end{figure}
\begin{figure}[t]
	\centering
	\mbox
	{
		\hspace{-0.1in}
		\begin{subfigure}[Classification Results]{
				\centering
				\includegraphics[width = 0.5 \linewidth]{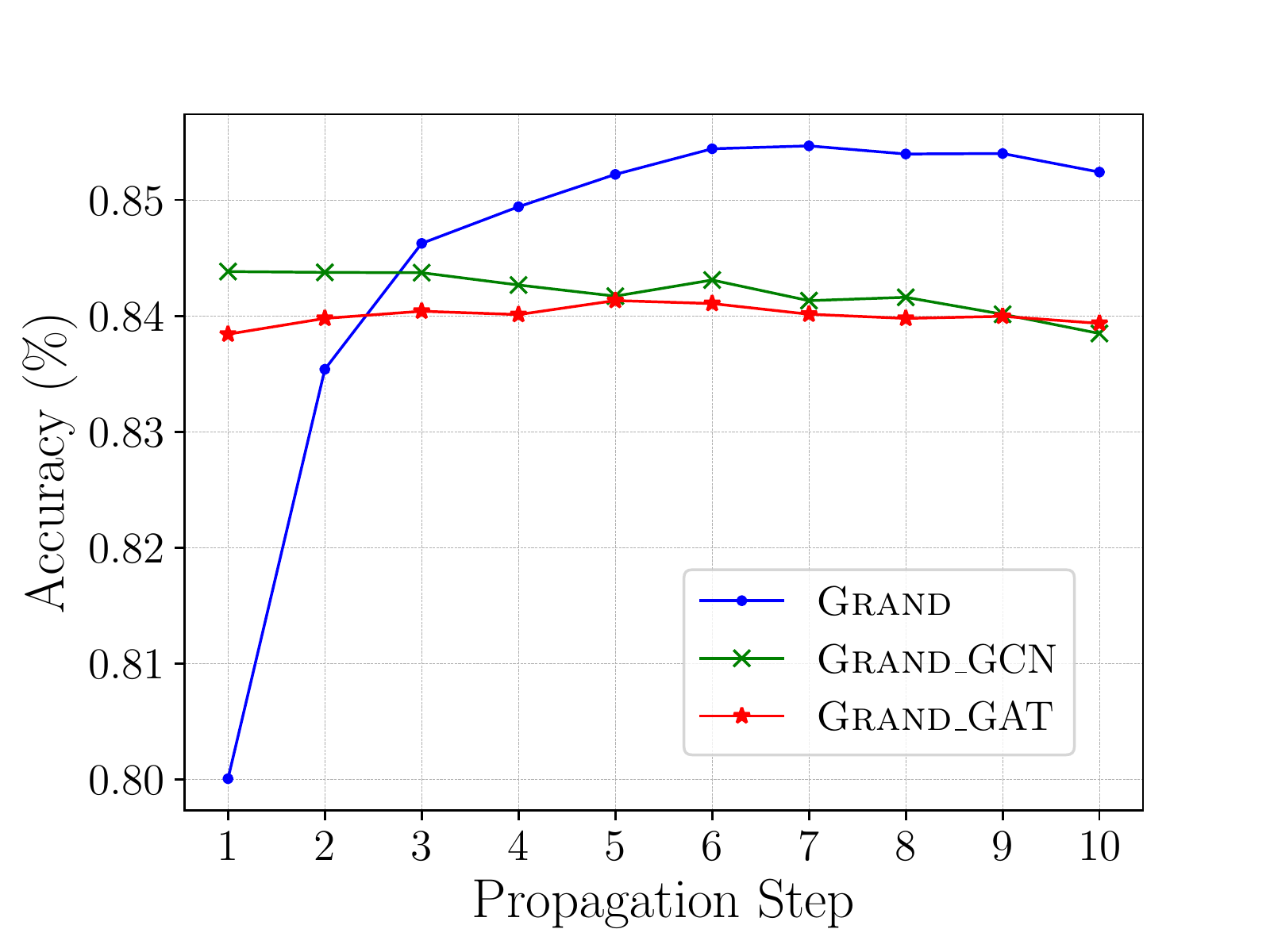}
			}
		\end{subfigure}
		\hspace{-0.1in}
		\begin{subfigure}[MADGap]{
				\centering
				\includegraphics[width = 0.5 \linewidth]{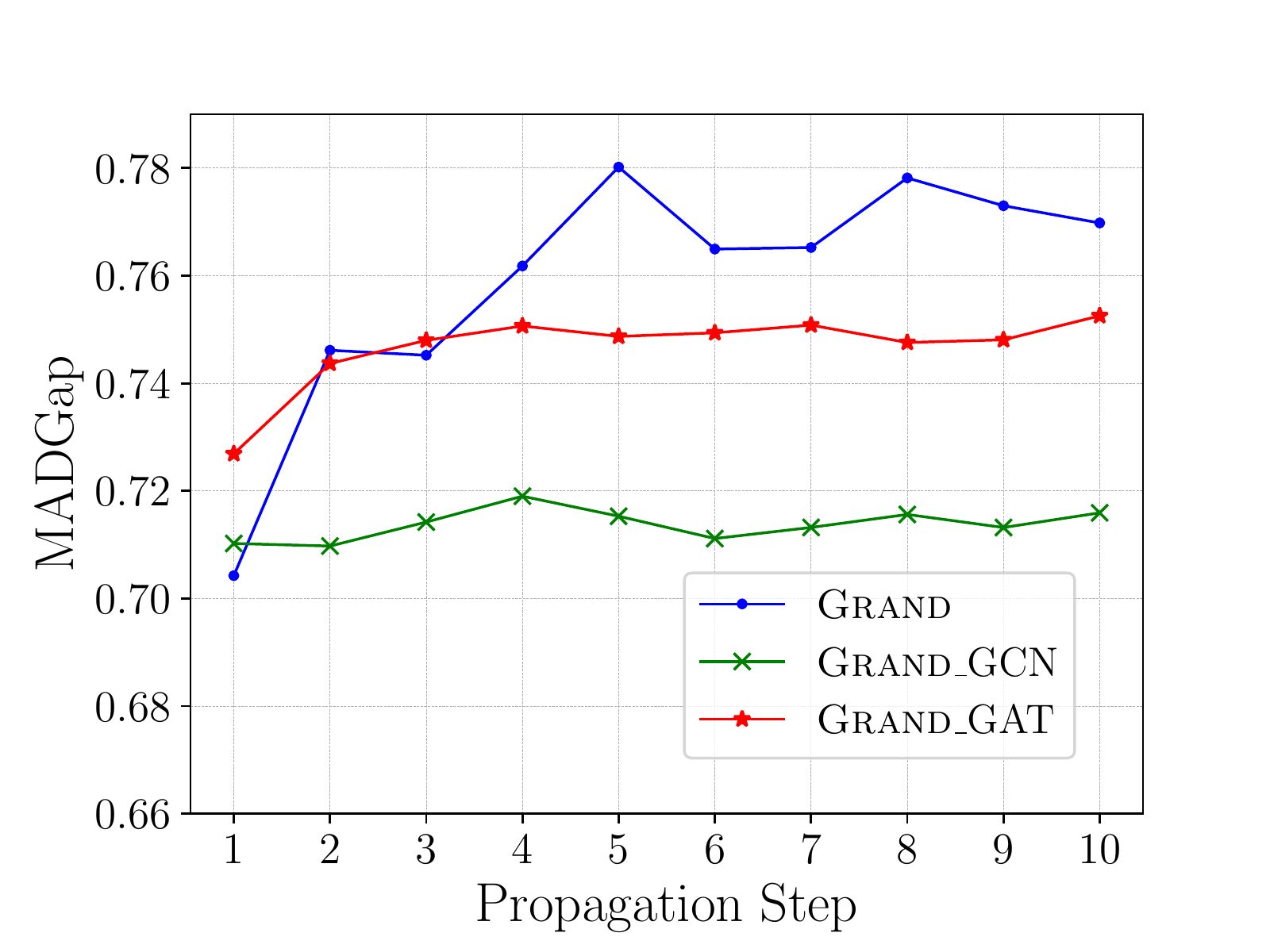}
			}
		\end{subfigure}
	}
	\caption{Comparisons among \model, \model\_GCN and \model\_GAT under different propagation steps.}
		\label{fig:grandgcn_vs_mlp}
\end{figure}
The over-smoothing problem, i.e., \textit{nodes with different labels  become indistinguishable}, exists in many GNNs when enlarging the feature propagation step. Here we provide several quantitative analyses to show \model\ has a lower risk of over-smoothing. Here we adopt MADGap~\cite{chen2019measuring} to measure the over-smoothness of node representations. That is, the cosine distance gap between the remote nodes and neighboring nodes. Hence the smaller MADGap value indicates the more indistinguishable node representations and the more severe over-smoothing issue.
 Here the remote nodes are defined as the nodes have different labels, neighboring nodes are the nodes with the same labels. 
 
 We first compare \model\ with vanilla GCN and GAT on Cora dataset. For each method, we examine both the classification results and MADGap of hidden representations of the last layer under different propagation steps. For GCN and GAT, we adjust the propagation step by adding or removing hidden layers. While in \model, the propagation step is controlled by the hyperparameter $K$. The results are shown in Figure \ref{fig:grandmlp_vs_gcn}. For \model, we report the results with and without using consistency regularization loss.
 As the propagation step increases, we observe the classification performance of GCN and GAT decrease dramatically because of the over-smoothing issue, and the corresponding values of MADGap nearly reduce to zero. While for \model, the performance and MADGap increase gradually when enlarging the propagation steps. This indicates the proposed \model\ model is much more powerful to relieve over-smoothing compared with traditional representative GNNs.

 As shown in Table \ref{tab:classification}, \model\_GCN and \model\_GAT get worse performances than \model, indicating GCN and GAT perform worse than MLP with the random propagation module. Here we conduct a series of experiments analyze this phenomenon. 
 Specifically, we fix the classifier layer numbers of  \model, \model\_GCN and \model\_GAT and compare their performance under different values of propagation step $K$. 
 The results are shown in Figure \ref{fig:grandgcn_vs_mlp}. We find \model\_GCN and \model\_GAT achieve better performance when $K$ is small. However, as the propagation step increases, their results have little improvements or even decreases, and \model\ exceeds the them gradually. On the other hand, different from \model,
 the MADGap values of \model\_GCN and \model\_GAT have no significant increment when increasing the value of propagation step. Considering the main difference among these classifiers is that GCN and GAT have an extra propagation in each layer, we conclude that the built-in propagation in GNNs will bring additional risk of over-smoothing if node features are fully-propagated previously.
 
 
 

\hide{
\begin{figure*}[t]
	\centering
	\mbox
	{
		\hspace{-0.1in}
		\begin{subfigure}[Cora]{
				\centering
				\includegraphics[width = 0.25 \linewidth]{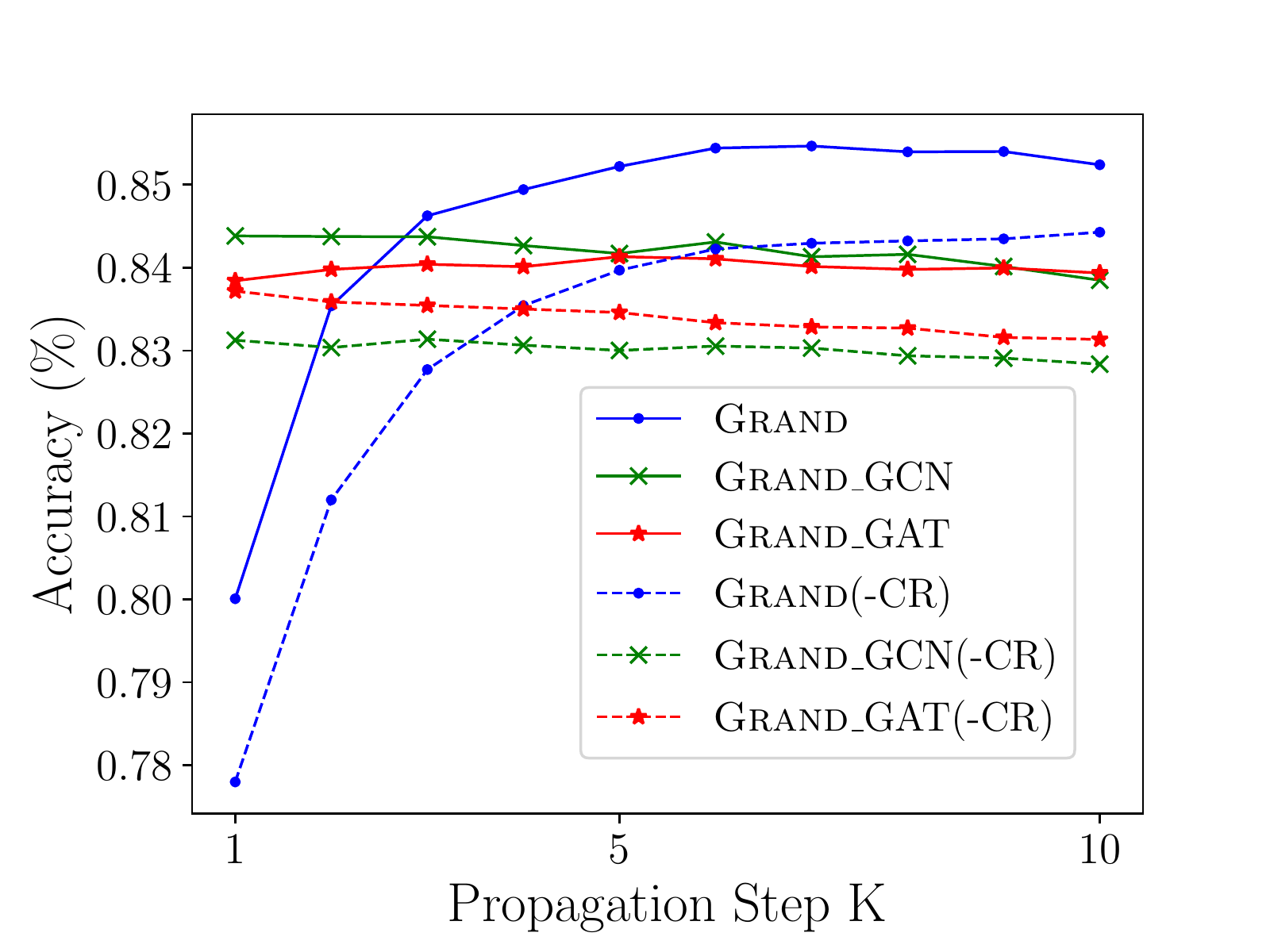}
				\label{fig:mlp_vs_gcn_cora}
			}
		\end{subfigure}
		\hspace{-0.1in}
		\begin{subfigure}[Citeseer]{
				\centering
				\includegraphics[width = 0.25 \linewidth]{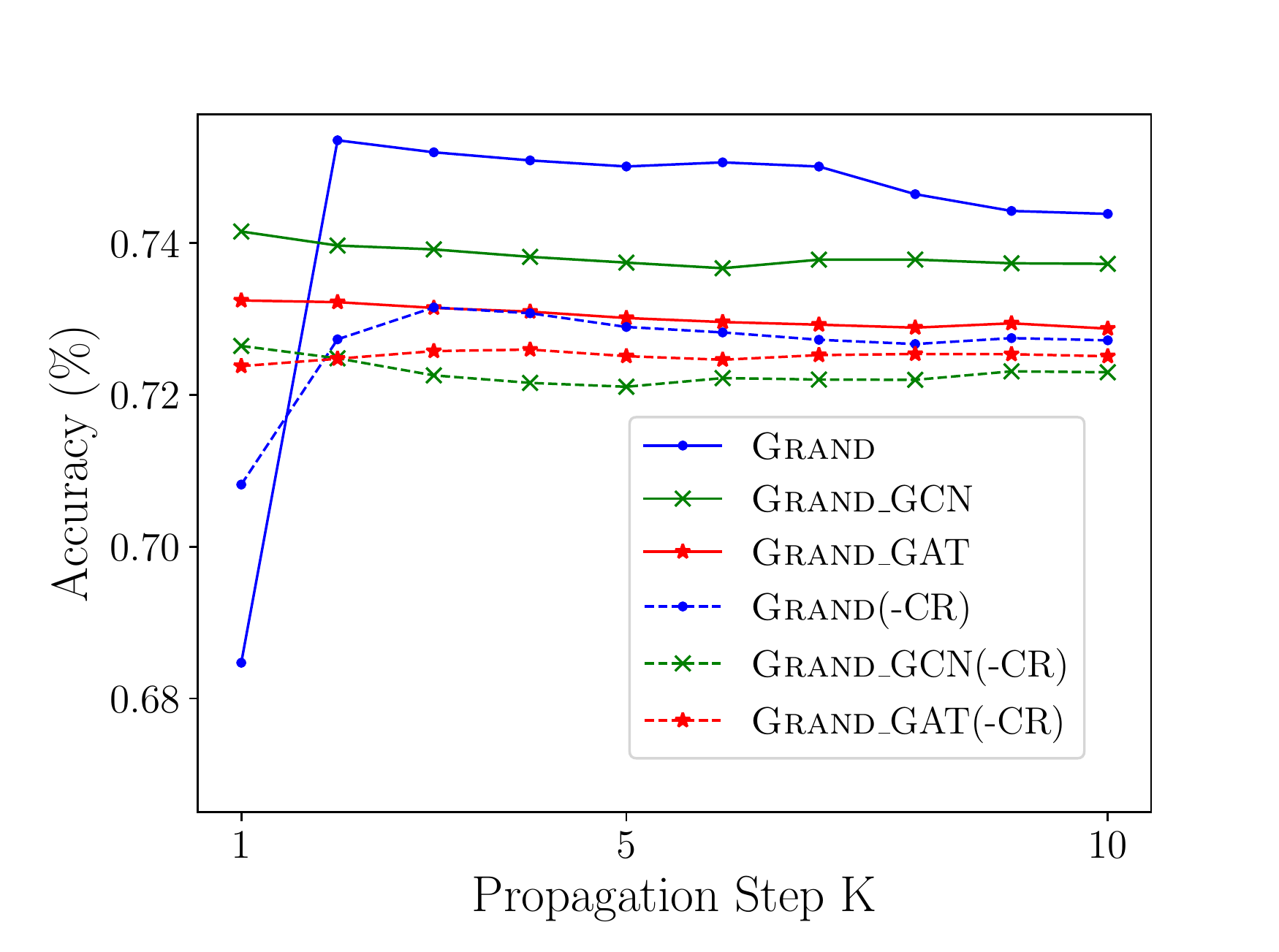}
				\label{fig:mlp_vs_gcn_cite}
			}
		\end{subfigure}
		\hspace{-0.1in}
		\begin{subfigure}[Pubmed]{
				\centering
				\includegraphics[width = 0.25 \linewidth]{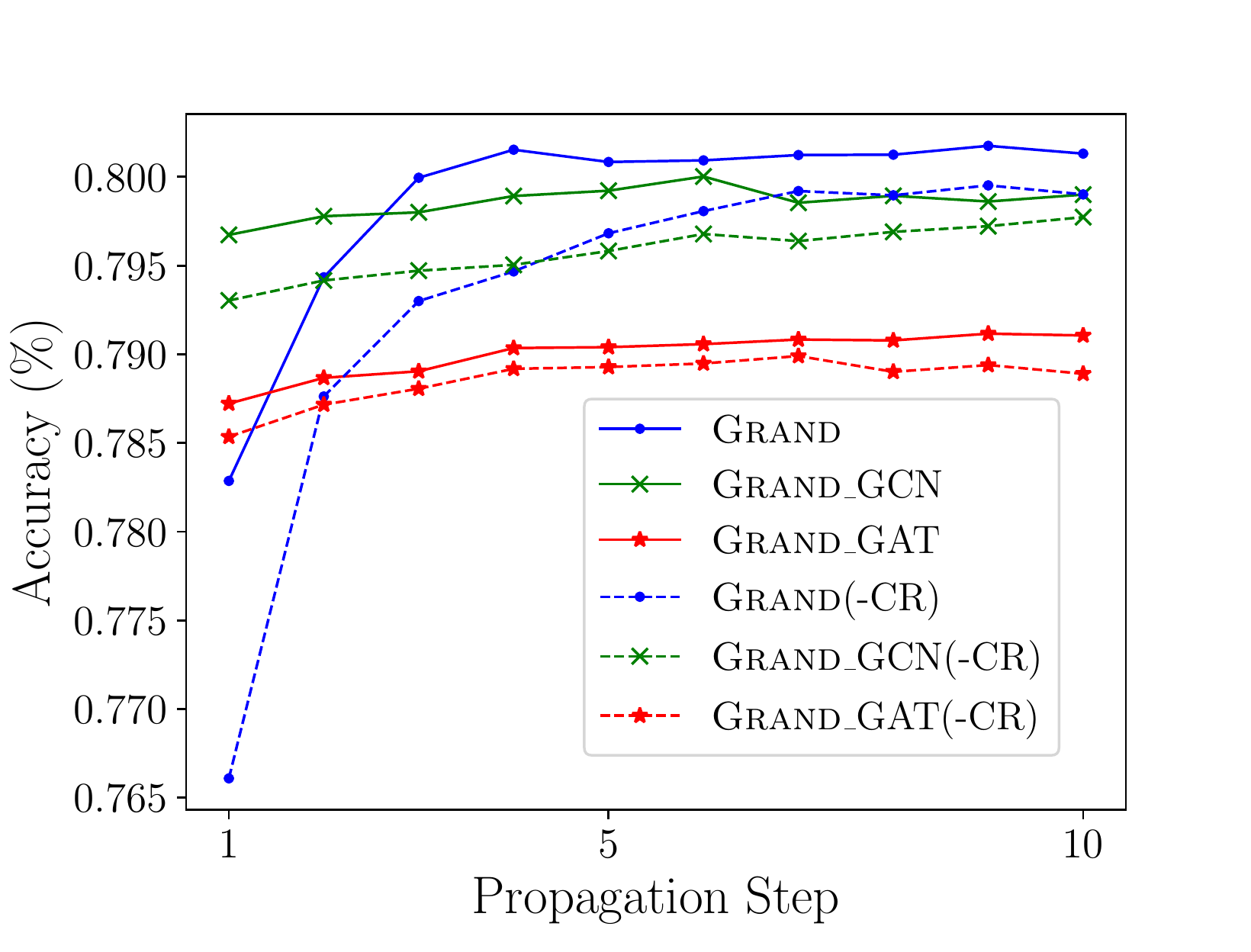}
				\label{fig:mlp_vs_gcn_pub}
			}
		\end{subfigure}
	}
	
	\caption{Performance of \model, \model\_GCN and \model\_GAT under different propagation step K\protect\footnotemark[3].}
	\label{fig:mlp_vs_gcn}
\end{figure*}
\footnotetext[3]{``(-CR)'' denotes not using consistency regularization loss in training.}
}

\subsection{Generalization Analysis}
\begin{figure}[t]
	\centering
	\mbox
	{
			 \hspace{-0.15in}
		\begin{subfigure}[\model]{
				\centering
				\includegraphics[width = 0.35 \linewidth]{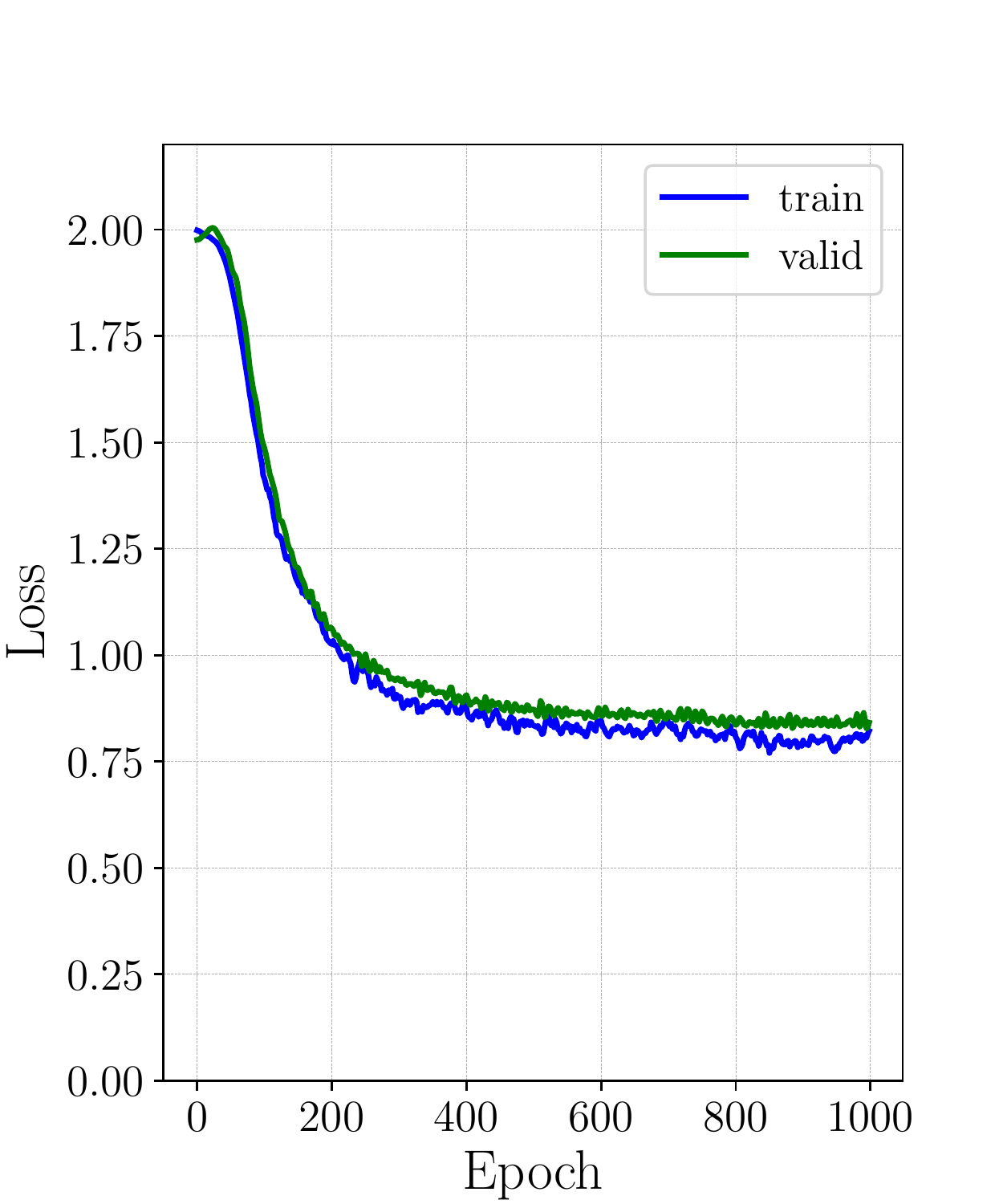}
			}
		\end{subfigure}
			 \hspace{-0.15in}
		\begin{subfigure}[Without CR]{
				\centering
				\includegraphics[width = 0.35 \linewidth]{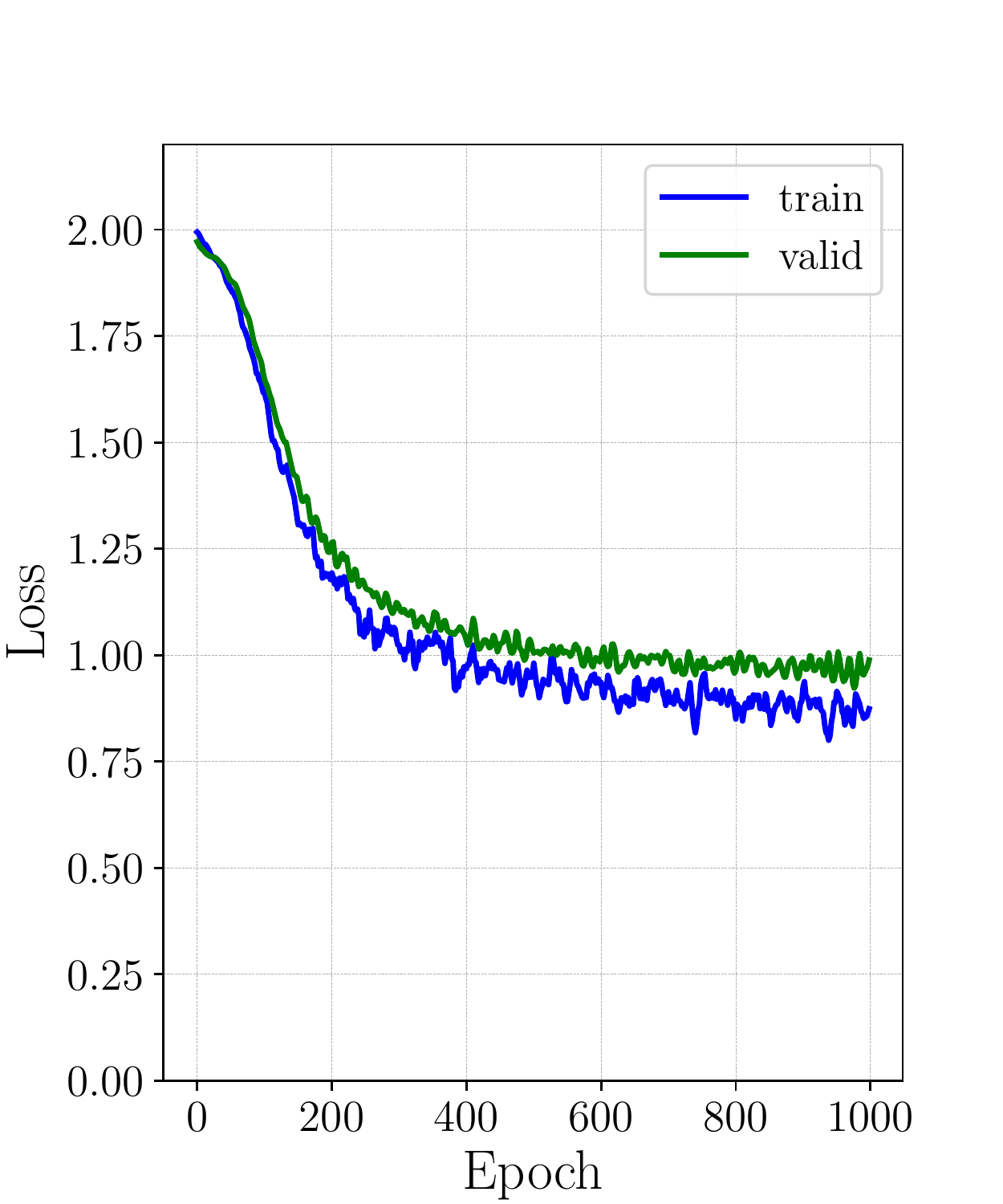}
			}
		\end{subfigure}
	 \hspace{-0.15in}
		\begin{subfigure}[Without CR and dropnode]{
				\centering
				\includegraphics[width = 0.35 \linewidth]{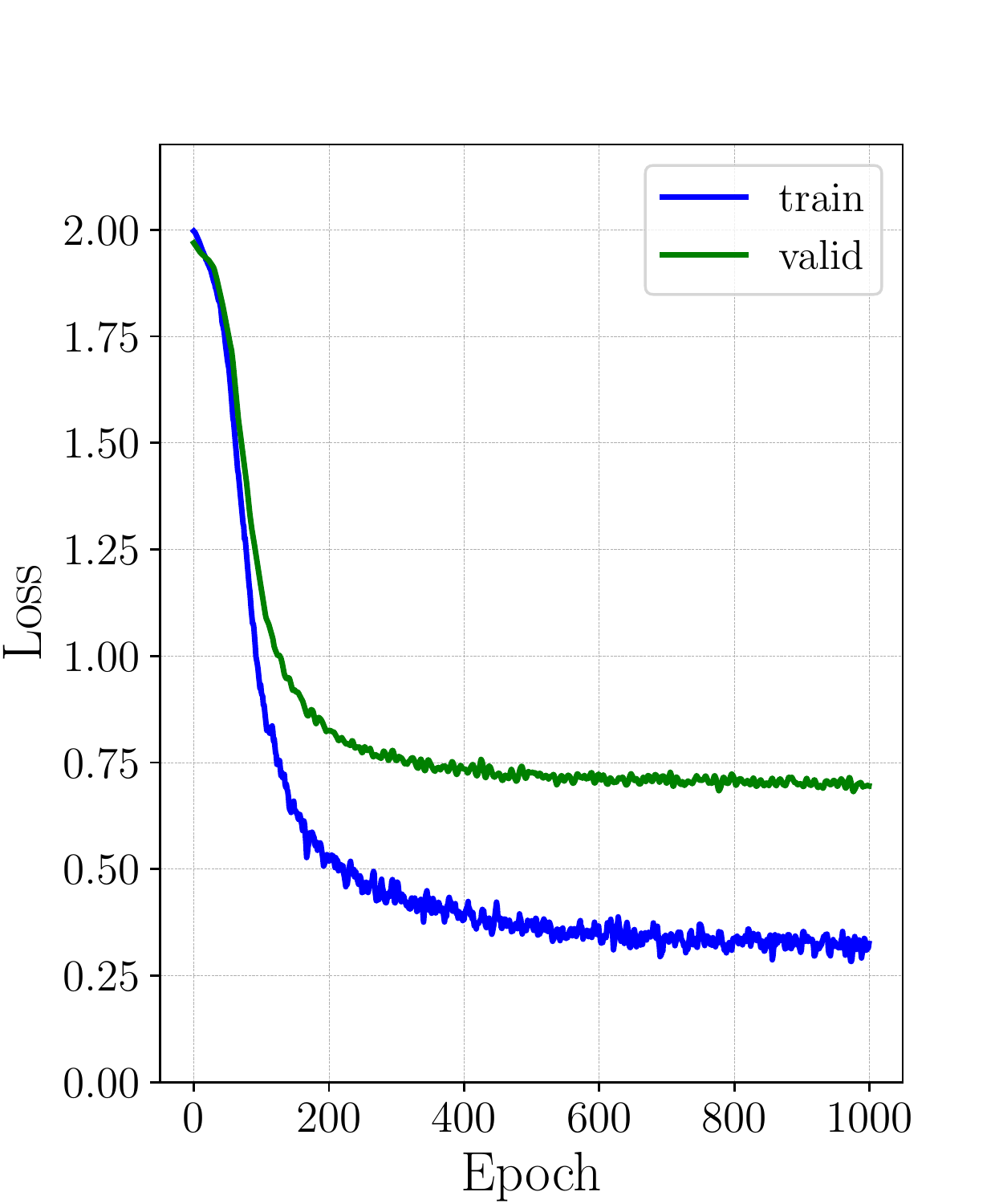}
			}
		\end{subfigure}
	}
	\caption{Training loss and validation loss in different settings on Cora.}
	\label{fig:loss}
\end{figure}
In this section, we examine the effect of random propagation and consistency regularization on improving generalization capacity. Using Cora dataset, we analyze model's cross-entropy losses on training set and validation set in three different settings: the original \model, \model\ without consistency regularization loss (without CR), \model\ without consistency regularization loss and dropnode (without CR and dropnode).
The results are illustrated in Figure \ref{fig:loss}. As can be seen, without using consistency loss and dropnode, model's training losses are much lower than the validation loss, indicating the overfitting problem. When applying the dropnode (without CR), the gap between training loss and validation loss becomes smaller. And when further adding the consistency regularization loss, training loss and validation loss become much closer and more stable, suggesting both random propagation and consistency regularization can improve model's generalization capability. 

\subsection{Efficiency Analysis}
\begin{figure}[t]
	\centering
	\mbox
	{
		\hspace{-0.1in}
		\begin{subfigure}[Per-epoch Training Time]{
				\centering
				\includegraphics[width = 0.5 \linewidth]{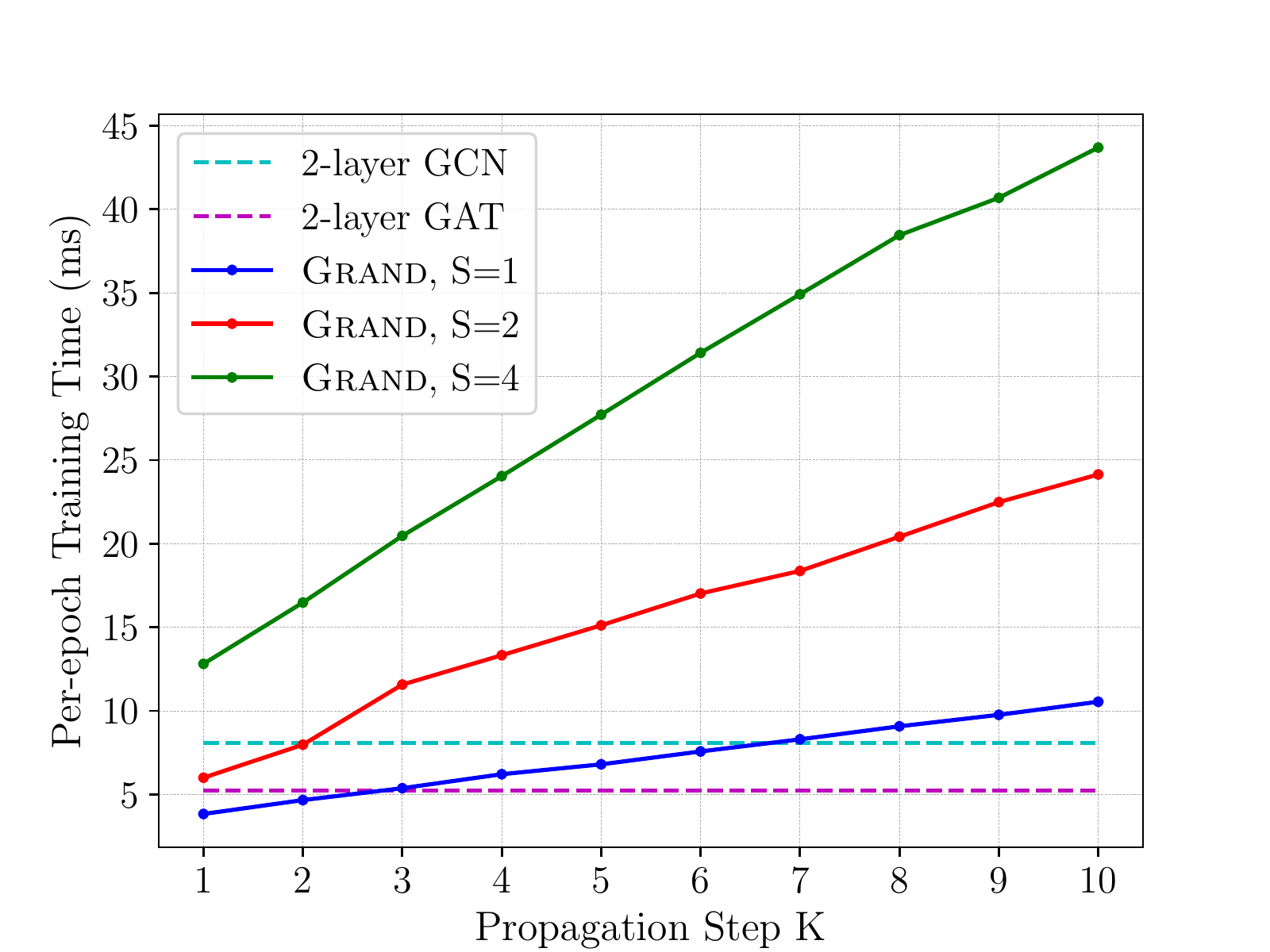}
			}
		\end{subfigure}
		\hspace{-0.1in}
		\begin{subfigure}[Classification Accuracy]{
				\centering
				\includegraphics[width = 0.5 \linewidth]{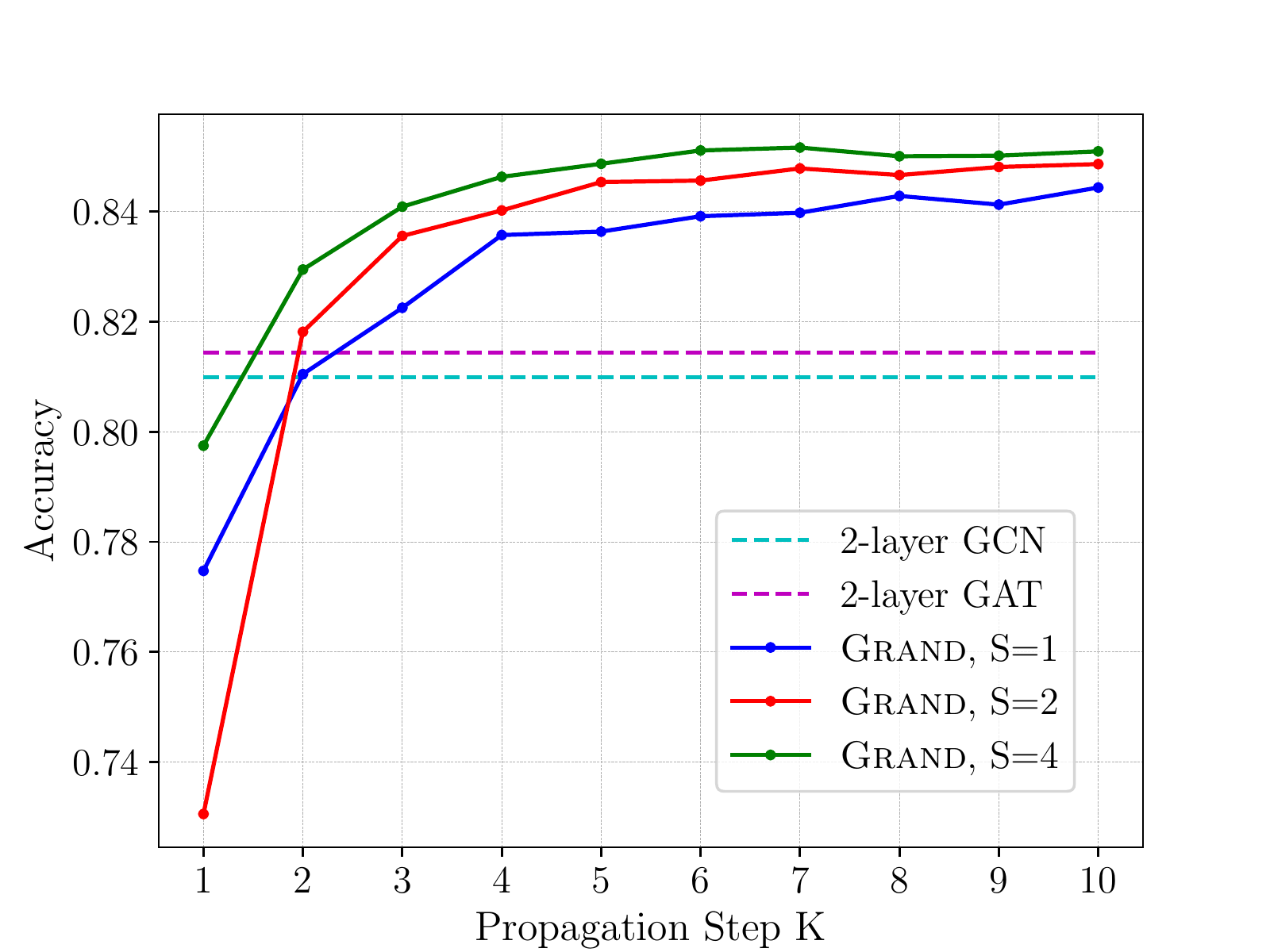}
			}
		\end{subfigure}
	}
	\caption{Efficiency Analysis for  \model.}
		\label{fig:efficiency}
\end{figure}

The efficiency of \model\ is mainly influenced by two hyperparameters: propagation step $K$ and augmentation times $S$. Figure \ref{fig:efficiency} presents the average per-epoch training time and classification accuracy of \model\ under different values of $K$ and $S$ when the number of training epochs is fixed to 1000 on Cora. We also report the same results of two-layer GCN and two-layer GAT with the same learning rate and hidden layer size as \model. As we can see, either increasing $K$ or increasing $S$ can promote model's classification accuracy, but it will also slow down the training efficiency. In practical applications, we can the adjust the values of $K$ and $S$ to balance the trade-off between performance and efficiency. We also observe that \model\ can outperform GCN and GAT with a lower time overhead when $K$ and $S$ are set to appropriate values, e.g., $K=2, S=1$ vs. GCN and $K=5, S=1$ vs. GAT. 

\hide{
\subsection{Parameter Analysis}
In this section, we analyze how the hyperparameters influence the performance of the \model\ framework. Specifically, we conduct parameter analyses for propagation step $K$, sampling times $S$, sharpeness temperature $T$, drop rate $\sigma$ and unsupervised consistency loss coefficient $\lambda$ on three benchmark datasets. When performing parameter analysis on a specific parameter, we fix other parameters and vary the parameter value to examine the performance of \model.\\ 

\vpara{Propagation Step $K$.} Random propagation layer lets each node randomly interact with its multi-hop neighbors, which acts as a neighborhoods-aware regularization methods based on the theoretical analyses in Section \ref{Sec:theory}. As the number of propagation steps increases, the model will incorporate more information of higher-order neighborhoods into prediction and regularization. Here we analyze the influence of propagation step $K$ to the performance by varying $K$ from 0 to 4. The results is shown in Figure 
\ref{fig:step}. When the classifier is chosen as MLP, i.e., \model\_MLP, the  classification accuracy is greatly improved as $K$ turns large. However, $K$ is not sensitive to \model\_GCN. This is because that GCNs have propagation operation inherently, which can substitute the function of random propagation layer to some extent.
\\
\vpara{Sampling Times $S$.}
As mentioned in Section \ref{sec:randpro}, random propagation layer offers an economic methods for graph data augmentation, and sampling times $S$ controls the number of augmentations used for unsupervised consistency regularization. To analyze the affect of sampling times $S$ in each epoch, we change the value of $S$ from 1 to 5. The results is presented in Figure \ref{fig:sampling}. As we can see, increasing the value of $S$ will promote model's performance in most cases. 
\\
\vpara{Temperature $T$.}
Temperature $T$ controls the ``sharpness'' of calculated distribution center, which implicitly affects the entropy of model's prediction distribution. Setting $T$ to a small value is equivalent to minimize the  model's output entropy (Cf. \ref{sec:consis}). Here we change the value of $T$ from $0.1$ to $1$. The corresponding results are shown in Figure \ref{fig:temp}. We observe that the sharpness trick can really help promote model's prediction performance. 
\\
\vpara{Drop Rate $\delta$.} Drop rate $\delta$ determines the proportion of information lost each time of sampling. To explore the influence of drop rate, we report the classification performance of models when drop rate changes from 0 to 0.9 in Figure \ref{fig:droprate}. We observe that models achieve best results when the droprate is set around 0.5. This is because the augmentations produced by random propagation layer in that case are more stochastic and will make model generalize better with the help of consistency regularization. What's more, by comparing dropnode with dropout, we can see that dropnode is more sensitive but always achieve better performance under optimal droprate.
}
\hide{
\subsection{Unsupervised Setting.}
\begin{table}
	\caption{Classification Accuracy in Unsupervised setting(\%).}
	\label{tab:res_1}
	\setlength{\tabcolsep}{1mm}\begin{tabular}{c|ccccc}
		\toprule
		Avaliable data &  Method &Cora & Citeseer & Pubmed \\
		\midrule
        $\mathbf{X}$ & Raw features & 47.9$\pm$0.4 & 49.3$\pm$0.2 & 69.1$\pm$0.3 \\
         $\mathbf{A}$ & DeepWalk & 67.2 & 43.2 & 65.3 \\
         $\mathbf{X},\mathbf{A}$ & DeepWalk + features& 70.7$\pm$0.6 & 51.4$\pm$0.5 & 74.3$\pm$0.9 \\ \hline

		$\mathbf{X},\mathbf{A}$&DGI   & \textbf{82.3$\pm$0.6} & \textbf{71.8$\pm$0.7} & 76.8$\pm$0.6  \\
		$\mathbf{X},\mathbf{A}$&\model\_{GCN}   & $82.2\pm$0.2 & 71.2$\pm$0.3 & \textbf{79.2$\pm$0.3}  \\ \hline
		$\mathbf{X},\mathbf{A}, \mathbf{Y}$& GCN & 81.5 &70.3 & 79.0 \\ 
		$\mathbf{X},\mathbf{A}, \mathbf{Y}$& Planetoid & 75.7 &64.3 & 77.2 \\ 

		\bottomrule
	\end{tabular}
\end{table}
}


}
\section{Conclusions}
\label{sec:conclusion}

In this work, we study the problem of semi-supervised learning on graphs and present the \full\ (\model). 
In \model, we propose the random propagation strategy to stochastically generate multiple graph data augmentations, based on which we utilize consistency regularization to improve the model's generalization on unlabeled data. 
We demonstrate its consistent performance superiority over fourteen state-of-the-art GNN baselines on benchmark datasets. 
In addition, we theoretically illustrate its properties and empirically demonstrate its advantages over conventional GNNs in terms of robustness and resistance to over-smoothing. 
To conclude, the simple and effective ideas presented in \model\ may generate a different perspective in GNN design, in particular for semi-supervised graph learning. In future work, we aim to further improve the scalability of \model\ with some sampling methods.





\hide{
In this work, we propose a highly effective method for for semi-supervised learning on graphs. 
Unlike previous models following a recursive deterministic aggregation process, we explore graph-structured data in a random way and the random propagation method can generate data augmentations efficiently and achieve a model by implicit ensemble with a much lower generalization loss. In each epoch, we utilize consistency regularization to minimize the differences of prediction distributions among multiple augmented unlabeled nodes. We further reveal the effects of random propagation and consistency regularization in theory. 
\model\ achieves the best results on several benchmark datasets of semi-supervised node classification.
Extensive experiments  also suggest that \model\ has better robustness and generalization,  as well as lower risk of over-smoothing. 
Our work may be able to draw more academic attention to the randomness in graphs, rethinking what makes GNNs work and perform better.

}

\hide{

Importantly, these refinements can be concisely concluded as the diagonal matrix and adaptive encoders, resulting in our \rank\ model.  
Second, we empirically reveal information redundancy  in graph-structured data and propose a graph dropout-based sampling mechanism for the ensemble of exponential subgraph models, resulting in our \nsgcn\ model. 

Excitingly, our node ranking-aware propagation can cover many network attention mechanisms, including node attention, (multiple-hop) edge attention, and path attention, and especially, some existing attention-based GCNs, e.g., GAT, can be reformulated as special cases. Some previous network sampling-based GCNs, e.g., GraphSAGE and FastGCN, may also benefit from our network ensemble and disagreement minimization ideas. 
The graph convolutional modules proposed in this work, including the diagonal matrix module, node-wise non-linear activation encoders, and the training and ensemble framework of sampling-based GCNs, can be reused in other graph neural architectures, and we would like to explore these applications of our model for future work.

Our primary contribution to graph convolutional networks originates from neighborhood aggregation and network sampling. 
There has also been a lot of progress in improving GCNs in various ways. 
For example, taking adversarial attack on graphs into account can make GCNs more robust~\cite{zugner2018adversarial, dai2018adversarial}; 
leveraging sub-graph selection algorithms can address the memory and computational resource problems on much larger-scale graph data~\cite{gao2018large}. 
} 

\clearpage

\section*{Broader Impact}

Over the past years, GNNs have been extensively studied and widely used for semi-supervised graph learning, with the majority of efforts devoted to designing advanced and complex GNN architectures. 
Instead of heading towards that direction, our work focuses on an alternative perspective by examining whether and how simple and traditional machine learning (ML) techniques can help overcome the common issues that most GNNs faced, including over-smoothing, non-robustness, and weak generalization.

Instead of the nonlinear feature transformations and advanced neural techniques (e.g., attention), the presented \model\ model is built upon dropout (and its simple variant), linear feature propagation, and consistency regularization---the common ML techniques. 
Its consistent and significant outperformance over 14 state-of-the-art GNN baselines demonstrates the effectiveness of our alternative direction. 
In addition, our results also echo the recent discovery in SGC~\cite{wu2019simplifying} to better understand the source of GCNs' expressive power. 
More importantly, these simple ML techniques in \model\ empower it to be more robust, better avoid over-smoothing, and offer stronger generalization than GNNs. 

In light of these advantages, we argue that the ideas in \model\ offer a different perspective in understanding and advancing GNN based semi-supervised learning. 
For future research in GNNs, in addition to designing complex architectures, we could also invest in simple and traditional graph techniques under the regularization framework which has been widely used in (traditional) semi-supervised learning.


 \hide{

Despite many efforts have been devoted to advancing graph neural networks (GNNs), the existing GNNs suffer from the limitations of non-robustness and over-smoothing. For example, it has been shown that GNNs are very susceptible to adversarial attacks, and the attacker can indirectly attack the target node by manipulating long-distance neighbors~\cite{zugner2018adversarial}. In our work, we show that we can greatly mitigate these problems through some simple operations. This 
might bring new inspirations in advancing GNNs. What's more, the proposed \model\ empirically shows  good  robustness to noise, which will also help downstream applications better cope with complex and noisy data in the real world.

In \model, we propose random propagation, an effective graph data augmentation method. With the help of consistency regularization, this strategy serves as a general technique to improve GNNs' generalization ability to unseen samples. This technique can also be applied into self-supervised learning on graphs by combining the recent developments of contrastive learning~\cite{He2019Momentum}.

 \model\ achieves strong results in  semi-supervised learning on graphs.  
 Compared with supervised methods, \model\ needs fewer labeled data during training. This really facilities many applications since labeled data is usually costly to obtain in the real world. On the other hand, in some specific scenarios (like finance, health), the data label often contains the user's private information.  Thus each access to training data means a potential leakage of user privacy. From this point of view, \model\ can better protect the privacy of users when applied to these applications. 
 
 }
 
 \hide{
Graph-structured data is ubiquitous in the real world. In this paper, we study the problem of semi-supervised learning on graphs, aiming to learn from a large amount of samples with only a few supervision information. We provide a simple method called \model\ to solve this problem. The impacts of our work lie in two different aspects:

 First, with the help of random propagation and consistency regularization, \model\ beats other advanced GNNs with only using a simple MLP model as classifier. 
This brings this research area new insight --- Except for exploring complex model architectures, we could alternatively improve GNNs by designing powerful regularization methods.

  Second, as a semi-supervised learning method, \model\  also will also bring many practical impacts. Compared with supervised methods, \model\ needs fewer labeled data than supervised method during training. This really facilities many applications since labeled data is usually costly to obtain in the real world. On the other hand, in some specific scenarios (like finance, health), the data label often contains the user's private information.  Thus each access to training data means a potential leakage of user privacy. From this point of view, \model\ can better protect the privacy of users when applied to these applications. What's more, \model\ empirically shows  good  robustness to noise, which will also help downstream applications better cope with complex and noisy data in the real world.
}
\hide{
\begin{ack}
	Use unnumbered first level headings for the acknowledgments. All acknowledgments
	go at the end of the paper before the list of references. Moreover, you are required to declare 
	funding (financial activities supporting the submitted work) and competing interests (related financial activities outside the submitted work). 
	More information about this disclosure can be found at: \url{https://neurips.cc/Conferences/2020/PaperInformation/FundingDisclosure}.

	Do {\bf not} include this section in the anonymized submission, only in the final paper. You can use the \texttt{ack} environment provided in the style file to autmoatically hide this section in the anonymized submission.
\end{ack}
}

\bibliographystyle{plain}
\bibliography{main.bib}

\clearpage
\begin{appendices}

\section{Reproducibility}
\label{sec:reproduce}

\subsection{Datasets Details}
Table~\ref{tab:data} summarizes the statistics of the three benchmark datasets --- Cora, Citeseer and Pubmed.  Our preprocessing scripts for Cora, Citeseer and Pubmed is implemented with reference to the codes of Planetoid~\cite{yang2016revisiting}. We use exactly the same experimental settings---such as features and data splits---on the three benchmark datasets as literature on semi-supervised graph mining~\cite{yang2016revisiting, kipf2016semi, Velickovic:17GAT} and run 100 trials with 100 random seeds for all results on Cora, Citeseer and Pubmed reported in Section 4. We also evaluate our method on six publicly available and large datasets, the statistics and results are summarized in Appendix~\ref{exp:large_data}.

\begin{table}[h]
\centering
\caption{Benchmark Dataset statistics.}
\small
\label{tab:data}
\setlength{\tabcolsep}{1.0mm}\begin{tabular}{c|ccccc}
	\toprule
	Dataset &  Nodes &  Edges & Train/Valid/Test Nodes & Classes & Features \\
	\midrule
	Cora & 2,708 & 5,429 & 140/500/1,000 & 7&1,433  \\ 
	Citeseer & 3,327 & 4,732 & 120/500/1,000 & 6 & 3,703 \\
	Pubmed & 19,717 & 44,338 & 60/500/1,000 & 3 & 500 \\
	\bottomrule
\end{tabular}
\vspace{-0.1in}
\end{table}
\subsection{Implementation Details}

We make use of PyTorch to implement \model\ and its variants. The random propagation procedure is efficiently implemented with sparse-dense matrix multiplication. The codes of GCN and \model\_GCN are implemented referring to the PyTorch version of GCN~\footnote{\url{https://github.com/tkipf/pygcn}}. As for \model\_GAT and GAT, we adopt the implementation of GAT layer from the PyTorch-Geometric library~\footnote{\url{https://pytorch-geometric.readthedocs.io}} in our experiments. The weight matrices of classifier are initialized with Glorot normal initializer~\cite{glorot2010understanding}.
We employ Adam~\cite{kingma2014adam} to optimize parameters of the proposed methods and adopt early stopping to control the training epochs based on validation loss. Apart from DropNode (or dropout~\cite{srivastava2014dropout})  used in random propagation, we also apply dropout on the input layer and hidden layer of the prediction module used in \model\ as a common practice of preventing overfitting in optimizing neural network.
For the experiments on  Pubmed, we also use batch normalization~\cite{ioffe2015batch} to stabilize the training procedure. All the experiments in this paper are conducted on a single NVIDIA GeForce RTX 2080 Ti with 11 GB memory size. Server operating system is Unbuntu 18.04. As for software versions, we use Python 3.7.3, PyTorch 1.2.0, NumPy 1.16.4, SciPy 1.3.0, CUDA 10.0.



\subsection{Hyperparameter Details}
\vpara{Overall Results in Section \ref{sec:overall}.}
\model\ introduces five additional hyperparameters, that is the DropNode probability $\delta$ in random propagation, propagation step $K$, data augmentation times $S$ at each training epoch, sharpening temperature $T$ when calculating consistency regularization loss and the coefficient of consistency regularization loss $\lambda$  trading-off the balance between  $\mathcal{L}_{sup}$ and $\mathcal{L}_{con}$. In practice, $\delta$ is always set to 0.5 across all experiments. As for other hyperparameters,  we perform hyperparameter search for each dataset. Specifically, we first search $K$ from \{ 2,4,5,6,8\}. With the best selection of $K$, we then search $S$ from \{2,3,4\}. Finally, we fix $K$ and $S$ to the best values and take a grid search for $T$ and $\lambda$ from \{0.1, 0.2, 0.3,0.5\} and \{0.5, 0.7, 1.0\} respectively.
For each search of hyperparameter configuration, we run the experiments with 20 random seeds and select the best configuration of hyperparameters based on average accuracy on validation set. Other hyperparameters used in our experiments includes learning rate of Adam, early stopping patience, L2 weight decay rate, hidden layer size, dropout rates of input layer and hidden layer. We didn't spend much effort to tune these hyperparameters in practice, as we observe that \model\ is not very sensitive with those. 
Table \ref{tab:hyper} reports the best hyperparameters  of \model\ we used for the results reported in Table \ref{tab:overall}.
\begin{table}[h!]
    \centering
\small
 \caption{Hyperparameters of \model\ for results in Table \ref{tab:overall}}
    \begin{tabular}{c|ccc}
    \toprule
    Hyperparameter & Cora & Citeseer& Pubmed \\
 \midrule    
 DropNode probability $\delta$ & 0.5& 0.5 &0.5 \\
Propagation step $K$ &8 &2 & 5\\
Data augmentation times $S$ &4 & 2& 4\\
CR loss coefficient $\lambda$ & 1.0 & 0.7&   1.0\\
Sharpening temperature $T$ & 0.5 & 0.3 & 0.2\\
Learning rate & 0.01& 0.01& 0.2 \\
Early stopping patience &200& 200& 100 \\
Hidden layer size &32 & 32& 32\\
L2 weight decay rate & 5e-4&5e-4 & 5e-4\\
Dropout rate in input layer &0.5 &0.0 &  0.6\\
Dropout rate in hidden layer &0.5 & 0.2& 0.8\\
\bottomrule
 \end{tabular}
 \label{tab:hyper}
 \vspace{-0.1in}
\end{table}

\vpara{Robustness Analysis in Section \ref{sec:robust}.}
For random attack, we implement the attack method with Python and NumPy library. The propagation step $K$ of \model\ (with or without CR) is set to 5. And the other hyperparameters are set to the values in Table \ref{tab:hyper}.
As for Metattack~\cite{zugner2019adversarial}, we use the publicly available implementation\footnote{\url{https://github.com/danielzuegner/gnn-meta-attack}} published by the authors with the same hyperparameters used in the original paper. We observe \model\ (with or without CR) is sensitive to the propagation step $K$ under different perturbation rates. Thus we search $K$ from \{5,6,7,8\} for each perturbation rate. The other hyperparameters are fixed to the values reported in Table \ref{tab:hyper}.

\vpara{Other Experiments.}For the other results reported in Section \ref{sec:overall} --- \ref{sec:oversmoothing}, the hyperparameters used in \model\ are set to the values reported in Table \ref{tab:hyper} with one or two changed for the corresponding analysis. 

\vpara{Baseline Methods.}
For the results of GCN or GAT reported in Section \ref{sec:robust} --- \ref{sec:oversmoothing}, the learning rate is set to 0.01, early stopping patience is 100, L2 weight decay rate is 5e-4, dropout rate is 0.5. The hidden layer size of GCN is 32. For GAT, the hidden layer consists 8 attention heads and each head consists 8 hidden units.


\section{Theorem Proofs}
\hide{
\subsection{Random propagation w.r.t supervised classification loss}
\label{sec:suploss}
We discuss the regularization of random propagation w.r.t the superivised classification loss. We follow the assumption settings expressed in Section 3.3. 
Then the supervised classification loss is:
$
\small
\mathcal{L}_{sup} =\sum_{i=0}^{m-1} -y_i\log(\tilde{z}_i) - (1-y_i)\log(1-\tilde{z}_i).
$ 
Note that $\mathcal{L}_{sup}$ refers to the perturbed classification loss with DropNode on the node features. 
By contrast, the original (non-perturbed) classification loss is defined as:
$
\small
\mathcal{L}_{org} =\sum_{i=0}^{m-1} -y_i\log(z_i) - (1-y_i)\log(1-z_i),
$
where $z_i = \text{sigmoid}(\overline{\mathbf{A}}_i \mathbf{X} \cdot \mathbf{W})$ is the output with the original feature matrix $\mathbf{X}$. 
Then we have the following theorem with proof in Appendix \ref{them:proof1}. 

\textbf{Theorem 2.}
\textit{
	In expectation, optimizing the perturbed classification loss $\mathcal{L}_{sup}$ is equivalent to optimize the original loss $ \mathcal{L}_{org}$ with an extra  regularization term $ \mathcal{R}(\mathbf{W})$, which has a quadratic approximation form  $  \mathcal{R}(\mathbf{W})\approx \mathcal{R}^q(\mathbf{W})= \frac{1}{2}\sum_{i=0}^{m-1} z_i(1-z_i) \textnormal{Var}_\epsilon \left(\overline{\mathbf{A}}_i \widetilde{\mathbf{X}} \cdot \mathbf{W} \right)$.}

This theorem suggests that DropNode brings an extra regularization loss to the optimization objective. Expanding the variance term, this extra quadratic regularization loss can be expressed as: 
\begin{equation}
\small
\label{equ:Rq}
\begin{aligned}
\mathcal{R}^q_{DN}(\mathbf{W}) 
& = \frac{1}{2}\frac{\delta}{1-\delta} \sum_{j=0}^{n-1} \left[(\mathbf{X}_j \cdot \mathbf{W})^2  \sum_{i=0}^{m-1} (\overline{\mathbf{A}}_{ij})^2\ z_i(1-z_i)\right].
\end{aligned}
\end{equation}

Different from $\mathcal{R}^c_{DN}$ in Eq. 5, the inside summation term in Eq. \ref{equ:Rq} only incorporates the first $m$ nodes, i.e, the labeled nodes. 
Thus, \textit{the dropnode regularization with supervised classification loss can enforce the consistency of the classification confidence between each node and its labeled multi-hop neighborhoods}.

}

\subsection{Proof for Theorem \ref{thm1}}
\label{them:proof1}
\begin{proof}

The expectation of $\mathcal{L}_{con}$ is:
\begin{equation}
\small
\begin{aligned}
\frac{1}{2}\sum_{i=0}^{n-1} \mathbb{E} \left[(\tilde{z}_i^{(1)}-\tilde{z}_i^{(2)})^2\right] = \frac{1}{2}\sum_{i=0}^{n-1} \mathbb{E}  \left[\left((\tilde{z}_i^{(1)} - z_i) - (\tilde{z}_i^{(2)} -z_i)\right)^2\right].
\end{aligned}
\end{equation}
Here $z_i = \text{sigmoid}(\overline{\mathbf{A}}_i\mathbf{X}\cdot \mathbf{W})$, $\tilde{z}_i = \text{sigmoid}(\overline{\mathbf{A}}_i\widetilde{\mathbf{X}}\cdot \mathbf{W})$. For the term of $\tilde{z}_i - z_i$, we can approximate it with its first-order Taylor expansion around $\overline{\mathbf{A}}_i\mathbf{X}\cdot \mathbf{W}$, i.e., $\tilde{z}_i - z_i \approx z_i(1-z_i)(\overline{\mathbf{A}}_i(\widetilde{\mathbf{X}} - \mathbf{X}) \cdot \mathbf{W})$.
Applying this rule to the above equation, we have:
\begin{equation}
\small
\begin{aligned}
\frac{1}{2}\sum_{i=0}^{n-1} \mathbb{E} \left[(\tilde{z}_i^{(1)}-\tilde{z}_i^{(2)})^2\right]
&\approx \frac{1}{2}\sum_{i=0}^{n-1} z^2_i(1-z_i)^2 \mathbb{E} \left[(\overline{\mathbf{A}}_i(\widetilde{\mathbf{X}}^{(1)} - \widetilde{\mathbf{X}}^{(2)})\cdot \mathbf{W})^2\right] \\
&= \sum_{i=0}^{n-1} z^2_i(1-z_i)^2 \text{Var}_\epsilon \left(\overline{\mathbf{A}}_i \widetilde{\mathbf{X}}\cdot \mathbf{W} \right).
\end{aligned}
\end{equation}
\end{proof}
\vspace{-0.2in}

\subsection{Proof for Theorem \ref{thm2}}
\label{them:proof2}
\begin{proof}
Expanding the logistic function, 
$\mathcal{L}_{org}$ is rewritten as:

\begin{equation}
\small
\label{equ_loss}
\begin{aligned}
\mathcal{L}_{org} 
& = \sum_{i=0}^{m-1} \left[ -y_i\overline{\mathbf{A}}_i \mathbf{X}\cdot \mathbf{W} + \mathcal{A}(\overline{\mathbf{A}}_i, \mathbf{X}) \right],
\end{aligned}
\end{equation}
where $\mathcal{A}(\overline{\mathbf{A}}_i, \mathbf{X}) = - \log\left(\frac{\exp(-\overline{\mathbf{A}}_i \mathbf{X} \cdot \mathbf{W})}{1+ \exp(-\overline{\mathbf{A}}_i \mathbf{X} \cdot \mathbf{W})} \right)$. 
Then the expectation of perturbed classification loss can be rewritten as:

\begin{equation}
\small
\begin{aligned}
\mathbb{E}_\epsilon ( \mathcal{L}_{sup} ) 
 = \mathcal{L}_{org}+ \mathcal{R}(\mathbf{W}),
\end{aligned}
\end{equation}
where 
$\mathcal{R}(\mathbf{W}) = \sum_{i=0}^{m-1} \mathbb{E}_\epsilon \left[\mathcal{A}(\overline{\mathbf{A}}_i, \widetilde{\mathbf{X}}) - \mathcal{A}(\overline{\mathbf{A}}_i, \mathbf{X}) \right]$.
Here $\mathcal{R}(\mathbf{W})$ acts as a regularization term for $\mathbf{W}$. To demonstrate that, we can take a second-order Taylor expansion of $\mathcal{A}(\overline{\mathbf{A}}_i,\widetilde{\mathbf{X}})$ around $\overline{\mathbf{A}}_i \mathbf{X}\cdot \mathbf{W} $:

\begin{equation}
\small
\begin{aligned}
 \mathbb{E}_\epsilon \left[ \mathcal{A}(\overline{\mathbf{A}}_i, \widetilde{\mathbf{X}}) -  \mathcal{A}(\overline{\mathbf{A}}_i, \mathbf{X}) \right]  
 \approx \frac{1}{2} \mathcal{A}^{''}(\overline{\mathbf{A}}_i, \mathbf{X}) 
\text{Var}_\epsilon \left(\overline{\mathbf{A}}_i \widetilde{\mathbf{X}}\cdot \mathbf{W}\right).
\end{aligned}
\end{equation}
Note that the first-order term $\mathbb{E}_\epsilon\left[ \mathcal{A}^{'}(\overline{\mathbf{A}}_i, \mathbf{X})(\widetilde{\mathbf{X}}-\mathbf{X})\right]$ vanishes since $\mathbb{E}_\epsilon( \widetilde{\mathbf{X}}) = \mathbf{X}$. We can easily check that 
$\mathcal{A}^{''}(\overline{\mathbf{A}}_i, \mathbf{X}) 
= z_i (1-z_i)$.
Applying this quadratic approximation to $\mathcal{R}(\mathbf{W})$
, we get the quadratic approximation form of $\mathcal{R}(\mathbf{W})$:
\begin{equation}
\mathcal{R}(\mathbf{W}) \approx \mathcal{R}^q(\mathbf{W}) = \frac{1}{2} \sum_{i=0}^{m-1}  z_i(1-z_i) \text{Var}_\epsilon (\overline{\mathbf{A}}_i \widetilde{\mathbf{X}}\cdot\mathbf{W}).
\end{equation}
\vspace{-0.05in}
\end{proof}
\hide{
\subsection{Training Algorithm of \model}
\label{sec:alg}
 Algorithm~\ref{alg:2} summarizes the training process of \model.
}

\section{Additional Experiments}

\subsection{Results on Large Datasets}
\label{exp:large_data}
 \begin{table}[h!]
\small
\centering
	\caption{Statistics of Large Datasets.}
	\label{tab:large_data}
	\begin{tabular}{crrrr}
		\toprule
		~& Classes& Features & Nodes & Edges\\
		\midrule
		Cora-Full  & 67 & 8,710 & 18,703 & 62,421 \\
		Coauthor CS  & 15 & 6,805&18,333&81,894\\
		Coauthor Physics & 5  & 8,415&34,493 & 247,962 \\
        Aminer CS & 18 & 100 & 593,486 & 6,217,004 \\
        Amazon Computers& 10 & 767 & 13,381 & 245,778   \\
		Amazon Photo& 8 & 745 & 7,487 & 119,043   \\
		\bottomrule
	\end{tabular}
\end{table}

We also evaluate our methods on six relatively large datasets, i.e., Cora-Full, Coauthor CS, Coauthor Physics, Amazon Computers, Amazon Photo and Aminer CS. \label{exp:large_data}
The statistics of these datasets are given in Table \ref{tab:large_data}. Cora-Full is proposed in~\cite{Bojchevski2017DeepGE}. Coauthor CS, Coauthor Physics, Amazon Computers and Amazon Photo are proposed in~\cite{shchur2018pitfalls}. We directly use the processed versions of the five datasets\footnote{\url{https://github.com/shchur/gnn-benchmark}} in our experiments.

Aminer CS is conducted by us based on DBLP citation network\footnote{ \url{https://www.aminer.cn/citation}}. In Aminer CS, each node corresponds to a paper in computer science, and edges represent citation relations between papers. 
We manually categorize these papers into 18 topics based on their publication venues. The corresponding topic of each class is described in Table~\ref{tab:aminer_topic}.
We use averaged GLOVE-100~\cite{pennington-etal-2014-glove} word vector of paper abstract as the node feature vector. The goal is to predict the corresponding topic of each paper based on feature matrix and citation graph structure.


Following the evaluation protocol used in~\cite{shchur2018pitfalls}, we run each model on 100 random train/validation/test splits and 20 random initializations for each split (with \textbf{2000} runs on each dataset in total). For each trial, we choose 20 samples for training, 30 samples for validation and the remaining samples for test. 
We ignore  3 classes with less than 50 nodes in Cora-Full dataset as done in~\cite{shchur2018pitfalls}. 
The results are presented in Table \ref{tab:large_res}. The results of GCN and GAT on the first five datasets are taken from~\cite{shchur2018pitfalls}.
We can observe that \textit{\model\ significantly outperforms GCN and GAT on all these datasets.}

\begin{table}[htbp]
    \centering
 \caption{Topics of Aminer CS.}
    \label{tab:aminer_topic}
    \begin{tabular}{ccc|ccc}
    \toprule
        Class & Topic & \#Nodes & Class & Topic & \#Nodes \\
    \midrule
        0 & Artificial Intelligence& 83172 & 9 & Robotics & 16457 \\
        1 & Information System & 17419 & 10 & Computational Theory &18196 \\
        2 & Parallel Computing & 44841 & 11 &Computer-aided Design & 30514\\
        3 & Computer Network & 137328 & 12 & Computer Vision &35729 \\ 
        4 & Information Security & 18263 & 13 & Natural Language Processing &21034 \\
        5 & Database and Data Mining &  52401 & 14 & Computer Graphics & 6820\\ 
        6 & Software Engineering  & 13631 & 15 & Machine Learning & 24900\\
        7 & Multimedia & 18187 & 16 & Bioinformatics & 8873\\
        8 & Human-computer Interaction&  13970 & 17 & Signal Processing & 31751\\
        \bottomrule
    \end{tabular}
\end{table}
\begin{table}[h!]
\small
\label{tab:res}
\centering
	\caption{Results on large datasets.}
	\label{tab:large_res}
	\setlength{\tabcolsep}{0.5mm}\begin{tabular}{ccccccc}
		\toprule
		Method & \tabincell{c}{Cora \\Full} & \tabincell{c}{ Coauthor\\ CS} & \tabincell{c}{Coauthor\\ Physics} & \tabincell{c}{Amazon \\Computer} & \tabincell{c}{Amazon \\Photo} & \tabincell{c}{Aminer \\ CS}\\
		\midrule
		GCN  & 62.2 $\pm$ 0.6 & 91.1 $\pm$ 0.5 &  92.8 $\pm$ 1.0& 82.6 $\pm$ 2.4 & 91.2 $\pm$ 1.2 & 49.9 $\pm$ 2.0\\
		GAT  & 51.9 $\pm$ 1.5 & 90.5 $\pm$ 0.6 & 92.5 $\pm$ 0.9&78.0 $\pm$ 19.0  & 85.7 $\pm$ 20.3 & 49.6 $\pm$ 1.7\\
        \midrule
		\model & \textbf{63.5 $\pm$0.6} & \textbf{92.9 $\pm$ 0.5} &\textbf{94.6 $\pm$ 0.5} & \textbf{85.7 $\pm$ 1.8}  &\textbf{92.5 $\pm$ 1.7} & \textbf{52.8 $\pm$ 1.2} \\
		\bottomrule
	\end{tabular}
\end{table}

\hide{
\subsection{Robustness Analysis.}
\label{sec:robust}

We study the robustness of the proposed \model\ model. 
Specifically, we utilize the following adversarial attack methods to generate perturbed graphs, and then examine the model's classification accuracies on them. 
\begin{itemize}
\small
	\item \textbf{Random Attack.} Perturbing the structure by randomly adding fake edges.
\item \textbf{Metattack~\cite{zugner2019adversarial}.} Attacking the graph structure by removing or adding edges based on meta learning.
\end{itemize}
Figure \ref{fig:robust} presents the classification accuracies of different methods with respect to different perturbation rates on the Cora dataset. 
We observe that \model\ consistently outperforms GCN and GAT across all perturbation rates on both attacks. 
When adding 10\% new random edges into Cora, we observe only a 7\% drop in classification accuracy for \model, while 12\% for GCN and 37\% for GAT. Under Metattack, the gap between \model\ and GCN/GAT also enlarges with the increase of the perturbation rate. 
This study suggests the robustness advantage of the \model\ model (with or without) consistency regularization over GCN and GAT. 

\begin{figure}[t]
	\centering
	\mbox
	{
		\begin{subfigure}[Random Attack]{
				\centering
				\includegraphics[width = 0.45 \linewidth]{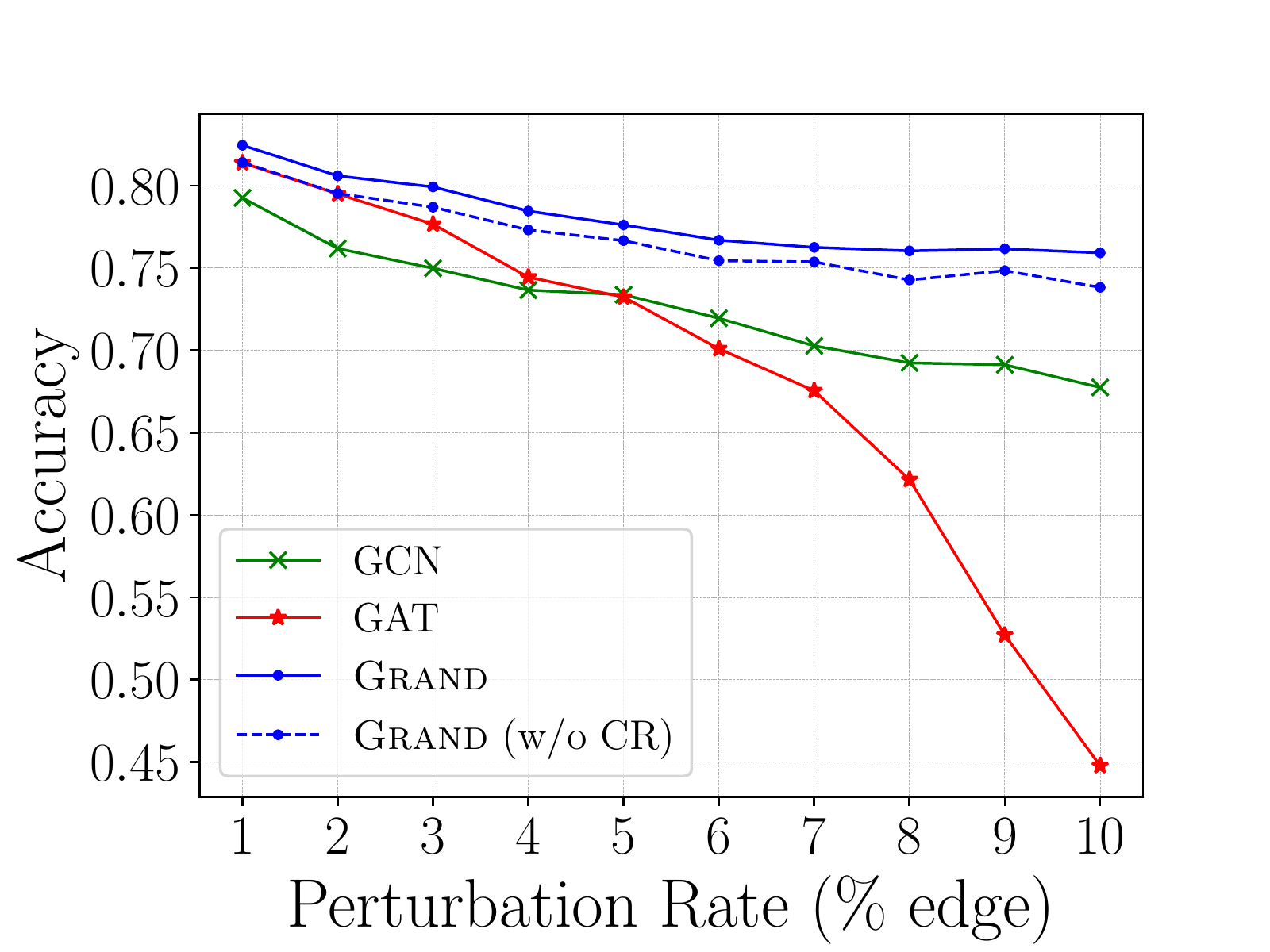}
			}
		\end{subfigure}
		\begin{subfigure}[Metattack]{
				\centering
				\includegraphics[width = 0.45 \linewidth]{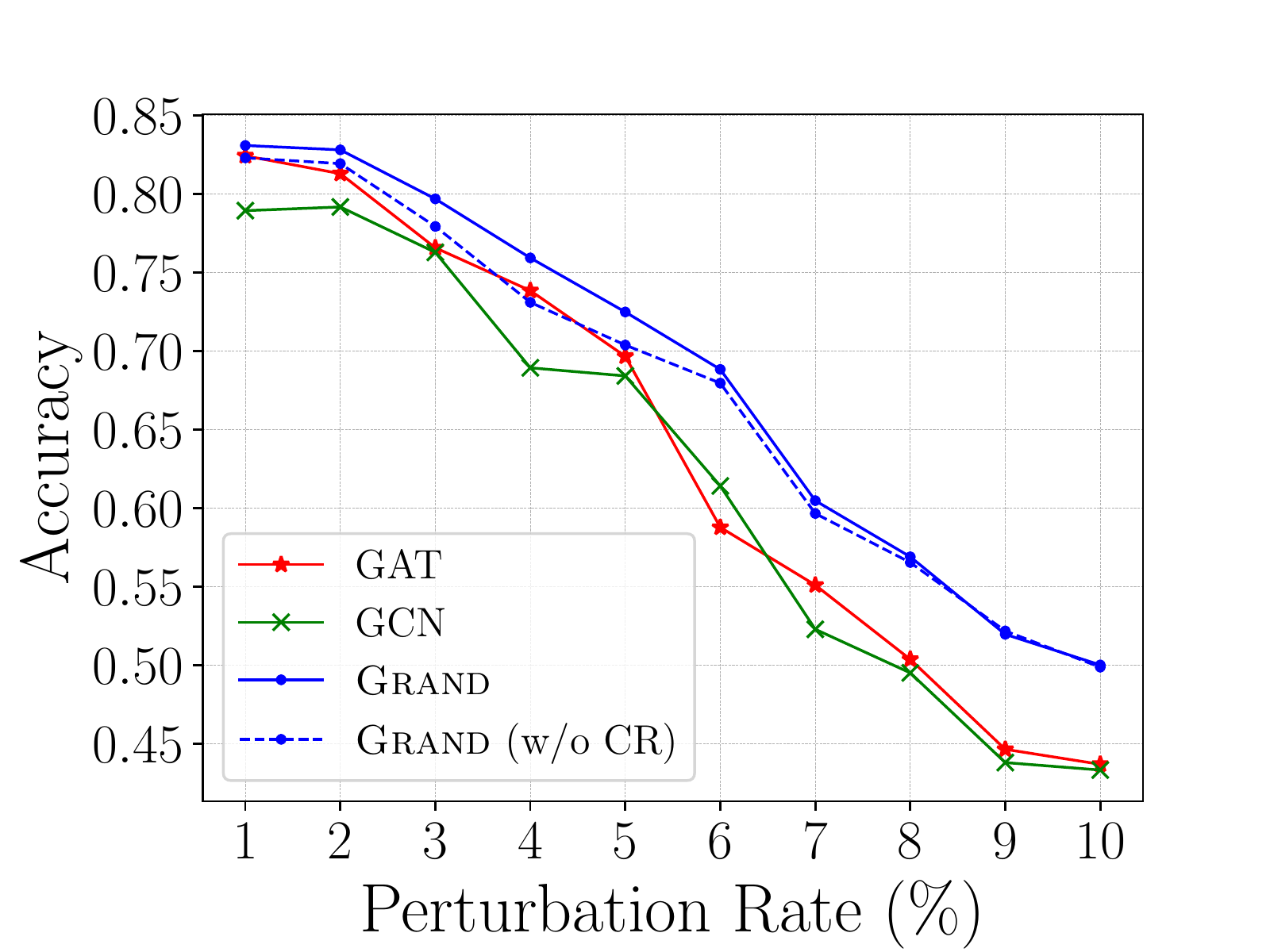}
			}
		\end{subfigure}
	}
	\caption{Robustness: results under  attacks on Cora.}
	\label{fig:robust}
\end{figure}

\label{sec:oversmoothing}
\begin{figure}[t]
	\centering
	\mbox
	{
		\begin{subfigure}[MADGap]{
				\centering
				\includegraphics[width = 0.45 \linewidth]{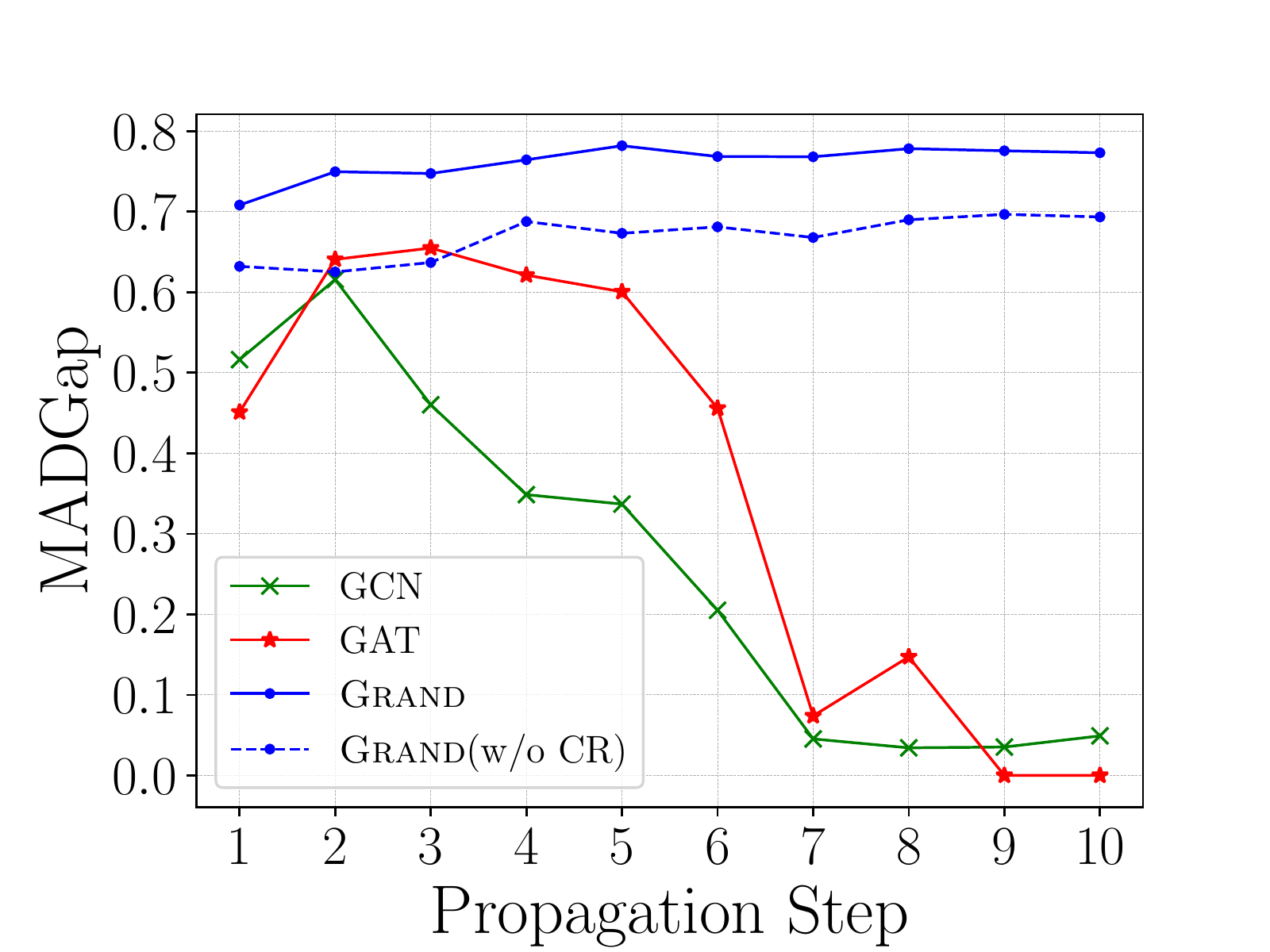}
			}
		\end{subfigure}
		\begin{subfigure}[Classification Results]{
				\centering
				\includegraphics[width = 0.45 \linewidth]{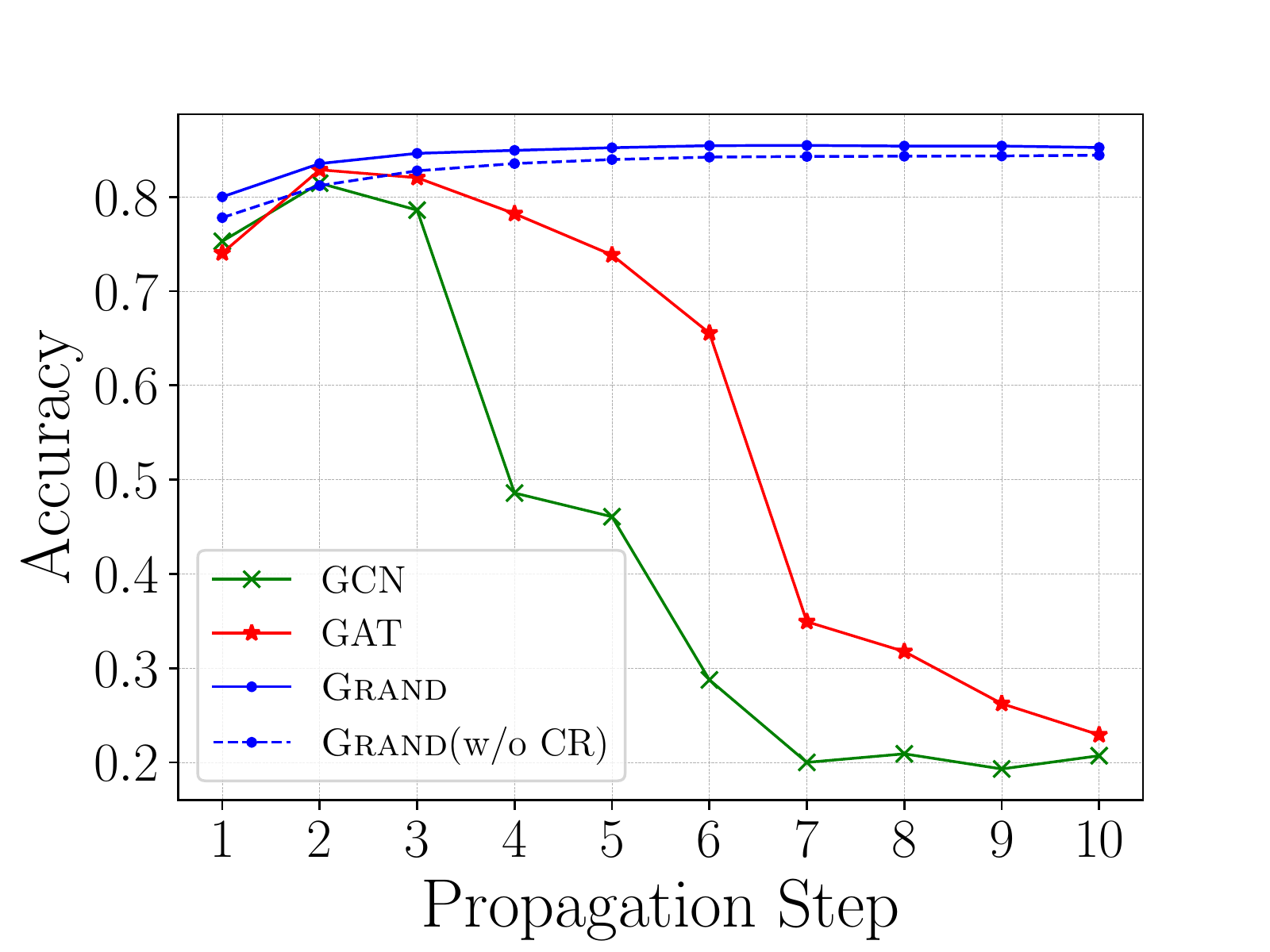}
			}
		\end{subfigure}
	}
	\caption{Over-smoothing: \model\ vs. GCN \& GAT on Cora.}
	\label{fig:grandmlp_vs_gcn}
\end{figure}


\subsection{Relieving Over-smoothing.}
\label{sec:oversmoothing}
Many GNNs face the over-smoothing issue---\textit{nodes with different labels  become indistinguishable}---when enlarging the feature propagation step~\cite{li2018deeper,chen2019measuring}.
We quantitatively study how \model\ is vulnerable to this issue by using MADGap~\cite{chen2019measuring}. 
MADGap measures the over-smoothness of node representations---the cosine distance between the remote and neighboring nodes, wherein the remote nodes are defined as those with different labels, and the neighboring nodes are those with the same labels. 
A smaller MADGap value indicates the more indistinguishable node representations and thus a more severe over-smoothing issue. 
Figure \ref{fig:grandmlp_vs_gcn} shows both the MADGap values of the last layer's representations and classification results of each GNN model with respect to different propagation steps.    
In \model, the propagation step is controlled by the hyperparameter $K$, while for GCN and GAT, we adjust the propagation step by stacking different hidden layers. 
The plots suggest that as the propagation step increases, both metrics of GCN and GAT decrease dramatically---MADGap drops from $\sim$0.5 to 0 and accuracy drops from 0.75 to 0.2---due to the over-smoothing issue. 
However, \model\ behaves completely different: both the performance and MADGap benefit from more propagation steps. 
This indicates that \model\ is much more powerful to relieve over-smoothing, when existing representative GNNs are very vulnerable to it. }

\subsection{Efficiency Analysis}
\label{sec:efficiency}
\label{sec:efficiency}
\begin{figure}[t]
	\centering
	\mbox
	{
		\begin{subfigure}[Per-epoch Training Time]{
				\centering
				\includegraphics[width = 0.45 \linewidth]{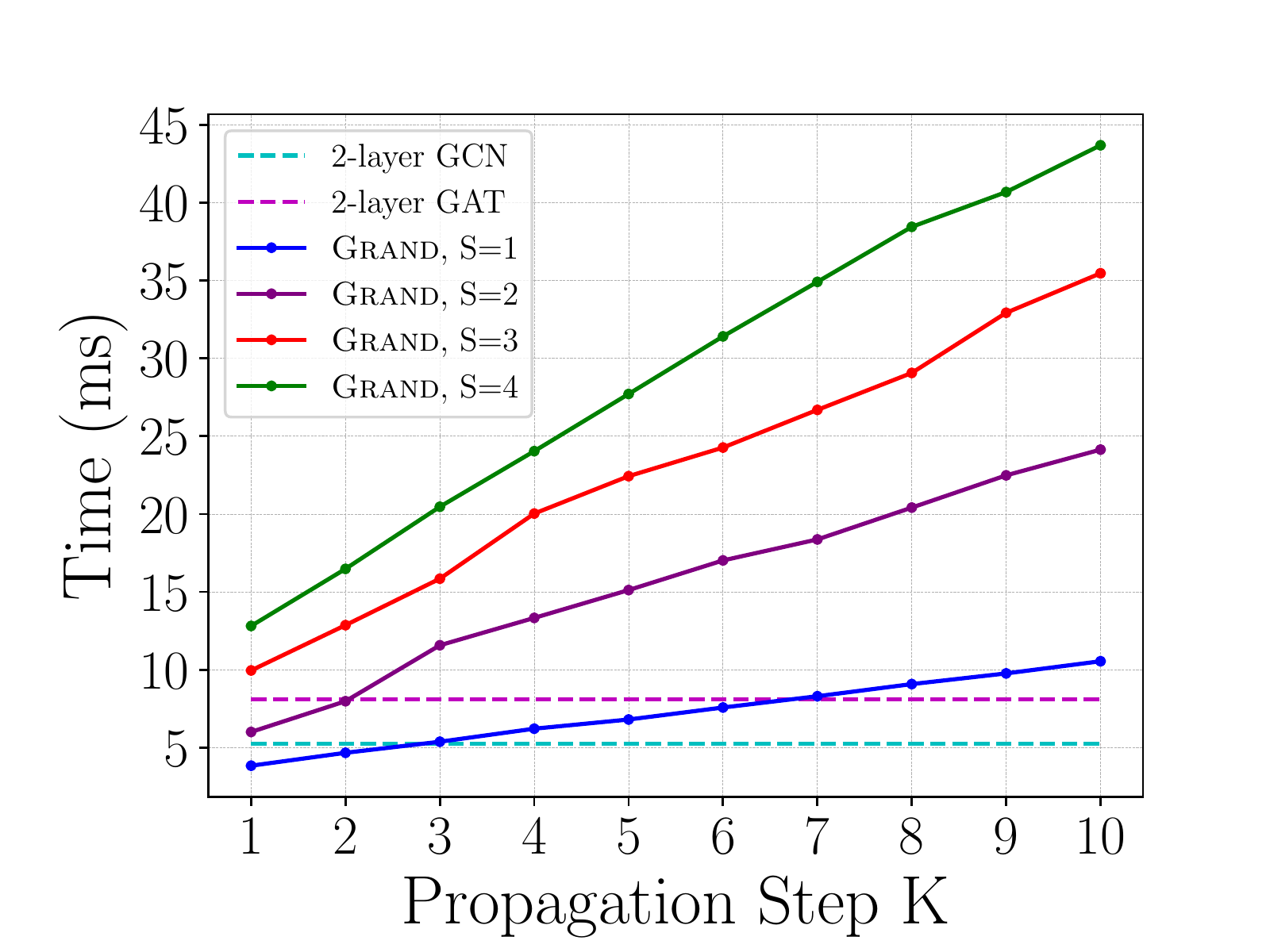}
			}
		\end{subfigure}
		\begin{subfigure}[Classification Accuracy]{
				\centering
				\includegraphics[width = 0.45 \linewidth]{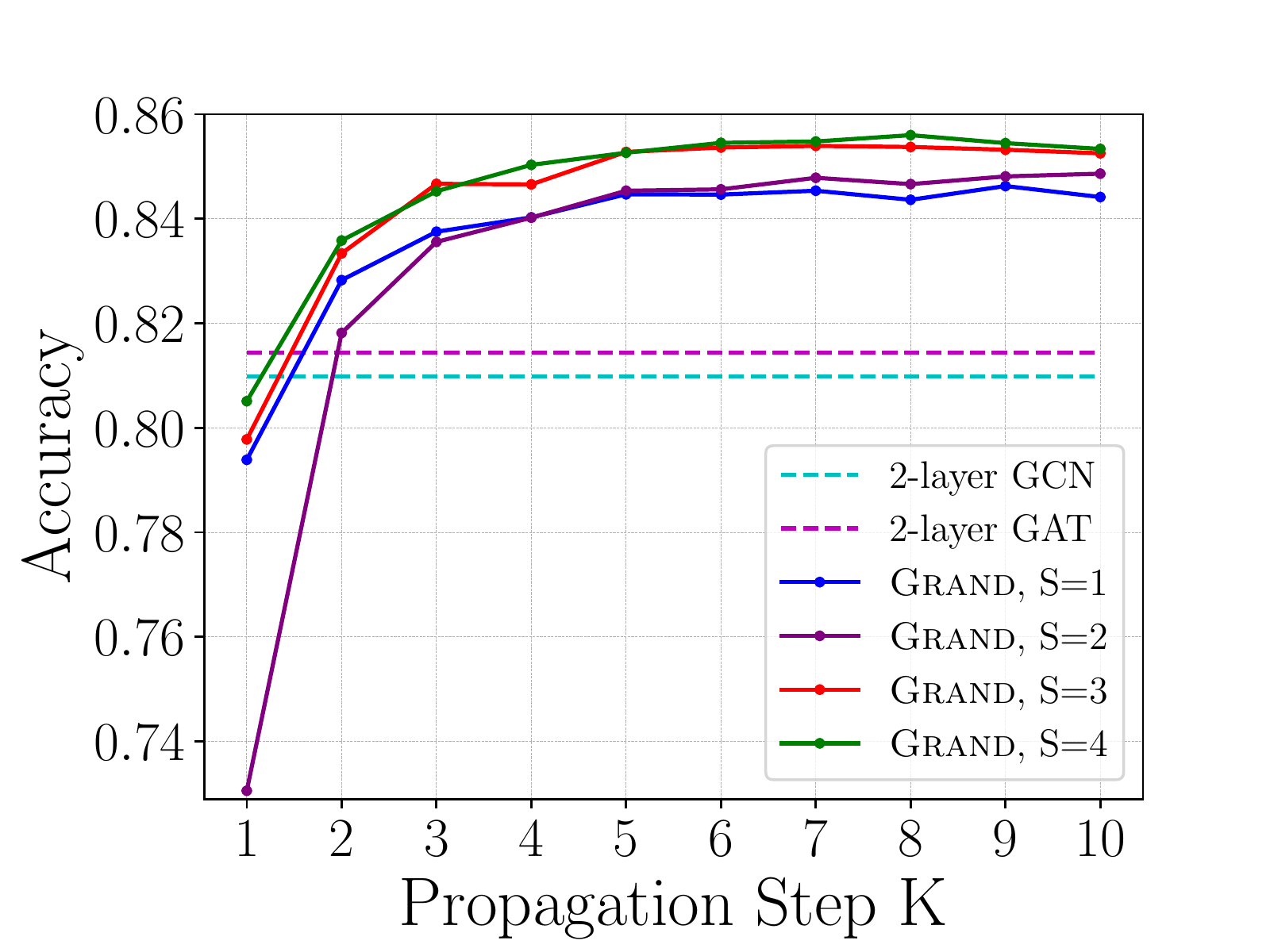}
			}
		\end{subfigure}
	}
	\caption{Efficiency Analysis for  \model.}
\label{fig:efficiency}
\end{figure}

The efficiency of \model\ is mainly influenced by two hyperparameters: the propagation step $K$ and augmentation times $S$. 
Figure \ref{fig:efficiency} reports the average per-epoch training time and classification accuracy of \model\ on Cora under different values of $K$ and $S$ with \#training epochs fixed to 1000. 
It also includes the results of the two-layer GCN and two-layer GAT with the same learning rate, \#training epochs and hidden layer size as \model. 

From Figure \ref{fig:efficiency}, we can see that when $K=2, S=1$, \model\ outperforms GCN and GAT in terms of both efficiency and effectiveness. 
In addition, we observe that increasing $K$ or $S$ can significantly improve the model's classification accuracy at the cost of its training efficiency. 
In practice, we can adjust the values of $K$ and $S$ to balance the trade-off between performance and efficiency. 


\subsection{Parameter Sensitivity}
\label{sec:paramsens}

\begin{figure}[t]
	\centering
	\mbox
	{
		\begin{subfigure}[CR loss coefficient $\lambda$]{
				\centering
				\includegraphics[width = 0.45 \linewidth]{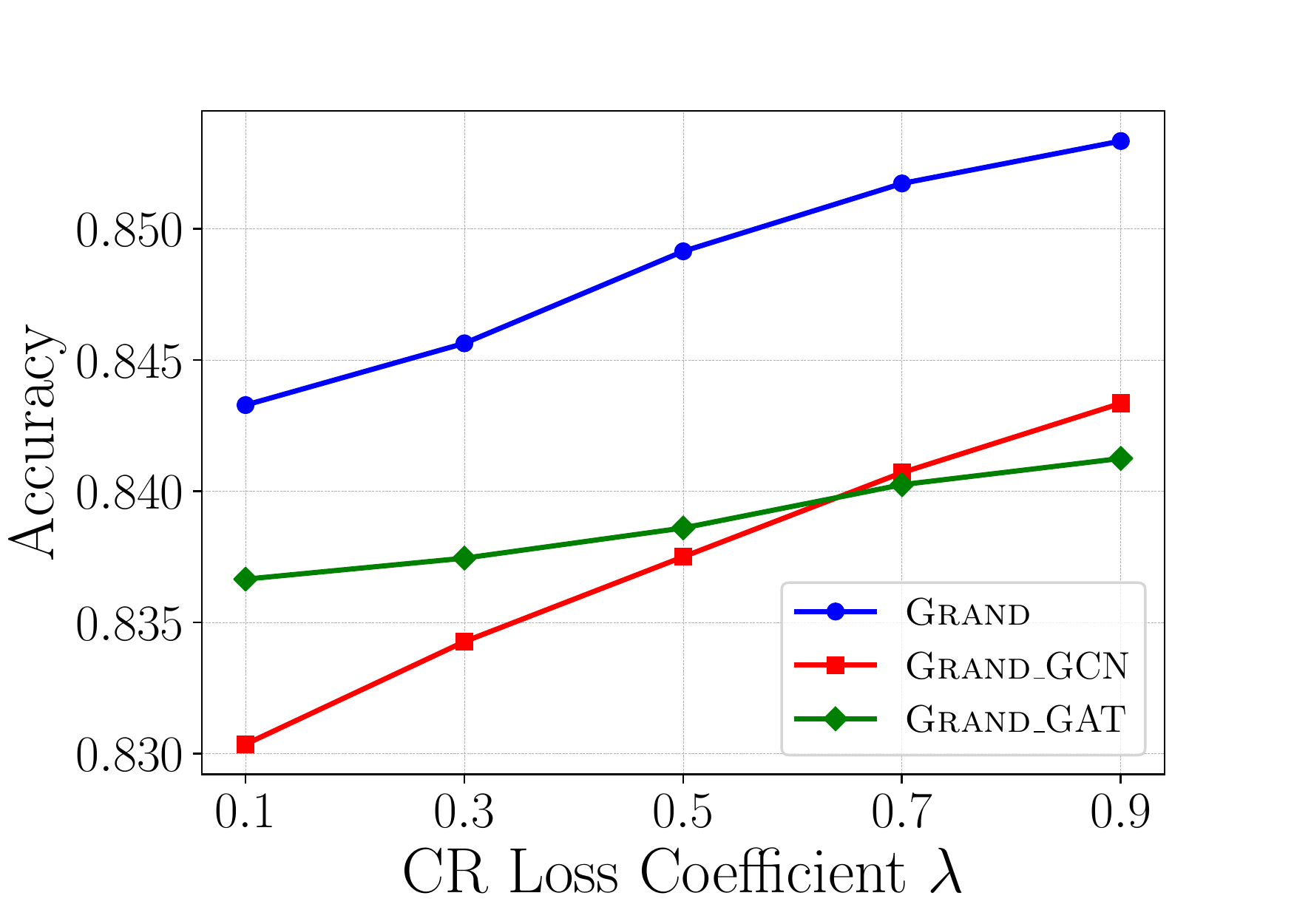}
			}
		\end{subfigure}
		\begin{subfigure}[DropNode probability $\delta$]{
				\centering
				\includegraphics[width = 0.45 \linewidth]{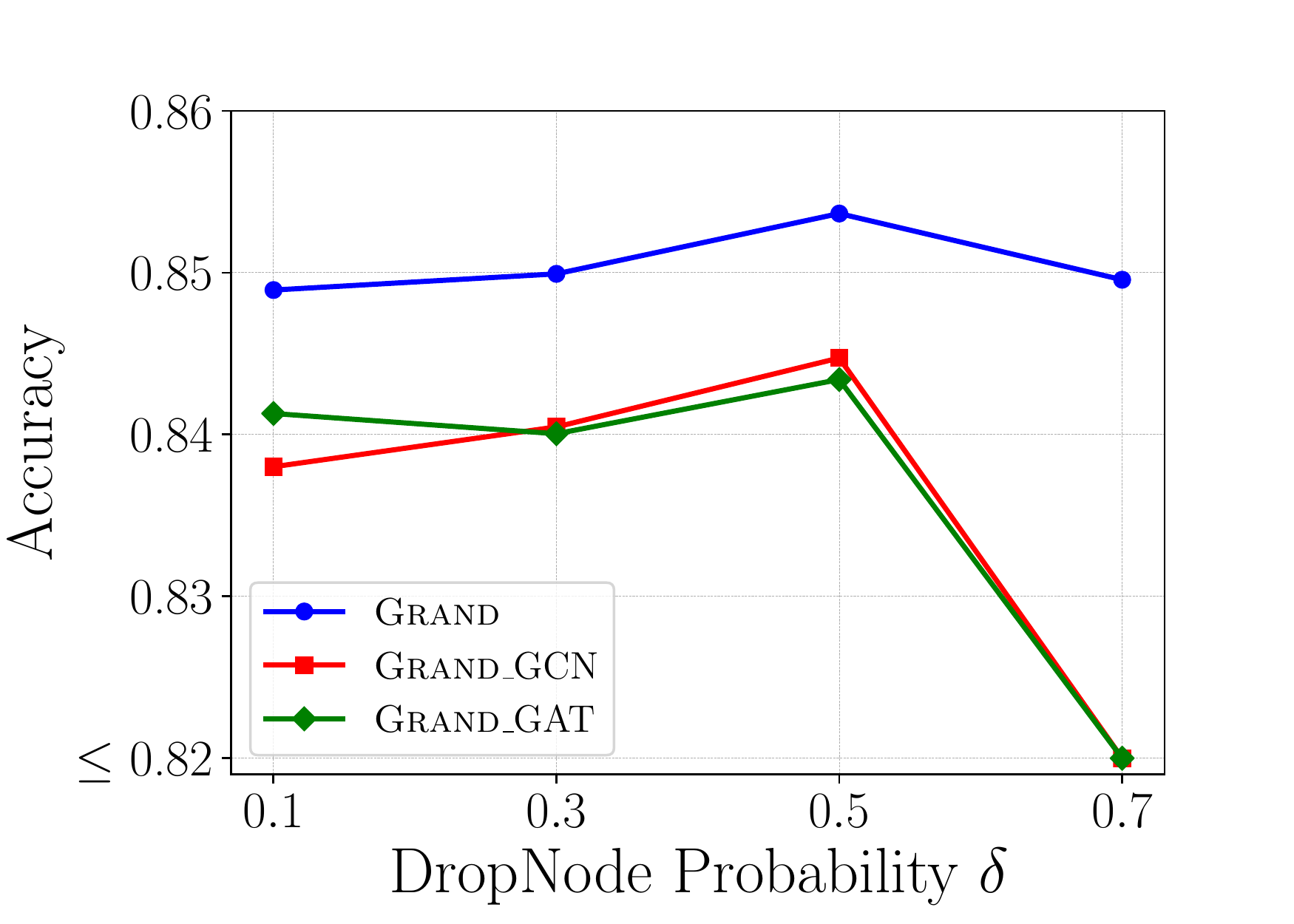}
			}
		\end{subfigure}
	}
	\caption{Parameter sensitivity of $\lambda$ and $\delta$ on Cora. 
	}
	\label{fig:para}
\end{figure}
We investigate the sensitivity of consistency regularization (CR) loss coefficient $\lambda$ and DropNode probability $\delta$ in \model\ and its variants on Cora. The results are shown in Figure~\ref{fig:para}. We observe that their performance increase when enlarging the value of $\lambda$. 
As for DropNode probability, \model, \model\_GCN and \model\_GAT reach their peak performance at $\delta = 0.5$. This is because the augmentations produced by random propagation in that case
are more stochastic and thus make \model\  generalize better with the help of consistency regularization.

\subsection{DropNode vs Dropout}
\label{sec:dropnode_vs_dropout}
\begin{figure}[ht]
	\centering
	\mbox
	{
		\begin{subfigure}[Cora]{
				\centering
				\includegraphics[width = 0.34 \linewidth]{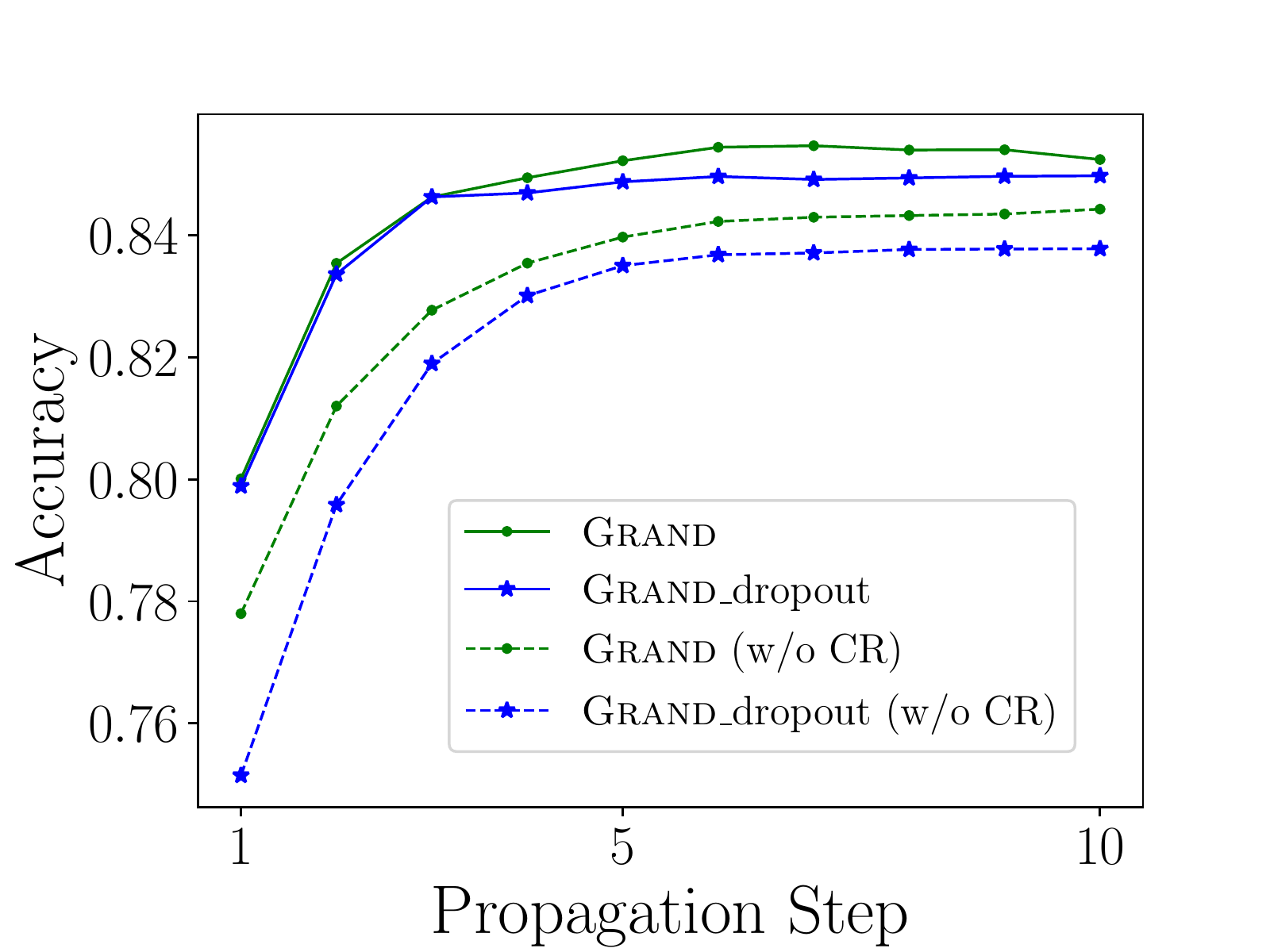}
				\label{fig:dropnode_vs_dropout_cora}
			}
		\end{subfigure}
		\begin{subfigure}[Citeseer]{
				\centering
				\includegraphics[width = 0.34 \linewidth]{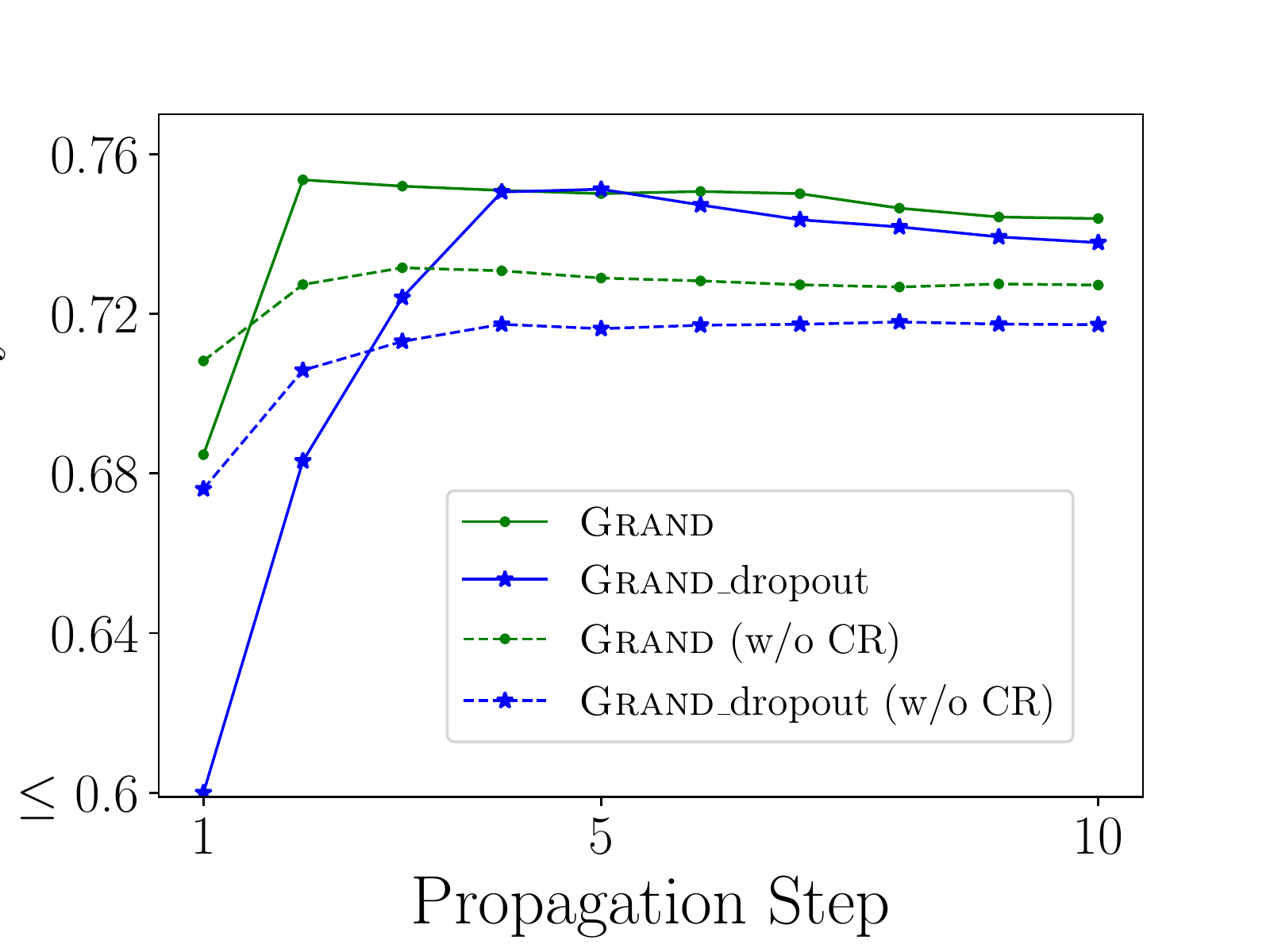}
				\label{fig:dropnode_vs_dropout_cite}
			}
		\end{subfigure}
		\begin{subfigure}[Pubmed]{
				\centering
				\includegraphics[width = 0.34 \linewidth]{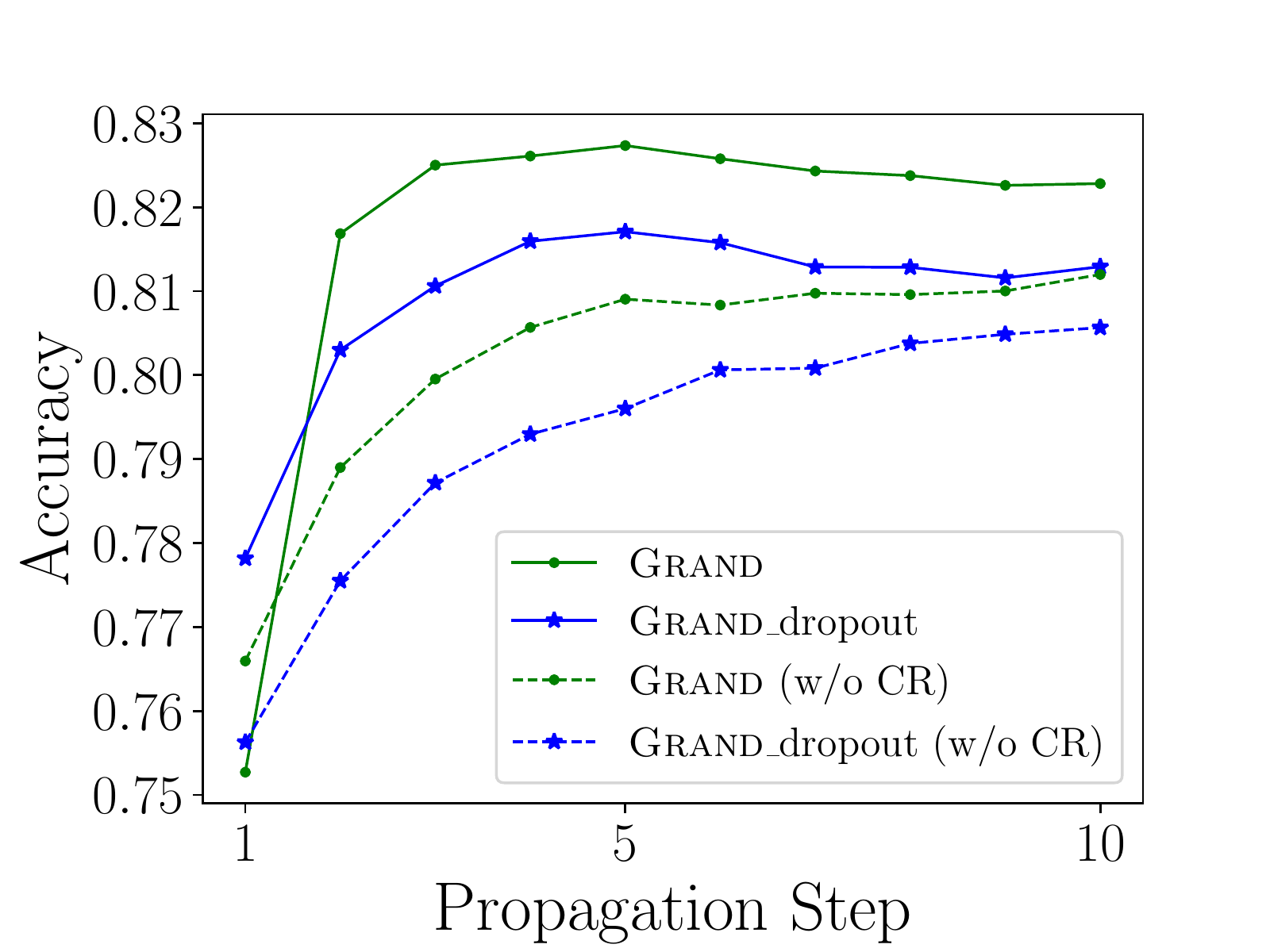}
				\label{fig:dropnode_vs_dropout_pub}
			}
		\end{subfigure}
	}
	
	\caption{\model\ vs. \model\_dropout.}
	\label{fig:dropnode_vs_dropout_2}
\end{figure}

We compare \model\ and \model\_dropout under different values of propagation step $K$. 
The results on Cora, Citeseer and Pubmed are illustrated in Figure \ref{fig:dropnode_vs_dropout_2}. We observe \model\ always achieve better performance than \model\_dropout, suggesting \textit{DropNode is much more suitable for graph data augmentation}.

\begin{figure}[h!]
	\centering
	\mbox
	{
		\begin{subfigure}[MADGap]{
				\centering
				\includegraphics[width = 0.45 \linewidth]{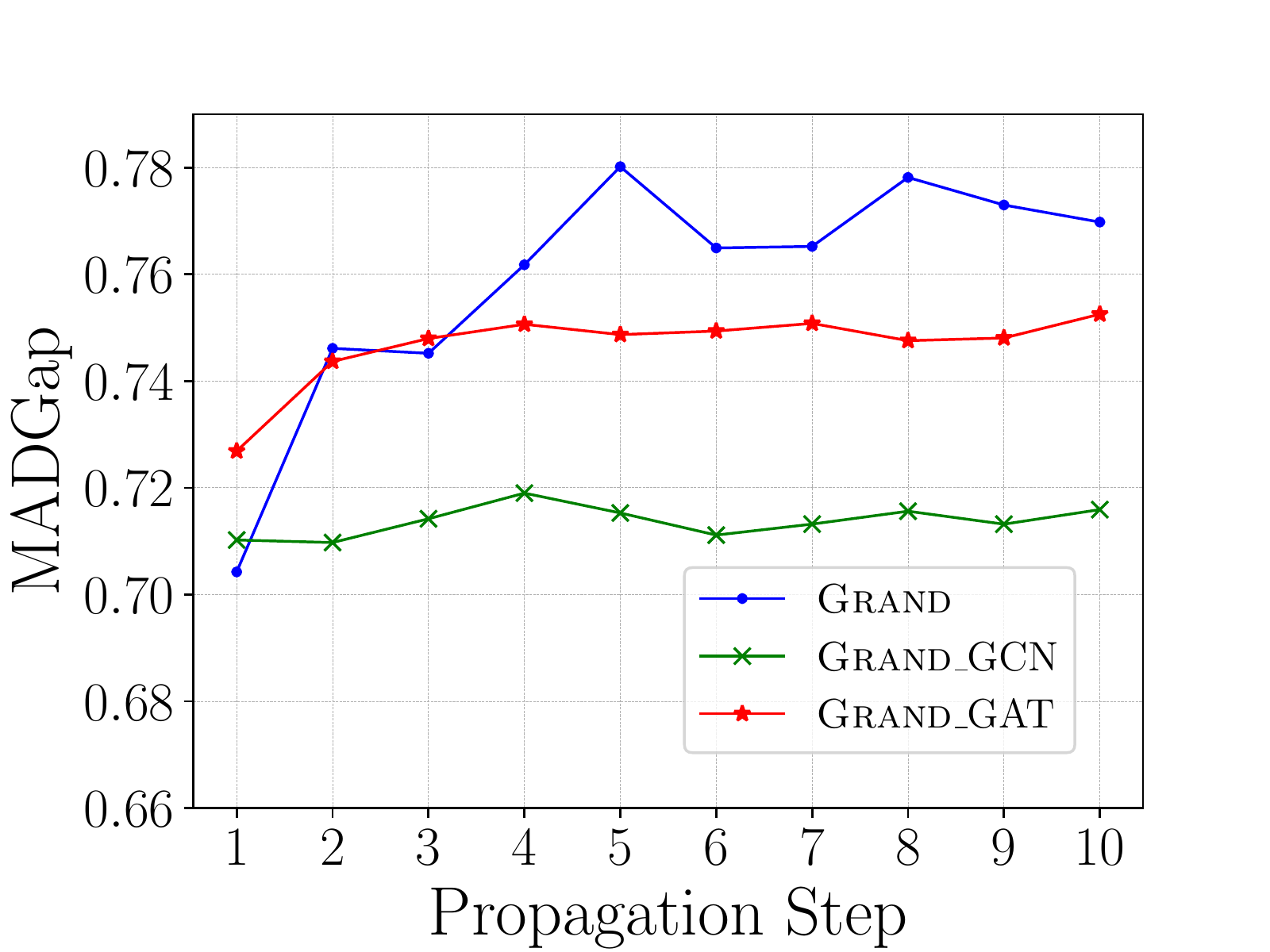}
			}
		\end{subfigure}
		\begin{subfigure}[Classification Results]{
				\centering
				\includegraphics[width = 0.45 \linewidth]{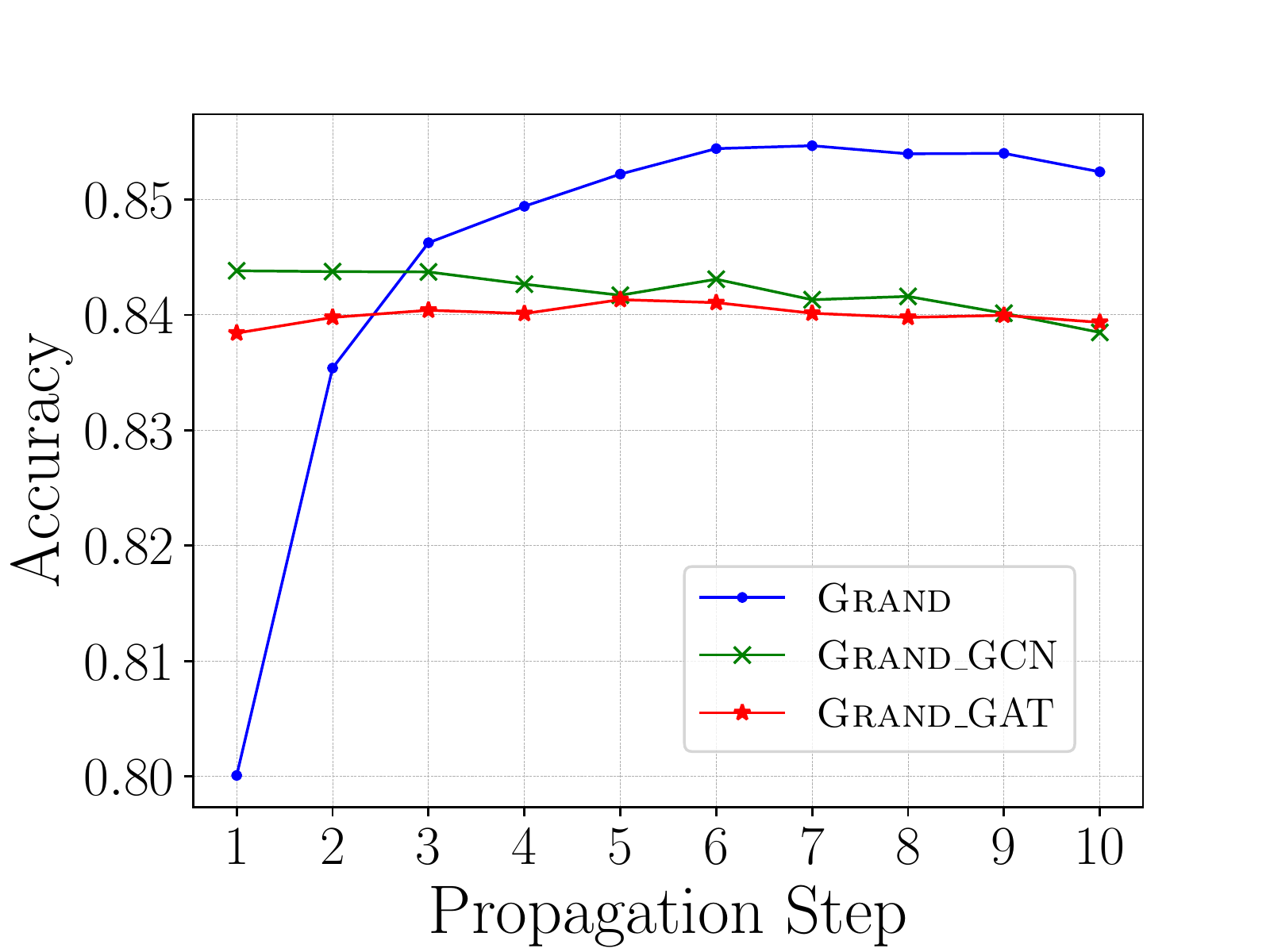}
			}
		\end{subfigure}
	}
	\caption{Over-smoothing: \model\ vs. \model\_GCN \& \model\_GAT on Cora.}
		\label{fig:grandgcn_vs_mlp}
\end{figure}

\subsection{\model\ vs. \model\_GCN \& \model\_GAT}
\label{sec:oversmoothing_grand}

As shown in Table \ref{tab:overall}, \model\_GCN and \model\_GAT get worse performances than \model, indicating GCN and GAT perform worse than MLP under the framework of \model. 
Here we conduct a series of experiments to analyze the  underlying reasons. 
 Specifically, we compare the MADGap values and accuracies \model, \model\_GCN and \model\_GAT under different values of propagation step $K$ with other parameters fixed. 
 The results are shown in Figure \ref{fig:grandgcn_vs_mlp}. We find that the MADGap and classification accuracy of \model\ increase significantly when enlarging the value of $K$. However, both the metrics of \model\_GCN and \model\_GAT have little improvements or even decrease. This indicates that \textit{GCN and GAT have higher over-smoothing risk than MLP}.
 
 \subsection{Performance of \model\ under different label rates}
 \label{sec:diff_label}
 We have conducted experiments to evaluate \textsc{Grand} under different label rates. For each label rate setting, we randomly create 10 data splits, and run 10 trials with random initialization for each split. We compare \textsc{Grand} with GCN and GAT. The results are shown in Table \ref{tab:label_rate}. We observe that \textsc{Grand} consistently outperforms GCN and GAT across all label rates on three benchmarks. 
\begin{table}[h]
	\centering
	\small
	\setlength{\tabcolsep}{1.0mm}
	\caption{Classification Accuracy under different label rates (\%).}
	\begin{tabular}{c|ccc|ccc|ccc}
		\toprule 
		Dataset &  \multicolumn{3}{|c|}{Cora} & \multicolumn{3}{|c|}{Citeseer} & \multicolumn{3}{|c}{Pubmed} \\ \midrule
		Label Rate &1\%&3\%&5\%&1\%&3\%&5\%&0.1\%&0.3\%&0.5\% \\
		\midrule
		GCN &62.8$\pm$5.3&76.1$\pm$1.9&79.6$\pm$2.1&63.4$\pm$2.9&70.6$\pm$1.7&72.2$\pm$1.1&71.5$\pm$2.1&77.5$\pm$1.8&80.8$\pm$1.5 \\
		\midrule
		GAT &64.3$\pm$5.8&77.2$\pm$2.4&80.8$\pm$2.1&64.4$\pm$2.9&70.4$\pm$1.9&72.0$\pm$1.3&72.0$\pm$2.1&77.6$\pm$1.6&80.6$\pm$1.2 \\
		\midrule
		\textsc{Grand}&\textbf{69.1$\pm$4.0}&\textbf{79.5$\pm$2.2}&\textbf{83.0$\pm$1.6}&\textbf{65.3$\pm$3.3}&\textbf{72.3$\pm$1.8}&\textbf{73.8$\pm$0.9}&\textbf{74.7$\pm$3.4}&\textbf{81.4$\pm$2.1}&\textbf{83.8$\pm$1.3}\\
	\bottomrule
	\end{tabular}
	\label{tab:label_rate}
	\vspace{-0.15in}
\end{table}

\hide{
\begin{table}[htbp]
	\caption{Results on Citation-CS.}
	\label{tab:aminer-cs}
	\begin{tabular}{ccccc}
		\toprule
		Method & GCN & GAT & \model\ \\
		\midrule
		Accuracy (\%) & 49.2 $\pm$ 1.5 & 49.6 $\pm$ 1.7 & \textbf{51.8 $\pm$ 1.3}  \\
		\bottomrule
	\end{tabular}
\end{table}
}

\hide{
\begin{table*}[]
    \centering
   \setlength{\tabcolsep}{0.01mm} \begin{tabular}{c|ccccccccc}
    	\toprule
        Class   &Artificial Intelligence& Information System & Parallel Computing& Computer Network & Information Security & Database and Data Mining & Software Engineering & Multimedia & Human-computer Interaction \\
        
        \# node  & 83172 &  17419 & 44841& 137328&  18263& 52401 &  13631&  18187&  13970 \\
    	\midrule
        Class & Robotics & Computational Theory & Computer-aided Design & Computer Vision & Natural Language Processing & Computer Graphics & Machine Learning & Bioinformatics & Signal Processing\\
        \# node  &16457&  18196&  30514&  35729&  21034&   6820&  24900&   8873&  31751 \\
   	\bottomrule
    \end{tabular}
    \caption{Statistics of different classes in Citation-CS.}
    \label{tab:my_label}
\end{table*}
}

\vspace{-0.1in}



\hide{
\section{Reproducibility}
\subsection{Running Environment}

\subsection{Baseline Code Sources}
GCN, GAT, FastGCN, GraphSAGE etc. are the official codes. Notably, GraphSAGE code for PPI is the old official version implemented in Tensorflow, and GraphSAGE for Cora, Citeseer and Pubmed is the new official version implemented in Pytorch, which performs better than the old in relatively small graphs. 

The GitHub link of most of the baselines listed in our paper can be found in their original papers.



\subsection{Datasets and Data Split}
The citation datasets, Cora, Citeseer, and Pubmed, are graph benchmarks and available in the Github websites of Planetoid, GCN,  GAT, FastGCN, etc. The training/validation/test split is conducted by Planioid and widely adopted by subsequent graph semi-supervised models. Note that the original data format is not friendly hence  there is another dataset version of Cora, Citeseer, and Pubmed in GitHub of Pytorch GCN (official) and Pytorch GAT(not official). But we confirm that toy dataset version has different data split compared with the original benchmarks. Thus we use the original  datasets and data split in our experiments, which are exactly the same as Planioid, GCN, GAT, etc.

The PPI dataset and its training/validation/test split are released in GraphSGE and also adopted as the graph inductive learning benchmark. We download GraphSAGE PPI dataset, and transform Json format into numpy format using the code provided in GAT GitHub. The dataset is suitable for inductive learning as the test data is unseen during training.

\subsection{Accuracy Mean and Standard Deviation}
When we run GCN, GAT, and other graph semi-supervised methods on Cora, Citeseer, and Pubmed, the resultant accuracy is very sensitive to the random initializations, even the mean accuracy averaged over 10 runs is usually slightly higher than the mean accuracy averaged over 100 runs. In Table~\ref{tab:res_1} we also report the maximum accuracy with different random seeds, only to intend to show the importance of repeated experiments. Hence we strictly follow the experimental setup in GCN and GAT, and every data point of our models are averaged over 100 runs with different random initializations. 

\subsection{\rank}
The hyper-parameters of \srank are mainly inherited from GAT, as GAT can be roughly reformulated in our model. Thus we almost didn't tune the hyper-parameters.
On Cora, Citeseer, and Pubmed datasets, we employ two-layer \rank. The architecture and hyper-parameters of models on Cora and Citeseer are exactly the same. 
The learning rate is 0.005 and l2 weight decay is 0.0005. The first layer consists of 16 propagation mechanisms with 1-hop and 2-hop propagation layers; the hidden dimension is 8; and the last layer only uses 1 propagation mechanism. Dropout mechanism with drop rate 0.6 is applied to the elements of the first layer diagonal matrix, and the adjacency matrix and inputs of each layer. 

On Pubmed, the learning rate is 0.01 and l2 weight decay is 0.001, the first layer consists of 16 propagation mechanisms with only 1-hop propagation layers, and the hidden dimension is 8. The output of the last layer is averaged over 8 propagation mechanisms. 

On PPI, a three-layer \srank is employed, with 4 propagation mechanism and 256 hidden dimension in the first and the second layers, and the output of the last layer is averaged by 6 propagation mechanism.

\subsection{NSGCNs}
The hyper-parameters of these sampling-based GCNs are mainly inherited from GCN with 2 layers and 32 hidden dimension. The learning rate is 0.01 and l2 weight decay is 0.0005. The dropout rate is 0.6. In Table~\ref{tab:res_2}, \shalf uses 10 propagation step for the sufficient validation of the network redundancy. But we suppose 5 is already sufficient so the propagation step of \sdrop and \sdm is set to 5. The sensitivity analysis of the propagation step is further conducted and Figure~\ref{fig:step} shows these sampling-based GCNs can be further optimized.

In \dm, we use $\hat{A} = D^{-1/2}AD^{-1/2}$ and $\bar{A} = D^{-1}A$ to further distinguish two GCNs in \dm, which, we suppose, may not be necessary. After propagating 5 times, the inputs of \sdm are dropout with rate 0.6. The $\lambda$ to weight the disagreement minimization loss in Equation (\ref{equ:loss}) is searched over  {0.01, 0.1, 1}, or according to the ratio of total node number to training node number(e.g., 2707/140 $\approx$ 20 in Cora). We set $\lambda =  20, 1, 0.01$ on Cora, Citeseer, and Pubmed datasets, respectively.

\subsection{Code}
We have all of our code on GitHub (private repository). We planned to release our code in a new GitHub account, but we're afraid this could not protect our anonymity completely. What's more, someone with affiliation information may leave some comments or even watch/fork our GitHub project, and thus reviewers would suspect s/he is related to us and even is one of the authors. After discussion, we decide to strictly follow the anonymity rules and promise to make our code available after the review is over.


\subsection{Complexity Analysis}
\label{secsub:complexity}

Computationally, \srank adds the diagonal matrix and multi-head propagation mechanism to the original GCN architecture, thus the computational complexity is also linear with the volume of the network, which is the same as the sparse version of GCNs.  Meanwhile, the calculation of network ranking can be parallelized across all nodes. 

\shalf and \sdrop only add the propagation preprocessing step in Equ \ref{equ:kAX} to the original GCN, and \sdm doubles the computation of \drop. 
Note that both the propagation preprocessing and convolutional operations in GCNs can be complemented via sparse matrix operations. Hence the computational complexity is linear with the volume of the network and the empirical running time is almost the same as the sparse version of GCNs. Practically, all the proposed models can also benefit from the recent development in GCNs' computational acceleration and scalable parallelization.
}
\end{appendices}

\end{document}